\documentstyle[a4,12pt,alltt,leqno,fleqn,twoside]{report}

\newcommand{\idx}[1]{#1\index{#1}}
\newcommand{\ttidx}[1]{{\tt #1}\index{{\tt #1}}}

\newcommand{\anglevr}[1]{$\langle${\it #1}$\rangle$}
\newcommand{\anglevar}[1]{$\langle #1 \rangle$}

\newcommand{\var}[1]{$\it \langle #1 \rangle$}

\newcommand{\form}{{\tt form}}
\newcommand{\term}{{\tt term}}
\newcommand{\forms}{{\tt form}s}
\newcommand{\terms}{{\tt term}s}

\newenvironment{quoteqlf}{\begin{small}\begin{list}{}%
{\setlength{\leftmargin}{0pt}\setlength{\rightmargin}{0pt}}\item\nopagebreak}%
{\end{list}\end{small}}

\newenvironment{quotenr}{\begin{list}{}%
{\settowidth{\leftmargin}{\tt xxx}\setlength{\rightmargin}{0mm}}\item\pagebreak[3]}%
{\end{list}}

\newcommand{\inx}[1]{#1}

{\begin{list}{$\bullet$}{\addtolength{\parsep}{-.25\parsep}
\setlength{\itemsep}{0in}
\setlength{\topsep}{0in}}}%
{\medskip \end{list}}

\begin{document}
\hoffset=-6mm
\setcounter{page}{0}
\pagestyle{empty}
\begin{center}
\vspace*{30mm}
\Large{CLARE: A Contextual Reasoning and Cooperative \\ Response
Framework for the Core Language Engine \\
\ \\
Final Report}

\vspace*{20mm}
\large{Hiyan Alshawi, David Carter, Richard Crouch, \\
Steve Pulman, Manny Rayner and Arnold Smith \\ 
{\it with identified additional contributions}}

\vspace*{20mm}
\large{December 1992}

\vspace*{10mm}
Prepared for CLARE Natural Language Processing Club \\
DTI Information Technology Division \\
\ \\
IEATP Project IED4/1/1165 \\
SRI Project 8468 \\

\vspace*{20mm}
\large{\copyright{} 1992 SRI International} \\
Cambridge Computer Science Research Centre \\
23 Millers Yard, Mill Lane, Cambridge CB2 1RQ, UK \\
Tel +44-223-324146, Fax +44-223-460402 \\
E-mail: \verb!sriclare@cam.sri.com!
\end{center}

\newpage
\pagestyle{myheadings}
\markboth{\sc CLARE final report}{\sc CLARE final report}

\section*{Executive Summary}

This document is the final report of CLARE, a project involving
BP Research, British Aerospace, British Telecom, Cambridge University, 
SRI Cambridge and the UK Defence Research Agency. The project received
a grant from the UK Department of Trade and Industry.
The report mainly describes the research, design and implementation work 
carried out in building the CLARE system at SRI, with some discussion
of experimentation with the software by the other partners. Low-level
interfacing issues and a guide to using the system are covered in
a manual provided with the final release of the software.
The project also involved a study by Cambridge University Computer 
Laboratory on evaluating natural language processing systems. 
A digest of the report for this study appears at the end of the
present report.

CLARE was designed as a natural language processing system with 
facilities for reasoning and understanding in context and for
generating cooperative responses.
The work plan for the project required both further development of the
Core Language Engine (CLE) natural language processor and the design
and implementation of new components for reasoning and response generation.
All the milestones set in the project plan were achieved, the final
system including the following capabilities:

\begin{itemize}

\item Wider coverage of English syntax and semantics than the original
CLE system. This is quantified in the report.

\item Lattice based lexical, phrasal and syntactic analysis. Partial
syntactic and semantic analysis of sentences outside coverage.

\item Gradual refinement of meaning in context according to a theory 
of ``monotonic interpretation'' employing an improved Quasi Logical Form 
representation.

\item Ranking alternatives produced by sentence processing with the use of
an extensible set of preference metrics, including semantic collocations.

\item Generation of informative paraphrases that indicate the system's
interpretation of English inputs.

\item Use of an abductive reasoning mechanism to support the 
processes of interpretation and of interfacing to an application.

\item Declarative specification of database domain models, including the 
specification of equivalence axioms used by the reasoner to translate
between linguistic and database representations.

\item Generation of natural language statements from database assertions,
and questions from incomplete assertions.

\end{itemize}

For all of these capabilities, the CLARE system has advanced the state
of the art either through the use of novel techniques developed on
the project, or by extending the coverage or scale of known techniques.
On the negative side, we had to specialize our development in the interfacing
part of the project to structured database system interfaces. The reason
for this is that general, or unstructured, knowledge-based systems currently 
appear too varied to allow the degree of integration between the language 
and back-end systems achieved in CLARE with database systems.
Nevertheless, the language components remain application-independent 
and provide interfaces for the development of new types of application.

CLARE was used during the project to build a number of database interface
systems: the example Project Resource Management application at SRI,
a similar database interface at BT Research Laboratories, and a personnel 
database application at BP.
The language processing components of CLARE are currently being used
in experimental work on spoken language systems and text processing:
an interface to a road map routing system at DRA Malvern, an English-Swedish 
speech translation project at SRI, and research on
processing specification texts at Cambridge University.

\tableofcontents

\cleardoublepage

\chapter{Components of CLARE}
\label{clarecomponents}

\section{CLARE System Architecture}

This chapter serves as a short introduction to the architecture
of the CLARE system. Readers unfamiliar with earlier NATTIE and
CLARE project reports are likely to find it particularly useful.

CLARE is a natural language processing system with built-in
facilities for domain modeling and reasoning. The linguistic
processing components, which together are referred to as the
Core Language Engine (CLE), are application independent, while
the domain modeling and reasoning capabilities are aimed at building
interactive natural language interfaces to knowledge based systems.
For example, the CLE has been used to build an experimental
machine translation system for English-Swedish car hire dialogues
(Alshawi et al. 1991),
and CLARE has been used to build an interface to a Project Resource
Management (PRM) database. 
Much of the work on reasoning is discussed in this report in the
context of the PRM application. The approach to language-application
interfacing depends on translating logic expressions using a method
we call ``abductive equivalential translation'' (Chapter~\ref{translation}).

CLARE can be viewed as consisting of five subsystems which
in turn consist of processing and data components. The subsystems
perform the following functions:

\begin{itemize}

\item Language analysis

\item Language generation

\item Lexical acquisition

\item Contextual interpretation

\item Domain modeling and reasoning.

\end{itemize}

The language analysis and lexical acquisition components are similar
in operation to those used in the version of the CLE resulting from
the NATTIE project, except that coverage has been considerably increased
under CLARE. (Chapter~\ref{clarecoverage} describes experiments carried
out to assess linguistic coverage.) Contextual interpretation has been 
largely redesigned under CLARE as described in Chapter~\ref{clareinterp}.
Language generation and domain modeling and reasoning were developed
as new subsystems under CLARE.

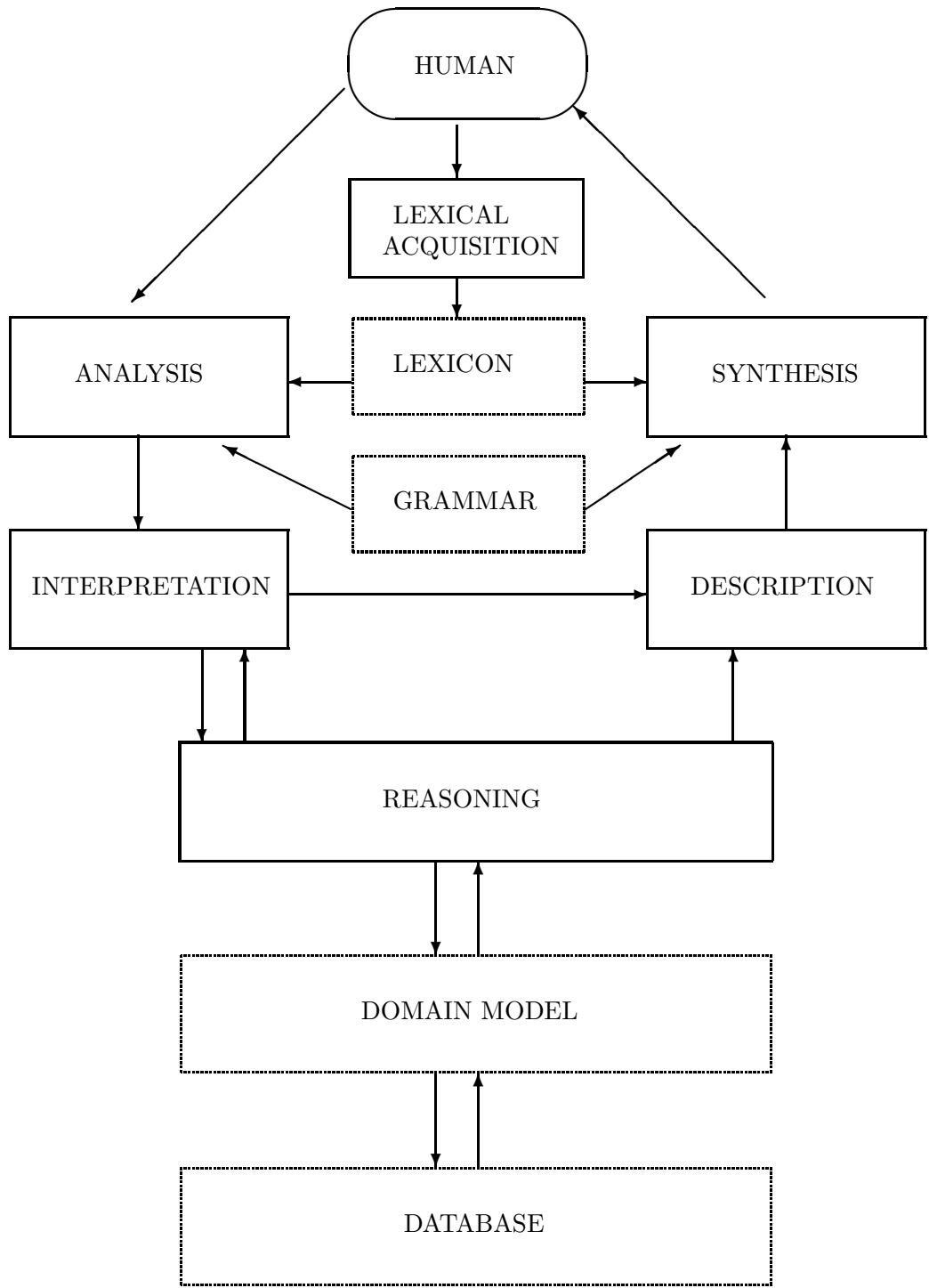
\begin{figure}
\setlength{\unitlength}{0.0125in}%
\begin{picture}(430,601)(20,185)
\thicklines
\put(235,760){\oval(112,52)}
\put(385,540){\vector( 0, 1){ 45}}
\put(320,485){\framebox(130,55){}}
\put(180,550){\vector(-2, 1){ 60}}
\put(290,550){\vector( 3, 2){ 45}}
\put(181,530){\dashbox(109,45){}}
\put(180,660){\framebox(109,45){}}
\put(290,610){\vector( 1, 0){ 30}}
\put(180,610){\vector(-1, 0){ 30}}
\put(230,660){\vector( 0,-1){ 20}}
\put(320,585){\framebox(130,55){}}
\put(181,595){\dashbox(109,45){}}
\put(240,240){\vector( 0, 1){ 45}}
\put(220,285){\vector( 0,-1){ 45}}
\put(240,340){\vector( 0, 1){ 45}}
\put(220,385){\vector( 0,-1){ 45}}
\put(130,440){\vector( 0, 1){ 45}}
\put(375,650){\vector(-1, 1){ 90}}
\put(177,748){\vector(-1,-1){100}}
\put(230,731){\vector( 0,-1){ 25}}
\put(150,510){\vector( 1, 0){170}}
\put(360,440){\vector( 0, 1){ 45}}
\put(110,485){\vector( 0,-1){ 45}}
\put( 80,585){\vector( 0,-1){ 45}}
\put(100,385){\framebox(278,55){}}
\put(100,185){\dashbox(278,55){}}
\put(100,285){\dashbox(278,55){}}
\put( 20,585){\framebox(130,55){}}
\put( 20,485){\framebox(130,55){}}
\put(340,510){\makebox(0,0)[lb]{\raisebox{0pt}[0pt][0pt]{\small DESCRIPTION}}}
\put(350,610){\makebox(0,0)[lb]{\raisebox{0pt}[0pt][0pt]{\small SYNTHESIS}}}
\put(200,550){\makebox(0,0)[lb]{\raisebox{0pt}[0pt][0pt]{\small GRAMMAR}}}
\put(200,615){\makebox(0,0)[lb]{\raisebox{0pt}[0pt][0pt]{\small LEXICON}}}
\put(210,755){\makebox(0,0)[lb]{\raisebox{0pt}[0pt][0pt]{\small HUMAN}}}
\put(205,210){\makebox(0,0)[lb]{\raisebox{0pt}[0pt][0pt]{\small DATABASE}}}
\put(195,410){\makebox(0,0)[lb]{\raisebox{0pt}[0pt][0pt]{\small REASONING}}}
\put(185,310){\makebox(0,0)[lb]{\raisebox{0pt}[0pt][0pt]{\small DOMAIN MODEL}}}
\put(195,670){\makebox(0,0)[lb]{\raisebox{0pt}[0pt][0pt]{\small ACQUISITION}}}
\put( 30,510){\makebox(0,0)[lb]{\raisebox{0pt}[0pt][0pt]{\small INTERPRETATION}}}
\put( 50,610){\makebox(0,0)[lb]{\raisebox{0pt}[0pt][0pt]{\small ANALYSIS}}}
\put(200,685){\makebox(0,0)[lb]{\raisebox{0pt}[0pt][0pt]{\small LEXICAL }}}
\end{picture}
\caption{Information flow between CLARE data and processing components
(data components in dashed boxes)}
\label{clarearchfig}
\end{figure}

There are two main representations used in CLARE. The first, Quasi
Logical Form or QLF, is a logic-like language motivated by the needs
of language processing. It extends first order logic by the use of
generalized quantifiers, higher order operators, and constructs
for expressing underspecification of contextually dependent aspects
of interpretation. Language analysis produces QLFs which are then
refined by contextual interpretation. QLFs are also the input to
the algorithm for linguistic generation. The second representation,
Target Reasoning Language (or TRL), is a logic aimed at tractable
reasoning; it is much simpler than QLF and close to ``standard'' 
first order logic. Contextually resolved logical forms are translated
into TRL, and the domain model is also expressed in TRL.
Figure~\ref{clarearchfig} shows, in schematic form, the relationship
between CLARE subsystems.

\section{Processing Phases}

The language analysis subsystem  makes use of a general purpose 
unification grammar for English syntax and semantics, and a core
lexicon with around 2000 common English word senses. The lexical acquisition
tool allows application developers to augment the core lexicon with
application specific vocabulary. (An interface is also provided to
allow a large external lexicon to be consulted during linguistic
processing.)
Linguistic processing employs a staged approach with the following
phases:
\begin{enumerate}
\item  Segmentation
\item  Morphological analysis
\item  Syntactic analysis
\item  Semantic analysis
\item  Sorts and preference ordering
\item  User interaction (optional).
\end{enumerate}
The current approach to segmentation (i.e. low-level lexical processing)
is presented in Chapter~\ref{clarelexproc}.
Linguistic analysis produces a set of contextually underspecified
QLFs ordered by a preference metric (the preference mechanism is
described in Chapter~\ref{clarepref}).

Contextual interpretation takes underspecified QLFs,
and increases their information content by the processes of reference
resolution (resolving noun phrase referents, vague relations, ellipsis,
and so on) and quantifier scoping. Further preference metrics are applied,
and natural language generation allows user verification of some of the 
choices made during interpretation. The phases of contextual 
interpretation are as follows:

\begin{enumerate}
\item  Reference resolution
\item  Sorts and preference ordering
\item  Paraphrasing substantive resolutions
\item  User interaction (optional)
\item  Quantifier scoping
\item  Preference ordering.
\end{enumerate}

We call the model for interpretation used in CLARE ``monotonic semantic 
interpretation'' (Chapter~\ref{monosem}) because it acts by
further instantiation of underspecified representations without destructive
manipulations or loss of information.  Interpretation, particularly
reference resolution, often makes use of the CLARE domain model and reasoning
components (Chapter~\ref{clareinterp}).

In parallel to the division of natural language understanding 
into analysis and interpretation, we also divide natural
language generation into {\it synthesis} (i.e. linguistic generation) 
and {\it description}. Both are covered in Chapter~\ref{generation}.
Synthesis takes place from QLF representations by ``reverse''
application of the unification-based grammar rules and lexical entries.
Description is the process of producing a QLF from which synthesis can 
take place. In CLARE, description is used for two purposes: generation of
paraphrases showing the results of interpretation and 
generating from database assertions
including the generation of questions from incomplete assertions. 
Generation from assertions can take place only if a domain
model for interfacing to a database has been defined.

For building a knowledge-based application with CLARE, a domain model
(Chapter~\ref{claredm}) needs to be specified. We have developed a methodology
for declarative specification of a domain model for database interface
applications. This methodology is described in detail in the CLARE
software manual. For such applications, much of the domain model takes
the form of logical equivalences in TRL which act as translation
rules between predicates corresponding to word senses and those corresponding
to database relations. Processing after contextual interpretation
comprises:
\begin{enumerate}
\item Translation of resolved QLFs to TRL
\item Translation of TRL to abstract database form
\item Further translation to concrete query or command (e.g. SQL)
\item Execution of query or assertion command.
\end{enumerate}
The translation of queries or assertions with linguistic 
predicates to expressions with database predicates (Chapter~\ref{translation})
makes use of a reasoner described in Chapter~\ref{ReasoningEngine}.
The reasoning process is abductive: it
can be used to derive a set of assumptions, given which the database
expression is equivalent to the original linguistic logical form. 
Consequently, assumptions implicit in the database
can be coded as conditions on the translation rules and fed back
to the user when they are invoked.

\section{Transcript of CLARE PRM Application}

Before describing the technical details of CLARE in the main chapters of
this report, we present a transcript of running the system with
the Project Resource Management (PRM) application. This should give an
informal impression of how the system operates and motivate some
of the technical problems addressed later. Although
the transcript session is very application-specific, it should be 
born in mind that CLARE, and especially its language processing 
components, have been used in several other test applications.

In the PRM application, CLARE is acting as an interface to an ORACLE 
database containing a dozen or so relations about projects and payments.
The structure
of the database has its roots in an actual system used at SRI
Cambridge, but for confidentiality reasons the original database
records have been replaced by fictitious ones.
The transcript is edited for brevity, though another version of
it that preserves more of the original output layout is given in
the CLARE software manual.  System and user output is shown in
{\tt typeface} font, comments in {\em italic} font.

\begin{verbatim}
user>> Show me the projects.
\end{verbatim}
{\em A Simple query.
After semantic analysis, a bracketing of the sentence is produced, plus a
list of word senses.  Preference measures normally select the appropriate
analysis first,  but the user has the option to accept or reject.

After reference resolution, a paraphrase of the resolved QLF is produced if the
resolutions are non-trivial.  Here it is shown that `the projects'
has been understood as meaning `all projects'.

Beneath the paraphrase is a list of noun phrases generated from the resolved
sentence, and a list of new relations.  The noun phrase list often
repeats what is already in the paraphrase.  But sometimes, generation of
the paraphrase fails, and generation of its component noun phrases can provide
valuable feedback to the user.  The new relations (in this case empty)
show what new word senses have been introduced by resolution.

After scoping and conversion to SQL, the answer is printed out, with a key
below.  In this case, the key says that the first item in every pair is
the project, and the second item its project number.}
\begin{verbatim}
   Complete sentence with bracketing:

     "{show {me} {the projects}}."

   Word senses (unordered):

     show: cause to see (rather than "reveal")
     project: activity

Confirm this analysis? (y/n/c/p/?): y

system>>   "show me all projects!"

Confirm this interpretation? (y/n): y

Answer:
        [Archie, 2989]
        [AstroSUD, 1254]
        [Bland Conversation Sprinkler, 8744]
        [CASE Workbench, 8743]
        [CLAM-BAKE, 8468]
        [CandyFloss, 8306]
        [Life Study, 7518]
        [RFO-1990, 1152]
        [WHIZ, 1392]

Key for answer fields:    
        [A, B]
   A: project
   B: account number, number, project number

user>> When did CLAM-BAKE start?

system>>   "when did CLAM-BAKE (the project) start?"

Answer: 20/11/1989

user>> When did it end?

system>>   "when did CLAM-BAKE (the project) end?"

Answer: 19/11/1992

user>> How many CLAM-BAKE payments were made in 1990?
\end{verbatim}
{\em This sentence illustrates the resolution of compound nouns.  Given the
domain model, a plausible interpretation of `CLAM-BAKE payments' is 
`payments on CLAM-BAKE'.  This contrasts with an expression like
`Marble Systems payments', which means `payments to Marble Systems'.

The list of the assumptions states that only payments made by CRI are
counted in the answer.  If anyone else (not covered by the database)
made payments on CLAM-BAKE, they would not be counted.}
\begin{verbatim}
system>>   "how many payments on CLAM-BAKE (the project) were made
            in 1990 (the year)?"

Assumptions used in query translation:

  all_transactions_referred_to_are_from_CRI

Answer: 45

user>> And last year?
\end{verbatim}
{\em Resolution of temporal noun phrases, and ellipsis resolution:  The
paraphrase illustrate how this elliptical question is understood in full.

The warning after the answer is printed because the
database is only complete for CRI payments up until 1/4/1991, as stated
in the list of assumptions.  Since this period ends half way through last
year, there may be some payments made last year that are not included in 
the database, and so not counted in the answer.}
\begin{verbatim}
system>>   "how many CLAM-BAKE (the project) payments were made 
            in 1991 (the year)?"

Assumptions used in query translation:

  transactions_referred_to_made_between_1989/8/17_and_1991/4/1
  all_transactions_referred_to_are_from_CRI

Answer: 10

(Warning: the response is based on possibly incomplete information).
 

user>> And 1989?
\end{verbatim}
{\em Throughout most of this transcript we have chosen the intended
interpretations of sentences; here, we diverge for a moment from
this principle and show some of the other less
plausible resolutions that this elliptical question can be given.
One reason for the large number
of resolutions (more are proposed than shown here),
is that `1989' can be resolved to
a year, a project number or just a number.  Generation makes the
type of resolution explicit.}
\begin{verbatim}
system>>   "how many CLAM-BAKE (the project) payments were made 
            in 1989 (the year)?"

Confirm this interpretation? (y/n): n

system>>   "how many 1989 (the year) payments were made 
            in 1991 (the year)?"

Confirm this interpretation? (y/n): n

system>>   "how many CLAM-BAKE (the project) payments were made 
            in account 1989?"

Confirm this interpretation? (y/n): n

user>> Can you show me the employees who worked on CLAM-BAKE.
\end{verbatim}
{\em A meta-question: In order to establish that the system can show
the employees, it must determine that it can translate the embedded
command (`show me the employees...') into an SQL query.  If it can
translate, then it can show the employees (whether or not there are any
to be shown is another matter).  In a spirit of co-operation, as well as
giving a literal answer to the question, it also sends of the query
constructed in establishing the answer.}
\begin{verbatim}
system>>   "can you show me all employees that worked on 
            CLAM-BAKE (the project)?"

Assumptions used in query translation:

  people_who_work_on_projects_do_so_from_their_start_to_their_finish
  all_employees_referred_to_are_at_CRI

Answer: Yes

    for instance
        Zorba Greek
        Sammy Davis Junior
        Peter Piper
        Maid Marion
        Gordon Bennett
        Geoff Steiner
        Darth Vader


user>> Which of them have cars.
\end{verbatim}
{\em Illustrates paraphrasing of more complex anaphora resolution and of bare plurals:
`Them' is resolved to refer to the employees mentioned in the previous question.
In this case, it would have been possible generate a paraphrase ``Which of the employees
that ...', but this is not always syntactically possible and so for robustness
a parenthetical appositive is generated instead.

Similarly, the paraphrase `have some cars' in place of the bare plural `cars' may
seem clumsy.  Bare plurals like `cars' may resolve either to `all cars' or `some
cars', and in this case we clearly do not wish to ask which employees have all cars.
But for a query like `Show current projects', one would presumably wish all current
projects to be shown.  

Note also that in translation an assumption is made that the query is about company
cars.  Since the database only knows about company cars, it could not translate the
query without making this assumption.  A warning is issued to the effect that this assumption 
may differ from that made by the user.}
\begin{verbatim}
system>>   "which of them (the employees that worked on 
            CLAM-BAKE (the project)) have some cars?"

Assumptions used in query translation:

  all_cars_referred_to_are_company_cars

Answer:
        Peter Piper 
        Zorba Greek 

Warning: a possibly incorrect assumption was made.

user>> Do you know what the largest payment on CLAM-BAKE was?
\end{verbatim}
{\em Another meta-question, and also illustrates superlatives
like `the first', `the
largest', etc.}
\begin{verbatim}
Answer: Yes

Do you want me to tell you? (y/n): y

Answer:
        [100661, 31/10/1990, Marble Systems Ltd, pounds-3395]

Key for answer fields:    
        [A,B,C,D]
   A: number, transaction number
   B: date, time
   C: payee
   D: amount, money

user>> How much was it for?

system>>   "how much money was transaction 100661 for?"

  money_is_measured_in_pounds_sterling

Answer: 3395

user>> How many cheques over #1000 did Marble Systems receive?
Answer: 1

user>> Over #200?

system>>   "how many cheques over 200 pounds sterling did 
            Marble Systems Ltd (the payee) receive?"
Answer: 6

user>> Show them.
\end{verbatim}
{\em CLARE does not always get it right first time.  This example shows how paraphrasing
is necessary for interactive disambiguation.  

The desired resolution
would be more effectively paraphrased as `Show them (the cheques over 200 that Marble
Systems Ltd received)'.  This paraphrase was in fact attempted from the resolved 
QLF, but unfortunately failed to produce an English string in this case.  Therefore,
a simpler paraphrase was attempted.}
\begin{verbatim}
Initial Resolution. 
system>>   "show Marble Systems Ltd (the payee)!"
Confirm this interpretation? (y/n): n

system>>   "show them (the cheques over 200 pounds sterling)!"
Answer:
        [100222, 6/10/1989,  Marble Systems Ltd, pounds-270]
        [100280, 1/12/1989,  Marble Systems Ltd, pounds-360]
        [100338, 9/1/1990,   Marble Systems Ltd, pounds-270]
        [100465, 14/5/1990,  Marble Systems Ltd, pounds-270]
        [100651, 11/10/1990, Marble Systems Ltd, pounds-270]
        [100661, 31/10/1990, Marble Systems Ltd, pounds-3395]

Key for answer fields:    
        [A,B,C,D]
   A: number, transaction number
   B: date, time
   C: payee
   D: amount, money

(Warning: the response is based on possibly incomplete information).

user>> How much have we paid Marble Systems?
\end{verbatim}
{\em Aggregation: the query is interpreted as as requesting the total amount paid.}
\begin{verbatim}
system>>   "how much money has Cornerstone Research Institute 
            paid Marble Systems Ltd (the payee)?"

Answer: 4295

user>> Tell me about CASE Workbench.
\end{verbatim}
{\em Generation from the database.  English sentences are generated direct from ORACLE
database records.  Generation does not proceed by slotting names into predefined
templates, but emerges naturally from the more general definition of the domain.
Checks are made to ensure that (a) all possible information from each record is
printed out, and (b) that no sentence merely repeats information that has already been
given.}
\begin{verbatim}
------------------------------------------------
Trying to describe a database record:
system>>   "CASE Workbench (the project)'s end date is 
          31/1/1992 (the day)."
system>>   "CASE Workbench (the project)'s number is 8743."
system>>   "CASE Workbench (the project)'s start date is 
          1/2/1990 (the day)."
(The record was fully described)

------------------------------------------------
Trying to describe a database record:
system>>   "Geoff Steiner (the employee) is a person."
system>>   "CASE Workbench (the project) has Geoff Steiner (the employee)."
(The record was fully described)

user>> FOO is a project.
\end{verbatim}
{\em Making assertions:  As well as querying the database, one can also add new
information.  Moreover, this information does not have to be added in such a way
that database records are completed in one fell swoop.  As the assertion below
shows, three of the for fields in the \verb!CRI_PROJECT! (\verb!C,D,E!) have not been
instantiated yet.}
\begin{verbatim}
Assertion:

CRI_PROJECT(FOO,_,_,_),

user>> .ask
(try to complete partial database records by asking questions [clare])
\end{verbatim}
{\em CLARE includes a facility for asking the questions needed to complete database
records.  As with generation from database records, no special templates need to
be predefined.  The form of the question is determined automatically from the general
domain definition.

Answers to the questions are treated as ordinary English sentences, which may be
elliptical.  These go through the normal process of analysis and translation
without being treated in any special way.

CLARE keeps asking questions until it has completed all the incomplete database records,
whereupon it returns to the normal CLARE loop, allowing the user to ask the questions
instead of the system.}
\begin{verbatim}
system>> what is FOO (the project)'s account?

user>> 1234.

system>>    "account 1234 is FOO (the project)'s account."

Confirm this interpretation? (y/n): y
Assertion:

  CRI_PROJECT(FOO,1234,_,_)

system>> what is FOO (the project)'s start date?

user>> It started on 1/1/91.

system>>   "it (the start date of FOO (the project)) 
            started on 1/1/1991 (the day)."

\end{verbatim}
{\em Answer is a full sentence rather than an elliptical noun phrase.  Unfortunately
the first resolution of the pronoun `it' is not the one required, and the second
reading is accepted.}
\begin{verbatim}
Confirm this interpretation? (y/n): n

system>>   "FOO (the project) started on 1/1/1991 (the day)."

Confirm this interpretation? (y/n): y

Assertion:
  CRI_PROJECT(FOO,1234,1-JAN-91,_)

system>> what is FOO (the project)'s end date?

user>> Tomorrow.

system>>   "27/11/1992 (the day) is FOO (the project)'s end date."

Assertion:
CRI_PROJECT(FOO,1234,1-JAN-91,27-NOV-92)
\end{verbatim}
{\em New record is now complete.}

\chapter{Interfacing by Abductive Equivalential Translation}
\label{translation}

\section{Introduction}
\label{AETIntro}

In this and the next chapter, we will describe the modules in CLARE
that allow it to be used as a natural-language interface to a knowledge
base, and in particular to a relational database. Before going any
further, we will establish more precisely what we mean by ``natural-language
interfacing''. Normally, the problem is
thought of as that of producing answers to natural-language questions posed 
to a knowledge base (cf. Perrault and Grosz, 1988). 
Here, we will adopt a more general view;
we will regard it as the problem of formalizing the relationship 
between 1) the contents of a database, and 2) the way in which people use
language to talk about it. Asking questions about the contents of
databases is far from being the only way in which language and database
can be related; thus apart from question-answering,
we wish the model to support reasoning that will allow at least the
following.
\begin{itemize}
\item Responding to natural-language commands.
\item Accepting statements describing new records for the database.
\item Reasoning about whether the database contains the information
necessary to answer an NL question.
\item Indicating whether answers to WH-questions are complete
or not, and distinguishing between ``No'' and ``Don't know'' answers to
Y-N questions.
\item Describing existing records in the database using NL.
\item When records are incomplete, formulating NL questions to 
elucidate the information needed to complete them.
\end{itemize}
We want the information needed to support these functionalities to be expressed
declaratively, using some sort of predicate logic; the result will be
a {\it linguistic domain theory} for the database, a theory of 
how language and database are related. Languages are normally based on
vague, common-sense concepts, databases on precise, formal ones. 
Relating these two disparate systems raises the usual problem
of common-sense reasoning: interpretation of vague NL concepts
in terms of precise database concepts is often only possible if
additional assumptions, not implied by the explicit linguistic content, are made.
These assumptions are in general defeasible. In some cases, it may
be desirable to relay them back to human users. If the scheme is to be 
practically useful, we also want to be able to specify a methodology for 
constructing 
linguistic domain theories. For the usual software engineering reasons, they
should also, as far as possible, be modular and re-usable. These, in brief,
are our top-level goals. We will now make them more specific, and relate
them to the concrete architecture of the Core Language Engine.

The CLE, being a general natural language system (rather than one
tailored to a specific application), constructs representations of utterances
that essentially mirror their linguistic content. 
Normally, every content word in the utterance will correspond to an occurrence
of a word-sense predicate in that utterance's TRL representation. Thus for example
in the CLARE Project Resource Management domain the sentence
\begin{description}
\item[(S1)] List all payments made to BT during 1990.
\end{description}
gets a TRL representation which contains predicates corresponding directly to
{\it list}, {\it payment}, {\it make} and {\it during}. In order to carry out the
command, however, CLARE needs to construct an SQL 
{\tt SELECT} statement which searches for
{\tt TRANS} tuples where the {\tt payee} field is filled by
{\tt bt}, and the {\tt cheque\_\-date} field by a date constrained
to be between 
1st January and 31st December, 1990. The differing nature
of the two representations can lead to several possible kinds of
difficulties, depending on how the ``linguistic'' and ``database''
representations are connected. There are three in particular
that we will devote most of our attention to in what follows:

\begin{enumerate}
\item A query can be {\it conceptually} outside the database's domain.
For example, if ``payments'' in (S1) is replaced by ``phone-calls'',
the interface should be able to indicate to the user that it is unable
to relate the query to the information contained in the database.

\item A query can be {\it contingently} outside the database's domain.
Thus if ``1990'' is replaced by ``1985'', it may be possible
to derive a query; however, if the database only
contains records going back to 1989, the result will be an empty list.
Presenting this to the user without explanation is seriously misleading.

\item A query may need additional implicit assumptions to be translatable
into database form. Asking (S1) in the context of our example Project Resource
Management domain, it is implicitly understood that all payments referred
to have been made by SRI. If the user receives no feedback 
describing the assumptions that have been made to perform the translation,
it is again possible for misunderstandings to arise.
\end{enumerate}

With the problems just mentioned born in mind, we will consider 
previous methodologies used to attack the NL interfacing problem.
The currently most popular approach is perhaps
to attempt to effect the connection between LF and database query
by encoding the database as a set of unit clauses, and then building
an interpreter for the logical forms which captures the relations
between linguistic and database predicates as ``rules'' or ``meaning
postulates'' written in Horn-clause form (cf. e.g. McCord 1987,
Pereira and Shieber 85).
Anyone who has experimented with this scheme 
will, however, know that it tends to suffer from all three of the
types of problem listed above. This is hardly surprising, when one
considers that Horn-clauses are ``if'' rules; they give conditions
for the LF's being true, but (as pointed out in Konolige 1981), they
lack the ``only if'' half that says when they are false. It is 
of course possible to invoke the Closed World Assumption (CWA); in this 
interpretation, finite failure is regarded as equivalent to
negation. Unfortunately, experience also shows that it is extremely difficult
to write meaning postulates for non-trivial domains that are valid
under this strict interpretation.

For these reasons, Scha (1983)
argues that approaches which express the connection
between LF and database query in terms of first-order logic formulas
are unpromising. Instead, previous approaches to query derivation which 
attempt to justify equivalence between queries and semantic representations
have been limited (at least in implemented systems) to employing
restricted forms of inference. Examples are the type inference used
in PHLIQA (Bronnenberg et al 1980) and Stallard's `recursive terminological
simplification' (Stallard 1986).

In what follows, we will show how a more general deductive
approach can be taken.  This depends on coding the relationship between LF and
database forms not as Horn-clauses but as ``definitional equivalences'', 
conditional if-and-only-if rules of a particular form.  
Our approach  retains computational tractability
by limiting the way in which the equivalences can take part in
deductions, roughly speaking by only using them 
to perform directed ``translation'' of predicates. However we still permit
nontrivial goal-directed domain reasoning in justifying query derivation,
allowing, for example, the translation of an LF conjunct to be influenced
by any other LF conjuncts, in contrast to the basically local translation
in PHLIQA. 
This approach deals with the
first two points above without recourse to the CWA and simultaneously allows a
clean integration of the ``abductive'' reasoning needed to take care of point
3. As we shall see, it also makes it possible to use substantially the same
framework to achieve interfacing functionalities other than question-answering.
The main technical problems to be solved
are caused by the fact that the left-hand sides of the equivalences are
generally not atomic.

The rest of this chapter is structured as follows. In the next section, we 
give an overview of the interface system, concentrating on the denotational and
functional aspects. We describe roughly what information will be found in
a linguistic domain theory (LDT), and how the interface functionalities listed at 
the beginning of the section can be formalized as tasks of the form `` given
a formula $P$, find a formula $P^\prime$ of a specified type, such that 
$P$ and $P^\prime$ are equivalent given the LDT and some permitted assumptions.''
We will refer to tasks of this kind as ``performing abductive equivalential
translation (AET) on $P$''. In section~\ref{AET}, we describe AET as a 
computational process. Section~\ref{Simplification} describes the
{\it simplifier}, a module that plays an important supporting role
in the translation process; section~\ref{Closed-World} considers the
connection between AET and the Closed World Assumption.
In the following chapter, we describe in detail the
construction of LDTs for relational database interfacing.

\section{Ontological issues}\label{Ontology}

We start by defining the basic ontology underlying the linguistic
domain theory; the 
ideas are closely related to those proposed in (Konolige 1981).
We assume as usual that there is a set of objects, $\bf O$, and a 
set of relations obtaining between them, $\bf R$. $\bf O$ will contain
all the things that the predicates
in logical forms range over --- the things, like 
suppliers, parts, projects, payments, 
deliverables, etc. that natural language refers to.
This includes events. Similarly, $\bf R$ will contain all the relations 
obtaining between elements of $\bf O$ that we will wish to refer to.

There will be no particular need in what follows to be much more
specific about exactly what can and cannot belong to $\bf O$ and $\bf R$,
though this is obviously not a trivial matter. We will however
be interested in picking out certain distinguished subsets of
these two sets which have direct relevance to database interfacing. We take
the position that databases are also objects in the world,
consisting of finite collections of rows of marks; this is not
a common viewpoint at the moment, but we think it corresponds well
to the naive view of ``what a database is''. We consequently
assume the existence of a subset $\bf O_{db}$ of $\bf O$ consisting of 
{\it database objects}, which will be 
numbers, strings, date representations, and other things that can
be found filling fields in database records. 
Similarly, we will assume the existence of a subset $\bf R_{db}$ of $\bf R$,
which will be the {\it database relations}.
Database relations are defined by
the database's internal structure in terms of tuples: a database
relation $D$ holds of a list of arguments $Arg_i$ iff there is a tuple
of type $D$ in the database whose fields are filled by the $Arg_i$. 
Since little can be done with database objects without recourse to
elementary arithmetic and an ability to display results, we assume
two more distinguished subsets of $\bf R$. $\bf Arith$ will be the relevant arithmetic relations 
(like ``being numerically less than'') that can obtain between members
of $O_{db}$. $\bf Exec$ will be a primitive set of relations
which the interface can cause to hold by performing actions. 
A typical relation in $\bf Exec$ would be the one that holds between
a database-object, and a time and place at which the interface displayed it.

In what follows,
the distinction between database objects, the non-database objects
they name, and the terms in the linguistic domain theory that refer to
both of them will be central. Thus for example we distinguish
between 
\begin{itemize}
\item a transaction (a non-database object, an event in the exterior world)
\item the term referring to the transaction in the linguistic domain theory
\item the database object (a number) that is the transaction's ID
\item the term in the theory referring to the database object. 
\end{itemize}
Although these distinctions may not immediately seem necessary, they
are in fact motivated by typical properties of real databases.
Database objects are often codes or non-unique names, that cannot simply
be treated as names in a first-order theory; for example, the same number,
occurring in different fields in a relation, can be related to distinct
real-world entities, or (if used as a code value), to a property of
an entity. 

The linguistic domain theory relates database objects
with real-world objects, allowing us to draw conclusions about the properties 
of the real-world objects by examining the database records.
Less obviously, it can also allow us
to draw conclusions about properties of real-world objects 
from the {\it lack} of records of a particular type
in the database. This amounts to a principled generalization of the 
``closed world assumption'', and is described further in 
section~\ref{Closed-World}. The linguistic domain theory can also 
partially model the interaction between system and user, by
defining a relation between predicates holding
in the ``utterance situation'' (the real-world situation in which 
the user is interacting
with the interface), and the executable relations. For example
there are real-world predicates which correspond to verbs like 
``show'' and ``list'', which are commonly used in commands
(e.g. ``Show me all payments over \pounds 500''); these are related 
by the linguistic domain theory to the primitive executable relation of 
displaying characters on the screen. Treating the utterance situation
uniformly as part of the domain adds both conceptual elegance and a
real increase in the interface's scope; so it is for instance possible
to deal with queries like ``Which of the payments that you showed me
in answer 12 were made this year?'' in a principled way.

\section{Formalizing NL interfacing functionalities}

\subsection{Preliminaries}

At a pre-theoretical level, our characterization of the NL interfacing
functionalities is actually a very simple one: it is only the technical
details that will prove complex. The key notion
is that of ``translation''. Given an interfacing task, described in one
vocabulary, we wish to {\it translate} it into an equivalent task, described in
another vocabulary. Thus for example the ``question-answering'' task
is of the form ``produce an answer to the question whose logical form is $P$'';
we wish to translate this into an equivalent task of the form ``execute
the database query $P^\prime$, and print the results in a readable way''.
Here, the original task is expressed using predicates corresponding to
word-senses; the problem is to translate it into a task expressed using 
predicates corresponding to database relations, arithmetical operations 
and primitive executable relations. What we want to do now is
to express these ideas in formal terms, so that we can then realize them
as an inference problem. In particular, we want to introduce linguistic
domain theories so that translation can be regarded as logical equivalence
with respect to a theory.
We will start by sketching the appearance of a linguistic domain theory,
and defining the formal concept of {\it effective translation}.
Throughout most of the chapter, we will regard it as sufficient to
allow the target representation (the result of performing the translation)
to be a predicate calculus expression 
that treats database relations as predicates;
in section~\ref{SQL-convert}, we briefly review the module that 
performs the final conversion into executable SQL queries.

\subsection{Basic form of linguistic domain theories}\label{Basic-LDT}

Since we are primarily interested in establishing equivalences (rather than
implications), it makes sense to write the LDT as far as possible using
formulas which themselves are equivalences. For various reasons, it turns out that it is
convenient to allow these equivalences to be conditional; to limit
computational complexity, we will only allow their left- and right-hand sides
to be existentially quantified conjunctions of atomic formulas. Thus
in summary, the greater part of an LDT will be composed of formulas 
of the general appearance
$$\forall\vec{x}.(Left\equiv Right) \leftarrow Conds$$
where $Left$, $Right$ and $Conds$ are existentially quantified conjunctions.
Linguistic domain theories
typically also contain Horn-clause axioms and declarations of 
functional relationships between arguments of predicates (these
are described in detail in section~\ref{Functional-relations}). 
It is permitted during inference
to assume goals abductively, at a cost depending both on the goal
and the context in which the assumption is made. We do not attempt
to define a formal semantics for the notion of abductively justified
inference: we merely assume that costs are assigned according to some
scheme that generally makes low-cost proofs intuitively more appealing than
high-cost ones. Since abductive assumptions are in any event made explicit
to the user of the system, we view the costs essentially as
heuristics to control the order in which the space of possible proofs
is searched. The range of goals that may be abductively assumed is controlled by 
declarations of the form 
\begin{verbatim}
assumable(Goal,Cost,Justification,Type,Cond)
\end{verbatim}
where {\tt Goal} and {\tt Cond} are atomic formulas, 
{\tt Cost} is a non-negative integer,
{\tt Justi\-fic\-ation} is a unique tag that names the assumption-type,
and {\tt Type} is the assumption-type (this is defined below).
The intended semantics are that {\tt Goal} may be assumed at 
cost {\tt Cost} in an environment where
{\tt Cond} holds. The ``justification'' 
can be used to identify the assumption to the user. 
The inference engine (which is basically a backwards-chaining
Horn-clause interpreter) contains no general treatment of negation, but 
in order to deal with cases where {\tt Goal} is explicitly 
contradicted by its context it is possible to define rules of the form
\begin{verbatim}
neg(Goal) <- Body
\end{verbatim}
where {\tt Goal} is an atomic formula and {\tt Body} is an arbitrary formula.
A rule of this type has the content ``the negation of {\tt Goal}
is implied by {\tt Body}''. It is only invoked to attempt to identify inconsistent
uses of assumptions. The framework thus allows in effect a limited use
of normal defaults.

Experimentation with CLARE seems to indicate that one can profitably
divide acceptable assumptions into at least three distinct types;
we illustrate using the example PRM domain, which covers a project and 
payment database for
SRI Cambridge. In the PRM domain, it is reasonable to assume 
(lacking evidence to the contrary) that ``project'' is equivalent
with ``project at SRI Cambridge''. The content of assumptions of 
this kind is that the speaker means something more specific than
what was actually said. We consequently refer to them as ``specializations''.
In contrast, it is also reasonable to assume that ``payment'' means 
``SRI payment made during the period covered by database records''. 
In this case, however, it seems intuitively less clear that the speaker
intended to use the word in the more restricted sense, and it is more
appropriate to assume that the database is limited in a way which the 
speaker may not be fully aware of. We will call assumptions of this kind
``limitations''.
Finally, it may sometimes be appropriate to make assumptions that can actually
be incorrect: for example, we assume that completion dates for future
deliverables have been correctly estimated. We call these assumptions ``approximations''.
Distinguishing between different kinds of assumption will become important
later on, when we consider issues regarding the completeness of answers
to questions.

\subsubsection{Effective translation}\label{Effective-translation}

We now introduce the key formal definition.
We write $\Gamma$ to symbolize a linguistic domain theory, and
let $F_{source}$ be an arbitrary formula; then we define
$F_{target}$ to be an {\it effective translation} of $F_{source}$
iff there exists a set of abductive assumptions $A$ such that
\begin{eqnarray}
\Gamma\cup A \Rightarrow (F_{source} \equiv F_{target})\label{Basic-equiv}
\end{eqnarray}
and
\begin{itemize}
\item Each assumption in $A$ is acceptable in the context in which it is made.
\item There is a finite proof procedure for determining the truth of $F_{target}$.
\end{itemize}
The question of whether or not a finite proof procedure exists for a formula,
given a theory, is of course undecidable in general; CLARE only attempts to
solve a simple subcase of this problem.

The first criterion that must be met is that all predicates in
$F_{target}$ must be {\it potentially finite}: that is to say, they should have the 
property that for at least some combinations of instantiation of their arguments
there are only finitely many instantiations of the remaining arguments that
make them true, and that these instantiations can be found within bounded time.
Thus the minimum requirement is that if all the arguments are ground it is
possible to determine the truth of the relation within bounded time.
There are in practice three kinds of potentially finite predicates in a domain
model: database predicates, arithmetic relation predicates, and primitive 
executable predicates. We examine each of these in turn:
\begin{description}
\item[Database predicates] These are predicates directly corresponding to the database
relations $\bf R_{db}$; a database predicate holds of its arguments iff the database relation
has a row with those arguments. The finiteness of database predicates follows from
the finiteness of the corresponding database relations. Database predicates are
consequently always finite, irrespective of their instantiation.
\item[Arithmetic relation predicates] These correspond the arithmetic relations 
in $\bf Arith$, such
as addition, subtraction, and inequality. In general, arithmetic relation predicates
are only finite if all or all but one of their arguments are instantiated. For example,
if $X$ is instantiated to a given number, there is an infinite set of values for $Y$
such that $X < Y$ holds. If both $X$ and $Y$ are fixed, however, the truth or falsity
of $X < Y$ can in practice be determined in bounded time.
\item[Primitive executable predicates] In order to be able to reason about CLARE's
ability to perform actions like displaying of objects, there is a set of predicates
which correspond to the primitive executable relations $\bf Exec$.
We assume that these predicates are finite for {\it instantiated} actions.
\end{description}

Since the finiteness of some predicates depends on their instantiation, the existence of
a finite proof procedure generally depends on being able to find a evaluation order
which ensures that conditionally finite predicates are sufficiently instantiated by
the time they are queried; CLARE can search for suitable evaluation orders by permuting
the order of evaluation of conjunctions. For example, if $TRANS/3$ is a
database predicate then the strategy ``find an $X$ 
such that $X > 100$, then find values of
$Y$ such that $TRANS(john,X,Y)$'' is not a finite strategy; however, reversing
the order to make the strategy ``find $X$ and $Y$ such that $TRANS(john,X,Y)$, then
determine whether $X > 100$ holds'' is finite. The search is carried out using 
a simple abstract interpretation method, which uses meta-information 
about finiteness of 
predicates with respect to different instantiation patterns to 
mimic the behaviour of the real execution module. This part of the system
is described further in section~\ref{Qopt}.

\subsection{Yes/No questions}

We are now in a position to describe formally the various interfacing
functionalities. The simplest one to begin with is that of answering a Y-N question.
The formal statement of the problem is as follows. 
We are given a formula,
$F_{ling}$, which is the logical form of a Y-N question to  be answered.
We wish to find a formula
$F_{eval}$ and a set of permissible assumptions $A$, such that 
$F_{eval}$ is an effective translation of $F_{ling}$ modulo $A$
in the sense defined in section~\ref{Effective-translation}.
There are several possibilities.
\begin{itemize}
\item No such $F_{eval}$ can be found: the answer is {\bf Don't know}.
\item $A$ is empty and $F_{eval}$
can be proved: the answer is {\bf Yes}.
\item $A$ is empty and $F_{eval}$
cannot be proved: the answer is {\bf No}.
\item $A$ is non-empty and $F_{eval}$
can be proved: the answer is {\bf Yes, conditional on the validity of the assumptions}.
\item $A$ is non-empty and $F_{eval}$
cannot be proved: the answer is {\bf No, conditional on the validity of the assumptions}.
\end{itemize}
If assumptions have been made, the user is told what they were.
An example of each case, taken from the PRM domain, follows.
\begin{itemize}
\item The question is {\it Does Peter have a dog?}. There is no effective translation
of the logical representation of this question, so the answer is {\bf Don't know}.
\item The question is {\it Does Peter have a car?}. Peter is an employee,
and there is an effective translation to a query that accesses the {\tt SRI\_EMPLOYEE}
relation. The translated query can be proved, so the 
answer is {\bf Yes}. 
\item The question is {\it Does Gordon have a car?}. This is the same as the previous
case, except that Gordon does not have a car according to the database, and the
effective translation cannot thus be proved. The answer is {\bf No}.
\item The question is {\it Has Peter booked less than 200 hours to CLARE?} 
Peter is an employee, and CLARE is a project, so there is an effective translation
to a query that accesses the {\tt BOOKING\_TO\_PROJECT} relation under the
assumption that the time referred to was booked during the period for which 
records were kept. The database reveals that only 165 hours were booked during this
period, so the answer is {\bf Yes, conditional on the validity of the assumptions}.
\item The question is {\it Has Peter booked less than 100 hours to CLARE?} 
The assumption used when translating is the same as in the last example, and
the answer is {\bf No, conditional on the validity of the assumptions}.
\end{itemize}

\subsection{Commands and WH-questions}\label{Command-functionality}

The functionality of responding to a natural language command
can be captured by a fairly simple extension of the previous definition.
We will assume that the logical form representation of a command is
a formula which is true if the command will be carried out
at some future time. Once again, we let
$F_{ling}$ be the logical form of the command, and the task is to
find an effective translation $F_{eval}$ modulo a set of assumptions $A$.
$F_{eval}$ will usually be a formula that contains primitive evaluable
predicates. 

Answering WH-questions can be treated as a special case of responding
to commands, if we assume that the logical form of a WH-question is
of the type $\lambda x.Q(x)$,
where $Q(x)$ holds iff $x$ is an object satisfying the question.
(cf. Rayner and Janson 1987).
Thus for example the logical form for {\it Which payments in June were
over \pounds 1000?} will be roughly
\begin{verbatim}
X^and(payment1(X),
      and(in1(X,june1),
          and(over1(X,1000))))
\end{verbatim}
The asking of a WH-question can be regarded as a command to display
all the objects that would be answers to it. So if $\lambda x.Q_{ling}(x)$
is the logical form of the question, $displayed(x,t)$ is a predicate
that holds if $x$ is displayed at time $t$, and $in\_future(t)$ is a predicate
that holds of future times, then the question can be regarded as equivalent
with a command whose logical form is
$$\forall x.Q(x) \rightarrow \exists t.displayed(x,t)\wedge in\_future(t)$$
The different cases that can arise in responding to a command or WH-question
are fairly similar to those for responding to Y-N questions. We summarize,
this time without examples:
\begin{itemize}
\item No such $F_{eval}$ can be found: the answer is {\bf Don't know how to respond}.
\item $A$ is empty, or only contains ``specialization'' assumptions, and $F_{eval}$
can be proved: the response is to perform all the necessary actions 
(e.g. displaying objects) and 
inform the user that the response is complete.
\item $A$ is empty, or only contains ``specialization'' assumptions, and $F_{eval}$
cannot be proved: the response is to inform the user that it is impossible to
carry out the command. (This cannot occur with questions).
\item $A$ contains ``limitation'' or ``approximation'' assumptions, and $F_{eval}$
can be proved: the response is to perform all the necessary actions and inform
the user that the completeness and/or accuracy of the answer depends on the
assumptions.
\item $A$ contains ``limitation'' or ``approximation'' assumptions, and $F_{eval}$
cannot be proved: the response is to inform the user that it is impossible to
carry out the command if the assumptions are correct.
\end{itemize}

\subsection{Meta-questions}

Both Y-N and WH questions can occur embedded in meta-questions of the
form ``Do you know {\it Q}?''. (E.g. ``Do you know how many people 
work on CLARE?'', ``Do you know whether there were any payments on 1/11/91?'').
If $F_{ling}$ is the logical form of Q, 
it follows from the preceding discussion that 
a sufficient criterion for an affirmative answer
is that there is an effective translation $F_{eval}$ modulo a set of assumptions $A$,
where $A$ contains only
``specialization'' assumptions. There are also several ways in which
inability to meet this criterion can be meaningful to a user.
If translation can only be carried out by assuming
a ``limitation'' or ``approximation'' assumption, or by making an
assumption whose negation is implied by its context, then the
offending assumption normally constitutes a good explanation of 
``why the system doesn't know''. 

Thus for example, if database records only extend back to 17/8/88
it may only be possible to translate the predicate corresponding
to the word ``payment'' into an expression involving the transaction
relation if it is assumed that the payment was made after 17/8/88. 
In this case the
meta-question ``Do you know whether there were any payments on 5/5/86?''
will receive the answer {\bf No, because this violates the assumption that
all payments referred to were made after 17/8/88}.

\subsection{Generating statements and questions}

Finally, we consider the functionalities of generating statements and questions.
If $S_{db}$ is a database formula
that we wish to realize in language, the problem is to find a set
of NL statements $S_i$ ($i = 1...n$) with logical forms $L_i$,
and an acceptable set of assumptions $A$ such that
$$\Gamma\cup A \Rightarrow (\bigwedge L_i \equiv S_{db})$$
This is fairly obvious; more interestingly, a very similar treatment
works for questions. For example, if $Q_{db}$ is a formula composed
of database predicates, and containing a free variable $x$, then 
$\lambda x.Q_{db}$ can reasonably be thought of as an abstract 
representation of the WH-question ``What $x$ satisfies $Q_{db}$?''.
Assuming as before that the logical form for a WH-question
is a lambda-abstraction, the problem of
finding a linguistic realization of $\lambda x.Q_{db}$ can be 
formalized as that of finding a single
natural-language question $Q$ with logical form $\lambda x. Q_{ling}(x)$,
and an acceptable set of assumptions $A$ such that
$$\Gamma\cup A \Rightarrow \forall x.(Q_{ling}(x) \equiv Q_{db}(x))$$
The issues involved are discussed at greater length in 
chapter~\ref{genass}.

\section{Abductive Equivalential Translation}
\label{AET}

\subsection{Translation schemas}\label{Translation-schemas}

We will now consider the actual computational mechanisms used to 
effect the task of carrying out abductive translation. Recall that
the main body of the declarative knowledge used is coded as a set of 
conditional equivalences, formulas of the type\footnote{Quantification 
over the $\vec{x}$ on the left-hand side will 
often in practice be vacuous. In this and other formulas, we assume implicit
universal quantification over free variables.}
\begin{equation}
Conds \rightarrow (\exists \vec{x}.P_1\wedge P_2\wedge P_3 ...) \equiv P^\prime \label{MP}
\end{equation}
In this and the next chapter,
these rules are written in a notation illustrated by the following example,
\begin{verbatim}
exists([Event],
  and(work_on1(Event,Person,Project),
      project1(Project))) <->
DB_PROJECT_MEMBER(Project,Person)
\end{verbatim}
in which \verb!work_on1! and \verb!project1! are linguistic predicates
and \verb!DB_PROJECT_MEMBER! is a database relation (we will adhere
to the convention of capitalizing names of database relations).

The attractive aspect of this type of equivalence stems from the
fact that it can be given a sensible interpretation 
in terms of the procedural notion of ``translation''.
We will neglect for the moment the existential quantification, i.e.
we will assume that~(\ref{MP}) has the simplified form
\begin{equation}
Conds \rightarrow (P_1\wedge P_2\wedge P_3 ...) \equiv P^\prime \label{Equiv-univ-LHS}
\end{equation}
(We will return to considering the general form~(\ref{MP}) in section~\ref{Existential}).
The intuitive idea is that is that~(\ref{Equiv-univ-LHS}) can be read as 
``$P_1$ can be expanded to $P^\prime$ if it occurs in an environment
where $P_2\wedge P_3 ...\wedge Conds$ can be inferred''. The ``environment'' is
provided by the conjuncts occurring together with
$P_1$ in the original logical form, other meaning postulates,
and the contents of the database. This provides a framework in which
arbitrary domain inference can play a direct role in justifying the 
validity of the translation of an LF into a particular database query.

We hasten to give our claim some formal backing; our strategy will be to
present it first in a fairly specific form, then in a more general one.
Suppose then that $P$ is a ground goal that unifies with $P_1$ with most general unifier
$\theta$, i.e. $\theta(P_1) = P$. Suppose also that $P$ occurs in a context
where it is conjoined with another formula $R$, such that $R$ implies 
$\theta(P_2\wedge P_3 ...\wedge Conds)$. Then we claim that
\begin{equation}
R \wedge P \equiv R \wedge \theta(P^\prime) \label{Eq-schema-0}
\end{equation}
The proof is simple. In the $\Leftarrow$ direction, we have that
$R$ implies $\theta(Conds)$ by hypothesis, and that $\theta(P^\prime \wedge Conds)$
implies $\theta(P_1), which by hypothesis is equal to P$, 
from~\ref{Equiv-univ-LHS}. Hence $R \wedge \theta(P^\prime)$
implies $P$ and thus $R \wedge P$.

In the $\Rightarrow$ direction we have
by hypothesis that $P = \theta(P_1)$ and $R$ implies $\theta(P_2\wedge P_3 ...\wedge 
Conds)$. Hence $R \wedge P$ implies $\theta(P_1 \wedge P_2\wedge P_3 ...\wedge Conds)$
which by~(\ref{Equiv-univ-LHS}) implies $\theta(P^\prime)$. Hence $R \wedge P$ implies
$R \wedge \theta(P^\prime)$. We can summarize~(\ref{Eq-schema-0}) as the
inference rule
\begin{eqnarray}
& & (Conds \rightarrow (P_1\wedge P_2\wedge P_3 ...) \equiv P^\prime) \wedge \nonumber \\
& & R \rightarrow \theta(P_2\wedge P_3 ... \wedge Conds) \nonumber \\
& \Rightarrow & R \wedge \theta(P_1) \equiv R \wedge \theta(P^\prime) \label{Eq-schema-1}
\end{eqnarray}

The next step is to generalize~(\ref{Eq-schema-1}) by making explicit the concept
of the ``conjunctive context''. We do this by splitting it into two inference 
rules,~(\ref{Eq-schema-base}) and~(\ref{Eq-schema-conj}), as follows:
\begin{eqnarray}
& & (Conds \rightarrow (P_1\wedge P_2\wedge P_3 ...) \equiv P^\prime) \wedge \nonumber \\
& & Context \rightarrow \theta(P_2\wedge P_3 ... \wedge Conds) \nonumber \\
& \Rightarrow & Context \rightarrow (\theta(P_1) \equiv \theta(P^\prime)) \label{Eq-schema-base} \\
& & \nonumber \\
& & {Context \wedge Q \rightarrow (P \equiv P^\prime)} \nonumber \\
& \Rightarrow & Context \rightarrow (P \wedge Q \equiv P^\prime \wedge Q) \label{Eq-schema-conj}
\end{eqnarray}
The proofs of (\ref{Eq-schema-base}) and~(\ref{Eq-schema-conj}) follow from that
of~(\ref{Eq-schema-1}). The two rules can be used recursively together 
to translate formulas composed using an arbitrary number of occurrences of
the conjunction operator. In both rules, the formulas
before the $\Rightarrow$ are the premises, and the formula after the conclusion.
(\ref{Eq-schema-base}) is the base case: it gives sufficient
conditions for using~(\ref{Equiv-univ-LHS}) to translate $P_1$ to $P^\prime$.  The other 
formula, (\ref{Eq-schema-conj}), is the recursive case;
it expresses translation of a conjunction in terms of translation
of one of its conjuncts, adding the other conjunct to the
conjunctive context as it does so. The advantage of splitting~(\ref{Eq-schema-1})
into (\ref{Eq-schema-base}) and~(\ref{Eq-schema-conj}) 
is that it is then possible to add rules for other logical 
operators, each of which passes around the conjunctive context; we will
call rules of this kind {\it translation-schemas}. We also show
the translation-schemas for quantified expressions and implications.
(\ref{Eq-schema-exists}) and~(\ref{Eq-schema-forall})
express translation of a quantified form in terms of translation of its body.
In each of them, $\theta$ substitutes a unique constant for the $\vec{x}$. 
\begin{eqnarray}
& & {Context \rightarrow (\theta(P) \equiv \theta(P^\prime))} \nonumber \\
& \Rightarrow & Context \rightarrow (\exists\vec{x}.P \equiv \exists \vec{x}.P^\prime) \label{Eq-schema-exists} \\
& & \nonumber \\
& & {Context \rightarrow (\theta(P) \equiv \theta(P^\prime))} \nonumber \\
& \Rightarrow & Context \rightarrow (\forall\vec{x}.P \equiv \forall \vec{x}.P^\prime) \label{Eq-schema-forall} 
\end{eqnarray}
(\ref{Eq-schema-impl1}) and~(\ref{Eq-schema-impl2}) express translation of an 
implication in terms of translations of its left- and right-hand sides. Note that 
the left-hand side is added to the context when translating the 
right-side side in~(\ref{Eq-schema-impl2}), but not {\it vice versa}. The proofs
of~(\ref{Eq-schema-exists})--(\ref{Eq-schema-impl2}) are straight-forward.
\begin{eqnarray}
& & {Context \rightarrow (P \equiv P^\prime)} \nonumber \\
& \Rightarrow & Context \rightarrow (P \rightarrow Q \equiv P^\prime \rightarrow Q) \label{Eq-schema-impl1}\\
& & \nonumber \\
& & {Context \wedge P \rightarrow (Q \equiv Q^\prime)} \nonumber \\
& \Rightarrow & Context \rightarrow (P \rightarrow Q \equiv P \rightarrow Q^\prime) \label{Eq-schema-impl2}
\end{eqnarray}

The central use of the equivalences is thus as 
truth-preserving conditional rewriting rules, which licence translation
of one of the conjuncts on the left-hand side into the right-hand side,
in suitable conjunctive contexts.
There is a second use of the equivalences as normal
Horn-clauses, which as we soon shall see is also essential to the
translation process. An equivalence of the form
\begin{displaymath}
\forall\vec{x}.(P_1\wedge P_2\wedge \ldots \equiv \exists\vec{y}.Q_1\wedge Q_2\wedge \ldots) \leftarrow Conds
\end{displaymath}
implies the validity, for any $k$, of all Horn-clauses either of the form
\begin{displaymath}
\forall\vec{x}.\forall\vec{y}.(P_k \leftarrow Q_1\wedge Q_2\wedge \ldots \wedge Conds)
\end{displaymath}
or
\begin{displaymath}
\forall\vec{x}.(\theta(Q_k) \leftarrow P_1\wedge P_2\wedge \ldots \wedge Conds)
\end{displaymath}
where $\theta$ replaces the $\vec{y}$ with Skolem functions of the $\vec{x}$.
We will refer to these, respectively, as {\it normal} and {\it backward}
Horn-clause readings of the equivalence. For example, the rule
\begin{verbatim}
and(man1(X),employee1(X)) <->
exists([HasCar],employee(X,m,HasCar))
\end{verbatim}
produces two normal Horn-clause readings,
\begin{verbatim}
man1(X) <- employee(X,m,HasCar).

employee1(X) <- employee(X,m,HasCar).
\end{verbatim}
and one backward Horn-clause reading,
\begin{verbatim}
employee(X,m,sk1(X)) <- man1(X),employee1(X).
\end{verbatim}
where \verb!sk1! is a Skolem function. In the implementation, 
each \verb!equiv! rule 
is compiled in three different ways, to yield the 
normal Horn-clause, backward Horn-clause and equivalential readings.
The Horn-clause readings are used to support proofs from the 
conjunctive context.

We can now define the basic translation process,
still neglecting for the time being the problems associated with
abductive proof and existential quantification on the left-hand side
of equivalences. 
Our strategy is to use (\ref{Eq-schema-base}) and the translation-schemas
as the kernel of a system that allows translation of logical forms,
using the equivalences as expandable complex definitions.

The actual process of 
translation of a complex formula $F$ is a series of single 
translation steps, each of which consists of the translation of 
an atomic constituent of $F$. A translation step contains the following
sub-steps:
\begin{description}
\item[Recurse:] descend through $F$ using the translation-schemas, until
an atomic sub-formula $A$ is reached. During this process, a conjunctive 
context $E$ has
been accumulated in which conditions will be proved, and some bound variables
will have been replaced by unique constants.
\item[Translate:] find a rule $(H\wedge R \equiv B) \leftarrow C$
such that $H$ unifies with $A$ with m.g.u. $\theta$.
If it is then possible to prove $\theta(R\wedge C)$ in $E$, replace $A$ with $\theta(B)$.
When carrying out a {\bf translate} step, we will sometimes refer to $H$ as
the {\it head} of the rules used, and $R\wedge C)$ as its {\it conditions}.
\item[Simplify:] if possible, apply simplifications to the resulting formula.
\end{description}

\subsection{A simple example}\label{Simple-example}

An example follows to illustrate how the process works. In the interests
of expositional clarity we use a simplified version of the actual CLARE domain
rules. We start with the sentence (S2), 
\begin{description}
\item[(S2)] Do any women work on CLARE?
\end{description}
which receives the LF
\begin{verbatim}
exists([Person,Event],
       and(woman1(Person),
           work_on1(Event,Person,clare)))                (LF1)
\end{verbatim}
{\tt (LF1)} has to be mapped to a query which accesses two 
database relations, {\tt SRI\_\-EMPLOYEE(Empl,Sex,HasCar)} and
{\tt SRI\_\-PROJECT\_\-MEMBER(Empl,Project)} 
({\tt Sex} can be \verb!w! or \verb!m!, and {\tt HasCar} can be {\tt y}
or {\tt n}). The desired result is thus {\tt (LF2)}:
\begin{verbatim}
exists([Empl,HasCar],
       and(SRI_EMPLOYEE(Empl,w,HasCar),
           SRI_PROJECT_MEMBER(clare,Empl)))              (LF2)
\end{verbatim}
The most clearly non-trivial 
part is justifying the conversion between the linguistic relation
{\tt woman1(X)} and the database relation 
{\tt SRI\_\-EMPLOYEE(X,w,\_)}.
Even in the limited PRM domain, it is incorrect to
state that ``woman'' is equivalent to ``employee classed as being of
female sex''; there are
for example several women who are listed in the \verb!PAYEE!
relation as having 
been the recipients of payments. It is more correct to say that
a tuple of type {\tt SRI\_\-EMPLOYEE(X,w,\_)} is equivalent to
the conjunction of two pieces of
information: firstly that \verb!X! is a woman, and secondly that she
is an employee. This can be captured in the rule
\begin{verbatim}
and(woman1(Person),
    employee1(Person)) <->
exists([HasCar],
  SRI_EMPLOYEE(Person,w,HasCar))                         (EQ1)
\end{verbatim}
In the left-to-right direction, the rule can be read as
``{\tt woman1(X)} translates to {\tt SRI\_\-EMPLOYEE(X,w,\_)},
in contexts where it is possible to prove {\tt employee1(X)}.''
For the rule to be of use in the present example, we must therefore
provide a justification for {\tt employee1(X)}'s holding
in the context of the query.
The simplest way to ensure that this is so is to provide a 
Horn-clause meaning postulate,
\begin{verbatim}
employee1(X) <-
  SRI_PROJECT_MEMBER(Project,X).                        (HC1)
\end{verbatim}
which encodes the fact that project members are employees.  

Similarly, we will need
an equivalence rule to convert between {\tt work\_on1}
and {\tt SRI\_\-PROJECT\_\-MEMBER}. Here the fact we want to state
is that project-members are precisely people who work on projects,
which we write as follows:
\begin{verbatim}
exists([Event],
  and(work_on1(Event,Person,Project),
      project1(Project))) <->
SRI_PROJECT_MEMBER(Project,Person)                      (EQ2)
\end{verbatim}
We will also make indirect use of the rule that states that 
projects are objects that can be found in the first field of
an {\tt SRI\_PROJECT} tuple,
\begin{verbatim}
project1(Proj) <->
exists([ProjNum,Start,End],
  SRI_PROJECT(Proj,ProjNum,Start,End))                  (EQ3)
\end{verbatim}
since this will allow us to infer (by looking in the database) that
the predicate {\tt project1} holds of {\tt clare}

Two translation steps now produce the desired transformation; in 
each, the schemas (\ref{Eq-schema-exists}) and (\ref{Eq-schema-conj}) 
are used in turn to reduce to the base case of expanding an atom. 
Remember that schema (\ref{Eq-schema-exists}) replaces variables with 
unique constants; when displaying the results of such a transformation, 
we will consistently write \verb!X*! to symbolize the new constant 
associated with the variable \verb!X!.

The first atom to be expanded is \verb!woman1(C*)!, and
the corresponding conjunctive context is 
{\tt work\_on1(E*, C*,clare)}.
{\tt woman1(C*)} unifies with the first conjunct on the left-hand side
of the \verb!(EQ1)!,
making its conditions \verb!employee1(C*)!. Using the
Horn-clause meaning postulate \verb!(HC1)!, this can be reduced
to {\tt SRI\_\-PROJECT\_\-MEMBER(Project,C*)}. Note that \verb!C*!
in this formula is a constant, while {\tt Project} is a variable.
This new goal can be reduced again, by applying the rule
\verb!(EQ2)! as a backwards Horn-clause, to
\begin{verbatim}
and(work_on1(Event,C*,Project),
    project1(Project))),
\end{verbatim}
The first conjunct can be proved from the assumptions, instantiating
\verb!Project! to \verb!clare!; 
the second conjunct can then be derived from the normal Horn-clause
reading of rule \verb!(EQ3)!, together with
the fact that \verb!clare! is listed as a 
project in the database. This completes the reasoning
that justifies expanding \verb!woman1(C)! in the context of this query, to
\begin{verbatim}
exists([HasCar],
  and(SRI_EMPLOYEE(C,w,HasCar)))
\end{verbatim}
The second translation step is similar. The atom to be expanded here is
{\tt work\_on1(E*, C*,clare)}, 
and the conjunctive context 
is {\tt woman1(C*)}. Now the rule \verb!(EQ2)! 
can be used; its conditions after
unification with the first conjunct in the LHS
are \verb!project1(clare)!,
the validity of which follows from another application
of \verb!(EQ3)!. So {\tt work\_on1(E,C,clare)} can be expanded
to {\tt SRI\_\-PROJECT\_\-MEMBER(clare,C)}, giving the desired
result. 

\subsection{Existential quantification}\label{Existential}

We finally consider cases where conditional equivalences contain
existential quantifiers on their left-hand sides. It will be
sufficient to restrict ourselves to unconditional equivalences where the LHS is
an existentially quantified atomic formula, that is equivalences of
the form
\begin{equation}
(\exists\vec{x}.P_1) \equiv P^\prime \label{Equiv-ex-LHS}
\end{equation}
since an expression of the form
\begin{equation}
Conds \rightarrow (\exists\vec{x}.P_1\wedge P_2\wedge P_3\ldots) \equiv P^\prime 
\end{equation}
can be rewritten, introducing the new predicate $P_{123}$,
as the two equivalences
\begin{equation}
\forall\vec{x}.(Conds \rightarrow (P_1\wedge P_2\wedge P_3\ldots) \equiv P_{123}(\vec{x}))
\end{equation}
(of type~\ref{Equiv-univ-LHS}), and
\begin{equation}
(\exists\vec{x}.P_{123}) \equiv P^\prime
\end{equation}
(of type~\ref{Equiv-ex-LHS}).

Equivalence of type~\ref{Equiv-ex-LHS} become important in connection with the
so-called ``Doctor on Board'' problem (Perrault and Grosz, 1988),
which in our domain can be illustrated by a query like (S3),
\begin{description}
\item[(S3)] Does Mary have a car?
\end{description}
This receives the LF
\begin{verbatim}
exists([Event,Car],
       and(car1(Car),have1(Event,mary,Car))))          (LF3)
\end{verbatim}
for which the intended database query is
\begin{verbatim}
exists([Sex],
       SRI_EMPLOYEE(mary,Sex,y))                       (LF4)
\end{verbatim}
if Mary is listed as an employee; the relevant equivalence from the
LDT could be
\begin{verbatim}
exists([Event,Car],
       and(employee1(Empl),
           and(car1(Car),
               have1(Event,Empl,Car))) <->
exists([Sex]
       SRI_EMPLOYEE(Empl,Sex,y))
\end{verbatim}
which we rewrite as 
\begin{verbatim}
and(employee1(Empl),
    and(car1(Car),
        have1(Event,Empl,Car))) <->
employee_has_car(Event,Empl,Car)                       (EQ4)

exists([Event,Car],
       employee_has_car(Event,Empl,Car)) <->
exists([Sex],
       SRI_EMPLOYEE(Empl,Sex,y))                       (EQ5)
\end{verbatim}
{\tt (EQ4)} and {\tt (EQ5)} illustrate the coding trick described above;
{\tt (EQ4)} is of type~\ref{Equiv-univ-LHS}, {\tt (EQ5)} is of
type~\ref{Equiv-ex-LHS}, and the role of predicate $P_{123}$ is played 
by {\tt employee\_\-has\_\-car}.

We now describe briefly how {\tt (EQ4)} and {\tt (EQ5)} can be used
to translate {\tt (LF3)}.
If we assume that {\tt employee1(mary)} can somehow be proved, we 
can apply {\tt (EQ4)} to {\tt (LF3)} to translate the sub-expression
{\tt car1(Car)}, using the 
schemas~\ref{Eq-schema-exists},~\ref{Eq-schema-conj}
and ~\ref{Eq-schema-base}; the details are similar to those of the
derivation in section~\ref{Simple-example}, and the result is {\tt (LF5)}.
\begin{verbatim}
exists([Event,Car],
       and(employee_has_car(Event,mary,Car),
           have1(Event,mary,Car)))                      (LF5)
\end{verbatim}
The second conjunct of {\tt (LF5)} can be simplified away 
(see section~\ref{Functional-relations}); since
one of the normal Horn-clause readings
of {\tt (EQ4)} is 
\begin{verbatim}
have1(Event,Empl,Car) <- employee_has_car(Event,Empl,Car)
\end{verbatim}
we have that {\tt have1(Event,mary,Car)} is implied by its
conjunctive context {\tt employee\_\-has\_\-car(Event, mary, Car)}
and can thus be removed.
This reduces {\tt (LF5)} to {\tt (LF6)}
\begin{verbatim}
exists([Event,Car],
       employee_has_car(Event,mary,Car))                (LF6)
\end{verbatim}
after which {\tt (EQ5)} can be applied directly, yielding
{\tt (LF4)} as required.

\section{Simplification}\label{Simplification}

AET, like many systems based on the idea
of rewriting, tends to suffer from the problem that expressions can
rapidly grow in size as they pass through successive stages of translation.
The standard solution to the problem is to include a {\it simplifier}, a 
module which takes the output from a translation stage and attempts
to reduce it to a more compact form. This section will describe 
CLARE's simplifier. Although the simplifier's primary use is in conjunction
with the translator, it has turned out to provide a functionality that
several other parts of the system utilise.
In section~\ref{Simplify-assertions}
we describe how the simplifier is used in the processing of assertions; here,
the problem is to take the conjunction of the TRL representations
of several assertions, and combine them into a compact form.

The simplifier consists of two main pieces. The first, described in
section~\ref{Normal-simplification} is a collection of standard
logical rewriting techniques, none of which use inference, that
have the common function of reducing expressions to a canonical
form. Simplification can also be carried out by using the inference-based
equivalential translation methods described earlier in this chapter,
in which inference justifies the replacement of sub-expressions by
other forms the equivalence of which is implied by the context.
This second type of simplification is described in 
section~\ref{Functional-relations}. 

\subsection{Non-inferential simplification}\label{Normal-simplification}

The non-inferential simplification module consists of the set of
rewriting methods listed below. Most of them are fairly obvious,
though there are a few tricky points. We present the methods in
the order in which they are applied.

\subsubsection{Moving existential quantifiers outwards}

This method essentially consists of two kinds of rewriting rule.
The first simplifies an expression built out of conjunction
and existential quantification operators into an equivalent form with a
single existential operator on the outside binding all the variables;
thus for example
\begin{verbatim}
and(exists([X],
      p(X)),
    exists([Y],
      q(Y)))
\end{verbatim}
will get rewritten to
\begin{verbatim}
exists([X,Y],
  and(p(X),Q(Y)))
\end{verbatim}
The other rewriting rule is applied when
an existentially quantified form occurs on the LHS of an implication;
in this case, the existentially quantified variables are moved upwards
to become universally quantified with wide scope over the implication.
The justification for this manoeuvre is the equivalence
\begin{displaymath}
((\exists x.P) \rightarrow Q)) \equiv \forall x.(P \rightarrow Q)
\end{displaymath}
which is easily proved. The result of recursively applying the two
rules is to give existential quantifiers as wide scope as possible,
if necessary turning them into universals on the way.

\subsubsection{Simplification of equalities}

The next simplification step removes equalities from the expression 
when this is possible. The basic equivalences that licence this
type of simplification are the following:
\begin{displaymath}
(\exists x.\exists y.P\wedge (x = y)) \equiv (\exists x. P[y/x])
\end{displaymath}
\begin{displaymath}
(\forall x.\exists y.P\wedge (x = y)) \equiv (\forall x. P[y/x])
\end{displaymath}
and 
\begin{displaymath}
(\exists y.P\wedge (y = a)) \equiv P[y/a]
\end{displaymath}
where $a$ is a constant. The methods are implemented by recursively
descending through the expression to find equality sub-forms, and
in suitable cases unifying together their two arguments, replacing
the equality with an occurrence of the identically true predicate.
After this, a second pass removes the occurrences of $true$ and the
vacuous and repeated quantifications introduced.

If the expression contains an equality whose arguments consist of
non-unifiable constants, its value can be known to be identically
false (assuming non-identity of reference of distinct constants).
In this case, the simplifier replaces the equality with the 
form
\begin{quote}
\begin{verbatim}
mismatch(Arg1,Arg2)
\end{verbatim}
\end{quote}
Subsequent behaviour depends on the nature of the expression being manipulated.
If it is a question, the {\tt mismatch} form is later replaced with an
occurrence of the identically false predicate, which will generally lead
to a ``No'' answer being produced. If the expression however represents
an assertion, the mismatch normally represents a presupposition failure.
In this case, the interface handler will inform the user that the
expression was translated to an identically false form due to the occurrence
of the mismatch, and attempt to print the incompatible arguments in
an informative way.

\subsubsection{Simplification of ground sub-expressions}

The final type of simplification involves ground sub-expressions, 
that is to say sub-expressions containing no variables. Since
these can be assigned a truth-value irrespective of their context,
the expression can be simplified by computing it and replacing
by an occurrence of either $true$ or $false$. At present, the method is
restricted in two ways: ground sub-expressions are only replaced
if their predicates are declared to be executable relations
(cf. section~\ref{Effective-translation}); also, only {\it true}
ground expressions are substituted. 

\subsection{Inferential simplification}\label{Functional-relations}

We will now consider
methods that use equivalential translation to carry
out simplification; the notion of {\it conjunctive context},
defined in section~\ref{Translation-schemas}, will play a key role.
We can immediately describe one way to achieve this end. Suppose
that $F$ is a formula containing an occurrence
of the subformula $F^\prime$, and suppose further
that $F^\prime$ is implied by its conjunctive context. It follows that
replacing $F^\prime$ with $true$ in $F$ will result in a formula
equivalent with $F$. This gives a simple method for removing 
duplicated conjuncts and the like.

A similar, but more refined, form of simplification can also be 
performed, which exploits functional relationships between arguments
in predicates. We start by re-examining our example (S1) from the 
beginning of the chapter,
reproduced here for convenience.
\begin{description}
\item[(S1)] List all payments made to BT during 1990.
\end{description}
In section~\ref{Examples}, we will see that
the logical form originally derived from (S1)
contains after a few translation steps three separate instances of 
the \verb!transaction! relation,
one from each of the original linguistic predicates 
\verb!payment1!, \verb!make2! and 
{\tt during1}. Assuming that the arguments of the {\tt transaction}
relation are
\begin{verbatim}
transaction(Payer,Transaction,Date,Payee,Amount)
\end{verbatim}
{\tt payment1(Payment)} expands, roughly speaking, to
\begin{verbatim}
transaction(_,Payment,_,_,_)
\end{verbatim}
{\tt make2(Event,Agent,Payment,Payee)} expands to
\begin{verbatim}
transaction(Agent,Payment/Event,_,Payee,_)
\end{verbatim}
and {\tt during1(Payment,Interval)} expands to
\begin{verbatim}
and(transaction(_,Payment,Date,_,_),
    time_during(Date,Interval))
\end{verbatim}
The database query will
conjoin all three instances together. It is clearly preferable, if
possible, to merge them instead,
yielding a composite predication 
\begin{verbatim}
transaction(Agent,Payment/Event,Date,Payee,_)
\end{verbatim}
The information that licences this step as a valid simplification is 
that \verb!transaction! is a function from its second argument
(the payment) to the remaining ones (the agent, the date,
the payee and the amount); in other words, a given payment is made by a unique
agent, on a unique date, to a unique payee, for a unique amount.
The system allows the information
to be entered as a ``function'' meaning postulate in the form
\begin{verbatim}
function(transaction(Payer,Transaction,Date,Payee,Amount),
         [Transaction] -> [Payer,Date,Payee,Amount])
\end{verbatim}
which is treated as a concise notation for the conditional equivalence
postulate
\begin{verbatim}
(transaction(Payer,Transaction,Date,Payee,Amount) <->
 and(Payer = Payer1,
     and(Date = Date1,
         and(Payee = Payee1,
             Amount = Amount1)))))  <-
 transaction(Payer1,Transaction,Date1,Payee1,Amount1)
\end{verbatim}
It is thus possible to view
``merging'' simplification of this kind as just another instance
of equivalential translation. In the current version of
the system, the transformation process operates in a cycle,
alternating normal translation followed by simplification
using the same basic interpreter; simplification consists
of functional ``merging'' followed by reduction of equalities 
where this is applicable.

\subsection{Simplification for processing assertions}\label{Simplify-assertions}

The simplification process also plays an important role in the processing of
assertions. Consider, for example, what would happen to
the pair of sentences (S6) - (S7) without simplification:
\begin{description}
\item[(S6)] Clara is an employee who has a car.
\item[(S7)] Clara is a woman.
\end{description}
(S6) translates into the database form
\begin{verbatim}
exists([A,B],
       SRI_EMPLOYEE(clara,A,y))
\end{verbatim}
(The second field in \verb!SRI_EMPLOYEE! indicates sex, and the
third whether or not the employee has a company car).
This can then be put into Horn-clause form as 
\begin{verbatim}
SRI_EMPLOYEE(clara,sk1,y)
\end{verbatim}
and asserted into the Prolog database. Since Clara is now known 
to be an employee, (S7) will produce the unit clause
\begin{verbatim}
SRI_EMPLOYEE(clara,w,sk2)
\end{verbatim}
The two clauses produced would contain all the information entered,
but they could not be entered into a relational database as they 
stand; a normal database has no interpretation for the Skolem 
constants \verb!sk1! and \verb!sk2!.
However, it is possible to use function information to merge them
into a single record. The trick is to arrange
things so that the system can when necessary recover the
existentially quantified form from the Skolemized one;
all assertions which contain Skolem constants are kept together
in a ``local cache''.
Simplification of assertions then proceeds according to
the following sequence of steps:
\begin{enumerate}
\item Retrieve all assertions from the local cache.
\item Construct a formula $A$, which is their logical conjunction.
\item Let $A_0$ be $A$, and let $\{sk_1\ldots sk_n\}$ be the
Skolem constants in $A$. For $i = 1\ldots n$, let
$x_i$ be a new variable, and let $A_i$ be
the formula $\exists x_i.A_{i-1}[sk_{i}/x_i]$, i.e. the result of
replacing $sk_i$ with $x_i$ and quantifying existentially over
it. 
\item Perform normal function merging on $A_n$, and call
the result $A^\prime$.
\item Convert $A^\prime$ into Horn-clause form, and replace
the result in the local cache.
\end{enumerate}
In the example above, this works as follows. After (S6) and (S7)
have been processed, the local cache contains the clauses
\begin{verbatim}
SRI_EMPLOYEE(clara,sk1,y)

SRI_EMPLOYEE(clara,w,sk2)
\end{verbatim}

$A = A_0$ is then the formula
\begin{verbatim}
and(SRI_EMPLOYEE(clara,sk1,y)
    SRI_EMPLOYEE(clara,w,sk2))
\end{verbatim}
and $A_2$ is
\begin{verbatim}
exists([X1,X2]
  and(SRI_EMPLOYEE(clara,X1,y)
      SRI_EMPLOYEE(clara,w,X2))
\end{verbatim}
Since \verb!SRI_EMPLOYEE! is declared functional on its first 
argument, the second conjunct is reduced to two equalities, giving
the formula
\begin{verbatim}
exists([X1,X2]
  and(SRI_EMPLOYEE(clara,X1,y)
      and(X1 = w,
          y = X2))
\end{verbatim}
which finally simplifies to $A^\prime$,
\begin{verbatim}
SRI_EMPLOYEE(clara,w,y)
\end{verbatim}
a record without Skolem constants, which can be added to a normal
relational database.

\section{The Closed World Assumption}
\label{Closed-World}

This section will briefly consider the question of 
how the Closed World Assumption fits into
the picture we have constructed; to make the discussion concrete, we will
concentrate on the case of Y-N questions. We start with a logical form,
$F_{ling}$, for which we assume that we can find an effective translation 
$F_{eval}$, and assume further that it turns out that there is no proof
of $F_{eval}$. Inability to prove $F_{eval}$ will imply that it is false, 
since we have assumed the existence of a finite proof procedure; 
this in turn implies (if the abductive
assumptions are granted) that $F_{ling}$ is
false, since the two are equivalent. 
With regard to the CWA, the interesting thing to consider is the part 
played by the database predicates that occur in $F_{eval}$.
Knowledge of the database predicates is complete,
since they have ``trivial''  semantics, only
referring to the existence of database tuples and not directly to
the rest of the world; they are closed by construction and the 
Closed World Assumption may safely be used on them.
However, the predicates in
$F_{ling}$ are in general not closed. It is interesting to examine
in more detail how the connection between the ``closed'' predicates in
$F_{eval}$ and the ``open'' predicates in $F_{ling}$ is defined.

In practice, the key step in 
proving the equivalence between $F_{ling}$ and $F_{eval}$
is usually of the following form.
There are two predicates $P(x,y)$ and $P_{rec}(n_{x},n_{y})$, whose intended
semantics are, respectively, ``$x$ and $y$ stand in relationship $P$''
and ``It is recorded in the database that objects named
$n_{x}$ and $n_{y}$ stand in relationship 
$P$''. Thus $P_{rec}$ is closed, but $P$ isn't. The two predicates
are related by a domain axiom of the form
\begin{eqnarray}
C(y) \rightarrow
(P(x,y) \equiv \exists n_{x},n_{y}.name_1(x,n_{x}) \nonumber \\
\wedge name_2(y,n_{y}) \nonumber \\
\wedge P_{rec}(n_{x},n_{y})) \label{CW-key-equiv}
\end{eqnarray}
Here, each $name_i(x,n_{x})$ ($i = 1,2$) 
is a predicate that relates an object $x$ and the identifier $n_{x}$ assigned
to it according to naming
convention $i$. (As pointed out above in section~\ref{Ontology},
it is important to take account of the fact that the same identifier 
can name different objects in different contexts).
$C(y)$ defines a sufficient condition for $P(x,y)$ to have been
recorded; for example, $y$ might be a date and $C(y)$ could say that $y$ is within
the period for which records have been kept.

Now if $P(x,y)$ occurs in $F_{ling}$ conjoined with some expression 
$D(y)$, where $D(y)$ can be proved to imply $C(y)$, 
then~(\ref{CW-key-equiv}) and~(\ref{Eq-schema-conj}) can be invoked to justify
replacing $P(x,y)$ in $F_{ling}$ with $\exists n_{x},n_{y}.name_1(x,n_{x})
\wedge name_2(y,n_{y})\wedge P_{rec}(n_{x},n_{y})$ while maintaining
equivalence. In this way, equivalential translation allows the
CWA in effect to be made conditional on the context of use.
Equivalence of type~(\ref{CW-key-equiv}) are discussed further 
in section~\ref{Equiv-dbconnect}.

\section{Translating to SQL}\label{SQL-convert}

This section briefly describes the module responsible for 
synthesis of actual SQL queries. The conversion module takes as 
input a TRL form $P$, assumed to represent a question; it
outputs a form $P^\prime$, such that $P$ and $P^\prime$ are equivalent
and as much as possible of $P$ has been replaced by calls to
the SQL interface. The interface is mediated through the predicate
\begin{quote}\begin{verbatim}
trl_select(Vars,SQLQuery)
\end{verbatim}\end{quote}
where {\tt SQLQuery} is a term representing an SQL {\tt SELECT} statement, and
{\tt Vars} is a list whose length is equal to the number of selected
variables in {\tt SQLQuery}. The intended semantics are that 
\begin{quote}\begin{verbatim}
trl_select(Vars,SQLQuery)
\end{verbatim}\end{quote}
holds iff {\tt Vars} are a row selected by {\tt SQLQuery}. 
{\tt SQLQuery} is a logical
term representing the abstract SQL syntax; conversion to concrete
SQL syntax is handled by the low-level SQL interface, and is straightforward.
We now describe how the conversion process is carried out; we will
illustrate with the extended example from section~\ref{Examples}
\begin{description}
\item[(S1)] List all payments made to BT during 1990.
\end{description}
This receives the final translated TRL representation (slightly
simplified in the interests of readability)
\begin{verbatim}
forall([TransId,Date,DBDate,AMNum],
       impl(and(TRANS(TransId,DBDate,bt,AMNum)
                and(db_date_convert(Date,DBDate),
                    and(t_precedes(date([1990,1,1]),Date),
                        t_precedes(Date,date([1990,12,31])))))
            x([Id,DisplayT],
              and(execute(DisplayEv,
                          display([TransId,DBDate,bt,AMNum]),
                          clare,
                          DisplayT),
                   t_precedes(<Now>,DisplayT)))))
\end{verbatim}
which can be glossed as 
\begin{quote}
``Find all {\tt TRANS} tuples with {\tt trn\_id} field {\tt TransId},
{\tt cheque\_date} field {\tt DBDate}, {\tt payee} field {\tt bt} and {\tt amount}
field {\tt Amt}, such that {\tt DBDate} represents a date after 1/1/90 and before
31/12/90; and for each one display the list {\tt [TransId, DBDate, bt, Amt]}.''
\end{quote}
Taking the conclusion first, the translation to SQL form produced is 
\begin{verbatim}
forall([TransId,DBDate,AMNum],
       impl(trl_select([TransId,DBDate,AMNum],
                        SELECT([t_1.trn_id,t_1.amount,t_1.cheque_date],
                               FROM([alias(TRANS,t_1)]),
                               WHERE([t_1.payee=bt,
                                      sql_date_=<(1-JAN-90,
                                                  t_1.cheque_date)),
                                      sql_date_=<(t_1.cheque_date,
                                                  31-DEC-90))]))),
            x([Id,DisplayT],
              and(execute(DisplayEv,
                          display([TransId,DBDate,bt,AMNum]),
                          clare,
                          DisplayT),
                   t_precedes(<Now>,DisplayT)))))
\end{verbatim}
Here, the abstract SQL syntax is a hopefully transparent representation of the
query whose concrete syntax will be
\begin{verbatim}                         
"SELECT DISTINCT t_1.trn_id , t_1.cheque_date , t_1.amount 
 FROM TRANS t_1
 WHERE t_1.payee = 'bt'
 AND '1-JAN-90' <= t_1.cheque_date
 AND t_1.cheque_date <= '31-DEC-90"
\end{verbatim}

We now explain in more detail how this result can be produced.
Processing traces the following series of top-level steps:
\begin{enumerate}
\item Translate representation-independent evaluable predicates such
as {\tt t\_precedes} into database access language dependent primitives
using AET.
\item Insert database column names into database predicates.
\item Make column names unique by associating a ``relation alias'' with each
      occurrence of a database predicate.
\item Group together conjunctions of goals suitable for turning into
      {\tt SELECT} statements. Goals suitable for inclusion are goals
      representing database lookups and goals that map onto conditions
      in {\tt WHEN} clauses. For each such conjunction, do the following:
      \begin{enumerate}
      \item Extract the variables that are to be selected. 
      \item Replace variables by column descriptions.
      \item Extract the variables that are constrained to have fixed values,
            and add these as suitable equalities to the {\tt WHEN} clause.
      \item If more than one relation instance is referenced, add equalities
            to the {\tt WHEN} clause to represent the join.
      \end{enumerate}
\item Simplify the result.
\end{enumerate}
We use the example to illustrate. The first step is to use AET to translate
the occurrences of {\tt t\_precedes}. The relevant equivalence is
\begin{verbatim}
t_precedes(Date1,Date2) <->
exists([DBDate1,DBDate2],
  and(db_date_convert(Date1,DBDate1),
  and(db_date_convert(Date2,DBDate2),
      sql_date_=<(DBDate1,DBDate2))))
\end{verbatim}
and applied twice; after some simplification (exploiting the fact that 
{\tt db\_\-date\_\-convert} is a function from each of its arguments
to the other one, see section~\ref{Functional-relations}), the result is
\begin{verbatim}
forall([TransId,Date,DBDate,AMNum],
       impl(and(TRANS(TransId,DBDate,bt,AMNum)
                and(db_date_convert(Date,DBDate),
                    and(sql_date_=<(1-JAN-90,DBDate)
                        sql_date_=<(DBDate,31-DEC-90))))),
            x([Id,DisplayT],
              and(execute(DisplayEv,
                          display([TransId,DBDate,bt,Amt]),
                          clare,
                          DisplayT),
                   t_precedes(<Now>,DisplayT)))))
\end{verbatim}
Note that {\tt 1-JAN-90} and {\tt 31-DEC-90} are atomic SQL date representations.
The next two steps are fairly trivial in nature, and involve substituting
SQL column names in the {\tt TRANS} relation and creating a unique relation alias
{\tt t\_1} for it. (Since there is only one relation here, this step is not
actually necessary, but we show it for completeness). The result is
\begin{verbatim}
forall([TransId,Date,DBDate,AMNum],
       impl(and(tuple(t_1,TRANS([trn_id=TransId,
                                 cheque_date=DBDate,
                                 payee=bt,
                                 amount=AMNum])),
                and(db_date_convert(Date,DBDate),
                    and(sql_date_=<(1-JAN-90,DBDate)
                        sql_date_=<(DBDate,31-DEC-90))))),
            x([Id,DisplayT],
              and(execute(DisplayEv,
                          display([TransId,DBDate,bt,Amt]),
                          clare,
                          DisplayT),
                   t_precedes(<Now>,DisplayT)))))
\end{verbatim}
Conjunctions are now when possible turned into {\tt trl\_select} goals;
the only suitable conjunction is
\begin{verbatim}
and(tuple(t_1,TRANS([trn_id=TransId,
                     cheque_date=DBDate,
                     payee=bt,
                     amount=AMNum])),
     and(sql_date_=<(1-JAN-90,DBDate)
         sql_date_=<(DBDate,31-DEC-90)))
\end{verbatim}
Here, the database look-up goals are the singleton
\begin{verbatim}
{tuple(t_1,TRANS([trn_id=TransId,
                  cheque_date=DBDate,
                  payee=bt,
                  amount=AMNum]))}
\end{verbatim}
and the initial {\tt WHEN} clause goals are
\begin{verbatim}
{sql_date_=<(1-JAN-90,DBDate)
 sql_date_=<(DBDate,31-DEC-90))}
\end{verbatim}
Consulting the database look-up goals, the variables to be selected 
are {\tt TransId}, {\tt DBDate} and
{\tt AMNum}, so they end up being the list of variables in the
{\tt trl\_select} goal's first argument; then replacing the variables
inside the {\tt SELECT} with column descriptions, we replace
{\tt TransId} with {\tt t\_1.trn\_id}, {\tt DBDate} with 
{\tt t\_1.cheque\_date} and {\tt AMNum} with {\tt t\_1.amount}.
The {\tt WHEN} clause goals now become
\begin{verbatim}
{sql_date_=<(1-JAN-90,t_1.cheque_date)
 sql_date_=<(t_1.cheque_date,31-DEC-90))}
\end{verbatim}
There is a single column, {\tt payee}, constrained to have a fixed value, {\tt bt},
so it adds another goal
\begin{verbatim}
t_1.payee=bt
\end{verbatim}
to the {\tt WHEN} clause. The final result is the {\tt trl\_select} goal
\begin{verbatim}
trl_select([TransId,DBDate,AMNum],
           SELECT([t_1.trn_id,t_1.amount,t_1.cheque_date],
                  FROM([alias(TRANS,t_1)]),
                  WHERE([t_1.payee=bt,
                         sql_date_=<(1-JAN-90,
                                     t_1.cheque_date)),
                         sql_date_=<(t_1.cheque_date,
                                     31-DEC-90))])))))
\end{verbatim}
Replacing the conjunction with the {\tt trl\_select} goal, we reach the final
form,
\begin{verbatim}
forall([TransId,DBDate,AMNum],
       impl(trl_select([TransId,DBDate,AMNum],
                        SELECT([t_1.trn_id,t_1.amount,t_1.cheque_date],
                               FROM([alias(TRANS,t_1)]),
                               WHERE([t_1.payee=bt,
                                      sql_date_=<(1-JAN-90,
                                                  t_1.cheque_date)),
                                      sql_date_=<(t_1.cheque_date,
                                                  31-DEC-90))]))),
            x([Id,DisplayT],
              and(execute(DisplayEv,
                          display([TransId,DBDate,bt,AMNum]),
                          clare,
                          DisplayT),
                   t_precedes(<Now>,DisplayT)))))
\end{verbatim}

\chapter{Linguistic Domain Theories}

\label{claredm}

\section{Introduction}

From the examples presented in the previous chapter, the reader will
already have gained some impression of what a linguistic domain theory
looks like. In this chapter, we will consider the subject in more detail.
We describe the various types of information that compose an LDT,
and a methodology that can be used to construct LDTs for interfacing
to relational databases.

There are four main kinds of information in an LDT. The greater part 
of the theory is normally composed of conditional equivalences, which
have already been discussed at length in chapter~\ref{translation}. 
Apart from these, there are
Horn-clause formulas, declarations of functional relationships between
arguments of predicates, and declarations of what assumptions are
permissible. We will first review each of these briefly; later on in the
chapter, we explain exactly how they fit into the methodology for 
constructing an LDT.
\begin{description}
\item[Horn-clauses]
Certain relationships in the LDT are inherently ones of implication rather
than equivalence; in these cases, Horn-clauses, rather than equivalences,
are appropriate. For example, it can be necessary to capture the fact
that everyone who books time to a project is of necessity an employee.
This cannot easily be made into an equivalence, since being an employee
does not directly imply that one books time to any particular project. 
A suitable Horn-clause will be something of the approximate form
\begin{verbatim}
employee_Worker(Person) <-
  booking_to_project(Person,Project,NumberOfHours,Date)
\end{verbatim}

\item[Functional relationships]
Many relations have the property of being functional on some subset of
their arguments;
that is to say, selecting a particular given value for that subset determines
a unique value for the remaining arguments. In practice, the most important
case is that of {\it primary key} fields in database predicates. We shall
later see that there are a number of ways in which information about
functional relationships can prove very useful. Functional relationships
are defined by declarations of the form
\begin{verbatim}
function(<Template>,<FunctionalArgs> -> <RemainingArgs>)
\end{verbatim}
Most commonly, \verb!<FunctionalArgs>! is a list consisting of a single
argument. For example, if the first argument of the {\tt TRANS} relation is
an identifier that functions as a primary key, this can be captured by a
declaration like
\begin{verbatim}
function(TRANS(TransId,ChequeNum,Date,Payee,Account,Amount),
         [TransId] -> [ChequeNum,Date,Payee,Account,Amount])
\end{verbatim}

\item[Assumption declarations]
Assumption declarations are used to control the abductive proof
mechanism: it is only permissible to make an abductive assumption when
this is explicitly licenced by an assumption declaration. 
An assumption declaration has the general form
\begin{verbatim}
assumable(<Goal>,<Cost>,<Justification>,<Type>,<Conditions>)
\end{verbatim}
The intended semantics are that \verb!<Goal>! may be assumed at 
cost \verb!<Cost>! in an environment where
\verb!<Conditions>! hold. \verb!<Type>! can be either {\tt specialization},
{\tt limitation} or {\tt approximation}; this is explain further in
sections~\ref{Basic-LDT} and~\ref{Assumptions}.
\verb!<Justification>!
is a tag which can be used to identify instances of use of this
declaration. For example, if one wants to declare that an individual
which occurs in the first, ``payer'' argument place of a {\tt transaction}
relation can be assumed without cost
to be SRI, a suitable declaration would be
\begin{verbatim}
assumable(Payer = sri,
          0,
          all_transactions_referred_to_are_from_SRI,
          specialization,
          transaction(Payer,_,_,_,_,_,_))
\end{verbatim}
\end{description}
The rest of the chapter will explain in detail how these various kinds
of information can be used to build up an LDT; we will focus on the case
of relational databases, but much of the methodology will carry over to
other kinds of knowledge-base applications. We start in section~\ref{DBLDT}
by giving an overview of what an LDT for a relational database will
look like. In the following sections, we go into more detail about
specific modelling problems.

\section{LDTs for relational databases}
\label{DBLDT}

This section gives an overview of what an LDT for a relational
database looks like, based on our experience with the example
LDT for the PRM domain. The PRM application, which is based on a real SRI 
projects and payments database, contains about 280 equivalences,
60 Horn clauses, 45 declarations of functional relationships,
and 25 assumption declarations.
The axioms are stratified into three distinct groups,
listed in order of increasing specificity: this reflects a
strategy which translates the original logical form predicates
into predicates which are increasingly closer to the database.
We refer to the groups as {\it general}, {\it domain-specific} and
{\it database-specific}, and describe each in turn.

\subsection{General axioms}\label{Basicdomain}

About 175 of the axioms fall into the first group, which we refer to
collectively as ``the general LDT''.
About 45 of them relate the words used in the
utterance situation (``show'', ``question'', ``answer'' etc.) to the
primitive executable relations. Thus for example a typical equivalence 
of this type would be the one that translates the ditransitive verb ``show'' (``Show 
me the payments'') into an {\tt action} of type {\tt display}:
\begin{verbatim}
show2(Event,Agent,X,Audience) <->
exists([Time],
  action(Event,display(X),Agent,Audience,Time))
\end{verbatim}
Of the remaining ``general'' axioms,
about 100 translate basic word senses, like temporal prepositions and 
size expressions, into a small set of ``primitive predicates'' that
associate events and objects with characteristic
attributes such as time of occurrence and magnitude. 
Thus for example the temporal preposition
``during'' is translated by the following equivalence:
\begin{verbatim}
during1(E1,E2) <->
exists([T1,T2,Gran1,Gran2],
  and(associated_time(E1,T1,Gran1),
  and(associated_time(E2,T2,Gran2),
      time_during([T1,Gran1],[T2,Gran2]))))
\end{verbatim}
This expresses the fact that an event {\tt E1} can be regarded as being
``during'' another event {\tt E2} if the time associated with the first
event is contained in the time associated with the second one. The
arguments {\tt Gran1} and {\tt Gran2} express the conceptual
level of granularity of the events {\tt E1} and {\tt E2}; at present they
can be either {\tt days} (appropriate for database events) or {\tt seconds}
(appropriate for utterance situation events).
Note that the definition of the {\tt associated\_time} predicate is provided
by domain-specific axioms; the equivalence performs the domain-independent
function of translating the linguistic
predicate {\tt during1} into a more convenient form expressed in terms 
of the general predicates {\tt associated\_\-time} and 
{\tt time\_\-during}.

The final set of about 30 general rules deal with translations of predicates
like {\tt time\_\-during}, which relate times; the 
complications are caused by the fact that ``times'' can be either points
or intervals, and can also be expressed at different levels of granularity.
The temporal axioms 
translate all the temporal predicates into a uniform representation in
terms of a single predicates, {\tt time\_\-point\_\-precedes}
that compares time-points with regard to 
temporal precedence.

\subsection{Domain-specific axioms}

The axioms of the second group, which for the PRM domain are slightly less
numerous than the first, are domain-specific but not database-specific.
They translate predicates representing senses of domain words 
(``project'', ``transaction'', ``deliverable'' etc.)
and semantic primitive predicates like {\tt associated\_\-time}
into a set of target predicates which we refer to as the ``conceptual''
predicates: the intention is that the conceptual predicates should be
roughly the ones that would occur in a conceptual database schema for
the domain. This means in practice that each distinct type of object and
event will have its own conceptual predicate, whose arguments 
correspond to the various entities associated with the object. 
``Conceptual predicate'' has
of course much in common with standard AI notions like ``Minsky frame''.
The PRM domain has about ten conceptual predicates, corresponding to 
concepts like ``project'', ``transaction'' and ``deliverable'';
a conceptual predicate typically has between three and seven arguments.
For example, the conceptual predicate for ``transaction''
has arguments corresponding to the transaction itself,
the payer, the payee, the cheque used,
the amount, the account on which the payment was made and the date.

Translation of the ``primitive predicates'' is normally dependent
on the nature of their arguments, and the corresponding equivalences
are thus either conditional or have complex left-hand sides.
For example, the time associated
with a payment is the transaction date, but that associated with a
project is the interval spanned by the start- and end-dates. The
equivalences that express this are
\begin{verbatim}
and(associated_time(Trans,Date,Granularity),
    transaction1(Trans)) <->
exists([Payer,Cheque,Payee,Acc,Amt],
   and(transaction(Payer,Trans,Cheque,Date,
                   Payee,Acc,Amt),
       Granularity = days))

and(associated_time(Project,Interval,Gran),
    project1(Project)) <->
exists([StartDate,EndDate],
    and(project(Org,Project,Account,
                StartDate,EndDate),
        Interval = interval(StartDate,EndDate))
\end{verbatim}

\subsection{Database-specific axioms}

The third group of axioms relate the ``conceptual
predicates'' to the associated concrete database predicates.
They capture specific
coding peculiarities which relate database-objects and their 
associated real-world counterparts, which as explained above
in section~\ref{Closed-World} includes contingent limitations
on the availability of recorded data. The database-specific group is
small, consisting of about 25 axioms; these consist of one equivalence
linking each conceptual predicate to its concrete counterpart,
together with some axioms that define the ``contingent limitations''
when necessary. For example, the axioms for the {\tt book\_\-time\_\-to\_\-project}
relation (slightly simplified) are
\begin{verbatim}
(exists([Ev],
   book_time_to_project(Ev,Emp,Amt,D,Acc) <->
 exists([EmpId,T,TNum,DateNum,AC],
   and(SRI_TIMESHEET(EmpId,Amt,DateNum,AccId),
       and(employee_id(Emp,EmpId),
           and(date_id(D,DateNum),
               account_id(Acc,AccId)))))) <-
  timesheet_data_available(D,Acc)          

timesheet_data_available(D,<CLAREAccount>) <-
  and(time_point_precedes(nonstrict,(<CLAREStartDate>,D),
      time_point_precedes(nonstrict,(D,<CLARELastRecordedDate>))

timesheet_data_available(D,<SLTAccount>) <-
  and(time_point_precedes(nonstrict,(<SLTStartDate>,D),
      time_point_precedes(nonstrict,(D,<SLTLastRecordedDate>))
...
\end{verbatim}
where \verb!<CLAREAccount>! is a ground term representing the 
CLARE account, and so on.
Here {\tt employee\_id}, {\tt date\_id} and {\tt account\_id} are ``naming
predicates'' which relate objects and their identifiers in the PRM database.
{\tt timesheet\_\-data\_\-available} is a predicate which
defines for each account the period for which timesheet records
are complete.

Most of the permissible assumptions that can be made are related to
the database-specific axioms, and are of one of two types.
The first type, a ``specialization'' (see section~\ref{Basic-LDT}), 
permits the assumption that arguments like ``employer'' in the
{\tt project} conceptual relation or ``payer'' in the {\tt transaction}
relation are filled by {\tt SRI} unless explicitly filled by something
else. We illustrate with the {\tt project} relation: the equivalence
that translates it into the associated database predicate is 
schematically of the form
\begin{verbatim}
(project(Org,Proj,Acc,StartDate,EndDate) <->
 ... <Right-hand side> ...) <-
  Org = organization(SRI)
\end{verbatim}
and there is an assumption declaration (see section~\ref{Basic-LDT})
\begin{verbatim}
assumable(Org = organization(SRI),
          0,
          all_projects_referred_to_are_at_SRI,
          specialization,
          project(Org,_,_,_,_))
\end{verbatim}
The content of the declaration is that it is permissible to assume
that a term $X$ is equal to {\tt organization(SRI)} if it occurs in a 
conjunctive context where $X$ is the first argument in an occurrence
of the {\tt project} relation. The atom 
{\tt all\_\-projects\_\-referred\_\-to\_\-are\_\-at\_\-SRI},
the ``justification'', is the
tag used to identify the assumption to the user.

The second type of assumption, a ``limitation'', is related to 
temporal incompleteness of database records: here, the declaration
is to the effect that it is permissible to assume that data is 
within the period for which records have been kept unless the
conjunctive context implies the contrary. We illustrate with
the {\tt timesheet\_\-data\_\-available} predicate above. There
is one assumption declaration for each account. The declaration for
the CLARE account is for example
\begin{verbatim}
assumable(
  timesheet_data_available(D,<CLAREAcc>)
  15,
  time_was_booked_after_<CLAREStartDate>_and_\\
                  before_<CLARELastRecordedDate>,
  limitation,
  book_time_to_project(_,_,_,D,<CLAREAcc>))
\end{verbatim}
Use of the declaration in cases where it is contradicted by
its context is prohibited by ``negated goal''
rules (see section~\ref{Basic-LDT})
\begin{verbatim}
neg(timesheet_data_available(D,<CLAREAcc>)) <-
  time_point_precedes(strict,D,<CLAREStartDate>)

neg(timesheet_data_available(D,<CLAREAcc>)) <-
  time_point_precedes(strict,<CLARELastRecordedDate>,D)
\end{verbatim}

The remaining sections in this chapter describe in more detail
the methodology we have developed for constructing LDTs 
for interfacing to relational databases. Much of the material
is taken directly from the corresponding chapter in the CLARE
software manual, though we have attempted where possible to
raise the level of abstraction by removing references to
essentially implementation-level constructs.

The basic procedure for construction of an LDT for a relational 
database is shown in Figure~\ref{domsteps}, which breaks it
down into a sequence of steps. More detailed descriptions of the
work involved in each step are given under the sections indicated 
in the figure.

\begin{figure}
\hrule
\begin{enumerate}

\item For each database relation, construct an associated 
``conceptual relation''. Declare each database and conceptual 
relation as such, and declare
the SQL column names associated with each database relation.
Declare any fields in conceptual relations which can only be filled
by one of a finite set of code values, together with a list of those
values. (Section~\ref{DB-rel-defs}).

\item For all of the conceptual and database relations in which some
fields are functions of other fields, write appropriate
\verb!function! declarations. (Section~\ref{DBfunctions}).

\item For each word sense that needs to be related to the database,
write the necessary equivalence (\verb!equiv!) meaning postulates.
(Section~\ref{Equivs}.) These equivalences, which can be thought of as
translation rules, are the heart of the interfacing process.
It will usually be necessary to add a number of Horn-clauses to
supplement the \verb!equiv! rules. (Section~\ref{DomPostulates}).

\item Declare any assumptions used in the equivalences with
\verb!assumable! declarations. (Section~\ref{Assumptions})

\end{enumerate}
\hrule
\caption{Steps for constructing a domain model for a database
interface}\label{domsteps}
\end{figure}

\section{Declaring database and conceptual relations}\label{DB-rel-defs}

The first step is to identify the tables used in the
actual \idx{SQL database}. Each table will correspond to two 
logical predicates: the {\it \idx{database relation}}, $R_{db}$, and the
associated {\it \idx{conceptual relation}}, $R_{conc}$. $R_{db}$ is a predicate which
exactly corresponds to the database table. The arguments of $R_{db}$ are 
{\it TRL \idx{database objects}}, which is to say 
TRL terms standing in direct correspondence to the actual
contents of the database tuple fields. 
(See also section~\ref{Ontology}).
Since the database relations and database objects are normally 
inconvenient to work with directly, the domain modelling procedure
requires the definition of an associated {\it \idx{conceptual
predicate}} $R_{conc}$ for each database predicate $R_{db}$; the connection
between language and database will always go through the 
conceptual predicates.
There are several important differences between $R_{db}$ and $R_{conc}$.
(The precise relationship between them is discussed later, in 
section~\ref{Equivs}, especially~\ref{Equiv-dbconnect}). 
The only one that we will consider for the moment is however that
$R_{conc}$ will normally have
more arguments than $R_{db}$. There are two reasons why it may be necessary
to add arguments to $R_{db}$. 

Firstly, it may be the case that the
conceptual relation corresponding to $R_{db}$ has one or more ``implicit''
arguments\index{implicit arguments}, 
which can be referred to in language, but which would 
always have the same value in the database. For example, the
{\tt TRANS} relation has an implicit ``payer'' argument, corresponding
to the agent who performs the transaction. Since the database only
contains payments made by SRI, there is no reason to include this as
a database column; however, language permits phrases like
``payments from SRI'', implying that the ``payer'' exists at least
at the level of conceptual modelling. 

The second reason for including extra arguments in $R_{conc}$ is related to
the standard database concept of ``primary key''. Most database
relations will have a primary key field: that is to say, for any
given constant, there is guaranteed to be at most one
record where the primary key has that value. In the case of the 
{\tt TRANS} relation, the {\tt transaction\_id} field is the natural
choice as the primary key. However, some relations will lack a 
primary key. For example, the {\tt SRI\_\-PROJECT\_\-MEMBER}
relation has two fields, {\tt employee\_\-name} and 
{\tt proj\_\-name}, neither of which can be primary; a project
normally has more than one member, and most people work on
more than one project. For a number of reasons, 
it is usually desirable for conceptual relations
to have a primary key argument; if there is none in the 
database relation $R_{db}$, it is normal to add one
$R_{conc}$. Thus the conceptual relation {\tt works\_\-on\_\-project}
(corresponding to {\tt SRI\_\-PROJECT\_\-MEMBER}) has an artificial
third argument to give it a primary key. This can be thought of as
the ``event'' of the person's working on the project.

\section{Declaring \idx{functional relationships}}\label{DBfunctions}

As described in the preceding section, 
database relationships usually tend to be functions on at least
one of their arguments (those that correspond to \idx{primary keys}); 
conceptual relations, 
should always have this property.
The translation mechanism is able to make use of information
about functional relationships
in several ways (see e.g.~\ref{Functional-relations}).
Functional relations are defined by declarations
of the form
\begin{verbatim}
function(<Template>,<From> -> <To>)
\end{verbatim}
This is to be read as stating that in 
\verb!<Template>!, an assignation of values to the variables making up
the list \verb!<From>! uniquely determines the values of the variables in
the list \verb!<To>!. 

Function declarations should be supplied both for database and
conceptual predicates: for example, the appropriate declarations 
for {\tt TRANS} and {\tt transaction} are
\begin{verbatim}
function(TRANS(Event,C,D,Payee,AC,AM),
         ([Event] -> [C,D,Payee,AC,AM]))
function(TRANS(Event,C,D,Payee,AC,AM),
         ([C] -> [Event,D,Payee,AC,AM]))

function(transaction(Agent,Event,C,D,Payee,AC,AM),
         ([Event] -> [Agent,C,D,Payee,AC,AM]))
function(transaction(Agent,Event,C,D,Payee,AC,AM),
         ([C] -> [Agent,Event,D,Payee,AC,AM]))
\end{verbatim}
Either the transaction ID or the cheque number could be used as 
a primary key, and hence there are two functional relationships.

\section{Equivalence rules}\label{Equivs}

The major part of the information needed to describe
the domain model is expressed using \idx{equivalence rules}. These come
in two forms: \idx{conditional equivalences} and \idx{unconditional
equivalences}. The abstract syntax we use 
for unconditional equivalences is
\begin{verbatim}
<LHS> <-> <RHS>
\end{verbatim}
For conditional equivalences, we write
\begin{verbatim}
(<LHS> <-> <RHS>) <- Conds
\end{verbatim}
Here \verb!<LHS>!, \verb!<LHS>! and \verb!Conds! are TRL expressions, which
may not use logical operators other than conjunction ({\tt and})
and existential quantification ({\tt exists}). The declarative 
semantics of equivalence rules are extensively discussed in 
chapter~\ref{translation}.

Writing equivalences is the non-trivial part of constructing
the domain model. In the following subsections, we will describe in
more detail how this is done. The main steps are summarized in
figure~\ref{equivsteps}. 

\begin{figure}
\hrule
\begin{enumerate}

\item For each conceptual predicate $R_{conc}$, collect together 
the word-senses and other predicates that are directly associated with
$R_{conc}$ (Section \ref{Equivs-basic}).

\item For each conceptual predicate $R_{conc}$,
write \verb!equiv! rules that relate the predicates associated
with $R_{conc}$ to the conceptual predicate. We have subdivided this
step into several sections (\ref{Equiv-nouns} -- \ref{Equiv-codes}) for
different types of predicate.

\item Collect predicates that map on to expressions which include
several database predicates, and write \verb!equiv! rules to
map them onto the corresponding conceptual predicates (\ref{Equiv-multiple}).

\item For each conceptual predicate $R_{conc}$, write \verb!equiv! 
rules that relate $R_{conc}$ to its associated database predicate.
(Section~\ref{Equiv-dbconnect}). This often involves conditions
which should be assumed rather than proved.

\item Write \verb!assumable! declarations for the ``assumable''
predicates introduced in the last step. (Section~\ref{Assumptions}).

\end{enumerate}
\hrule
\caption{Steps for writing equivalences}\label{equivsteps}
\end{figure}

Illustrative examples will be taken from the PRM application.
The conceptual predicates we will use, and their corresponding 
database relations, are the following:
\begin{itemize}
\item Transactions
\begin{verbatim}
transaction(Payer,TransId,ChequeId,Date,Payee,Acc,Amt)

TRANS(TransId,ChequeId,Date,Payee,Acc,Amt)
\end{verbatim}
\item Projects
\begin{verbatim}
project(Organization,ProjectName,ProjectNum,StartDate,EndDate)

SRI_PROJECT(ProjectName,ProjectNum,StartDate,EndDate)
\end{verbatim}
\item Payees (classified as men, women, companies or universities)
\begin{verbatim}
payee(Type,Payee)

PAYEE(Type,Payee)
\end{verbatim}
\item Employees
\begin{verbatim}
employee(Employer,Employee,Sex,HasCar),

SRI_EMPLOYEE(Employee,Sex,HasCar),
\end{verbatim}
\item Bookings of time to projects
\begin{verbatim}
booking_to_project(BookingEvent,Person,Time,Date,Project)

BOOKING_TO_PROJECT(Person,Time,Date,Project)
\end{verbatim}
\end{itemize}

We will assume that the LDT is built on top of the
general LDT, described in the section~\ref{Basicdomain}.
The general LDT supplies in particular axioms that translate
linguistic predicates having to do with {\it time} and {\it quantity}
into a more uniform representation. This is
explained in more detail in sections~\ref{Basicdomain} and
~\ref{Equiv-time}.

\subsection{Finding predicates associated with conceptual predicates}\label{Equivs-basic}

The first step is to collect, for each conceptual predicate $R_{conc}$,
the predicates occurring in untranslated TRL that are directly 
associated with it. It is difficult to give an exhaustive list of
the ways in which a linguistic predicate can be associated with $R_{conc}$;
the following are the most common cases. 

\begin{enumerate}
\item Nouns and adjectives describing objects that fill a single 
argument slot in $R_{conc}$.
For example, the word-sense predicates {\tt tran\-sac\-tion\_\-Ev\-ent/1} 
and {\tt payment\_Pay\-ing\-Event/1} describe
the object filling the {\tt TransId} argument slot in 
{\tt transaction}, 
while {\tt cheque\_Document/1} describes objects filling the {\tt ChequeId}
one. (See also section~\ref{Equiv-nouns}). The appropriateness of these
predicates can often be conditional on the value of another slot being
filled by a suitable code; for example, the predicate {\tt man\_MalePerson/1}
maps onto the {\tt Employee} field of {\tt employee} if 
the {\tt Sex} field is filled by an {\tt m}. This is explained further in
section~\ref{Equiv-codes}.

\item Prepositions describing the relationship between pairs of 
argument slots in $R_{conc}$, or between an argument slot in $R_{conc}$ and some
other type of object. For example, {\tt for} relates the {\tt ChequeId}
and {\tt Amount} slots in {\tt transaction} (one can say ``cheque
{\bf for} over 200 pounds''), while {\tt to} relates the {\tt TransId} and
{\tt Payee} slots (``payment {\bf to} Maxwells''). 
See also section~\ref{Equiv-prepositions}.
Note that many preposition senses, in particular temporal ones,
are translated by rules in the general LDT.

\item Verbs that describe the relationship between an event and one or
more argument slots in $R_{conc}$. (See also section~\ref{Equiv-verbs}).
Normally, the sense entry for a verb will
relate an event, the object denoted by the verb's subject, and zero or
more other arguments. For example, the predicate corresponding to
the ditransitive sense of ``pay'' (``SRI paid Maxwells 100 pounds'') is 
{\tt pay\_\-GiveTo/3}.
Here, the second argument maps into the {\tt Payer} slot,
the third argument (the object) maps onto the {\tt Amount} slot in
{\tt transaction}, and the fourth argument maps to the
{\tt Payee} slot. The first argument (the paying event),
is regarded as being equal to the transaction. It is in 
general a natural idea to map events onto primary keys, since
an event will often uniquely determine the objects participating in it.

\item Relational nouns that describe the relationship between pairs of 
argument slots in $R_{conc}$. These are similar to prepositions. For example,
the relational predicate deriving from the compound noun {\it project
number} is {\tt Project\_\-Number\_\-Account\_\-Number\_\-For\_\-ProjectOf}, and
this relates the {\tt ProjectName} and {\tt ProjectNum} slots in the
{\tt project} relation. See also section~\ref{Equiv-relational}.

\item Temporal relations between slots. These arise when there
is an argument slot in $R_{conc}$ that associates a time 
with the object filling another argument slot. The relation in question can
be {\tt associated\_\-time}, i.e. time of occurrence
(e.g. the relation between the {\tt TransId} and {\tt Date} slots in
{\tt transaction} relation); {\tt associated\_\-start\_\-time},
i.e. time of starting,
(e.g. the relation between the {\tt Project} and {\tt StartDate} slots in
{\tt project}); or {\tt associated\_\-end\_\-time}, i.e. time of
ending. See section~\ref{Equiv-time}.

\item The {\tt associated\_\-size} relation; this arises when one
slot in $R_{conc}$ describes the size or magnitude of another. For example,
the {\tt Amount} slot in {\tt trans} is the size of
the {\tt TransId} expressed as a quantity of {\tt pounds\_\-sterling}.
See section~\ref{Equiv-size}.

\end{enumerate}

\subsection{Common noun describing a conceptual slot}\label{Equiv-nouns}

In this and the following sections, we will describe how to
write the equivalences that translate the linguistic and primitive 
predicates into conceptual predicates. For convenience of
reference, we will divide up the material by different types of
linguistic predicate. 
We begin with the simplest case: a \idx{common noun} which describes a
conceptual slot. (We will use the term ``conceptual slot'' as a handy
abbreviation for ``argument of a conceptual predicate'').
In the following example, the rule says that an object
which fills the third argument place in the conceptual 
{\tt transaction} relation is a {\tt cheque\_\-Document}:
\begin{verbatim}
cheque_Document(Cheque) <->
exists([Payer,Trans,Date,Payee,Acc,Amt],
  transaction(Payer,Trans,Cheque,Date,Payee,Acc,Amt))
\end{verbatim}
Note that the variable {\tt Cheque} is implicitly universally quantified,
but the remaining variables are existentially quantified on
the RHS of the equivalence.

\subsection{Preposition\index{preposition} describing relationship between slots}
\label{Equiv-prepositions}

Equivalences that translate preposition senses are also straight-forward.
The following are the equivalences that
cover ``cheque for amount'' (``cheques for more than \pounds 200'')
and ``cheque to payee'' (``cheque to Maxwells''). 
\begin{verbatim}
and(for(Cheque,Amt),
    cheque_Document(Cheque)) <->
exists([Payer,Trans,Date,Payee,Acc,Amt],
  transaction(Payer,Trans,Cheque,Date,Payee,Acc,Amt))

and(to(Cheque,Payee),
    cheque_Document(Cheque)) <->
exists([Payer,Trans,Date,Payee,Acc,Amt],
  transaction(Payer,Trans,Cheque,Date,Payee,Acc,Amt)) 
\end{verbatim}
It can be helpful to think of the
first atom in the LHS in a rule of this kind (the ``head'') 
as a pattern to be matched; the remaining atoms on the 
LHS are conditions of applicability, while the RHS is the result of 
applying the rule. Thus in the first rule, the pattern to be matched is
{\tt for(Cheque, Amount)}, and the rule is applicable if the context
implies that {\tt Cheque} satisfies the predicate {\tt cheque\_\-Document}.

It may sometimes be necessary to use more than one condition, as for
example in the following rule, which would be used for an expression 
like ``payments on account 8468''. Here, one must specify both
that the first argument is a cheque, and that the second argument is
an account, thus:
\begin{verbatim}
and(on(Cheque,Acc),
and(cheque_Document(Cheque),
    account_ChargeNumber(Acc))) <->
exists([Payer,Trans,Date,Payee,Acc,Amt],
  transaction(Payer,Trans,Cheque,Date,Payee,Acc,Amt))
\end{verbatim}

\subsection{Verb describing relationship between slots}
\label{Equiv-verbs}

Rules for translating \idx{verb} senses are similar to those for translating
preposition senses. The following, for example, is the
rule that covers ``pay money to'', as in ``SRI paid \pounds 120 to Maxwells''.
\begin{verbatim}
and(pay_GiveTo(Trans,Payer,Amt,Payee),
    money_AbstractCredit(Amt)) <->
exists([Payer,Date,Acc],
  transaction(Payer,Trans,Cheque,Date,Payee,Acc,Amt))
\end{verbatim}
Note that the ``paying event'' is regarded as being the same as the
transaction, and that the agent of the paying event becomes the 
{\tt Payer} in the conceptual {\tt transaction} relation.

Support verb\index{support verb} constructions like ``make a payment''
can be treated in the same way.
The next example shows the relevant equivalence:
\begin{verbatim}
and(make_Commit(MakeEvent,Payer,Trans),
    payment_PayingEvent(Trans)) <->
exists([Cheque,Date,Payee,Acc,Amt],
  and(transaction(Payer,Trans,Cheque,Date,Payee,Acc,Amt)
      MakeEvent = Trans))
\end{verbatim}
Here the ``making'' event is once again regarded as being the same
as the payment. In cases like this, where two variables from the LHS
({\tt MakeEvent} and {\tt Trans}) are being mapped onto a single event
on the RHS, it is important to write the rule in the way shown, with
an equality on the right. It is easy to show that the following variant 
\begin{verbatim}
and(make_Commit(Trans,Payer,Trans),
    payment_PayingEvent(Trans)) <->
exists([Cheque,Date,Payee,Acc,Amt],
  transaction(Payer,Trans,Cheque,Date,Payee,Acc,Amt))
\end{verbatim}
is not equivalent. The easiest way to convince oneself of the truth
of this statement (which may seem counter-intuitive to readers used
to Horn-clauses) is to consider the pair of formulas
\begin{verbatim}
p(X,Y) <-> and(q(X),X = Y)
\end{verbatim}
and
\begin{verbatim}
p(X,X) <-> q(X)
\end{verbatim}
The first of them means that {\tt p(a,b)} implies {\tt q(a)}, but 
not the second.

\subsection{Relational nouns}\label{Equiv-relational}

Rules for translating \idx{relational noun} senses are of a form similar
to those for prepositions and transitive verbs. 
For example, the rule for the relational sense of the compound noun
{\it project number} (e.g. ``CLAM-BAKE's project number'') is
\begin{verbatim}
project_number_AccountNoForProjectOf(Number,Project),
exists([Org,Start,End,Account],
  and(project(Org,Project,Account,Start,End),
      numbered_object(Account,account_ChargeNumber,Number)))
\end{verbatim}
Note the use of the predicate {\tt numbered\_\-object} on the RHS.
This expresses the fact that {\tt Account} is the numbered object
of type {\tt account\_\-ChargeNumber} with number {\tt Number}, i.e.
the object referred to by the English expression ``account {\tt Number}''.
See also section~\ref{Equiv-dbconnect}.

\subsection{Primitive predicates}
\label{Equiv-time}\label{Equiv-size}

If a relation has a field that can be interpreted as associating a date or time
with the tuple, it is possible to use temporal expressions, involving
words like ``when'', ``during'', ``after'', ``start'' and ``end''. 
Equivalences in the general LDT (see section~\ref{Basicdomain}) 
should translate the linguistic predicates deriving from expressions
like these into forms which use one of the ``primitive predicates''
\ttidx{associated\_\-time}, \ttidx{associated\_\-start\_\-time} and 
\ttidx{associated\_\-end\_\-time}. Each of these is a three-place predicate, 
whose arguments are, in order
\begin{itemize}
\item the object (which will often be an event)
\item the associated date or time 
\item the ``granularity'' of the date or time, which at present must be either 
{\tt days} or {\tt seconds}
\end{itemize}
Similarly,
some relations may have a field that can be interpreted as denoting
the ``size'' or ``magnitude'' of one of the objects, and can be used
to assign a meaning to expressions that compare objects with reference
to their extent; expression of this form
include comparatives (``larger'', ``smaller'')
superlatives (``largest'', ``smallest''), and some prepositions like
``over'' (``over \pounds 250''). All of these are
translated by equivalences in the general LDT into a form expressed in
terms of the primitive predicate {\tt associated\_\-size}.

Domain-dependent equivalences must then be provided to translate 
the primitive predicates in the contexts provided by
the domain word-senses: the RHS of these equivalences will
usually contain conceptual predicates.
The following rule, for example,
translates the {\tt associated\_\-time} predicate in a context 
where its first argument is a transaction; it
states that the fourth field of a {\tt transaction} relation is 
the time (at the level of granularity of days)
associated with the transaction. 
\begin{verbatim}
and(associated_time(Trans,Date,Granularity),
    transaction_Event(Trans)) <->
exists([Payer,Cheque,Payee,Acc,Amt],
  and(transaction(Payer,Trans,Cheque,Date,Payee,Acc,Amt),
      Granularity = days))
\end{verbatim}
Similarly, some relations, representing events that occur over a period of
time, may contain fields denoting the start and end of this period. 
The following pair of rules state that the start and end dates of
a project are encoded in the fourth and fifth fields of the
{\tt project} relationship respectively.
\begin{verbatim}
and(associated_start_time(Project,Start_date,Granularity),
    project_Activity(Project)) <->
exists([Org,ProjNum,End_date],
  and(project(Org,Project,ProjNum,Start_date,End_date),
      Granularity = days))

and(associated_end_time(Project,End_date,Granularity),
    project_Activity(Project)) <->
exists([Org,ProjNum,Start_date],
  and(project(Org,Project,ProjNum,Start_date,End_date),
      Granularity = days))
\end{verbatim}
The \ttidx{associated\_\-size} 
primitive predicate takes two arguments: the first is
the object in question, and the second is its ``size'', which should be
a number. Finding a suitable number will often involve including an
equality on the RHS, which extracts the number from a term representing
an amount. In the 
following example, which defines the ``size'' of a cheque, 
the extraction is performed using the 
{\tt amount\_object} predicate, defined in the general LDT.
\begin{verbatim}
and(associated_size(Cheque,N),
    cheque_Document(Cheque)) <->
exists([Payer,Trans,Date,Payee,Acc,Amt],
  and(transaction(Payer,Trans,Cheque,Date,Payee,Acc,Amt),
      amount_object(AM,sterling_unit,N)))
\end{verbatim}

\subsection{Expressions that map onto \idx{code} values}\label{Equiv-codes}

The filler of a field in a database relationship is often a code, and
an English expression will map onto a value, or set of values, for this
code. Thus for example it might easily be the case that a word like
``man'' or ``woman'' ultimately maps onto a field being filled by an
{\tt m} or a {\tt w}. Two examples follow. The first states that
``man'' when applied to an object known to be a payee, maps onto
the first field of the {\tt payee} relation being filled by an {\tt m}.
The second states that an overhead payment is a transaction where
the ``account'' field is filled by a numbered object of type ``account'' 
whose number satisfies the predicate {\tt overhead\_\-code}.
\begin{verbatim}
and(man_MalePerson(Payee),
    payee_Recipient(Payee)) <->
payee(m,Payee)

overhead_payment_PaymentMadeOnOverhead(Trans) <->
exists([Payer,Cheque,Date,Payee,Acc,Amt,Num],
  and(transaction(Payer,Trans,Cheque,Date,Payee,Acc,Amt),
  and(numbered_object(Account,account_ChargeNumber,Num),
      overhead_code(Num))))
\end{verbatim}
A further definition is then needed to specify the range of permitted
values for overhead codes: here, they are defined to be exactly those
numbers which are members of the set {\tt \{775, 675\}}:
\begin{verbatim}
overhead_code(Num) <-> one_of(Num,[775,675])
\end{verbatim}
An important special case is that of the code which indicates 
existence or non-existence of an object related to
the filler of some other field in the relation. (See also 
section~\ref{Existential}). In the PRM database,
an example is provided by the {\tt employee} relation, whose third
argument is set to {\tt y} if the employee in question has a company car.
The following two rules (one for ``have'' and one for ``car'')
encode this fact.
\begin{verbatim}
and(employee1(Empl),
and(car1(Car),
    have1(Event,Empl,Car))) <->
employee_has_car(Event,Empl,Car)         

exists([Event,Car],
  employee_has_car(Event,Empl,Car)) <->
exists([Sex]
  SRI_EMPLOYEE(Empl,Sex,y))         
\end{verbatim}

\subsection{``Display'' predicates}
\label{Equiv-display}

The general LDT (see section~\ref{Basicdomain}) contains equivalences
that translate words like ``show'' and ``list'' into a uniform representation
using the predicates {\tt executable\_\-action/6} and 
{\tt display\_\-format/2}.
The second of these is a predicate which determines the appearance of the
term printed to represent a database object: the intention is that
{\tt \ttidx{display\_format}(Object,Format)} should hold if {\tt Format} is 
either a conceptual object, or a list containing one 
or more conceptual objects that can be used together as a representation
of {\tt Object}. For example, in the PRM domain it is sensible to print
out a transaction as a list consisting of its transaction ID, the date
of the payment, the payee, and the amount. This is captured in the
following equivalence,
\begin{verbatim}
and(display_format(Trans,Format),
    transaction_Event(Trans)),
exists([Id,Payer,C,Date,Payee,AC,Amount],
  and(transaction(Payer,Trans,C,Date,Payee,AC,Amount),
  and(numbered_object(Trans,transaction_Event,Id),
      Format = [Id,Date,Payee,Amount])))
\end{verbatim}
Normally, the domain implementor should provide one {\tt display\_\-format}
definition for each type of object. Many objects will be ``named objects''
of some type (see section~\ref{Equiv-dbconnect}),
whose names serve as adequate descriptions. 
This is for example the case for
sponsors in the PRM domain. If this is so, an equivalence of the 
following kind is sufficient:
\begin{verbatim}
(display_format(Sponsor,Format) <->
 named_object(Sponsor,sponsor_ContributorOfFunding,Format)) <-
  sponsor_ContributorOfFunding(Sponsor).
\end{verbatim}

\subsection{Mapping on to combinations of conceptual predicates}
\label{Equiv-multiple}

Sometimes, a linguistic predicate will relate slots taken from
more than one conceptual predicate. {\tt Equiv} rules of this
type are a little more complex than those which map onto a single 
conceptual predicate, which is why we have left this case for 
last; there are a couple of coding tricks that can be used,
however, to simplify the structure of the rules.

We illustrate with the example of the preposition {\it on}, used 
to relate a payment to a project (e.g. ``Which payments on
CLAM-BAKE were over \pounds 1000?''). The rule must in effect join
the {\tt TRANS} and {\tt SRI\_\-PROJECT} relations
through the {\tt Account} field. The simplest way to do so is
to write the following equivalence:
\begin{verbatim}
and(on(Trans,Account),
and(payment_PayingEvent(Trans),
    project_Activity(Project))) <->
exists([Org,Acc,StartDate,EndDate,
        Payer,Cheque,Date,Payee,Amt],
  and(project(Org,Project,Acc,StartDate,EndDate),
      transaction(Payer,Trans,Cheque,Date,Payee,Acc,Amt)))
\end{verbatim}
The rule as presented, is, however,
uneconomical, since it fails to exploit the general fact
that ``on a project'' means generally ``on that project's account'';
what we have here is actually a simple case of the linguistic phenomenon of
{\it metonymy}. The practical consequence is that a second equivalence
will have to be added to deal with the expression ``cheque on project''.
A better solution is to write a conditional equivalence which makes
explicit the general relation between ``on project'' and ``on account'',
thus:
\begin{verbatim}
(on(X,Project) <-> on(X,Account)) <-
  project(Org,Project,Account,Start,End).
\end{verbatim}

\subsection{Translating conceptual into database predicates}
\label{Equiv-dbconnect}\index{conceptual predicates}\index{
database predicates}

The final step when writing the equivalences is to add 
the rules which translate between 
the conceptual and database predicates. This is
intimately connected with the assumption mechanism, and it
will be helpful to review the ideas involved.

There are three
main reasons for distinguishing between 
``conceptual predicates'' and 
``data-base predicates''. The first has already been discussed
in section~\ref{DB-rel-defs}, 
and arises from the ``extra arguments'' in the conceptual
predicate, like the {\tt Organization} argument in {\tt project},
or the {\tt Payer} argument in {\tt transaction}. The fact we wish
to express here is that these arguments must have a single, fixed
value, in order for the predicate to be capable of being translated.
In the two examples quoted, the appropriate value is {\tt SRI}. 
If SRI is mentioned explicitly in the query, (because some expression
like ``project at SRI'' or ``SRI payment'' was used),
the slots in question will 
already be filled by the appropriate value. More frequently, though,
the value of the extra arguments will be left unspecified, and the
value of the filler should simply be assumed to be SRI. The intended
behaviour, then, is that an argument like {\tt Organization} or 
{\tt Payer} should be assumed to be equal to the appropriate value
unless this is explicitly contradicted by the information in the 
original query.

The second, related, point is the need to be able to deal with cases
where information in the database is systematically incomplete
(see also section~\ref{Closed-World}); the
most important case is that of a relation which lists records kept
over a specified period. For example, the {\tt TRANS}
relation only holds information about transactions carried out
over the 18-month period starting in August 1989. Questions about payments
carried out in 1988 or earlier can thus not be answered, and the desired
behaviour is for query translation to fail with an appropriate warning.
If the query makes no mention of time (e.g. ``List all payments over
\pounds 200  made on the WHIZ project''), then it is reasonable to
assume that the user meant only to refer to payments within the period
covered, as long as a message is printed to inform her that the 
assumption is being made. If the query explicitly refers to a time-span
within the one for which records exist, translation is unconditionally
valid.

The third point is the necessity of making a clear distinction between
the types of object occurring respectively as arguments to conceptual
and database predicates. (See also section~\ref{Ontology}).
The intention is that the arguments to the database
predicates should denote the {\it actual contents of the database records},
that is to say numbers, dates and strings.
The arguments to the conceptual predicates,
on the other hand, are intended to denote the {\it real-world objects} that 
the database records refer to. 
It may at first glance seem that making a
distinction between these two types of object is a needless form of
hair-splitting, but it is in fact primarily motivated by practical 
considerations. If the distinction is not made, problems can arise: 
in particular,
it may easily be the case that the same database object, used in different
database fields, can refer to different
objects. For example, in the PRM domain there are some numbers which are 
both cheque and transaction IDs; however, the fact that a cheque ID
happens to be identical with a transaction ID does {\it not} imply
that the cheque and transaction are identical, or even that they
are related. 
Because the problem with non-unique identifiers
is so common and important, CLARE
provides a special mechanism for dealing with it. Terms of the
form 
\begin{quote}
\begin{verbatim}
SF(c(x^[<PredicateName>,x],ID))
\end{verbatim}
\end{quote}
are treated by the system
as referring to ``the object of type \verb!<PredicateName>!, with
identifier \verb!ID!''. (An alternate notation, which we sometimes
use as an abbreviation, is
{\tt <PredicateName>\#ID}). Thus for example 
\verb!SF(c(x^[transaction_Event,x],123))! is the object that would
be referred to by the English noun-phrase {\it transaction 123}.
There is a predicate in the general LDT, {\tt named\_\-object/3},
which can be used to refer to terms of this type; thus for example
\begin{quote}
\begin{verbatim}
named_object(Trans,transaction_Event,123)
\end{verbatim}
\end{quote}
is equivalent to
\begin{quote}
\begin{verbatim}
Trans = SF(c(x^[transaction_Event,x],123))
\end{verbatim}
\end{quote}

All three types of functionality can be achieved in a uniform way,
by suitably defining the respective truth-conditions of the 
conceptual and database predicates. The key idea can be summarized
as follows:
\begin{itemize}
\item the conceptual predicate holds of the conceptual objects
if the event in question has happened
\item the database predicate holds of the corresponding database objects
if the event is recorded in the database.
\end{itemize}
Since the types of argument to the two predicates are in general
distinct, the equivalences
used to define the relationship between the conceptual and
database predicates will include a number of occurrences of
{\it conversion predicates}, which relate database and conceptual
objects. There are at present four such predicates in the
general LDT: {\tt named\_\-object/3} (defined above),
{\tt sql\_\-numbered\_\-object/3}, {\tt sql\_\-number\_\-convert/2},
and {\tt sql\_\-date\_\-convert/2}. The last three are defined as follows:
\begin{description}

\item[{\tt sql\_numbered\_object(Obj,Type,N)}] {\tt Obj} is a named 
object of type {\tt Type} whose identifier is a number, and {\tt N}
is the database representation of that number. 
This is the predicate used to relate a number field in a database
record with the object that number refers to, in cases where the number
is being used as an identifier.

\item[{\tt sql\_number\_convert(Num,DBNum)}] Converts between numbers
and database number objects (in the current implementation, Prolog atoms
whose print-names are strings representing numbers).

\item[{\tt sql\_date\_convert(Date,DBDate)}] Converts between dates
and database date objects (in the current implementation,
Prolog atoms whose print-names
are strings representing dates in SQL format).

\end{description}

Looking at a concrete example should make these ideas clearer. 
The database relation {\tt TRANS/6}
holds if an appropriate SRI transaction is recorded in the
database. The corresponding conceptual predicate,
{\tt transaction/7}, holds if an appropriate transaction
(not necessarily made by SRI) occurred. The predicates
agree if two conditions are fulfilled: firstly, that the
transaction was made by SRI, and secondly, that it was
made within the recorded period.
This can be expressed
(removing a few details irrelevant to the present discussion)
using the following {\tt equiv} rule:
\begin{verbatim}
(and(transaction(Payer,Trans,Cheque,Date,Payee,Acc,Amt),
     transaction_data_available(Date)),
 exists([Id,CNum,AC,AM,AMNum,DBDate],
   and('TRANS'(Id,CNum,DBDate,Payee,AC,AM),
   and(sql_numbered_object(Trans,transaction_Event,Id),
   and(sql_numbered_object(Cheque,cheque_Document,CNum),
   and(sql_date_convert(Date,DBDate),
   and(sql_numbered_object(Account,account_ChargeNumber,AC),
   and(amount_object(Amount,sterling_unit,AMNum),
       sql_number_convert(AMNum,AM))))))))) <-
  Payer = sri.
\end{verbatim}
This rule contains four kinds of predicate. Firstly, there are
the conceptual predicate ({\tt transaction(...)}), and the database
predicate ({\tt TRANS(...)}) that it is being mapped into.
Secondly, there are the ``conversion predicates'' 
({\tt sql\_\-numbered\_\-object} etc.);
as can be seen, they are used to link the arguments to the 
conceptual predicate ({\tt Trans}, {\tt Cheque}, {\tt Date} etc.)
with the corresponding arguments to the database predicate
({\tt Id}, {\tt CNum}, {\tt DBDate} etc.) Finally, there are the 
predicates {\tt =/2} and {\tt transaction\_\-data\_\-available/1},
which are used to express the contingent relationship between
{\tt transaction} and {\tt TRANS}: namely, that the {\tt Payer}
has to be {\tt SRI}, and that the transaction has to have been 
made during the specified period.

Here {\tt transaction\_data\_available/1} is a predicate supplied
by the user, which holds if its argument is a date between 
1/8/1989 and 31/3/1991. This is expressed by the Horn-clause
\begin{verbatim}
transaction_data_available(Date) <-
  time__precedes(nonstrict,date([1989,8,1]),Date),  
  time_precedes(nonstrict,Date,date([1991,3,31]))
\end{verbatim}
It is also necessary to add {\tt assumable} declarations
for both {\tt =/2} and
{\tt trans\-action\_\-data\_\-avail\-able/1}.
This is explained in the next section.

\section{Assumptions}\label{Assumptions}

The concept of {\it \idx{assumptions}} has already been referred to 
several times; this section, will consider the subject in more detail.

It will frequently turn out that the equivalences,
and in particular the equivalences which translate conceptual
predicates into database predicates, are not strict,
but rather depend on tacit assumptions related to the nature of
the domain. For example, in the PRM domain it is normally reasonable
to assume that the cheques, payments, projects and so on referred to
in queries are all related in direct way to SRI, and that queries about
all records refer to the time-periods for which they are available,
unless this is explicitly contradicted by the query.

The implementor indicates
which assumptions are permissible by including declarations of the form
\begin{verbatim}
assumable(<Goal>,<Cost>,<Justification>,<Type>,<Condition>)
\end{verbatim}
This states that the goal \verb!<Goal>! may be assumed, rather than being
proved, at a cost \verb!<Cost>! which should be a non-negative number,
for the reason \verb!<Justification>!, if the \verb!<Condition>! can be
inferred from the context in which the goal is being proved.
\verb!<Type>! should be one of the atoms {\tt specialization}, 
{\tt limitation} or {\tt approximation}. The choice of assumption-type
is related to the production of possible warning-messages if that assumption
is used: the implementor should decide whether it is most appropriate 
to 
\begin{itemize}
\item produce no warning message ({\tt specialization})
\item warn that the answer may depend on incomplete knowledge ({\tt limitation})
\item warn that the answer may depend on an incorrect assumption ({\tt approximation}).
\end{itemize}
The total cost of a proof is defined to be the number of Horn-clauses
used, plus the sum of the costs of all assumed rules; the 
search through the proof space is carried out using an iterated-deepening 
algorithm. By making the cost of an
assumption sufficiently large, it is thus possible to ensure that proofs 
which do not use it are preferred to ones that do.

To give a better idea of how to write {\tt assumable}
declarations, we will continue the example started in the previous 
section. Recall that we have the equivalence
\begin{verbatim}
(and(transaction(Payer,Trans,Cheque,Date,Payee,Acc,Amt),
     transaction_data_available(Date)),
 exists([Id,CNum,AC,AM,AMNum,DBDate],
   and('TRANS'(Id,CNum,DBDate,Payee,AC,AM),
   and(sql_numbered_object(Trans,transaction_Event,Id),
   and(sql_numbered_object(Cheque,cheque_Document,CNum),
   and(sql_date_convert(Date,DBDate),
   and(sql_numbered_object(Account,account_ChargeNumber,AC),
   and(amount_object(Amount,sterling_unit,AMNum),
       sql_number_convert(AMNum,AM))))))))) <-
  Payer = sri.
\end{verbatim}
linking the conceptual relation {\tt transaction} and the
database relation {\tt TRANS}. We want 
{\tt Payer = SRI} and {\tt trans\-action\_\-data\_\-avail\-able(Date)}
to be assumed in this context, as long as this is not explicitly
inconsistent with the query. The relevant {\tt assumable}
statements are
\begin{verbatim}
assumable(
      Payer = SRI,
      0,
      all_transactions_referred_to_are_from_SRI,
      specialization,
      transaction(Payer,_,_,_,_,_,_))

assumable(
      transaction_data_available(Date),
      15,
      transactions_referred_to_made_between_1988/8/17_\\
             and_1991/4/1,
      limitation,
      transaction(_,_,_,Date,_,_,_))
\end{verbatim}
In both cases, the condition for the validity of the rule is to be
inferred from the context, in particular from the LHS of the equivalence.
The zero cost on the first rule
reflects the intuition that it is normally permissible
to assume without further reasoning that payments are made by SRI.
However, the second rule costs 15 units to apply, since it is quite
frequently possible to infer from other conjuncts that the payment
actually is within the known range. Since it is preferable to
prove conditions rather than assume them, a charge is made for
the assumption. The ``justifications'' are in both rules the 
atoms that will be used as tags when printing out the lists 
of assumptions made during translation.

The system saves each assumption made during the proof 
together with its context of use. After completing the translation process,
an attempt is made to prove the {\it negation} of each assumption
from its associated conditions. If this can be done, query
processing is terminated, and the user is informed that 
the use of the relevant assumption is inappropriate in the given
context. The prover does not provide full support for \idx{negation};
it is however sufficient to provide 
Horn-clause meaning postulates of the form 
\begin{verbatim}
neg(<Goal>) <- <Conditions>
\end{verbatim}
which will allow normal Horn-clause proofs of goals of the form
{\tt neg(G)}. Since this makes {\tt neg} a one-place
predicate, it is important to realize that the prover has no way of
relating proofs of {\tt neg(G)} to proofs of {\tt G}.
Lack of this capability does not seem, however, to be a serious problem
for the simple proofs carried out in this context.

The rules to define the negation of {\tt =/2} and 
{\tt time\_\-precedes/3} are supplied in the 
general LDT; those for other 
assumed predicates may be added by the user. For example, the
appropriate definitions for {\tt transaction\_\-data\_\-available/1} are
\begin{verbatim}
neg(transaction_data_available(D)) <-
  neg(time_precedes(nonstrict,date([1989,8,17]),D))
negation_of(transaction_data_available(D)) <-
  neg(time_precedes(nonstrict,D,date([1991,3,31])))
\end{verbatim}

\section{Horn Clause axioms} \label{DomPostulates}

Typically a number of non-equivalential meaning postulates are 
also required for the domain.
These are represented as (interpreted) \idx{Horn clauses}.
We have already seen some examples of Horn-clause axioms in the
preceding section, used to define conditions under which 
conceptual and database predicates correspond. There are also
some more straight-forward cases where
predicates may not be definable by means of
equivalences, and where a Horn-clause definition may consequently
be appropriate. The most common example results from the
predicate {\tt personal/1}, which holds of entities that can
be referred to by the interrogative pronoun ``who''. 
In a normal context like {\it Who works on CLAM-BAKE?} or
{\it Who booked time to project 8744 this month?}, no information
is contributed by the occurrence of {\tt personal}; it is implicit
that entities who can work on projects, or book time to projects,
can be referred to as ``who''. The appropriate way to declare this
is by including the Horn-clauses
\begin{verbatim}
personal(Person) <-
  works_on_project(Event,Person,Project).

personal(Person) <-
  booking_to_project(Event,Person,Time,Date,Account).
\end{verbatim}
The translation process will be able to use these to deduce that
the occurrences of {\tt personal} are implied by their contexts,
and can safely be thrown away. In general, one such Horn-clause 
will be needed for every argument position in a conceptual 
predicate holding an object that {\tt personal} can hold of.

\section{AET with a domain theory: a full example}\label{Examples}

In this final section, we present a detailed example showing
how CLARE uses the PRM linguistic domain theory
to find an effective translation of
the TRL representation of sentence (S1) (repeated here for convenience).
\begin{description}
\item[(S1)] List all payments made to BT during 1990.
\end{description}
We will take a few liberties with the exact representation of 
the domain axioms in the interests of increased readability.
In particular we have suppressed the complications caused by the
treatment of granularity, and removed some arguments from
predicates when they were irrelevant to the example. To make
the formulas more compact, we also introduce two small notational
abbreviations. We write {\tt x} instead of {\tt exists} to represent
existential quantification, and we use the notation 
\begin{verbatim}
<Predicate>#<Id>
\end{verbatim}
to mean ``the object of type {\tt <Predicate>} with identifier {\tt <Id>}''.
We also shorten the names of the linguistic predicates in an obvious way;
thus for example we write {\tt payment1} instead of 
{\tt payment\_\-Paying\-Event}.

The original TRL representation of (S1) is
\begin{verbatim}
forall([Trans],
       impl(x([Agnt,Ev],
              and(payment1(Trans),
                  and(make2(Ev,Agnt,Trans,payee1#bt),
                      during1(Ev,interval(date([1990,1,1]),
                                          date([1990,12,31])))))),
            x([ShowEv],
              and(show1(ShowEv,clare,Trans),
                  before1(<Now>,ShowEv)))))
\end{verbatim}
(We write \verb!<Now>! to stand for the
TRL representation of the present moment).
The formula can be glossed as ``For all {\tt Trans} such that {\tt Trans}
is a payment made by some {\tt Agent} during the interval from
1/1/90 to 31/12/90, it is the case that CLARE will show {\tt Trans} at some
time after the present time''. 

The desired effective translation will be expressed in terms of
three evaluable predicates. The database information will be 
accessed by the predicate
{\tt TRANS(Trans\-Id,DB\-Date,Payee\-Name,Amount)}, which 
represents a database of SRI payments. Dates will be manipulated by the
arithmetic predicate {\tt t\_precedes}, which holds if its
arguments are two representations of calendar
dates such that the first is temporally before the second.
Execution of actions will be represented by the 
predicate 
\begin{verbatim}
execute(Event,Action,Agent,Time)
\end{verbatim}
The actions we are interested in here are displaying actions carried out
by CLARE: these will be represented by letting {\tt Agent} be equal to 
{\tt clare} and {\tt Action} be of the form {\tt display(FormatList)}
where {\tt FormatList} is a list of objects to be displayed.

The first step is to use equivalences which translate the linguistic
predicates \verb!payment1!, \verb!make2!, {\tt during1}, {\tt show1}
and {\tt before1} into conceptual predicates. We examine these in
turn. 
\begin{verbatim}
payment1(Trans*)
\end{verbatim}
is translated to
\begin{verbatim}
x([Payer,Date,Payee,Amt],
  transaction(Payer,Trans*,Date,Payee,Amt))
\end{verbatim}
using the domain-specific equivalence
\begin{verbatim}
transaction1(Trans) <->
x([Payer,Date,Payee,Amt],
  transaction(Payer,Trans,Date,Payee,Amt))).
\end{verbatim}
Then 
\begin{verbatim} 
make2(Ev*,Agnt*,Trans*,payee1#bt)
\end{verbatim}
is translated into
\begin{verbatim}
x([Date1,Amt1],
  and(transaction(Agnt*,Ev*,Date1,payee1#bt,Amt1),
      Ev*=Trans*))
\end{verbatim}
using the rule
\begin{verbatim}
and(make2(Event,Payer,Payment,Payee),
    transaction1(Payment)) <->
x([Date,Amt],
  and(transaction(Payer,Event,Date,Payee,Amt),
      (Event = Payment))
\end{verbatim}
The next rule to be applied 
is the one that translates \verb!during1!. The intended semantics of
{\tt during1(E1,E2)} are
``\verb!E1! and \verb!E2! are events, and the time
associated with \verb!E1! is inside that 
associated with \verb!E2!''. The relevant equivalences are
\begin{verbatim}
during1(E1,E2) <->
x([T1,T2],
  and(associated_time(E1,T1),
  and(associated_time(E2,T2),
      time_during(T1,T2))))
\end{verbatim}
composed with the equivalence 
\begin{verbatim}
and(associated_time(Trans,Date),
    transaction1(Trans)) <->
x([Payer,Payee,Amt],
  transaction(Payer,Trans,Date,Payee,Amt))
\end{verbatim}
which says that the {\tt associated\_\-time} of a {\tt transaction} is its 
third argument, and with a third equivalence from the general LDT which 
says that the time associated with an explicit
interval is the interval itself. Applying all three of these in turn,
\begin{verbatim}
during1(Ev*,interval(date([1990,1,1]),date([1990,12,31])))
\end{verbatim}
is translated into
\begin{verbatim}
x([Payer2,Date2,Payee2,Amt2],
  and(transaction(Payer2,Ev*,Date2,Payee2,Amt2),
      time_during(Date2,interval(date([1990,1,1]),date([1990,12,31])))))
\end{verbatim} 
After this,
\begin{verbatim}
show1(Agnt*,clare,Trans*)
\end{verbatim}
is translated by following two equivalences. The first,
\begin{verbatim}
show1(E,Agent,X) <->
x([DisplayT,Format],
  and(execute(E,display(Format),Agent,DisplayT),
      display_format(X,Format)))
\end{verbatim}
which is in the general LDT, can be glossed ``A showing of {\tt X} by
{\tt Agent} is the execution of a displaying of {\tt Format}, where
{\tt Format} is the display-format of {\tt X}''. The second,
\begin{verbatim}
and(display_format(Trans,Format),
    transaction1(Trans)) <->
x([TransId,PayeeId,Payer,Date,Payee,Amt],
  and(transaction(Payer,Trans,Date,Payee,Amt),
  and(Trans = transaction#TransId,
  and(Payee = payee#PayeeId,
      Format = [TransId,Date,PayeeId,Amt]))))
\end{verbatim}
(see section~\ref{Equiv-display}) is a context-sensitive
domain-dependent
equivalence which defines the display-format for a transaction to
be a four-element list consisting of the transaction ID, the transaction
date, the payee's name, and the amount. The result of applying both
of these is
\begin{verbatim}
x([Id,Date3,Payee3,Amt3,DisplayT],
  and(execute(Agnt*,display([Id,Date3,Payee3,Amt3]),clare,DisplayT3),
      and(transaction(Payer3,Trans*,Date3,Payee3,Amt3),
          Trans*=transaction1#Id)))))
\end{verbatim}
Finally,
\begin{verbatim}
before1(<Now>,ShowEv*)
\end{verbatim}
is translated into
\begin{verbatim} 
x([Format1,DisplayAgent,DisplayT]
  and(execute(ShowEv*,Format1,DisplayAgent,DisplayT),
      time_before(strict,<Now>,DisplayT)))
\end{verbatim}
using rules similar to those used above to translate {\tt during1}.
When all the rules mentioned so far
have been applied, the translated form is
\begin{verbatim}
forall([Payer,Trans,Date,Payee,Amt,Payer1,Date1,Amt1,Payer2,
        Date2,Payee2,Amt2],
       impl(and(transaction(Payer,Trans,Date,Payee,Amt),
                and(transaction(Payer1,Trans,Date1,payee1#bt,Amt1)
                    and(transaction(Payer2,Trans,Date2,Payee2,Amt2)
                        time_during(Date2,
                                     interval(date([1990,1,1]),
                                              date([1990,12,31])))))),
            x([Id,Date3,Payee3,Amt3,Format1,DisplayAgent,DisplayT3,
               DisplayT],
              and(execute(DisplayEv,
                          display([Id,Date3,Payee3,Amt3]),
                          clare,
                          DisplayT1),
                  and(execute(DisplayEv,Format1,DisplayAgent,DisplayT),
                      and(time_before(<Now>,
                                       DisplayT),
                          and(transaction(Payer3,Trans,Date3,
                                          Payee3,Amt3),
                              Trans=transaction1#Id)))))))
\end{verbatim}        
Exploiting the fact that {\tt transaction} is functional on its
second argument, and {\tt execute} on its first,
this reduces after simplification involving functional merging
(see section~\ref{Functional-relations}) to
\begin{verbatim}
forall([Payer,Trans,Date,Payee,Amt],
       impl(and(transaction(Payer,Trans,Date,payee1#bt,Amt),
                time_during(Date,
                             interval(date([1990,1,1]),
                                      date([1990,12,31])))),
            x([Id,DisplayT],
              and(Trans=transaction1#Id,
                  and(execute(DisplayEv,
                              display([Id,DisplayT,payee1#bt,Amt]),
                              clare,
                              DisplayT),
                      time_before(<Now>,
                                   DisplayT))))))
\end{verbatim}
Applying rules from the general LDT to translate {\tt time\_before}
and {\tt time\_during} into {\tt t\_precedes}, we get
\begin{verbatim}
forall([Payer,Trans,Date,Payee,Amt],
       impl(and(transaction(Payer,Trans,Date,payee1#bt,Amt),
                and(t_precedes(date([1990,1,1]),Date),
                    t_precedes(Date,date([1990,12,31])))),
            x([Id,DisplayT],
              and(execute(DisplayEv,
                          display([Id,Date,bt,Amt]),
                          clare,
                          DisplayT),
                  and(Trans=transaction1#Id,
                      t_precedes(<Now>,DisplayT))))))
\end{verbatim}
Now we apply the equivalence that translates the conceptual predicate
{\tt transaction} into the database predicate {\tt TRANS}, 
\begin{verbatim}
(and(transaction(Payer,Trans,Date,Payee,Amount),
     transaction_data_available(Date)) <->
 x([TransId,AMNum,PayeeName,DBDate],
   and(TRANS(TransId,DBDate,PayeeName,AMNum),
       and(Trans = transaction1#TransId,
           and(Payee = payee1#PayeeName,
               and(db_date_convert(Date,DBDate),
                   Amount = amount(sterling,AMNum))))))) <-
 Payer = sri
\end{verbatim}
When the equivalence is applied,
the conjunctive context is the set
\begin{verbatim}
{transaction(Payer*,Trans*,Date*,Payee*,Amount*),
 t_precedes(date(1990,1,1),Date*),
 t_precedes(Date*,date(1990,12,31)}
\end{verbatim}
There are two conditions that must be proved from the context for
the rule to be applicable, 
\begin{verbatim}
Payer* = sri
\end{verbatim}
and
\begin{verbatim}
transaction_data_available(Date*)
\end{verbatim}
The first of these cannot be proved, but the assumption declaration
\begin{verbatim}
assumable(Payer = sri,
          0,
          payments_referred_to_are_from_SRI,
          specialization,
          transaction(Payer,Trans,Date,Payee,Amount))
\end{verbatim}
(``{\tt X} can be assumed to be SRI if it occurs in the {\tt Payer}
argument of a {\tt transaction} relation'')
makes it permissible to assume it. The other condition must be 
inferred. The relevant axioms are now two Horn-clauses: one defines the period
for which transaction records exist, viz.
\begin{verbatim}
transaction_data_available(Date) <-
   and(t_precedes(date(1,8,89),Date2),
       t_precedes(Date,date(1,4,91)))
\end{verbatim}
and the other encodes the fact that
\verb!t_precedes! is transitive, 
\begin{verbatim}
t_precedes(Date1,Date3) <-
   and(t_precedes(Date1,Date2),
       t_precedes(Date2,Date3))
\end{verbatim}
By chaining backwards through these to the 
context, it is then possible to prove that
{\tt transaction\-\_data\-\_available(Date*)} holds,
and translate to the final form
\begin{verbatim}
forall([TransId,Date,DBDate,AMNum],
       impl(and(TRANS(TransId,DBDate,bt,AMNum)
                and(db_date_convert(Date,DBDate),
                    and(t_precedes(date([1990,1,1]),Date),
                        t_precedes(Date,date([1990,12,31])))))
            x([Id,DisplayT],
              and(execute(DisplayEv,
                          display([TransId,Date,bt,Amt]),
                          clare,
                          DisplayT),
                   t_precedes(<Now>,DisplayT)))))
\end{verbatim}
The final formula admits a finite proof procedure, (cf 
sections~\ref{Effective-translation} and~\ref{Qopt}), since its
truth can be ascertained by an algorithm that can be summarized as
\begin{enumerate}
\item
Search for all tuples from the finite {\tt TRANS} relation.
\item
Check each to find the ones where the {\tt Payee}
field is equal to {\tt bt} and the {\tt Date} field is suitably
constrained by the arithmetic predicates {\tt db\_date\_convert}
and {\tt t\_precedes}
\item
Assume that appropriate lists will
in each such case be displayed by the interface's causing the
corresponding instantiated instance of the {\tt execute} predicate
to hold in the future. 
\end{enumerate}
The system has also managed to prove,
under an abductive assumption whose justification is
{\tt payments\_\-referred\_\-to\_\-are\_\-from\_\-SRI},
that the final formula is equivalent to the original query.
Granted this assumption, the user can be informed that 
the response is complete (cf section~\ref{Command-functionality}).

\chapter{Evolution of the QLF Formalism}
\label{qlfevolve}

This chapter describes the Quasi Logical Form formalism as currently used in
CLARE, highlighting the differences between the current version of the
QLF formalism and QLF as it was at the end of the CLE project (Alshawi 1992,
Alshawi {\em et al} 1989).

The QLF formalism presented here provides one way, amongst many,
of implementing a monotonic approach to semantic interpretation.  
However, various
details of the QLF formalism arise in response to specific features of
the semantics of English, or because they have proved convenient in
the CLARE system.  These details do not form part of the core of
monotonic interpretation, and the reader is directed to Chapter~\ref{monosem}
for a more theoretical, implementation independent account of monotonic
semantic interpretation.  In this chapter we will concentrate on specific
features rather than the general theory, and hope to provide the reader
with sufficient background to enable them to understand particular semantic
analyses produced by the CLARE system.

\section{The QLF Language}
\label{qlflang}

\subsection{A Single Level of Logical Form}

In the CLE-3 system, three distinct semantic representation languages were
used: (i) Quasi Logical Form (QLF) to represent the results of the syntax 
driven semantic interpretation rules, 
(ii) Resolved Quasi Logical Form (RQLF) to represent
the results of raising quantifiers, replacing anaphoric terms by their
referents and so forth, and (iii) Logical Form (LF) representing the truth
conditional parts extracted from RQLFs.  Converting one representation into
another usually involved destructive manipulation of the logical forms
and/or loss of information contained in the earlier logical forms.
Moreover, the QLF and RQLF languages did not have a proper semantics; it was 
only through converting them to LFs that they acquired a derivative semantics.

In CLARE (CLE-6) only one semantic representation language is used ---
QLF.  QLF is a development of the CLE-3 QLF language\footnote{
To avoid confusion, we will henceforth
refer to the CLE-3 QLF language as QLF-3, leaving QLF to refer to the
current version of the language.}, but differs from it in
a number of important respects.

First, expressions in QLF may be underinstantiated.
Underinstantiation is represented by means of
meta-variables contained within a QLF expression.
Reference resolution or scoping instantiates the QLF by means
of instantiating some or all of its meta-variables to further QLF
(sub-)expressions.  Meta-variable instantiation is non-destructive, and
is implemented as straightforward Prolog unification.  Thus
destructively raising quantifiers out of their surface positions or
replacing anaphoric terms by their referents plays no role in scoping
or resolution.  Instead, a scoping constraint meta-variable may be instantiated
to show where a particular term is scoped, or the referent meta-variable
of an anaphoric term may be instantiated to an expression denoting the 
referent.

Second, the QLF language has a directly defined semantics.  This applies
even when QLF expressions contain uninstantiated meta-variables.  Semantically,
instantiating meta-variables has the effect of narrowing the class of
models in which a QLF formula counts as true (see Chapter~\ref{monosem}).  
As a simple illustration, the initial
QLF for a sentence like {\em He slept} would simply say that some male
entity, salient in context, slept.  The unresolved QLF
could thus be true if, say, John Smith slept, or if George Brown slept, or
if Tom Jones slept.  Resolving the pronoun might result in a QLF that
said that some male entity, salient in context and identical to John Smith,
slept.  This narrows the class of models in which the
QLF could count as true.  The resolution just adds the extra
information that the object denoted by the term is identical to John
Smith.

The QLF instantiation brought about by reference resolution and scoping
is both syntactically and semantically monotonic.  Syntactically
because it just instantiates variables, and semantically because it
narrows the class of possible models.  Since QLFs, whether completely
uninstantiated, partially instantiated, or fully instantiated, all have
truth conditions associated, there is no need to postulate either a
separate level of RQLF or a separate level of LF.  Both these
levels of representation have disappeared from the CLARE system.

However, an LF-like representation can be obtained, if necessary, as part
of the process of evaluating QLFs.  For example, in the PRM application
(Chapters~\ref{translation},~\ref{claredm}) 
expressions in a Target Reasoning Language (TRL)
are extracted from QLFs after they have been scoped and resolved.  One can look
on TRL as a form of meta-language in which the semantic definitions for QLF
are expressed\footnote{In giving the semantics for a language like the
propositional calculus, one finds definitions like ``$S\phi\vee\psi$
is true iff $\phi$ is true or $\psi$ is true'' where the object language
connective $\vee$ is defined in terms of the meta-language connective
``or''.  The TRL connectives and quantifiers behave as meta-language versions
of QLF constructs.}  It is generally more convenient to reason using a
TRL representation of a QLF's truth conditions rather than the QLF itself,
but from a purely linguistic point of view TRL is an eliminable level of
representation.

Finally, there are a number of more minor differences between QLF-3 and QLF. 
For example, the QLF-3 constructs \verb!qterm! and
\verb!a_term! have been reduced to a single \term\ construct, and
\verb!a_form! constructs are now \forms.  Constructs like \verb!term_coord!
have been eliminated thanks to a more satisfactory treatment of
NP conjunction, as have `union terms'.  Tense operators like \verb!past!
or \verb!pres! have also been dispensed with.  Other differences
will become apparent below.

\subsection{Syntax of QLF}
\label{qlfbnf}

The QLF language has the following BNF syntax:
\bigskip

\begin{tt}
\begin{tabbing}
\anglevr{qlf\_formula} $\rightarrow$ \=
 form(\=\anglevr{word\_string},
        \anglevr{category},
        \anglevr{form\_index},
        \anglevr{qlf\_restriction},\\
\> \>   \anglevr{form\_referent}) $|$\\
\>[\anglevr{functor},\anglevr{argument$_1$},\dots \anglevr{argument$_n$}] $|$\\
\>sc(\anglevr{scope\_constraint},\anglevr{qlf\_formula})\\[2ex]

\anglevr{qlf\_term} $\rightarrow$ \>
\anglevr{term\_index} $|$ \\
\> term(\anglevr{word\_string},
        \anglevr{category},
        \anglevr{term\_index},
        \anglevr{qlf\_restriction},\\
\> \>	\anglevr{quantifier\_referent},
        \anglevr{term\_referent}) $|$ \\
\anglevr{constant} $|$ \anglevr{variable} $|$ \anglevr{term\_index}\\[2ex]

\anglevr{qlf\_abstract} $\rightarrow$
 \anglevr{variable}\verb!^!\anglevr{qlf\_formula} $|$
 \anglevr{variable}\verb!^!\anglevr{qlf\_abstract}\\[1ex]

\anglevr{qlf\_restriction} $\rightarrow$
 \anglevr{variable}\verb!^!\anglevr{qlf\_formula}\\[1ex]

\anglevr{functor} $\rightarrow$ \anglevr{atom} \\[1ex]

\anglevr{argument} $\rightarrow$ \anglevr{qlf\_formula} \\[1ex]

\anglevr{argument} $\rightarrow$ \anglevr{qlf\_term} \\[1ex]

\anglevr{argument} $\rightarrow$  \anglevr{qlf\_abstract} \\[1ex]

\anglevr{argument} $\rightarrow$
[apply,\anglevr{qlf\_abstract},\anglevr{argument}] \\[1ex]

\anglevr{constant} $\rightarrow$ \anglevr{atom} \\[1ex]

\anglevr{constant} $\rightarrow$ 'N'(\anglevr{number}) \\[1ex]

\anglevr{constant} $\rightarrow$
'SF'(c(\anglevr{qlf\_restriction},\anglevr{atom}))\\[1ex]

\anglevr{variable} $\rightarrow$ \anglevr{prolog\_variable}\\[1ex]

\anglevr{term\_index} $\rightarrow$ \anglevr{variable} $|$
\anglevr{atom}\\[1ex]

\anglevr{form\_index} $\rightarrow$ \anglevr{variable} $|$
\anglevr{atom}\\[1ex]

\anglevr{quantifier} $\rightarrow$ \anglevr{distrib\_quant} $|$
                                 \anglevr{set\_quant} 

\end{tabbing}\begin{tabbing}
\anglevr{quantifier} $\rightarrow$ \= wh $|$ 
   \anglevr{variable}\verb!^!count(\anglevr{variable}) $|$ \\
\> \anglevr{variable}\verb!^!sum(\anglevr{variable}) $|$ \\
\> amount(\anglevr{variable}\verb+^+\anglevr{variable}\verb+^+\anglevr{qlf\_formula},
         \anglevr{unit}) \\[1ex]

\anglevr{distrib\_quant} $\rightarrow$ forall $|$ exists 
$|$ most $|$ $\ldots$ \\[1ex]

\anglevr{distrib\_quant} $\rightarrow$ 
  \anglevr{variable}\verb+^+\anglevr{variable}\verb+^+\anglevr{lf\_formula} \\[1ex]

\anglevr{set\_quant} $\rightarrow$ \verb+set(+\anglevr{distrib\_quant}\verb+)+ 
\\[1ex]

\anglevr{unit} $\rightarrow$  \anglevr{qlf\_restriction} \\[1ex]

\anglevr{word\_string} $\rightarrow$ l([\anglevr{word}*]) \\[1ex]

\anglevr{word} $\rightarrow$ \anglevr{atom} 

\end{tabbing}\begin{tabbing}
\anglevr{form\_referent} $\rightarrow$ \=
 \anglevr{meta\_variable} $|$ \anglevr{qlf\_formula} $|$\\
 \> app(\anglevr{qlf\_abstract},\anglevr{form\_index}\\[1ex]

\anglevr{quantifier\_referent} $\rightarrow$ \anglevr{meta\_variable} $|$ 
                                      \anglevr{quantifier}

\end{tabbing}\begin{tabbing}
\anglevr{term\_referent} $\rightarrow$ \= 
  \anglevr{meta\_variable} $|$\\
\> strict(\anglevr{term\_index}) $|$\\
\> sloppy(\anglevr{term\_index},\anglevr{qlf\_restriction}) $|$\\
\> ent(\anglevr{constant}) $|$\\
\> qnt(\anglevr{term\_index}) \\[1ex]

\anglevr{scope\_constraint} $\rightarrow$
 [\anglevr{term\_index}*] $|$ \anglevr{meta\_variable}\\[1ex]

\anglevr{meta\_variable} $\rightarrow$ \anglevr{prolog\_variable}\\[1ex]

\end{tabbing}

\end{tt}

\bigskip

\noindent
Below, an informal description of the meaning of various QLF constructions
is given.  See also Chapter~\ref{monosem}.  Note also that in 
Chapter~\ref{monosem} a variant of the QLF formalism is used.  This differs
primarily in (a) using round bracket notation for predicate argument formulas,
i.e.
\begin{verbatim}
Pred(Arg1,..,Argn)
\end{verbatim}
instead of \verb![Pred,Arg1,...,Argn]!, and (b) omitting the {\tt word\_string}
argument to \terms\ and \forms.

\subsubsection*{\terms\ and \forms}

The main components of the QLF language are \forms\ and \terms.  The first
four arguments to \form\ and \term\ expressions are: (1) A list of words
representing the surface string that the \form\ or \term\ represents; (2)
A category, which can be any Prolog {\tt functor(arguments...)} expression;
(3) An index uniquely identifying the \form\ or \term; (4) A restriction,
which is a one-place abstract.  For \terms, the restriction is a one-place
predicate, corresponding perhaps to the meaning of a common noun.  For
\forms, the restriction is typically a one-place higher-order predicate.

The remaining arguments to the \forms\ and \terms\ are normally left as
uninstantiated meta-variables after the initial stages of semantic
analysis.  The meta-variables can then be instantiated by reference
resolution.  

\subsubsection*{\form\ Resolutions}

\forms\ have just one unresolved / meta-variable argument.  This can
be instantiated in one of two ways.  It can either be instantiated to a
formula or to an application.  An application 
({\tt app(\anglevr{abstract},\anglevr{index})}) is a
way of stating that the restriction of the \form\ should be applied to 
the \anglevr{abstract} (the \anglevr{index} is always the index of the
\form).  When resolved to a formula, the formula represents the result
of such an application.  The formula (or result of the application)
constitutes the final meaning of the \form.

\subsubsection*{\term\ Resolutions}

\terms\ have two unresolved arguments.  The first is resolved to the
quantifier associated with the \term. The second (the `term referent')
provides a further
contextual restriction on the range of this quantification (the contextual
restriction being combined with the \term's restriction).  

\paragraph{Quantifiers}
Quantifiers are generalised determiners, i.e. two place predicates on set 
cardinalities.  
The cardinality may either be represented explicitly, or in abbreviated
form (e.g. {\tt exists} or {\tt forall}).  There is a difference between
ordinary (distributive) quantifiers and set quantifiers. A distributive
quantifier quantifies over individual objects satisfying the restriction
predicate, whereas the set quantifier quantifies over sets of such objects.

There also a few `non-standard' quantifiers.  The WH-quantifiers
({\tt wh, count, sum}, corresponding roughly to {\em which, how many} and
{\em how much}) are in fact abstraction operators (so that WH-questions
are semantically treated as lambda-abstracts and not quantified formulae).
The {\tt amount} quantifier is used for expressions like {\em three
pints of milk}, where the unit of measure for the cardinality predicate
needs to be specified.

The treatment of quantifiers / generalised determiners is broadly the
same as that described in Alshawi 1992.

\paragraph{Contextual Restrictions / Term Referents}
The \term's contextual restriction comes in a number of varieties.
The simplest is a {\tt qnt(\anglevr{index})} restriction.  This just says
that the term acts as an ordinary quantifier with no restriction on its range
of quantification beyond that specified by the \term's restriction.  The
\anglevr{index} in these case is always the \term's index.  For
terms (like proper names) referring to specific entities, an
{\tt ent(\anglevr{constant})} resolvent is used, where \anglevr{constant}
is a constant term referring to some object.  This means that the \term\
is treated as an (existential) quantifier over objects identical to that
denoted by the constant.  For anaphoric terms (like pronouns and some
definite descriptions), a {\tt strict(\anglevr{index})} resolution is 
appropriate.  Here, the index is that of the antecedent \term.  If
the anaphoric \term\ is within the scope of the antecedents quantification,
this index becomes bound by the antecedent.  Otherwise a predicate uniquely
describing the antecedent is obtained from context and used as a further
contextual restriction.  For things like `one-anaphora', a 
{\tt sloppy(\anglevr{index},\anglevr{predicate})} resolution is needed.
Here, the index is that of the antecedent term, and the description is
obtained from the antecedent restriction.

The {\tt qnt} style resolution could be extended to include a general
contextual restriction on quantification not deriving from any particular
antecedent term.  This would necessitate including another argument place
for a property expression.  This has not been implemented, however.

\subsubsection*{Scoping Constraints}

Scoping constraints are (possibly empty) ordered lists of term indices.
Each \term\ is `quantified in' at the point where its index occurs.
Scoping constraints are normally only instantiated by reference
resolution.  Chapter~\ref{monosem} provides a fuller account of scoping
in monotonic interpretation.

\subsubsection*{Sorted Constants}

Sorted constants of the form {\tt 'SF'(c(\anglevr{sort},\anglevr{object}))}
are a way of explicitly associating a sortal predicate with
some individual denoting constant.  They are usually introduced when
\terms\ are resolved to refer to {\tt ent}ities.  The {\tt 'N'} wrapper
also identifies constants referring to numbers.

\subsubsection*{Categories}
No syntax for category expressions has been specified.  Usually, they
are atoms or Prolog {\tt functor(arguments...)} expressions, where
the arguments are give the values of certain features.  Sometimes, as
in the case of some \term\ categories, one of these values can be a
QLF expression.  This most often occurs when a complex comparative
determiner is built up (e.g. for {\em more cows than Mary has sheep}).

\section{Example QLFs}

We now illustrate the way that the QLF formalism is used to handle
various semantic phenomena in English.  This is done by giving the QLFs 
assigned to various sentences.  It will usually be convenient first to show
the uninstantiated QLF initially produced by syntax driven interpretation,
and then an instantiated QLF as produced by reference resolution.

\subsection*{Proper Names and Tense}

A simple sentence like {\em John slept} illustrates the basic treatment
of proper names and tense.  The unresolved QLF for the sentence is
\begin{quoteqlf}
\begin{verbatim}
[dcl,
 form(l([John,slept]),
      verb(past,no,no,no,y),
      A,
      B^[B,[sleep, A,
            term(l([John]),proper_name(tpc),C,
                 D^[name_of,D,'John'],_,_)]],
      _)]
\end{verbatim}
\end{quoteqlf}
At the top level, this a formula with the mood operator {\tt dcl}
applying to a verb \form.  The verb \form\ has the category
{\tt verb(past,no,no,no,y)} which says that the \form\ corresponds to a
verb phrase, that it is in the past tense, is non-perfective, non-progressive,
contains no modal verb, and is in the active voice.  The \form\
restriction essentially says that some (contextually salient) property
needs to be applied to the formula {\tt [sleep,A,term(\ldots)]}
in order to resolve the \form.  The resolution meta-variable is
uninstantiated.  Unlike the QLF-3 representation of this sentence, there
are no tense operators, and no terms referring to events.  The latter, though,
are introduced by reference resolution.

The \term\ corresponding to the subject noun phrase occurs within
the restriction of the verb \form.  Its category specifies that 
the term corresponds to a proper name, and the {\tt tpc} feature indicates
that the NP occupies subject / topic position.  The \term's restriction
is simply the property of having the name `John'.  Both resolution
meta-variables are uninstantiated.

Resolving the QLF above might instantiate the meta-variables as follows
(QLF shown abbreviated)
\begin{quoteqlf}
\begin{verbatim}
[dcl,
 form(l([John,slept]), .., .., P^[P,[sleep,...]],
      sc([E],[sleep,
               term(.., .., E, 
                    X^[and,[event,X],[precedes,X,now]],
                    exists, 
                    qnt(E)),
               term(l([John]), .., C, ..,
                    exists,
                    ent('SF'(c(Y^[person,Y],john_brown))))]
         ))]
\end{verbatim}
\end{quoteqlf}
The verb \form\ is resolved to a formula rather than an application.
Since the resolution includes much of the structure of the \form's restriction,
and hence the bulk of the QLF, we show only this.  

The resolution of the verb \form\ essentially involves abstracting over
the \form\ index and replacing it with an event \term.  This results
in a quantified formula saying that there was an event preceding the
current time, and that it was an event of John sleeping.  The event \term\
is given scope over the {\tt sleep} predication.  

It should be pointed out that the use of quantification over events and times 
is not an inherent part of the QLF formalism. We could instead have used
tense operators, or some mixture of quantification over time and tense
operators.  Nor are we necessarily committed to the view that quantification 
over events and times is the past way of handling temporal reference in a 
language like English: it just happens to be the way it is done here.

The second argument to the resolved verbal predication, the \term\ referring 
to John, is identical to the \term\ as it occurs in the \form\ restriction.  
The \term\ is resolved
to be an existential quantification over objects identical to the
sorted entity {\tt john\_brown}, where {\tt john\_brown} is a person.
Since this \term\ can be given any scope without affecting the truth
conditions of the QLF, its scope is left uninstantiated.

\subsection*{NP Modification and PPs}

Modified NPs are dealt with by conjoining extra properties to the restriction
of the NP's \term.  A resolved \term\ corresponding to {\em A man in London} is
\begin{quoteqlf}\begin{verbatim}
term(l([a,man,in,London]),
     q(_,a,sing), M,
     X^[and,[man,X],
            form(l([in,London]),
                 prep(in), F,
                 P^[P,X,
                      term(l([London]),
                           proper_name(_),L, 
                           Y^[name_of,Y,'London'],
                           exists, ent('SF'(c(Z^[town,Z],london))))],
                 app(R^S^[in_spatial,R,S],F))],
      exist,qnt(M))
\end{verbatim}\end{quoteqlf}
The \term's category, {\tt q(\_,a,sing)} shows that it is a quantificational
term, with determiner {\em a} and in the singular.  The underscore is
for a subject/topic feature, and is uninstantiated since the \term\ is not
included in a wider sentence.  The \term\ is resolved to be a straight-forward
existential quantification.

The \term's restriction is a conjunction of the unmodified noun property
(\verb!X^[man,X]!) and a prepositional phrase property (\verb!X^form(...)!).
The category of the PP \form\ states which preposition is used, but does not
attempt to distinguish between different temporal, spatial or other uses.
The \form\ resolution indicates the specific sense in which the preposition
is to be understood.  Applying the \form\ restriction to the resolvent
property, \verb!R^S^[in_spatial,R,S]!, gives
\begin{quoteqlf}\begin{verbatim}
[in_spatial, X,
             term(l([London]),
                  proper_name(_),L, 
                  Y^[name_of,Y,'London'],
                  exists, ent('SF'(c(Z^[town,Z],london))))]
\end{verbatim}\end{quoteqlf}
By treating prepositions as having vague senses that must be further
resolved to spatial, temporal or other properties, one can avoid a
proliferation of sense entries for prepositions in the lexicon.

\subsection*{VP Modification}
Prepositional phrases can modifiy verb phrases as well as noun phrases.
Here the QLF representation is slightly different.  The PPs, as well as
other modifiers to the VP, are included in a list in the verb \form's
restriction preceding the verbal predication itself.  Thus, a sentence
like {\em John slept well on Tuesday} would receive the following QLF
\begin{quoteqlf}\begin{verbatim}
[dcl,
 form(l([John,slept,well,on,Tuesday]),
      verb(past,no,no,no,y),
      A,
      B^
      [B,
       form(l([on,Tuesday]),
            prep(on), C,
            D^[D, v(A),
                  term(l([Tuesday]),
                      time(day), E,
                      F^[day_name,F,tuesday_DayName],
                      _,_)],
            _),
       [good,v(A)],
       [sleep, A,
               term(l([John]),proper_name(tpc),R,
                    S^[name_of,S,John],_,_)]],
      _)]
\end{verbatim}\end{quoteqlf}
The adverb {\em well} is interpreted as being derived from the
adjective {\em good}, and gives rise to a predication on the verb \form's
index.  The verb \form's index also provides the first argument in the
prepoisitional phrase restriction.  When the verb \form\ is resolved,
occurrences of the \form's index will be replaced by the corresponding
event variable, and modifiying predications will be conjoined with the
main verbal predication.   The net result is a Davidsonian style treatment
of adverbial modification (Davidson 1967), where modifiers are predicates
of an event.

\paragraph{`Big' PPs}
When several prepositional phrases modify a verb phrase, as in {\em John slept
in London on Tuesday}, the prepositional phrases are parsed as forming a
single `big' PP acting as just one modifier to the verb phrase.  The
QLF reflects the `big' PP structure:
\begin{quoteqlf}\begin{verbatim}
[dcl,
 form(l([John,slept,in,London,on,Tuesday]),
      verb(past,no,no,no,y),A,
      B^
      [B,
       form(l([in,London,on,Tuesday]),
            conj(pp,implicit_and),_,
            C^
            [C,
             form(l([in,London]),prep(in),_,
                  D^
                  [D,v(A),
                   term(l([London]),proper_name(_),_,
                        E^[name_of,E,London],_,_)],
                  _),
             form(l([on,Tuesday]),prep(on),_,
                  F^
                  [F,v(A),
                   term(l([Tuesday]),time(day),_,
                        G^[day_name,G,tuesday_DayName],_,_)],
                  _)],
            _),
       [sleep,A,
        term(l([John]),proper_name(tpc),_,
             H^[name_of,H,John],_,_)]],
      _)]
\end{verbatim}\end{quoteqlf}
The big PP is in fact treated as an implicit conjunction of the component
prepositional phrases. After resolution it gives rise to the same conjunction 
of predications on the event variable as would arise if the PPs has been 
treated  as individual modifiers.

\subsection*{Coordination}

Conjunctions are treated as lists of the conjoined QLF expressions, forming
the restriction of a conjunction \form.  For example {\em John ate, drank
and slept} gets the QLF:
\begin{quoteqlf}\begin{verbatim}
[dcl,
 form(l([John,ate,,,drank,and,slept]),
      conj(vp,and),_,
      A^
      [A,
       form(l([slept]),verb(past,no,no,no,y),B,
            C^
            [C,[sleep,B,
              term(l([John]),proper_name(tpc),D,E^[name_of,E,John],
                   _,_)]],
            _),
       form(l([drank]),verb(past,no,no,no,y),F,
            G^[G,[drink,F,v(D)]],_),
       form(l([ate]),verb(past,no,no,no,y),H,
            I^[I,[eat,H,v(D)]],_)],
      _)]
\end{verbatim}\end{quoteqlf}
The category {\tt conj(vp,and)} states what type of constituent is
coordinated and the conjunction used.  Resolving the \form\ results in
a logical conjunction being built up from the resolved elements of
the conjunction list.

\paragraph{NP Conjunction}
Coordination of most other constituents is handled analagously to that
of VPs.  However, noun phrase conjunction is handled slightly differently,
since one needs to produces a conjoined \term\ rather than a conjoined
\form.  A sentence like {\em Every man, every woman and every child slept}
gets the following QLF:
\begin{quoteqlf}\begin{verbatim}
[dcl,
 form(l([every,man,,,every,woman,and,every,child,slept]),
      verb(past,no,no,no,y),A,
      B^
      [B,
       [sleep_BeNaturallyUnconscious,A,
        term(l([every,man,,,every,woman,and,every,child]),
             conjdet(tpc,and,np,C,_),C,
             D^
             form(_,conj(np,and),_,
                  E^
                  [E,D,
                   term(l([every,child]),q(tpc,every,sing),_,
                        F^[child_YoungPerson,F],_,_),
                   term(l([every,woman]),q(tpc,every,sing),_,
                        G^[woman_FemalePerson,G],_,_),
                   term(l([every,man]),q(tpc,every,sing),_,
                        H^[man_MalePerson,H],_,_)],
                  _),
             _,_)]],
      _)]
\end{verbatim}\end{quoteqlf}
Here, a conjunction \form\ serves as the restriction to a special
conjunction \term.  In resolving the conjunction, one first of all
constructs a group denoting term, such that every child is included within
the group, every woman is included within the group and (or) every
man is included within the group.  One then quantifies over individual
members of this group (distributive conjunction) or over the group
as a whole (collective conjunction).

The (distributive) resolution of the conjoined NP above would be something like
\begin{quoteqlf}\begin{verbatim}
term(..,conjdet(...),C
     D^form(...,
            [partof,D,
              term(_,group,G,
                   X^[and,sc([C1],[partof,X,
                                   term(l([every,child]),..,C1,..)]),
                      [and,sc([W1],[partof,X,
                                    term(l([every,woman]),..,W1,..)]),
                           sc([M1],[partof,X,
                                    term(l([every,man]),..,M1,..)])]],
                   exists,qnt(G))]),
     forall,qnt(C))
\end{verbatim}\end{quoteqlf}
The group term \verb!G! is always given wide scope relative to 
the conjunction term \verb!C!.

This treatment of NP coordination avoids use of earlier \verb!term_coord!
constructions and union terms.  It uses only propositional conjunction and
disjunction, and may be used to combine all forms of noun phrase to
give either a distributive or collective coordination.

\subsection*{Possessives and Partitives}

A possessive noun phrase, like {\em John's mother}, gets the following
QLF
\begin{quoteqlf}\begin{verbatim}
term(l([John,'s,mother]),q(_,poss_some,_),_,
     B^[and,[mother_FemaleParent,B],
            form(_,poss,_,
                 C^
                 [and,[mother_FemaleParent,B],
                  [C,B,
                   term(l([John]),proper_name(_),_,
                        D^[name_of,D,John],_,_)]],
                 _)],
           _,_)
\end{verbatim}\end{quoteqlf}
The embedded {\tt poss} \form\ is resolved to a particular relation
holding between people and their mothers.  For different possessives,
different relations might be appropriate.  In the PRM domain for example
a possessive noun phrase like {\em the project's payments}
would be understood as referring to payments on the project.

Partitive noun phrases, like {\em each of the men} are treated as
follows
\begin{quoteqlf}\begin{verbatim}
term(l([each,of,the,men]),
     ell(q(_,each,sing),l([])),A,
     B^[and,[entity,B],
        form(l([of,the,men]),
            genit(sub),_,
            C^[and,[entity,B],
               [C,B,
                  term(l([the,men]),
                       ref(def,the,plur,l([])),_,
                       D^[man_MalePerson,D],_,_)]],
             _)],
      _,qnt(A))
\end{verbatim}\end{quoteqlf}
Syntactically, partitives are treated as prepositional phrases ({\em of
the men}) modifying bare determiners acting as pronouns ({\em each, some,
all} etc).  However, rather than resolving the bare determiner NP in the
normal way as an anaphor, the initial semantic analysis forces it to be
quantificational by instantiating the term referent.  The {\tt genit(sub)}
\form\ is resolved to give the property of being an element that is
{\tt partof} the denotation of the (group denoting) NP argument 
(i.e. being one of the men).

Partitives differ from possessive NPs like {\em the mother of John} in that
(i) the genitive relation is not fixed as {\tt sub}, (ii) there
is no elliptical determiner NP, and (iii) the referent is not instantiated
by the semantics:
\begin{quoteqlf}\begin{verbatim}
term(l([the,mother,of,John]),ref(def,the,sing,l([])),_,
     B^[and,[mother_FemaleParent,B],
            form(l([of,John]),genit(_),_,
                 C^
                 [and,[mother_FemaleParent,B],
                  [C,B,
                   term(l([John]),proper_name(_),_,
                        D^[name_of,D,John],_,_)]],
                 _)],
      _,_)
\end{verbatim}\end{quoteqlf}
(In fact, leaving the sub-class of the {\tt genit} category uninstantiated
is not quite correct, and the grammar has since been altered to ensure that
the category is properly instantiated).

\chapter{Monotonic Semantic Interpretation}
\label{monosem}

\section{Introduction}

The monotonicity property of unification based grammar formalisms is
perhaps the most important factor in their widespread use for grammatical
description and parsing.  Monotonicity guarantees that the grammatical
analysis of a sentence can proceed incrementally by combining information
from rules and lexical entries in a nondestructive way.  By contrast, 
aspects of semantic interpretation, such as reference and quantifier
scope resolution, are often realised by non-monotonic
operations involving loss of information and destructive manipulation of
semantic representations. A `two-level' approach to semantic
interpretation tends to result (Bronneberg {\it et al.} 1980),
where an initial, underspecified representation is
transformed into a separate, specified, representation.

This chapter describes
 a model for semantic interpretation that is fully monotonic in
both linguistic and contextual aspects of interpretation, and which
employs just one level of semantic representation ---
Quasi Logical Form (QLF). Contextual resolution of underspecified QLF expressions 
involves the instantiation of QLF meta-variables.
The semantics for the QLF formalism makes the denotation of a QLF formula a 
partial function to truth-values, with resolution leading to a monotonic
extension of the denotation function.
We believe that there are several advantages to the approach taken,
including:
\begin{itemize}
\item
Order independence of resolution operations
\item 
Production of partial interpretations
\item
Simpler interactions between phenomena
\item
Reversibility for synthesis/generation
\end{itemize}

The QLF formalism is a development of (Alshawi 1990).  
As before, underspecified
QLFs are produced on the basis of a unification grammar.
Previously, QLF resolution was only partially monotonic;
full monotonicity required changes to the original QLF formalism and the 
resolution and scoping processes, as described in 
Chapter~\ref{qlfevolve}.

The chapter is organized as follows.  Section~\ref{monosem2} provides the
syntax of the QLF language and Section~\ref{monosem3} gives some
illustrative examples of monotonic QLF resolution.  Sections~\ref{monosem4} and
\ref{monosem5} present
the semantics of the QLF formalism.  Section~\ref{monosem6} discusses
the relationship between monotonic interpretation, Pereira's categorial
semantics (Pereira 1990), and context change approaches to semantics.
Section~\ref{monosem7} mentions some benefits of using QLF-like
representations in implementing natural language systems.

\section{Syntax of Monotonic QLF}
\label{monosem2}

We give here a fuller syntactic description of the QLF
constructs for
terms and formulas\footnote{The notation
we use in implementations is slightly different but equivalent to that
presented here.}.

A QLF term must be one of the following
\begin{itemize}

\item a term variable: \verb!X!, \verb!Y!, \dots

\item a term index: \verb!+i!, \verb!+j!, \dots

\item a constant term: \verb!7!, \verb!mary1!, \dots

\item an expressions of the form:\\
\verb!term(Idx,Cat,Restr,Quant,Reft)!

\end{itemize}

\noindent
The term index, \verb!Idx!, uniquely identifies the \verb!term! expression.
\verb!Cat! is a list of feature-value equations, for example
\verb!<type=pro,num=sing,...>!.
\verb!Restr! is a first-order, one-place predicate.
For a resolved \verb!term!, \verb!Quant! 
will be a generalized quantifier (a cardinality predicate holding of two 
properties)
and \verb!Reft!, the \verb!term!'s `referent', will be a constant or term
index. For an `unresolved' \verb!term!,
\verb!Quant! and \verb!Reft! may be meta-variables 
(\verb!_x!,\verb!_y!,\dots).  (QLF terms may also be functional applications,
though we will ignore these here).

A QLF formula must be one of the following

\begin{itemize}

\item the application of a predicate to arguments:\\
\verb!Predicate(Argument1,...,Argumentn)!

\item an expression of the form:\\
\verb!form(Category,Restriction,Resolution)!

\item a formula with scoping constraints:\\
\verb!Scope:Formula!

\end{itemize}

\noindent
\verb!Predicate!  is a first or higher-order predicate, including the
usual logical operators \verb!and!, \verb!not!, etc.  An argument may be a
term, a formula or a lambda abstract.  Lambda abstracts take the form
\verb!Var^Body!  where \verb!Body!  is a formula or an abstract and
\verb!Var!  is a variable ranging over individuals or relations.
 \verb!Restriction!  is a higher-order predicate.
\verb!Resolution! is a formula (the `referent' of the {\tt form} expression), or
is a meta-variable if the \verb!form! expression is unresolved.
\verb!Scope!  is either a meta-variable when scoping
information is underspecified or a (possibly empty) list of term
indices e.g. \verb![+i,+j]! if term \verb!+i! outscopes \verb!+j!.
The terms identified by the indices must occur within \verb!Formula!.

The degree to which a QLF is unresolved corresponds approximately
to the extent to which
meta-variables (appearing above as \verb!Quant!, \verb!Reft!, 
\verb!Scope!, and \verb!Resolution!) are instantiated to the 
appropriate kind of object level expressions (though see 
Section~\ref{monosem5} for an explicit characterization of unresolved
QLFs and partial interpretations.)

\section{Example QLF Resolutions}
\label{monosem3}

Resolution of QLFs through the instantiation of meta-variables has been applied
to a wide range of phenomena. These include pronouns, definite
descriptions, implicit or vague relations, ellipsis and temporal relations
(see Alshawi 1990 for an account of some kinds of reference resolution in an
earlier QLF formalism).  For concreteness, we present a few illustrative
examples of monotonic QLF resolution\footnote{
Although the QLF framework can support a variety
of alternative semantic analyses for specific phenomena, to provide concrete
illustrations one or other analysis needs to be chosen.  
In the following examples, it should be possible to separate 
particular analyses from the general points we
wish to make about monotonic interpretation.}.  We do not attempt to describe
the mechanism by which the resolutions are chosen.

It will become evident that the notation is closer to (the syntactic structure
of) natural language than is the case for traditional logical formalisms.  
For example, \verb!term!s usually correspond to noun phrases, with information
about whether e.g. they are pronominal, quantified or proper names included in
the \verb!term!'s category.  
This makes the QLF representation easier to read
than it might seem at first, once its initial unfamiliarity is overcome.

\paragraph{Quantification:}
{\it Every boy met a tall girl} illustrates the
representation of quantification. The basic QLF analysis
might be (ignoring tense):
\begin{quoteqlf}\vspace*{-2ex}\begin{verbatim}
_s:meet(term(+b,<type=q,lex=every>,boy,_q,_x),
        term(+g,<type=q,lex=a>,
             Y^and(girl(Y),tall(Y)),_r,_y)).
\end{verbatim}\end{quoteqlf}
A resolved structure could be obtained by instantiating 
the quantifier meta-variables \verb!_q! and \verb!_r! to 
\verb!forall! and \verb!exists!\footnote{
The benefits of being able to resolve determiners to quantifiers are
discussed in Alshawi 1990.  For example, determiners like {\it some} (plural)
could be resolved to collective or distributive quantifiers, {\it three}
could be interpreted as meaning either `exactly three' or `at least three',
and if need be, bare plurals like {\it dogs} could be variously interpreted
as meaning `some dogs', `all dogs' or `most dogs'.}, 
and the scoping meta-variable \verb!_s! to \verb![+b,+g]! for the 
`$\forall\exists$' reading:
\begin{quoteqlf}\begin{verbatim}
[+b,+g]:
 meet(term(+b,<type=q,lex=every>,
           boy,forall,+b),
      term(+g,<type=q,lex=a>,
           Y^and(girl(Y),tall(Y)),exists,+g)).
\end{verbatim}\end{quoteqlf}
In a restriction-body notation for generalized
quantifiers, the truth conditional content of this resolved expression
corresponds to
\begin{quoteqlf}\begin{verbatim}
forall(B,boy(B),
         exists(G,and(girl(G),tall(G)),
                  meet(B,G))).
\end{verbatim}\end{quoteqlf}

\paragraph{Anaphora:}
{\it Every boy claims he met her} illustrates the treatment of anaphora
(in a context where Mary is assumed to be salient)\footnote{Here we simplify
the issues arising out of the semantics of intensional, sentential complement
verbs like {\it claim}.}
\begin{quoteqlf}
Unresolved:\vspace*{-1ex}
\begin{verbatim}
_s1:claim(
     term(+b,<type=q,lex=every>,boy,_q1,_x),
     _s2:meet(term(+h1,<type=pro,lex=he>,
                   male,_q2,_y),
              term(+h2,<type=pro,lex=her>,
                   female,_q3,_z))).
\end{verbatim}

Resolved:\vspace*{-1ex}
\begin{verbatim}
[+b]:claim(
      term(+b,<type=q,lex=every>,
           boy,forall,+b),
      [+h1]:meet(term(+h1,<type=pro,lex=he>,
                      male,exists,+b),
                 term(+h2,<type=pro,lex=her>,
                      female,exists,mary))).
\end{verbatim}\end{quoteqlf}
The pronominal term for {\it her} is resolved so that it existentially quantifies
over female objects identical to \verb!mary!. The `bound variable' pronoun 
{\it he} has a referent coindexed with its antecedent, \verb!+b!. 
The scope of \verb!+h2!
is left unspecified, since exactly the same truth
conditions arise if it is given wide or narrow scope with respect to
{\it every boy} or {\it he}.

\paragraph{Vague Relations:}

An unresolved  QLF expression representing the noun phrase 
{\it a woman on a bus} might
be a \verb!term! containing a \verb!form! that arises from the the
prepositional phrase modification:
\begin{quoteqlf}\begin{verbatim}
term(+w,<lex=a,..>,
     X^and(woman(X),
           form(<type=prep,lex=on>,
                R^R(X,term(+b,<lex=a,..>,
                            bus,_q2,_b)),
                _f)),
     _q1,_w).
\end{verbatim}\end{quoteqlf}
Informally, the \verb!form! is resolved by applying its restriction, 
\verb!R^R(...)! to an appropriate salient predicate, and instantiating 
the \verb!form!'s meta-variable, \verb!_f!,
with the result. In this case, the appropriate predicate might be \verb!inside!,
so that \verb!_f! is instantiated to
\begin{quoteqlf}\begin{verbatim}
inside(X,term(+b,<lex=a,..>,bus,_q2,_b)).
\end{verbatim}\end{quoteqlf}

\paragraph{Tense:}
One way of treating tense is by means of a temporal relation \verb!form!
in the restriction of an event \verb!term!. For {\it John slept} we might have:
\begin{quoteqlf}\begin{verbatim}
_s:sleep(term(+e,<type=event>,
              E^form(<type=trel,tense=past>,
                        R^and(event(E),R(E)),
                        _t),
              _q1,_e),
         term(+j,<type=name>,
              J^name(J,'John'),_q2,_j)).
\end{verbatim}\end{quoteqlf}
Since the tense on the temporal relation category is {\tt past}, 
the resolution
says that the event occurred before a particular speech time, \verb!t7!:
\begin{quoteqlf}\begin{verbatim}
[+e]:
 sleep(
  term(+e,<type=event>,
       E^form(<type=trel,tense=past>,
              R^and(event(E),R(E)),
              and(event(E),precede(E,t7))),
       exists,+e),
  term(+j,<type=name>,
       J^name(J,'John'),exists,john1)).
\end{verbatim}\end{quoteqlf}
The resolution {\tt and(event(E),precede(E,t7))}
is the result of applying the {\tt form}'s restriction 
\verb!R^and(event(E),R(E))! to a contextually derived predicate, in this
case \verb!E1^precede(E1,t7)!.

QLF is not committed to an event based treatment of tense.  
An alternative that has also been implemented is to
treat the verbal predication {\tt sleep(...)} as a temporal \verb!form!, 
whose category specifies
tense and aspect information.

\paragraph{Ellipsis:}
A more complex example, involving ellipsis
and quantification, is provided by
\begin{quote}
{\it Each boy claimed he was clever, and so did John.}
\end{quote}
A partially resolved QLF, but one in which the ellipsis
is still unresolved, might be as follows (ignoring tense and event 
variables):

\begin{quoteqlf}\begin{verbatim}
and(
claim(term(+b,<lex=every>,
           boy,exists,+b),
      clever(term(+h,<lex=he>,
                  male,exists,+b))),
form(<type=vpellipsis>,
     P^P(term(+j,<type=name>,J^name(J,'John'),
              exists,john)),
     _e)).
\end{verbatim}\end{quoteqlf}
This is a conjunction of the QLF for the antecedent clause ({\it Each
boy claimed he was clever} under a bound pronoun reading) with a 
\verb!form! expression for the verb phrase ellipsis.
Solutions for instantiating the meta-variable 
\verb!_e! for the ellipsis are the result of applying a property \verb!P1!,
derived from the antecedent clause, to the \verb!term! with index \verb!+j!.
The sentence has two readings: a sloppy reading where John claims that he is 
clever, and a strict one where John claims that each of the boys is clever.  
The choice between a strict or sloppy reading depends on how the term 
{\it he} is reinterpreted in the ellipsis resolution. Intuitively, 
strict identity involves referring to the same object as before, whereas 
sloppy identity involves referring to a relevantly similar object.

In QLF, a strict reading results from re-interpreting the ellipsis pronoun as 
co-indexed with the original, i.e. taking \verb!P1! to be:
\begin{quoteqlf}\begin{verbatim}
X^claim(X,clever(+h)).
\end{verbatim}\end{quoteqlf}
Constraints on legitimate scoping (Section~\ref{monosem5}) force \verb!+b!
and \verb!+h! to take wide scope over both the antecedent and ellipsis.
The sloppy reading results from re-indexing the ellipsis pronoun so 
that it has the same restriction and category as the original, but is resolved
to \verb!+j! and has a new index \verb!+h1!.
This corresponds to taking 
\verb!P1! to be:
\begin{quoteqlf}\begin{verbatim}
X^claim(X,clever(term(+h1,<lex=he>,
                      male,exists,+j))).
\end{verbatim}\end{quoteqlf}

More generally, in Chapter~\ref{ellipsis} we explore the claim that
solutions to verb phrase ellipsis have the general form:
\begin{quoteqlf}\begin{verbatim}
P1 = X1^..Xi^S:{s1/X1,..,si/Xi,..,sn/tn}.
\end{verbatim}\end{quoteqlf}
That is, \verb!P1! is formed out of an antecedent clause QLF \verb!S!
by abstracting over the `parallel elements' \verb!s1..si!, perhaps with some
additional substitutions for terms \verb!si+1..sn! in \verb!S! (\verb!E:{a/b}!
is the expression \verb!E! with \verb!b! substituted for \verb!a!.
The substitutions are applied in a slightly non-standard way --- see
Chapter~\ref{ellipsis} and section~\ref{monosem5} below).
This seems to be sufficient to cover the range of examples treated
by Dalrymple, Shieber and Pereira (1991),
but that is a specific linguistic claim about verb phrase ellipsis
in English and not central to the present chapter.  Ellipsis is further
discussed in Chapter~\ref{ellipsis}.

\section{Semantics for QLF}
\label{monosem5}

In this section we outline the semantics of the QLF language in a way that
is as close as possible to classical approaches that provide the semantics in
terms of a function from models to truth values.  The main difference is
that denotation functions will be partial functions for some unresolved
QLF formulas, reflecting the intuition that these are `partial interpretations'.
The denotation of a QLF expression will be extended monotonically as it is further
resolved, a fully resolved formula receiving a total function as its
denotation. The semantics is not intended to describe the resolution process.

Before giving evaluation rules for the QLF language,
we first present a simplified version of the semantics for fully instantiated
QLF expressions. This is for expository purposes only; the full QLF semantics
does not depend on the simplified version.

\subsection{Simplified Semantics}

We will use the notation [[\verb!E!]]{\it m} for the 
truth value of an expression \verb!E! with respect to a model {\it m} 
(but will leave {\it m} implicit). 
{\it m} includes an interpretation function
{\it I} for mapping constants and predicates into domain individuals
and relations.  Also left implicit is a function assigning values to variables,
which is required for the evaluation of lambda abstracts as characteristic
functions. 

Constructs in the `standard' predicate logic subset of QLF receive their
semantics with the usual evaluation rules, for example:
\begin{itemize}
\item[$\bullet$]
[[\verb!P(a1,...,an)!]] = 1 iff I(\verb!a1!) \dots I(\verb!an!) are in the
relation I(\verb!P!), and 0 otherwise.

\item[$\bullet$]
[[\verb!and(F1,F2)!]] = 1 iff [[\verb!F1!]]=1 and [[\verb!F2!]]=1, and
0 otherwise.
\end{itemize}

The evaluation rule for a formula \verb!F! with a scoping
variable instantiated to \verb![I,J,...]! and containing a term 
\verb!T!=\verb!term(I,C,R,Q,A)! is as follows:
\begin{itemize}
\item[$\bullet$]
[[\verb![I,J,...]:F!]]=1 iff [[\verb!Q(R',F')!]]=1, and 0 otherwise,
where\\
\verb!R'! is \verb!X^(and(R(X),X=A))![\verb!I!/\verb!X!], and\\
\verb!F'! is \verb!X^([J,...]:and(F,X=A))![\verb!T!/\verb!X!, \verb!I!/\verb!X!]
\end{itemize}
This evaluation rule states that a formula with a scoping constraint
list may be evaluated by `discharging' the term for the first index on the
list with respect to a formula with a reduced scoping constraint.
The rule discharges the term by abstracting over
occurrences of the term and its index, and applying 
the generalized quantifier \verb!Q! to the term's restriction and the 
abstract derived from the formula.
In Section~\ref{monosem5} we will say more about the
ramifications of adopting this type of quantifier evaluation rule.
Note that this rule is also applicable to resolved
terms such as pronouns for which \verb!Q! has been resolved to 
\verb!exists! and \verb!T! is a constant or a scoped variable.

The denotation assigned to a resolved formula
\verb!form(C,R,F')! in which the resolution variable has been 
instantiated to a formula \verb!F'! is simply:
\begin{itemize}
\item[$\bullet$]
[[\verb!form(C,R,F')!]]=1 iff [[\verb!F'!]]=1, and 0 otherwise.
\end{itemize}

\subsubsection{Note on Substitutions}
The substitution notation,
\begin{verbatim}
E[I/X,T/X]
\end{verbatim}
in the semantic meta-language and
\begin{verbatim}
E:{I/X,T/X}
\end{verbatim}
in the object language, needs to be read in a slightly non-standard way.

In the usual notation, the substitutions are applied in left to right
order to the expression {\tt E}.  That is, first replace all occurrences
of {\tt I} bt {\tt X}, and then replace all occurrences of {\tt T} by
{\tt X}.  This may lead to problems if {\tt I} is a subexpression of
{\tt T} (e.g. the index of a \term), since after the first set of substitutions
is applied, the second is no longer applicable.  In the conventional notation,
the order of substitutions is important.

In monotonic interpretation, the substitutions are to read as directives
on semantic valuation. Thus {\tt I/X} means `occurences of {\tt I} are
to be evaluated as though they were occurrences of {\tt X}'.  Since
semantic evaluation applies recursively over the structure of the
QLF, it is the order in which {\tt I} and {\tt T} occur in the QLF expression
that is important, and not the order in which they occur in the substitution
list.

\subsection{QLF Semantics}

As mentioned earlier, the denotation of a formula \verb!F!  in the QLF
language will be a possibly partial function ([[\dots]]) from models to truth
values. Again we use the notation [[\verb!F!]]{\it m} for the truth value 
of a formula \verb!F! with respect to a model {\it m}
(explicit reference to a variable assignment function is again suppressed).
For interpretation
to be monotonic, we want [[\verb!G!]] to be an extension of [[\verb!F!]]
whenever \verb!G! is a more resolved version of \verb!F!, and in
particular for [[\verb!G!]] to be total if \verb!G! is fully 
resolved.

We will define [[\dots]] for QLFs in terms of a relation {\it W} between
formulas, models and truth values. Evaluation rules will be given 
for {\it W}(\verb!F!,{\it m},v), but since more than one rule may apply (or a 
rule may apply in more than one way), {\it W} will in general be a
relation. The relationship between [[\dots]] and {\it W} for a formula
\verb!F! is as follows:

\medskip
\noindent
\hspace*{2mm} $\bullet$ [[\verb!F!]]{\it m}=1 iff 
   {\it W}(\verb!F!,{\it m},1) but not {\it W}(\verb!F!,{\it m},0);\\[0.1mm]
\hspace*{2mm} $\bullet$ [[\verb!F!]]{\it m}=0 iff 
   {\it W}(\verb!F!,{\it m},0) but not {\it W}(\verb!F!,{\it m},1);\\[0.1mm]
\hspace*{2mm} $\bullet$ [[\verb!F!]]{\it m} undefined iff 
   {\it W}(\verb!F!,{\it m},1) and {\it W}(\verb!F!,{\it m},0).

\smallskip
Henceforth we will leave the model argument {\it m} implicit.
The evaluation rules for {\it W} will generally take the form

{\it W}(\verb!F!,v) if {\it W}(\verb!F'!,v)

\noindent
where \verb!F'! contains one fewer unresolved expression than \verb!F!
(so that it is possible for the process of rule application to
terminate). The use of {\it if} rather than {\it iff} in these rules
means that it is possible for rules producing more than one value v
to apply and hence for [[\verb!F!]] to be partial.

The model provides an interpretation function {\it I} mapping constants and
predicates to individual and relations. We will also need to assume
a relation {\it S}(\verb!C!,\verb!H!) (for `salient') between QLF 
categories \verb!C! and QLF expressions \verb!H! standing for individuals, 
quantifiers, or predicates,
but the precise nature of the salience relation and the way it changes 
during a discourse are not important for the evaluation rules for 
QLF given here. 
The intuitive motivation for {\it S} is that the category
in an unresolved QLF expression restricts the set of possible referents 
for that expression. {\it S} is discussed further in
Section~\ref{monosem5}.
We are now in position to present the evaluation
rules, which we number Q1, Q2, etc.

For standard connectives we have the obvious evaluation rules, for
example,
\begin{itemize}
\item[Q1]
{\it W}(\verb!and(F,G)!,1) if {\it W}(\verb!F!,1) and {\it W}(\verb!G!,1).
\item[Q2]
{\it W}(\verb!and(F,G)!,0) if {\it W}(\verb!F!,0) or {\it W}(\verb!G!,0).
\item[Q3]
{\it W}(\verb!not(F)!,1) if {\it W}(\verb!F!,0).
\item[Q4]
{\it W}(\verb!not(F)!,0) if {\it W}(\verb!F!,1).
\end{itemize}

\noindent
Two rules applicable to a formula \verb!F! containing a term with 
uninstantiated referent and quantifier meta-variables:
\begin{itemize}\item[Q5]
{\it W}(\verb!F!,v) if {\it W}(\verb!F![\verb!_q!/\verb!exists!,\verb!_r!/\verb!A!],v)
and {\it W}(\verb!R(A)!,1),\\
where:\\
\hspace*{2mm}\verb!F! is a formula containing the term\\
\hspace*{6mm} \verb!T!=\verb!term(I,C,R,_q,_r)!, and\\
\hspace*{2mm}\verb!A! is a term such that {\it S}(\verb!C!,\verb!A!).
\end{itemize}
\begin{itemize}\item[Q6]
{\it W}(\verb!F!,v) if {\it W}(\verb!F![\verb!_q!/\verb!Q!, \verb!_r!/\verb!I!],v),\\
where:\\
\hspace*{2mm}\verb!F! is a formula containing the term\\
\hspace*{6mm}  \verb!T!=\verb!term(I,C,R,_q,_r)!, and\\
\hspace*{2mm}\verb!Q! is a quantifier such that {\it S}(\verb!C!,\verb!Q!).
\end{itemize}
(The substitutions for the meta-variables \verb!_r! and \verb!_q!
are to be read as part of the evaluation rule.)

\medskip
\noindent
A rule applicable to a formula
\verb!F! in which a (possibly unscoped) quantified term occurs:
\begin{itemize}\item[Q7]
{\it W}(\verb!F!,v) if {\it W}(\verb!Q(R',F')!,v),\\
where:\\
\hspace*{2mm}\verb!F! is a formula containing the term\\
\hspace*{6mm}  \verb!T!=\verb!term(I,C,R,Q,A)!,\\
\hspace*{2mm}\verb!R'! is \verb!X^(and(R(X),X=A))![\verb!I!/\verb!X!], and\\
\hspace*{2mm}\verb!F'! is \verb!X^(and(F,X=A))![\verb!T!/\verb!X!,
\verb!I!/\verb!X!].
\end{itemize}

\noindent
A rule applicable to a formula with an instantiated scoping 
constraint
\begin{itemize}\item[Q8]
{\it W}(\verb![I,J,...]:F!,v) if {\it W}(\verb!Q(R',F')!,v),\\
where:\\
\hspace*{2mm}\verb!F! is a formula containing the term\\
\hspace*{6mm}  \verb!T!=\verb!term(I,C,R,Q,A)!,\\
\hspace*{2mm}\verb!R'! is \verb!X^(and(R(X),X=A))![\verb!I!/\verb!X!], and\\
\hspace*{2mm}\verb!F'! is \verb!X^([J,...]:and(F,X=A))![\verb!T!/\verb!X!,
\verb!I!/\verb!X!].
\end{itemize}

\noindent
We also need a trivial rule for a formula with an uninstantiated
scoping constraint so that it reduces to application of other
rules:
\begin{itemize}\item[Q9]
{\it W}(\verb!_s:F!,v) if {\it W}(\verb!F!,v).
\end{itemize}

\noindent
Two rules are applicable to \verb!form! expressions,
corresponding to the cases of an uninstantiated or instantiated
resolution meta-variable:
\begin{itemize}\item[Q10]
{\it W}(\verb!F!,v) if {\it W}(\verb!F![\verb!_r!/\verb!R(P)!],v)\\
where:\\
\hspace*{2mm}\verb!F! is a formula \verb!form(C,R,_r)!\\
\hspace*{2mm}\verb!P! is a predicate such that {\it S}(\verb!C!,\verb!P!).
\end{itemize}
\begin{itemize}\item[Q11]
{\it W}(\verb!form(C,R,F')!,v) if {\it W}(\verb!F'!,v)\\
where \verb!F'! is a QLF formula.
\end{itemize}

In a more complete description of the semantics we would also have
to state that the evaluation rules provided give the only way of
determining membership of the relation {\it W}.

\section{Notes on the Semantics}
\label{monosem4}

\paragraph{Monotonicity:}
We are using monotonicity in two senses which (by design)
turn out to be consistent. The first is a syntactic notion for QLF
representations (instantiation rather than destructive manipulation),
while the second is semantic: 
\begin{enumerate}

\item \verb!F1!  is a more resolved version of \verb!F2! if \verb!F1!
can be obtained by instantiating zero or more meta-variables
in \verb!F2!.

\item \verb!F1! is a less partial interpretation than \verb!F2!
if [[\verb!F1!]] is an extension of [[\verb!F2!]].

\end{enumerate}
The claim of monotonicity for QLF is that for formulas \verb!F1! and
\verb!F2!, if \verb!F1! is a more resolved version of \verb!F2!
then \verb!F1! is a less partial interpretation than \verb!F2!.

\paragraph{Scoping Constraints:}
\label{quantremarks}
The quantification rules, (Q7) and (Q8), (i) select a term from 
a formula,
(ii) discharge all occurrences of the term and its index in the formula
and the term's restriction, replacing them by a variable, and (iii) apply
the term's quantifier to the discharged restriction and formula.  The
difference between (Q7) and (Q8) is simply that the latter 
also discharges
the head of the scoping list, in this case by removing it rather than by
replacing it. (Keep in mind that the discharge and replacement operations take
place at the level of the evaluation rules for QLF; they are not applied to
the QLF expressions representing natural language meanings themselves).

As with Lewin's scoping algorithm, (Lewin 1990),
there are no constraints built explicitly into the QLF semantics
on where a quantification rule for
a term may be applied, or indeed on the number of times it may be applied.
However, several constraints arise out of (a) the {\em absence} of any 
semantic rules for evaluating isolated \verb!term!s, term indices or scope lists, and 
(b) the requirement that a \verb!term! be selected from a formula so that its quantifier
is known.

The emergent conditions on legitimate scoping are
\begin{enumerate}
\item
No term may be quantified-in more than once: The first application of the
quantifier rule discharges the term. Subsequent applications of the rule 
lower down in the evaluation would fail to select an undischarged term.
\item
When a term's index occurs in a scope list, the quantifier rule for the
term must be applied at that point: It must be applied to discharge the head
of the scope list, and by (1) above cannot additionally be applied anywhere else.
\item
All occurrences of a term's index must occur within the scope of the
application of the term's quantifier rule: The quantification rule will only
discharge indices within the formula to which it is applied.  Any occurrences
of the index outside the formula will be undischarged, and hence unevaluable.
\item
If a term R occurs within 
the restriction of a term H, and R is to be given wide scope over the 
restriction, then R must also be given wide scope over H: Otherwise, suppose
H is given wide scope over R.  Term H will first be discharged, replacing the
term, and with it its restriction, in the formula to which the rule
is applied.  Then the quantification rule for R needs to be applied to the 
discharged formula, but the formula will not contain an occurrence of the term R,
making the rule inapplicable.
\end{enumerate}

The last two constraints have often been attributed to restrictions on
free variables and vacuous quantification.  The attribution is problematic
since open formulas and vacuously quantified formulas are both logically
well defined, and without suspect appeal to the syntax of the logical formalism
they cannot be ruled out as linguistically ill-formed.
By contrast, QLF makes these violations semantically unevaluable.

\paragraph{Unscoped Terms:}
When a term's index is not mentioned in any scope list, the term may be quantified
in at any point within the formula.  For anaphoric terms whose referent has
been resolved to some individual constant, it does matter where
the quantification rule is applied; since the term existentially quantifies over
things identical to a single object, the scope of the quantification is
immaterial.  It is thus convenient to leave anaphoric terms like this unscoped
in QLF.  Although this makes the QLF look (syntactically) as though it is
not fully resolved, semantically it is.
For other unscoped terms, alternative applications of the quantifier rule
may well lead to distinct truth conditions, and in these cases the QLF is
genuinely unresolved.

\paragraph{Context Dependence:}
\label{context-dependence}
Fully resolved QLFs are  context-independent in the same sense
that holds for closed formulas in traditional predicate logic (i.e.
if the interpretation of the constant symbols in the
language is fixed). Unresolved QLFs behave more like open formulas, and there
is an analogy between assignments to unbound variables in predicate
logic and possible resolutions of meta-variables admitted by 
the salience relation {\it S}. {\it S}(\verb!C!,\verb!H!) should be 
thought of as providing QLF expressions whose denotations are possible
referents for unresolved expressions with category \verb!C!.
(It would have been possible to define {\it S} as a direct relation between 
categories and referents, but this complicates the statement of
its role in resolution and in the semantic definitions.)
We used {\it S} above in the definition of QLF semantics, but it
is also central to NL processing: being able to compute {\it S}
can clearly play an important role in the process of reference resolution
during NL interpretation and in the process of building descriptions during
NL synthesis. (The computational analogue of {\it S} was implemented
as a collection of `resolution rules' in Alshawi 1990.)

An important question is what to allow as possible expressions
in the range of {\it S}. One observation is that as the range is widened,
more NL resolution phenomena are covered. A rough summary is as follows:
\begin{itemize}

\item constants: 
intersentential pronouns

\item predicate constants: 
compound nouns, prepositions

\item quantifiers: 
vague determiners

\item indices:  
bound variables, intrasentential pronouns

\item predicates built from NP restrictions: 
one-anaphora

\item predicates built from previous QLFs: 
intersentential ellipsis

\item predicates built from current QLF: 
intrasentential ellipsis

\end{itemize}

\section{Related Approaches}
\label{monosem6}

Viewed from a slightly different perspective, 
monotonic interpretation has a
number of points of contact with Pereira's categorial
semantics (Pereira 1990).  Put briefly, in categorial semantics, semantic
evaluation is represented as deduction in a functional calculus that derives the
meanings of sentences from the meanings of their parts. Considerable emphasis
is placed on the nature of these semantic derivations, as well as on the final
results of the derivations (the `logical forms' of sentences).

One significant advantage of this approach is that constraints on legitimate
scoping emerge naturally from a consideration of permissible derivations of
sentence meaning, rather than arising artificially from syntactic
constraints imposed on  logical forms. Derivations involving
quantified terms first introduce an assumption that allows one to derive a
simple term from a quantified term.  This assumption is later discharged by the
application of a quantifier.  Conditions on the appropriate introduction and
discharge of assumptions in natural deduction systems impose restrictions
on the way that quantifiers may legitimately be applied.  For example, a
quantifier assumption may not be discharged if it depends on further assumptions 
that have not themselves been discharged.  This prevents
the occurrence of free variables in logical form, but without appeal to the
syntax of logical form.

The discharge of terms and term indices when evaluating QLF closely parallels
the discharge of quantifier assumptions in categorial semantics. Indeed, the
terms and the indices are precisely the assumptions introduced by quantified
expressions, and which need to be discharged.  Furthermore, the
different orders in which quantifier assumptions may be discharged in categorial
derivation correspond to the choices that the quantifier rules permit for
discharging quantified terms.

Where monotonic interpretation and categorial semantics part company is on
the degree of explicitness with which semantic derivations are represented.
In categorial semantics, derivation is a background process that builds
up logical forms, but is not explicitly represented in the semantic formalism.
By contrast, the annotation of QLFs with scope lists provides an extra
level of information about how the derivations proceed.  In particular, they
indicate which evaluation rules should be applied where.

QLF thus provides a (usually partial) specification of a semantic derivation,
showing (a) what the initial `premises' are (roughly, lexical meanings, although
these too may only be partially specified), and
(b) the rules by which the `premises' are combined.  QLF resolution amounts
to further instantiating this specification.  This view of QLF can be
contrasted with
Logical Form as it is normally understood, which represents the results
of carrying out a semantic derivation.

The difference between specifying a derivation and carrying it out is what
makes resolution order independent in monotonic interpretation. Making a
resolution to QLF only specifies when and how an expression should be evaluated
during semantic derivation; it does not carry out that part of the derivation.
Where no distinction is drawn between making a resolution and carrying out the
corresponding step of the derivation, the order of resolution can be important.
Thus, for Dalrymple, Shieber and Pereira (1991), where this distinction is not
drawn, the precise interleaving of scope and ellipsis resolution determines
the interpretation of the sentence.  In QLF, resolutions dictate the order
in which various steps of the derivation are carried out, but the resolution
order does not reflect the derivation order.

Distinguishing between specifying and performing a derivation also means that
a monotonic treatment of ellipsis resolution does not need to resort to
higher-order unification.  Dalrymple, Shieber and Pereira use
higher-order unification to `unpick'
the composition of constituent meanings obtained in the semantic
derivation from the ellipsis antecedent.
Some of these meanings are then put back together to produce a 
predicate that can be applied to the ellipsis arguments. Since monotonic
resolution does not carry out the final composition of meanings,
but merely sets out conditions on how it is to take place, there is no
need to unpick composed meanings and put them back together again.  

\medskip

It is worth pointing out that
monotonic interpretation is compatible with approaches to meaning
as a transition between contexts or information
states, and where the order in which transitions are made
is significant (e.g. Veltman 1991).  
In such a framework, monotonic interpretation would amount to 
making decisions about which transitions to take when, but would not
involve putting those decisions into action.  The monotonicity in monotonic
interpretation thus refers to the way in which alternative derivations of
sentence meanings may be chosen, but not to the semantic effects of those sentence
meanings.

\section{Implementation Benefits}
\label{monosem7}

A description of the language processing mechanisms to which we have applied
the monotonic semantics model is beyond the scope of this chapter. However, 
we believe that the QLF representation presented here brings significant
advantages to implementing mechanisms for reference resolution, scoping, 
preference and generation.

\paragraph{Reference and Scoping:}
The order independence of resolution operations allows for a variety of
control structures in implementing a resolution mechanism.  We find it
convenient to make a bottom up pass through QLFs making reference resolutions,
followed by a stage of scoping resolution, and to iterate over this
should any of the resolutions introduce further unresolved expressions.

The salience relation {\it S}
can be implemented as procedures that search for properties, objects or
indices in context.  Scoping proceeds simply by the non-deterministic
instantiation of scoping constraints, subject to the restrictions imposed on
evaluable QLFs (Section~\ref{monosem5}), plus techniques for ignoring logically
equivalent scopings, as for example described by Moran (1988).

\paragraph{Preference and Disambiguation:}
A resolved QLF preserves all the information in the original unresolved QLF,
and also records the correspondence between resolved and unresolved
expressions. This makes it possible to define preference metrics that
can be used for ranking alternative interpretations independently of the search
strategies used to derive them. For example, in the case of scoping, these 
metrics can combine information about how far a quantifier was `raised' with 
information about the surface form of its determiner.
Preference ranking over alternative resolutions facilitates automatic
disambiguation of input.  Interactive disambiguation can make use of generation
from resolved QLFs for confirmation by a user.

\paragraph{Generation:}
There is a strong connection between monotonicity and reversibility
in language processing systems. Monotonicity of unification means that algorithms
such as head-driven generation (Shieber {\it et al} 1990) can be applied to grammars
developed for analysis. We use a variant of this algorithm for generating from
QLFs, and the monotonicity of semantic interpretation means that the grammar used for 
generating from unresolved QLFs (the normal `output' of the grammar) can also be
used for generation from resolved QLFs.

In parallel to the distinction between grammatical analysis (of NL into 
unresolved QLFs) and interpretation, we make the distinction between grammatical 
synthesis (of NL from QLFs) and {\em description}. 
Description is the process of
deriving a QLF from which synthesis proceeds by taking a fact (e.g. a database 
assertion) as input.
However, one of the principles of QLF-based description is that 
while interpretation instantiates referent fields in 
underspecified QLFs, description involves instantiating category and
restriction fields for QLFs in which referent fields are already instantiated.
The preference metrics applied to rank alternative interpretations can be applied
equally well to ranking resolved QLFs produced by a nondeterministic description 
process, so there is a sense in which the preference mechanism can also be made 
reversible.  This matter is further discussed in Chapter~\ref{generation}.

\chapter{The Domain and Context Models in Interpretation}
\label{clareinterp}

In the language understanding model embodied in CLARE, semantic processing
is divided into {\it analysis} and {\it interpretation}. The first corresponds
to the semantic contribution of the grammar, and yields contextually
underspecified QLF (Quasi Logical Form) representations. Semantic
interpretation further specifies these QLF representations through the 
processes of scoping and reference resolution. This interpretation process 
can take into account linguistic and nonlinguistic aspects of context,
including the domain model described in the previous chapters of this report.
In this chapter we describe the role played by the domain model in
interpretation.

This chapter is organized as follows.  In section~\ref{context-model} we 
briefly describe the context model, which is the part of the domain model that
records information about the utterances made by the user and the CLARE system.
Section~\ref{AET-refres} presents an abstract characterisation of reference
resolution in terms of abductive equivalential translation, using the domain
model as the background theory.  Sections~\ref{dm-resmethods} 
and~\ref{scp-func} then describe
a number of resolution methods and preferences that require inference using 
the domain model.  Finally, section~\ref{dm-qlfinterp} explains how the
domain model is sometimes also used when resolved QLFs are evaluated
in terms of TRL expressions.

\section{The Domain and Context Models}
\label{context-model}

The context against which any natural language utterance is interpreted
includes both non-linguistic facts about the world (e.g. that two entities
stand in a certain relation to one another) and linguistic facts about the
world (e.g. that the previous utterance in the discourse was made at a
certain time and had a particular surface form).  For convenience we can
divide the overall context into (1) the context model more narrowly defined, 
which records facts about the utterances in the discourse, and (2) the
domain model, which records other facts about the world.

Domain models have been described in some detail in earlier chapters.
Here we will only note that it is not strictly accurate to characterise
the domain model as capturing all and only the non-linguistic aspects of 
context.  From at least one perspective, the definitional equivalences 
contained in a linguistic domain theory can be seen as a way of relating
linguistic predicates / word senses to the underlying model.  However, it
will do us no harm here if we talk of the domain model as though it dealt
only with non-linguistic facts.

\paragraph{Context Model}
A variety of facts about utterances are recorded in the context model.
Each utterance is indexed by a unique identifier, and the information
recorded includes
\begin{itemize}
\item The surface form of the utterance (a word string)
\item The (resolved) QLF assigned to the utterance
\item The time at which the utterance was made
\item Who made the utterance
\end{itemize}
The QLF formalism also lends itself to recording further information
about the constituents of an utterance.  Individual noun phrases
within the utterance correspond to uniquely indexed \terms\ within the
utterance's QLF, and so in the context model extra information is associated
with \term\ indices.  Of importance below (section~\ref{dm-qlfinterp})
is the way that \term\ indices can be associated with internal domain
model names (skolem constants) for entities newly introduced in the domain
model.

The context model is used in three main ways. (1) To provide information
necessary for certain kinds of reference resolution.  For instance, records of 
QLFs for previous utterances are used to extract possible antecedents
for ellipses (Chapter~\ref{ellipsis}). Similarly, a list of salient \terms\
or entities is maintained, which is exploited in the resolution of
anaphora. (2)  To answer certain kinds of meta-question, such as
 {\em How many questions have you been asked?
How many have you answered?  Who asked them?}. (3) To assist in the evaluation
of (resolved) QLFs in terms of TRL expressions.

\subsection{Reverse Compilation of Linguistic Domain Theories}

The Linguistic Domain Theory (LDT, Chapter~\ref{claredm}) is the part of the 
domain
model that links linguistic predicates to domain predicates.  For
converting resolved QLFs to TRL domain queries and assertions, this link
is exploited in the linguistic to domain direction.  Since the linkage is
expressed by means of equivalences, it is also possible to use it in the
opposite direction, to go from domain predicates to linguistic predicates.
The ability to use the linkage in the reverse direction is useful
both for certain aspects of reference resolution, and in the generation
of English sentences from domain facts (section~\ref{genass}).

In practice, linkages in the domain to linguistic predicate direction are
obtained by compiling out the LDT to obtain the following:

\subsubsection{TRL Lemmas}\label{TRL-lemmas}

These are implications of the form
\begin{verbatim}
Conditions -> (TRLPredications -> QLFPredication)
\end{verbatim}
Subject to the \verb!Conditions! holding, this says that whenever the TRL
predications hold, the QLF predication holds.  As an example, we might have
\begin{verbatim}
true -> 
 and(project(C,B,D,E,F),transaction(G,H,A,I,J,D,K))
 -> on(A,B)
\end{verbatim}
which says that whenever \verb!A! and \verb!B! occur in the {\tt project}
and {\tt transaction} (conceptual) domain relations as shown, then
linguistically \verb!A! may be said to be {\em on} \verb!B! (where \verb!A! is
a cheque and \verb!B! is a project).

As an example of a lemma with non-trivial conditions, we might have
\begin{verbatim}
 and(SF(amount(I,sterling_unit))=A,
 and(SF(amount(J,sterling_unit))=H,
 and(safe_<(K,0),
     measure_diff(I,J,K))))
->
 transaction(C,D,B,E,F,G,H) -> under(A,B)
\end{verbatim}
This says that (a cheque) \verb!B! may be described as being {\em under}
an amount
\verb!A!, provided that \verb!A! is an amount of money, and that the amount
the cheque is for, \verb!H!, is less than \verb!A!.

\paragraph{Compilation}
The TRL Lemmas form the backbone of the reverse compilation of the linguistic
domain theory.  Other compiled implications and relations are derived in
part from TRL lemmas plus other LDT relations.  Compilation of the TRL Lemmas
is described in Chapter~\ref{ReasoningEngine}.

\subsubsection{Sort Declarations}

For each argument to a domain relation, it is useful to identify a set
of linguistic predications that can be applied to that argument.  This
information is already implicit in the TRL lemmas, since there
will be lemmas like
\begin{verbatim}
true -> transaction(C,D,B,E,F,G,H) -> cheque_Document(B)
\end{verbatim}
Lemmas like these can be regarded as sort declarations.

\subsubsection{Scope Functional Constraints}

The LDT contains a number of functional declarations, e.g. in the PRM
domain it is declared that there is a functional relation from projects
to their start dates.  This functional information is expressed in
terms of relations between arguments of domain predicates.  However,
it is also useful to have the same information expressed in terms
of linguistic predicates.

Linguistic functional information may be obtained by (1) finding lemmas that
connect functionally related domain arguments, (2) determining the linguistic
sorts on those arguments, and (3) creating a function declaration that 
state that arguments to the relevant linguistic predication are functionally
related provided that the appropriate sortal predicate apply.

As a concrete example, suppose we have
\begin{verbatim}
function(project(A,B,C,D),([A]-->[C])

true -> project(A,B,C,D) -> start_date_of(A,C)

true -> project(A,B,C,D) -> project_Activity(A)

true -> project(A,B,C,D) -> date_Time(C)
\end{verbatim}
We can then create a new, purely linguistic, functional declaration as follows:
\begin{verbatim}
and(project_Activity(A),date_Time(C))
->
 function(start_date_of(A,C),([A]-->[C]))
\end{verbatim}

\section{Reference Resolution and Abductive (Equivalential) Translation}
\label{AET-refres}

The category of a QLF \term\ or \form\ expresses a relation between
context, the \term\ or \form's restriction, and the \term\ or \form's
resolvent(s).  We can express this more formally in terms of abductive
equivalential translation, as described in Chapter~\ref{AET}.

The basic idea behind using the translation schema for interpretation
is that interpretation is a process of reducing the underspecification
in a formula \verb!Q! to produce a (possibly fully specified) formula
\verb!Q'!. In other words, we want \verb!Q'! to be semantically
subsumed by \verb!Q! such that by suitably restricting the set of models
under consideration (through knowledge of the context, meaning
postulates and possibly additional assumptions made during interpretation),
\verb!Q'! can then be made equivalent to \verb!Q! with respect to this restricted 
set of models.

This leads to the following characterization of interpretation in
terms of the translation schema: Given \verb!Q!, a possibly underspecified
formula corresponding to a speaker utterance, and a context \verb!C!, 
 determine \verb!Q'! and \verb!A! such that
\begin{quote}
 {\tt C} $\cup$ {\tt A} $\Rightarrow$ ({\tt Q} $\leftrightarrow$ {\tt Q'}) 
\end{quote}
where \verb!Q'! is syntactically subsumed by \verb!Q! and \verb!A! is a set
of assumptions consistent with \verb!C!. For a plausible interpretation
we wish to find fully resolved solutions \verb!Q'! which 
``minimize the cost'' of the assumptions \verb!A!. 

As an example of this general schema in operation, consider the resolution
of a sentence like {\em The prototype disintegrated}:
\begin{verbatim}
context(i1,C) /\ most_salient(i1,C,X^prototype(X),zed9) ... =>
    disintegrate(term(i1,<det=the>,X^prototype(X),_,_))
    <->
    disintegrate(term(i1,<det=the>,X^prototype(X),exists,zed9)).
\end{verbatim} 
Here, \verb!C! is the context that is current for the utterance of the \term\
\verb!i1!.  Relative to this context, we are supposing the \verb!zed9!
is the most salient prototype.  This supposition may either (a) already 
follow from the context \verb!C!, or (b) be an extra assumption that we
make about the context.  The latter case shows our treatment permits a 
`relational' account of meaning --- interpreting an utterance may serve to
fix the context as much as it fixes the meaning of the
utterance.  Thus, if it is not already the case that \verb!zed9! is the
most salient prototype in \verb!C!, at the cost of making an assumption
we can resolve the \term\ \verb!i1! to refer to \verb!zed9! and force it to be
the most salient prototype (the cost of the assumption will obviously depend
on how salient \verb!zed9! already is in context \verb!C!).

For reference resolution solely in the interpretive direction (i.e. where
we are not interested in driving resolution backwards for the sake of
synthesis and generation), it is convenient to reduce the equivalential
schema to Horn clauses.  The Horn clause relating specifically to the 
resolution of the definite \term\ above would be something like
\begin{verbatim}
context(I,C) /\ det(I,C,the) /\ restriction(I,C,R) 
             /\ most_salient(I,C,R,Rft)
   -> referent(I,C,Rft) 
\end{verbatim}
Here, the functions \verb!det!, \verb!restriction! and \verb!referent!
serve to identify the different components of the \term\ indexed by
\verb!I!, and \verb!context! and \verb!most_salient! are as before.
Note that the Horn clause formulation is entailed by the equivalence,
but not vice versa.

The predicate \verb!most_salient! is in fact a {\em resolution method},
suitable for the resolution of definite noun phrases.  In practice,
the resolution method will be implemented as a Prolog procedure, which may well
call upon the services of the domain model inference engine.  Also, Horn clause
axioms of the form shown above are in fact represented in a short-hand form
as reference rule declarations.  These declarations, of the form
\begin{verbatim}
ref_rule(term(I,Cat,Restriction,Quant,Referent),
         Method).
\end{verbatim}
associate resolution methods, like \verb!most_salient! with \terms\ (or
\forms) of various categories.   Several different resolution methods may
be associated in this way with a single category.  When this is so, the 
relation the category expresses between the \term\ or \form's restriction
and the context is the union of the relations expressed by the
different resolution methods.

With the general translation schema for \term\ referents in place, we can
also slightly reformulate the QLF evaluation rules for unresolved \terms\
(Chapter~\ref{monosem}).
The revised version of the valuation rule is as follows (the valuation
relation V has the same arguments as W but also takes an 
assignment \verb!g!):
\begin{quote}
{\it V}(\verb!F(...term(I,R,exists,_)...)!,m,g,v) if\\
\hspace*{10mm}\verb!g(_)!=\verb!A! and\\
\hspace*{10mm}{\it V}(\verb!F(...term(I,R,exists,A)...)!,m,g,v) and\\
\hspace*{10mm}{\it V}(\verb!referent(I,A)!,m,g,1).
\end{quote}
This valuation rule together with the definite description postulate
given above allows the solution presented earlier for the interpretation
of {\it The prototype disintegrated}.
This approach has the advantage of making notions like salience part of
the linguistic theory of reference rather than of the valuation rules for
the logic.

One can also apply abductive translation to the resolution of quantifier
scope.  Consider a sentence like {\it A nut secures every bolt}, and the
following translation schema:
\begin{verbatim}
context(U,C) /\ 
function(C,X^Y^and(nut(X),and(bolt(Y),secures(X,Y))))... |=
      _:secures(term(i2,<det=a>,X^nut(X),_),
                term(i3,<det=every>,Y^bolt(Y),_,_))
       <->
[i3,i2]:secures(term(i2,<det=a>,X^nut(X),exists,i2),
                term(i3,<det=every>,Y^bolt(Y),forall,i3)).
\end{verbatim}
Here, \verb!U! identifies the utterance (\verb!U! will normally be the
index of a \verb!form!); 
\begin{verbatim}
function(X^Y^p(X,Y))
\end{verbatim}
is to be read as meaning
\begin{verbatim}
forall([X,Y1,Y2],
   impl(and(p(X,Y1),p(X,Y2)),
        Y1=Y2))
\end{verbatim}
The function assertion captures the domain fact that a nut 
cannot secure more than one bolt.
This, together with a cardinality assumption saying there is more than one 
bolt (left implicit here, and implied by the use of the determiner 
{\em every}), forces the universal to take wide scope.
The reverse scoping would violate the functional constraint by having at least
one nut that secures all the bolts.

Functional constraints are in practice also applied in an implicational rather
than equivalential form.  Moreover, they are generally applied in the
contrapositive direction in order to rule out scoping that violate
functional constraints rather than promote scopings that obey them.  The
details of this are given in section~\ref{scp-func} below.

\section{Domain Specific Resolution}
\label{dm-resmethods}

Deciding whether a particular resolution method relates a QLF index to
a referent via the context requires a certain amount of inference.  In
some cases the amount of inference required may be relatively small; for
example, most ellipsis resolution (see Chapter~\ref{ellipsis}) requires
only identifying an antecedent QLF and deciding on a range of substitutions
to be made to it.  In other cases, however, more extensive inference using
the domain model is required.  In this section we discuss four cases where
more extended inference is required: anaphora resolution, compound noun
and prepositional phrase resolution, scoping preferences, and general
domain preferences.

\subsection{Anaphoric and Definite Noun Phrases}

To resolve an anaphoric or definite noun phrase, one needs to find
an entity (or set of entities) that are salient in the current context,
and which satisfy some minimal description obtained from the \term's
restriction.  Entities may be salient in context in the following ways:
(a) they have recently been referred to by some \term\ in the discourse,
(b) they are functionally related to some previously mentioned entity 
(unimplemented),
(c) they are the only objects in the domain having a certain property.

\paragraph{Pronouns}
Pronouns like {\em it, he} or {\em she} look for recently mentioned
entities which satisfy the properties {\tt impersonal}, {\tt masculine} and
{\tt feminine} respectively.
Which entities have recently been mentioned is inferrable from the context
model simply by looking through the most recently uttered \terms\ as recorded
there.  A limited amount of inference using the
domain model is also required to determine whether the requisite
properties apply, in a way now described.

At the QLF level, pronoun resolutions are expressed by means of co-indexing
QLF \terms.  For a discourse like ``{\em John walked. He talked.}'' we
might therefore have the following pair of QLFs (simplified)
\begin{verbatim}
walk(term(i1,proper_name,X^name_of(X,'John'),exist,ent(john_smith)))

talk(term(i2,he,Y^masculine(Y),exists,strict(i1)))
\end{verbatim}
For the resolution of the second QLF to be acceptable, we must establish
that the entity referred to by the \term\ \verb!i1! satisfies the property
\verb!Y^masculine(Y)!.   Given that \verb!i1! is a directly referential
\term, there are two alternate queries that may be addressed to the
inference engine:
\begin{verbatim}
masculine(john_smith)

forall([X],impl(name_of(X,'John'),masculine(X)))
\end{verbatim}
(note that the queries addressed to the inference engine are first
converted from QLF to TRL). 

In a discourse like ``{\em A man walked. He talked.}'' the first query above
cannot be formulated, since {\em a man} will not be resolved to refer
to a specific named entity.  However, as discussed 
below, the indefinite antecedent will lead to the introduction of some 
(possibly) 
new entity into the domain model, having its own internal name.  Direct
use of internal names are not permitted in QLF, in part because
internal names keep changing.  However the domain model does maintain a list
of assertions of the form
\begin{verbatim}
ctx_idx_assoc(Idx,IntName)
\end{verbatim}
which associates \term\ indices (\verb!Idx!) with internal names for objects
(usually skolem constants or functions).  So in this case, there are 
again two queries that may be tried:
\begin{verbatim}
ctx_idx_assoc(i1,X) /\ masculine(X)

forall([X],impl(man(X),masculine(X)))
\end{verbatim}
(In fact the first query generalizes the more specific query {\tt 
masculine(john\_smith)} above, since in that case, the domain model
will imply {\tt ctx\_idx\_assoc(i1,john\_smith)}).

Pronouns may also be used in a non-referential way, e.g. {\em John gets a lot
of headaches, and I get them too}.  Here, the pronoun {\em them} does
not refer to the headaches that John gets but just to headaches in general.
In cases such as these, the pronoun is resolved in such a way that 
it quantifies over objects satisfying the same restriction property as
the antecedent (or a more general description entailed by the antecedent's
restriction), e.g.
\begin{verbatim}
term(i4,them,X^entity(X),exist,sloppy(i3,Y^headache(Y)))
\end{verbatim}
Strictly speaking, two queries should be addressed to the inference engine:
\begin{verbatim}
and(restriction(i3,X^P),
    forall([Y], impl(P(Y), headache(Y))))

forall([Y], impl(headache(Y),entity(Y)))
\end{verbatim}
In practice, only the second (trivial) query would be used, since the
resolution property is normally obtained by removing conjuncts from the 
antecedent restriction, syntactically ensuring the first entailment.  As it is,
this way of resolving pronouns is not currently implemented within CLARE.

\paragraph{Definite Noun Phrases}

One way of resolving definite noun phrases is very similar to that for
pronouns: select a salient antecedent \term, and ensure that any entities
it refers to are compatible with the restriction of the definite \term.

When the antecedent refers to a single named entity, as before one can simply 
apply the definite restriction to the entity and check against the domain
model that the resultant predication is true.  Thus, for {\em John walked.
The man talked.}, we would need to establish that
\begin{verbatim}
man(john_smith)
\end{verbatim}
But when the antecedent does not refer to a single named entity, the 
compatibility check needs to be more elaborate.  

In essence, we need
to check that a full description of the antecedent is not incompatible
with the restriction of the definite \term.  By full description, we
mean not just the description given by the restriction of the antecedent
\term, but a more specific description of the \term\ as it occurs in its
sentence. To illustrate why a full description is needed, consider
{\em Six people work on CLARE.  The people who do not work on CLARE work on 
BCI}.  Obviously, we would not wish to resolve the definite NP so that
it refers to (some of?) the six people who work on CLARE.  Yet if
we just look at the antecedent's restriction (i.e. people) there is nothing to
prevent such a resolution.  The full description of the antecedent 
is `people that work on CLARE', which is
clearly incompatible with the description `people who do not work on CLARE'.

For a fully correct treatment of definites, two queries would be required
(here \verb!A! stands for the full antecedent description, and \verb!R!
for the restriction of the definite term
\begin{verbatim}
not(forall([X], impl(A(X), not(R(X)))))

forall([X], impl(R(X), A(X)))
\end{verbatim}
The first query ensures consistency, and the second may be used to favour 
resolutions where the definite term restriction is as specific
or is more specific than the antecedent description.  The preference
query is required in cases such as {\em One man walked and another man
ran. The man that walked whistled} --- whistling is not inconsistent with
running.

In practical terms, however, these queries often turn out to be excessively
complex to answer, especially the consistency checking. Instead, much simpler
queries are formulated.  The head predicates (corresponding to the head
nouns) for the definite and antecedent are retrieved, and it is checked that
the antecedent head entails the definite head.  This is error prone when
the definite is modified, e.g. by a relative clause, but in such cases there
is a general preference in favour of attributive rather than anaphoric
resolutions.

Singular definites may also be resolved referentially if there is a single
object in the domain satisfying the definite restriction.  For example
{\em the start date of CLAM-BAKE} may be resolved to a particular date,
even if that date has not previously been mentioned.  In cases such as
these the definite \term's restriction is used to formulate a query 
\begin{verbatim}
extension_of(R,Ext)
\end{verbatim}
that determines the extension \verb!Ext! of the \term\ restriction \verb!R!
in the domain model.  If there is just a single object in the extension,
this may serve as the referent.

Functional resolutions of definites are not currently implemented.  To do so
would require using functional declarations in the LDT to link previously
mentioned entities to functional dependents satisfying the restriction
of the definite \term.  

In formulating queries for the resolution of definites, it is necessary
to ensure that the \term\ restriction is fully resolved and scoped,
and then translated into TRL.
(Though partially resolved QLFs do have well defined truth conditions,
no proof theory for them has yet been worked out).

\subsection{Compound Nouns and Vague Relations}

In the PRM domain, one would expect the following different resolutions
for two syntactically and semantically similar compound nouns
\begin{quote}
{\it company payments} $\Rightarrow$ payments to companies\\
{\it project payments} $\Rightarrow$ payments on projects
\end{quote}
The reason for the difference is that the domain model determines that
(a) the relation between companies and payments is one where the payments
are made to the companies and (b) the relation between projects and payments
is one where the payments are made on a project account.  

Information encoded
in the domain (in non-linguistic form) is thus used to resolve
compound nominals like the ones above.  In brief, the procedure is as 
follows.\\
(1) The head property of the compound noun (i.e. payments in the case above) 
is used to retrieve, via sortal TRL lemmas, the kind of domain relations that
payments can occur in. \\ 
(2) Chaining declarations, which specify how various
domain relations intersect, are then used to select further domain relations
that may be connected to the original.\\  
(3) Sortal TRL lemmas are then used again
to determine which arguments on the connected relations could be
described by the other property mentioned in the compound nominal (i.e.
projects or companies).\\
(4) At this point, one will have identified one or more two place
domain relations, e.g.
\begin{verbatim}
A^B^and(transaction(A,_,_,_,C,_,_),project(B,C,_,_))
\end{verbatim}
The TRL lemmas can once more be consulted to determine what
linguistic predicates (if any) correspond to these properties.

Vague preposition senses may be resolved in exactly the same way.  The only
significant difference is that in some cases the second argument to the
preposition may be a referential \term\ (e.g. for a proper name).  Here,
the domain model has to be consulted directly to see which relations that
named entity may occur in, rather than using sortal TRL lemmas.

\section{Functional Information and Scoping} \label{scp-func}

The domain model contains declarations concerning functional relations between
different objects.  For example, a project might uniquely define a start and end
date.  This information can be used to rule out linguistically possible quantifier
scopings that are impossible in the domain.  A query like
{\it List every start date of a project} has two linguistically possible scopings:
(a) for one project, list all of its start dates, and (b) list all the dates on
which some project started.  Given the functional relation between projects
and their start dates, only the latter reading is plausible.

In the functional relation above, projects are the {\it domain}
of the function, and start and end dates the {\it range}.  We need
to distinguish between {\it iterative} quantifiers (universals, distributive
plurals) and non-iterative quantifiers (existentials, plural collectives).
We also need to be able to say whether one term originates from the restriction
of another, i.e. whether they are in a  {\it head--restriction} relation.  We can
then specify functional constraints on scoping between two terms as follows.
\begin{enumerate}
\item
Domain and range terms not in head--restriction relation
\begin{enumerate}
\item
Iterated range term within scope of domain term disallowed
\item
Iterated range term with scope over iterated domain term disallowed
\end{enumerate}
\item
Domain term in restriction of range term 
\begin{enumerate}
\item
Iterated plural range term within scope of domain term disallowed. (Universals
allowable, since they can be used to refer to single object, though this is
dispreferred.  Numerical quantifiers etc. disallowed).
\item
Iterated range term with scope over iterated domain term disallowed
\end{enumerate}
\item
Range term in restriction of domain term
\begin{enumerate}
\item
Iterated range term within scope of domain term disallowed
\item
Iterated range term with scope over iterated domain term {\em is} allowed.
\end{enumerate}
\end{enumerate}

Examples of scopings affected by these restrictions are
\begin{quote}

{\it A project started on every date} disallowed if {\it a project} takes wide 
scope. (Violates 1a)

{\it Every start date of a project} dispreferred if {\it a project} takes wide 
scope. (Dispreferred by 2a)

{\it Three start dates of a project} disallowed if {\it a project} takes wide 
scope. (Violates 2a)

{\it A project starting on every date} disallowed if {\it a project} takes wide 
scope. (Violates 3a)

{\it Every project started on every date} disallowed on any scoping. 
(Violates 1a or 1b)

{\it Every start date of every project} disallowed if {\it every start date} 
has wide scope. (Violates 2b).

{\it Every project starting on every date} disallowed if {\it every project} takes
wide scope (violates 3a), but OK if {\it every date} has wide scope (3b).

{\it Every project started on a date} is OK under any scoping, though giving
{\it a date} wide scope is dispreferred on other grounds.

{\it A start date of every project} is OK under any scoping, though giving
{\it a date} wide scope is dispreferred on other grounds.

{\it Every project starting on a date} is OK under any scoping, though giving
{\it a date} wide scope is dispreferred on other grounds.

\end{quote}

The constraints on functional relations are applied during preference ranking
of scoped forms (Chapter~\ref{clarepref}).  In order to establish whether
a functional restriction applies to a given predication, a query to
this inference engine is constructed as follows:\\
(1) all one-place linguistic predications applying to the predication's
arguments in the QLF are conjoined with the predication itself.\\
(2) A query is produced to see if this conjunction implies the conditions
on the compiled scope function declaration for the predication.

\section{The Context Model in Evaluating Resolved QLFs}\label{dm-qlfinterp}

As well as providing information necessary for the resolution of 
certain QLF expressions or determining the preference rankings for
alternate resolutions, the context model also plays an important (though
limited) role when it comes to evaluating QLFs through conversion to
TRL expressions.  This applies particularly to the way that anaphorically
resolved \terms\ are evaluated, and in this connection two pieces of
contextual information recorded about (antecedent) \terms\ are of
importance --- the \term\ index's domain model association, and the full
description associated with the index.

\subsection{Index--Domain Entity Associations}

The role of index associations is best understood by considering the effect
that assertions involving indefinite noun phrases have on the domain model.
An assertion like
\begin{quotenr}
A woman works on FOO.
\end{quotenr}
will, after resolution and conversion to TRL, lead to the addition of 
skolemised clauses to the domain model, e.g.
\begin{quote}\begin{verbatim}
CRI_PROJECT_MEMBER(sk(89),FOO).
CRI_EMPLOYEE(sk(89),w).
\end{verbatim}\end{quote}
The skolem constant, \verb!sk(89)!, represents an internal name that the
domain model assigns to the as yet unidentified woman.

The assignment of an internal name to the referent of the indefinite noun
phrase occurs only after the resolution of the QLF for assertion.  At
the QLF level, all we have is some \term\ with a uniquely identifying
index.  It is 
important to associate the QLF \term\ index with the internal domain model
name.

The necessity for this association can be seen if we consider a follow-on 
assertion like
\begin{quotenr}
She also works on CLAM-BAKE.
\end{quotenr}
At the QLF level, the pronoun will be resolved so that it is co-indexed
with the antecedent \term:
\begin{quote}\begin{verbatim}
term(<pro=she..>,idx77,X^[female,X],exists,strict(idx73))
\end{verbatim}\end{quote}
where \verb!idx73! is the index of the \term\ corresponding to {\em a woman}.
The QLF resolution makes no reference to the internal names used in
the domain model.  (This is of vital importance, since the technique
de-skolemising and re-skolemising incomplete assertions means that the
internal names constantly change; see Chapter~\ref{translation}).  
However, we need to make sure that
when the QLF is converted to TRL we have the following (still incomplete)
clauses:
\begin{quote}\begin{verbatim}
CRI_PROJECT_MEMBER(sk(89),CLAM-BAKE).
CRI_PROJECT_MEMBER(sk(89),FOO).
CRI_EMPLOYEE(sk(89),w).
\end{verbatim}\end{quote}
so that it is the same woman who works on CLAM-BAKE and on FOO.

To achieve this effect, we proceed as follows.  During the conversion of
the first assertion to TRL, we will first arrive at an untranslated TRL
expression (i.e. before equivalential translation) of the form
\begin{quote}\begin{verbatim}
exists([X],and(and(woman(X),works_on(X,FOO)),
               ctx_idx_assoc(idx73,X))).
\end{verbatim}\end{quote}
That is, the QLF \term\ interpretation rules ensure that an extra conjunct
is added when interpreting all \terms\, associating the \term's index
with the bound variable produced in discharging occurrences of the \term.
After the untranslated TRL undergoes equivalential translation and
skolemisation, we will in fact have three clauses that are added to the domain
model:
\begin{quote}\begin{verbatim}
ctx_idx_assoc(idx73,sk(89)).
CRI_PROJECT_MEMBER(sk(89),FOO).
CRI_EMPLOYEE(sk(89),w).
\end{verbatim}\end{quote}
The first of these has the effect of including information in the domain model
about certain aspects of the utterance.

In discharging the resolved pronominal \term\ to a TRL quantification,
the index association for \verb!idx73! is used as follows.  The \term\
is strictly co-indexed with \verb!idx73!, which in turn is associated
with the skolem constant \verb!sk(89)!.  Therefore, in discharging the \term\
an extra conjunct is added:
\begin{quote}\begin{verbatim}
exists([X], ...and(and(female(X),eq(X,sk(89))),
                   ctx_idx_assoc(idx77,X))
            ....)
\end{verbatim}\end{quote}
After translating, simplif
ying identities and re-skolemising all the currently
incomplete assertions, this will lead to the required set of skolemised
clauses being produced. 

Should we make a third assertion such as
\begin{quotenr}
Her name is Clara Luser
\end{quotenr}
the way that incomplete assertions are de-skolemised and then re-skolemised
would ensure that we end up with the following domain model clauses:
\begin{quote}\begin{verbatim}
ctx_idx_assoc(idx73,Clara Luser).
ctx_idx_assoc(idx77,Clara Luser).
CRI_PROJECT_MEMBER(Clara Luser,FOO).
CRI_PROJECT_MEMBER(Clara Luser,CLAM-BAKE).
CRI_EMPLOYEE(Clara Luser,w).
\end{verbatim}\end{quote}
At this point, the internal name for the woman is replaced by a (unique)
external name.

In principle, this treatment can be extended to cases where indices are
associated with skolem functions, or implicitly universally quantified
TRL variables (re-skolemisation of the results would ensure these variables
are properly `bound').  However, this runs into a number of technical 
difficulties with regard to interpreting collective versus distributive
anaphor resolutions, and an alternative approach, described below, is adopted.

\subsection{Full \term\ Descriptions}

To understand what is meant by restricted and full descriptions of a \term,
it is easiest to present an example.  Consider the noun phrase {\em a dog that
bit him} in the following sentence
\begin{quotenr}
Every man kicked a dog that bit him
\end{quotenr}
A restricted description of the entities referred to by this noun phrase
is
\begin{quotenr}
dogs that bit a man
\end{quotenr}
A full description would be
\begin{quotenr}
dogs that bit a man and were kicked by him
\end{quotenr}

The full description can be used to interpret the resolution of the 
pronoun {\em they} in
\begin{quotenr}
They all ran away
\end{quotenr}
At the QLF level, the pronoun {\em they} is simply resolved to be 
co-indexed with the antecedent \term, e.g. for {\em a dog that bit him}. 
Supposing that the antecedent \term\ gave rise to a skolem
constant associated with the \term\ index, then this resolution could be
interpreted in the way described in the last section.  That is, we refer
directly to those entities referred to by the antecedent \term.

However, in this case the antecedent is associated with a skolem function,
and we employ a different method of interpretation.  This is simply to
impose the full antecedent \term\ description as an additional restriction
on the range pronominal \term's quantification.  That is, the sentence
is interpreted as meaning something like: the dogs that bit and were kicked
by a man all ran away.  Here we rely on a sufficiently detailed 
description to pick up exactly those entities referred to by the antecedent
\term.

Unfortunately, we cannot always guarantee that the description will be 
sufficiently constraining to pick up exactly the same entities.  This
is especially pressing with the case of indefinite / existential terms.
For example, in a sentence like {\em A man kicked a dog}, the full description
for {\em a dog} would be `dogs that were kicked by a man'.  Even if we
anchor the temporal reference of the sentence down to a particular time,
it is quite possible that there is more than one dog that was kicked by a man
at that time.  But the sentence itself refers to just one man and one dog.
Fortunately, it is exactly in cases such as these that the association of
indices with skolem constants can be used to interpret subsequent anaphoric
references, avoiding the problem of insufficiently constrained descriptions.

When descriptions are used it is because the antecedent \term\ either is
equal to 
or is within the scope of some \term\ that behaves like a universal
quantifier.  In such cases, it is much more likely that the \term\ description
will be sufficiently constrained.  So, for {\em a dog} in {\em Every man
kicked a dog}, the description `dogs that were kicked by a man' is more likely
to be sufficiently constrained, provided that the temporal reference of
the sentence and description is anchored down, and provided that any further
contextual restrictions on {\em every man} are also captured in the 
description.  Also, descriptive interpretations are less liable to lead to
inaccuracy in evaluating questions than they are for assertions.

A full description for a term {\tt term(i,..X\^R..)} in a formula
{\tt S(..i..)} is 
\begin{verbatim}
X^and(R'(X),S'(..X..))
\end{verbatim}
Here, \verb!R'! is the \term's restriction plus any other contextual 
restrictions imposed by the \term's resolution. \verb!S'! is the original
QLF with the \term\ and all occurrences of its index abstracted over.
Furthermore, all universally quantified \terms\ with wide scope over \verb!i!
are exchanged for existentials.  Full descriptions for \terms\  
correspond closely to Webber's (1983) discourse descriptions.

\chapter{Resolution of Ellipsis}
\label{ellipsis}

\section{Analysis of Ellipsis}
\label{ellipana}

Elliptical input, in the form of fragmentary 
input and certain \inx{coordination} constructions, 
is common in natural language \inx{discourse}, 
especially typed interactive discourse of the kind a
user of CLARE would engage in, e.g.
\begin{quote}
{\it John, Which house?, Inside, On the table, Difficult to do,\\
John doesn't, He might not want to.}
\end{quote}
Syntactically it is not essential to treat these fragments as instances of
S constituents, although for our purposes it is convenient to do so.

Many elliptical fragments can be analysed by allowing an
elliptical S to consist of one or more phrases (NP, PP, AdjP, AdvP) or their
corresponding lexical categories.  Most other commonly occurring patterns can
be catered for by allowing verbs that subcategorize for a nonfinite VP
(modals, auxiliary {\it do}, {\it to}) to appear without one, and by adding a
special lexical entry for a main verb {\it do} that allows it to constitute a
complete VP.  In our grammar, the latter two moves allow all of the following
to be analyzed:
\begin{quote}
{\it Will John?, John won't, He may do, He may not want to, Is he going to?
etc.}
\end{quote}

At the QLF level, elliptical fragments are analysed as \forms, where the
\form\ category gives information about the kind of ellided constituent.
For example, for an elliptical NP like ``John'' we would get the
following QLF (square bracket notation used):
\begin{quote}
{\em John}:
\begin{verbatim} 
form(l([John]), ell(np), +e,
     P^[P, term(l([John]), proper_name, +j
                X^[name_of,X,'John'],
                _,_)],
     _)
\end{verbatim}
\end{quote}
To resolve the elliptical \form, one needs to find a property salient in 
context, and apply the \form's restriction (\verb!P^[P,term(..+j..)]!)
to the salient property.   Either the property itself or the
result of its application is used to instantiate the \form's resolvent
meta-variable.

For ellipsis resolution thus construed, the hard part lies in determining
what counts as a contextually salient property suitable for
resolving the ellipsis.

\section{Ellipsis Resolution}

The nature of ellipsis in English is that 
its resolution is strongly dependent on the local linguistic context, as
provided by the current sentence, the preceding sentence, or the preceding
exchange in a turn-taking dialogue.
The basic claim we wish to defend in this chapter is as follows:
The properties suitable for ellipsis resolution can be obtained by 
the substitution of \terms, \forms\ and their indices in the QLFs of the
sentences contained in the local linguistic context.

In some ways this claim is too strong.  For example,
someone can shout `Stampede!', and without any prior linguistic context
this can be interpreted as meaning `Look out everybody, there is a cattle
stampede'.  However, we will ignore here cases of ellipsis with non-linguistic
antecedents.  More importantly, the claim may be too strong in that there
are some cases of ellipsis resolution that go beyond simple substitution
of \terms, \forms\ and their indices in QLFs.  

However, as we will see below,
a substitutional treatment covers a wide range of cases, and with a greater
degree of simplicity than many alternative approaches.  In particular, while
bearing a close connection to approaches using higher-order unification
(Dalrymple, Shieber and Pereira 1991), a substitutional treatment avoids
the computational complexity of using greater than second-order matching to
account for certain interactions between ellipsis and quantifier scope.

In this section, we will first present a number of examples 
illustrating a substitutional treatment of ellipsis.
We will then discuss the use of higher-order unification by Dalrymple, 
Shieber and Pereira (1991) to cover the same kinds of  example, and then
comment on the connections between the two approaches.

\subsection{Examples of Substitutional QLF Ellipsis}

\subsubsection{A Simple Example}

To present the basic ideas behind a substitutional  approach to ellipsis 
resolution in QLF, let us consider the following simple example
\begin{quote}
John slept.
So did Mary.
\end{quote}
The resolved QLF for the first sentence and the unresolved QLF for the
second are given below, in a heavily abbreviated and simplified form.
The reason for this simplification is that there are a number of complications
to do with mood and tense that we do not yet wish to mention.
\begin{quote}
Antecedent:
\begin{verbatim}
form(<type=verb,...>, +s, 
     P^[P,[sleep, term(+j,...,john)]],
     [sleep, term(+j,...,john)])
\end{verbatim}

Ellipsis:
\begin{verbatim}
form(<type=vp_ellipsis,...>, +e,
     R^[R,term(+m,...,mary)],
     _)
\end{verbatim}
\end{quote}
Since the ellipsis has the category of VP ellipsis, it is resolved by 
finding a verb \form\ from a previous QLF, and making the following
interpretive substitutions.

First, the subject \term\ of the antecedent QLF is interpreted as though
it were the subject of the ellipsis.  This means that wherever the
subject \term\ or its index occurs in the antecedent, it should be 
re-interpreted as though it were the ellipsis subject \term\ or its index.
Second, the index of the antecedent \form\ also needs to be replaced since
the ellipsis amounts to a re-interpretation of the original \form.  Since
the re-interpreted antecedent is intended to be the resolution of the 
elliptical \form, it makes sense to substitute the antecedent's index
by the ellipsis' index.

We can therefore represent the resolved elliptical QLF as follows:
\begin{quote}
\begin{verbatim}
form(<type=vp_ellipsis,...>, +e, R^[R,term(+m,...,mary)],
     form(<type=verb,...>, +s, 
          P^[P,...],
          [sleep, term(+j,...,john)]):
     {+j/+m,
      term(+j,...,john)/term(+m,...,mary),
      +s/+e})
\end{verbatim}
\end{quote}
The resolvent field is the antecedent verb \form\ followed by a list of
substitutions, \verb!{+j/+m,...}!.

\paragraph{Substitutions}
It should be emphasised that these are `interpretive' substitutions, rather
than syntactic substitutions to be made in place on the antecedent
\form.  The substitution \verb!term(+j,...,john)/term(+m,...,mary)! says
that during the course of re-interpreting the antecedent \form, whenever
you come across an interpretive rule that makes reference to
\verb!term(+j,...,john)!, you should re-cast it as making reference
to \verb!term(+m,...,mary)!, and similarly for the indices {\tt +j},
{\tt +m}, {\tt +s} and {\tt +e}.

More specifically, in the course of interpreting the original antecedent
\form\ we might have employed the following rule for an (unscoped) \term:

\begin{quote}
{\it W}(\verb!F!,v) if {\it W}(\verb!Q(R',F')!,v),\\
where:\\
\hspace*{2mm}\verb!F! is a formula containing the term\\
\hspace*{6mm}  \verb!T!=\verb!term(+j,C,R,Q,john)!,\\
\hspace*{2mm}\verb!R'! is \verb!X^(and(R(X),X=john))![\verb!X!/\verb!+j!], 
and\\
\hspace*{2mm}\verb!F'! is \verb!X^(and(F,X=john))![\verb!X!/\verb!T!,
\verb!X!/\verb!+j!].
\end{quote}
Under the extra interpretive substitutions imposed by the ellipsis resolution,
we now wish to treat the term and its index as though they were
{\tt term(+m,...,mary)} and {\tt +m} respectively.  To do so, we abstract
over the original \term\ and index as before, but also make sure we abstract
over the new \term\ and index, and also use the restriction of the new \term:
\begin{quote}
{\it W}(\verb!F!,v) if {\it W}(\verb!Q2(R2',F')!,v),\\
where:\\
\hspace*{2mm}\verb!F! is a formula containing the term\\
\hspace*{6mm}  \verb!T1!=\verb!term(+j,C1,R1,Q1,john)!,\\
\hspace*{2mm} \verb!T2!= \verb!term(+m,C2,R2,Q2,mary)!,\\
\hspace*{2mm}\verb!R2'! is \verb!X^(and(R2(X),X=mary))![\verb!X!/\verb!+j!,
\verb!X!/\verb!+m!], 
and\\
\hspace*{2mm}\verb!F'! is \verb!X^(and(F,X=mary))![\verb!X!/\verb!T1!,
\verb!X!/\verb!T2!,\verb!X!/\verb!+j!,\verb!X!/\verb!+m!].
\end{quote}
(Note that for each substitution in the semantic interpretation rule,
the ellipsis substitutions leads to an addition of a corresponding new one).

Provided that the ellipsis antecedent is fully resolved, it is possible
to treat the ellipsis substitutions as though they were ordinary
syntactic substitutions.  But the order of application of the substitutions
does not depend on the order in which they are listed.  Instead, they depend
on the order in which the substituees occur in (the recursive, top-down
descent through) the antecedent QLF.  This is unlike the normal notation
for substitutions, where {\tt \{+j/+m,term(+j,...,john)/term(+m,...,mary)\}}
would mean `first replace {\tt +j} by {\tt +m}, and then replace
{\tt term(+j,...,john)} by\\ {\tt term(+m,...,mary)}'.

If we were to spell these substitutions out syntactically, the resolved 
ellipsis would be equivalent to:
\begin{quote}
\begin{verbatim}
form(<type=vp_ellipsis,...>, +e, 
     R^[R,term(+m,...,mary)],
     form(<type=verb,...>, +e, 
          P^[P,[sleep,term(+m,...,mary)]],
          [sleep, term(+m,...,mary)]))
\end{verbatim}
\end{quote}
i.e. `Mary slept'.  In practice, it is often easier to see what the results
of a particular ellipsis resolution are if the substitutions are treated
as syntactic substitutions in this way.

\subsubsection{Strict and Sloppy Identity}

There are a number of distinct readings for a sentence like
\begin{quote}
John read a book he owned, and so did Bill.

(a) Bill read a book he (Bill) owned.\\
(b) Bill read a book he (John) owned.\\
(c) Bill read the book that John owned and read.
\end{quote}
These differences depend on whether the expressions {\em he} and {\em
a book he owned} are reinterpreted in the ellipsis as (i) strictly
referring to the same objects as in the antecedent, or (ii) sloppily to
the same kind of object.

A strict substitution for a \term\ involves replacing occurrences of that
\term\ by its index.  A sloppy substitution involves re-indexing the \term.
This can be demonstrated as follows:
\bigskip
\begin{itemize}
\item
Partially Resolved QLF:
\begin{verbatim}
[and, 
  [read, 
       term(+j,...,john),
       term(+b,B^[and,[book,B],
                      [own,term(+h,...,+j),B]],
             ...)),
  form(<type=vpellipsis,...>,
       R^[R,term(+w,...,bill)],
       _E)]
\end{verbatim}

\item
Ellipsis antecedent / source (\verb!S!):
\begin{verbatim}
S =
[read,term(+j,...,john),
      term(+b,B^[and,[book,B],
                 [own,term(+h,...,+j),B]],
           ...)]
\end{verbatim}
\medskip

\item[(a)]
Sloppy `he', sloppy `book':
\verb!_E! =
\begin{verbatim}
S:{term(+w,...)/term(+j,...), +w/+j,
   +h1/+h,
   +b1/+b}
\end{verbatim}

Equivalent QLF:
\begin{verbatim}
[and,
 [read,term(+j,...,john),
       term(+b,B^[and,[book,B],
                  [own,term(+h,..,+j),B]],
           ..)],
 [read,term(+w,...,bill),
       term(+b1,
            B1^[and,[book,B1],
                 [own,term(+h1,..,+w),B1]],
           ..)]].
\end{verbatim}

\item[(b)]
Strict {\it he}, sloppy {\it book}:\\
\begin{verbatim}
S:{term(+w,...)/term(+j,...), +w/+j,
   +h/term(+h,...),
   +b1/+b}
\end{verbatim}

Equivalent QLF:
\begin{verbatim}
[and,
 [read,term(+j,...,john),
       term(+b,B^[and,[book,B],
                  [own,term(+h,..,+j),B]],
           ..)],
  [read,term(+w,...,bill),
        term(+b1,B1^[and,[book,B1],
                     [own,+h,B1]],
              ...)]].
\end{verbatim}

\item[(c)]
Strict {\it book}:
\begin{verbatim}
S:{term(+w,...)/term(+j,...), +w/+j,
   +b/term(+b,...), ...}
\end{verbatim}

Equivalent QLF:
\begin{verbatim}
[and,
 [read,term(+j,...,john),
       term(+b,B^[and,[book,B],
                  [own,term(+h,..,+j),B]],
           ..)],
  [read,term(+w,...,bill),+b]]
\end{verbatim}
\end{itemize}
Whenever a \term\ is re-interpreted strictly it in effect leaves behind a
\term\ index.  For the QLF to be interpretable, it is necessary that the
antecedent term is given sufficiently wide scope so that it binds
this occurrence of the index.  However, this constraint does not need to
be explicitly enforced: any QLF that violates it would simply be 
uninterpretable in the same way that other ill-scoped QLFs are.
However, we have omitted explicit scoping constraints in this example.

\paragraph{Primary and Secondary Substitutions}

The possibility of making a choice between strict and sloppy substitutions
only arises with {\em secondary} noun phrases, i.e. {\em a book} and
{\em he}. The substitutions open to {\em primary} noun phrases ({\em John} and
{\em Bill}) are fixed --- replace the antecedent \term\ and all occurrences
of its index.

Which noun phrases are primary and which are secondary is largely determined
by the form of the ellipsis. Thus, in {\it John read a book he owned, and so 
did Bill} the fact that {\em Bill} occurs within the elliptical phrase
means that it is primary, and the fact that it is a verb phrase ellipsis means
that its parallel primary term must be the subject of the antecedent (the
QLF for the antecedent will carry enough information to enable one
to recognise its subject).  In other cases, there may be a choice
about which terms in the antecedent are to be taken as primary.  For example,
with a noun phrase ellipsis like
\begin{quote}
John greeted Mary before Bill
\end{quote}
either {\em John} or {\em Mary} may be taken as primary (i.e. {\em John greeted
Mary before Bill greeted Mary} or {\em John greeted Mary before he greeted
Bill}).

\subsubsection{Scope Parallelism}

In a sentence like
\begin{quote}
John gave every student a test and so did Bill.
\end{quote}
there is scope parallelism between antecedent and ellipsis.  That is, if
John gave every student a different test, then Bill also gave every student
a different test.  But if John gave just one test to all the students, then
Bill also gave just one test to all the students.  Readings where John
gave a different test to each student but Bill gave the same test to all
the students, or vice versa, are not available.

This parallelism of scope arises naturally through the way that sloppy 
substitutions are handled.  Essentially, scope constraints in the
antecedent are re-interpreted as applying to the re-indexed \terms\ in
the ellipsis resolution.  Whatever scoping is imposed on the antecedent
is thus automatically `copied across' to the ellipsis resolution.

Consider one of the possible scopings for the antecedent,
\begin{quote}
\begin{verbatim}
S =
 [+s,+t]:
   [give, term(+j,...,john),
          term(+s,..student..),
          term(+t,..test..)]
\end{verbatim}
\end{quote}
with both the secondary \terms\ given sloppy substitutions in the ellipsis
resolution:
\begin{quote}
\begin{verbatim}    
S:{term(+w,...,bill)/term(+j,...,john), +w/+j,
   +s1/+s
   +t1/+t}
\end{verbatim}
\end{quote}
Spelling these substitutions out, we get
\begin{quote}
\begin{verbatim}
[+s1,+t1]:
   [give, term(+w,...,bill),
          term(+s1,..student..),
          term(+t1,..test..)]
\end{verbatim}
\end{quote}
i.e. with the same scope order as in the antecedent.

It should be borne in mind that the ellipsis resolution is a re-interpretation
of the antecedent.  The fact that the antecedent and ellipsis have the same
scoping configurations is not due to an arbitrary decision to keep them the 
same.  Rather, the scope of the antecedent is an integral part of the
antecedent and the way it is re-interpreted.

\paragraph{Strict Identity and Parallelism}
A strict substitution for a secondary \term\ necessitates the antecedent
\term\ being given scope over both antecedent and ellipsis, on pain of
the QLF being uninterpretable.  One might wonder what happens if the
antecedent \term\ is given wide scope, but is given a sloppy substitution.

It turns out that under such circumstances, strict and sloppy substitutions
are equivalent.  For if the antecedent \term\ has wide scope over antecedent
and ellipsis, applying its quantifier rule will discharge all occurrences
of the \term, both in antecedent and ellipsis resolution.  
This means that when it comes to re-interpreting the antecedent as part of
the ellipsis resolution, there will be no occurrences of the original
index left to re-interpret: they will all have been discharged.  Hence it
makes no difference to the final interpretation of the ellipsis whether
the \term\ is treated as strict or sloppy; in either case the ellipsis
\term\ will be discharged by scoping the antecedent \term.

As a result, we can say that as far as scope parallelism is concerned
(a) either the antecedent \term\ is given wide scope, and the ellipsis
behaves as though there were a strict identity, or (b) if the \term\
has narrow scope, it is given a sloppy identity and scope parallelism
ensues.

\subsubsection{(No) Type-Raising for Quantifiers}

As we will see below, examples like
\begin{quote}
Every boy revised his paper, and then John did.
\end{quote}
create a number of complications for a higher-order unification treatment
of ellipsis.  Under a substitutional treatment, this example poses no
particular problems, and is covered by the principles already outlined.

Two readings for this sentence are possible, depending on whether a
strict or sloppy substitution is made for {\em his paper}.
A sloppy identification leads to the reading where John revises his
own paper, and a strict reading where he revises the paper of each boy.
\begin{quote}
\begin{verbatim}
S =
 [revise, 
       term(+b,<lex=each>,boy,forall,+b),
       term(+p,<type=possdet>,
               X^[and,[paper,X],
                   [poss,term(+h,..,+b),X]],
            exists,+p)]
\end{verbatim}

Strict:
\begin{verbatim}
S:{term(+j,..)/term(+b,...), +j/+b,
   term(+p,...)/+p}


[+b,+p]:
 [and,
  [revise, 
       term(+b,...),
       term(+p,...)]
  [revise,
       term(+j,...),
       +p]]
\end{verbatim}

Sloppy:
\begin{verbatim}
S[term(+j,..)/term(+b,...), +j/+b,
  +p1/+p, +h1/+h]


[and,
 [+b,+p]:
   [revise, 
       term(+b,...),
       term(+p,...)]
 [+j,+p1]:
   [revise,
       term(+j,...),
       term(+p1,
            X^[and,[paper,X],
                   [poss,term(+h1,..,+j),X]],
            exists,+p1)]]]
\end{verbatim}
\end{quote}

A substitutional treatment also lends itself readily to examples such as
\begin{quote}
A Canadian flag was hanging in front of each house, and an American one
was too.
\end{quote}
which were discussed by Hirshb\"uhler (1982).
\begin{itemize}
\item[(a)]
\begin{verbatim}
S =
[+h,+c]:[hang,term(+c,..canadian_flag..),
              term(+h,..house..)]
\end{verbatim}
Resolution: 
\begin{verbatim}
S:{term(+a,..)/term(+c,...), +a/+c,
   +h1/+h}
\end{verbatim}
That is:
\begin{verbatim}
[and,
 [+h,+c]:[hang,term(+c,..canadian_flag..),
               term(+h,..house..)]
 [+h1,+a]:[hang,term(+a,..american_flag..),
                term(+h1,..house..)]
\end{verbatim}
\medskip

\item[(b)]
\begin{verbatim}
S =
[+c,+h]:[hang,term(+c,..canadian_flag..),
              term(+h,..house..)]
\end{verbatim}
Resolution: 
\begin{verbatim}
S:{term(+a,..)/term(+c,...), +a/+c,
   +h1/+h}
\end{verbatim}
That is:
\begin{verbatim}
[and,
 [+c,+h]:[hang,term(+c,..canadian_flag..),
               term(+h,..house..)]
 [+a,+h1]:[hang,term(+a,..american_flag..),
                term(+h1,..house..)]
\end{verbatim}
\medskip

\item[(c)]
\begin{verbatim}
S =
[+c]:[hang,term(+c,..canadian_flag..),
           term(+h,..house..)]
\end{verbatim}
Resolution: 
\begin{verbatim}
S:{term(+a,..)/term(+c,...), +a/+c,
   +h/term(+h,...)}
\end{verbatim}
That is:
\begin{verbatim}
[+h]:
 [and,
    [+c]:[hang,term(+c,..canadian_flag..),
               term(+h,..house..)]
    [+a]:[hang,term(+a,..american_flag..),
               +h]
\end{verbatim}
\end{itemize}
(Note that cases (a) and (c) are logically equivalent).

\subsubsection{De Dicto / De Re}
It should be apparent that the treatment of strict and sloppy identity also
lends itself to a treatment of de re / de dicto ambiguities:
\begin{quote}
Bill wants to read a good book and John does too.
\end{quote}
\begin{itemize}
\item
\begin{verbatim}
S = [wants, term(+w,...,bill)
            [read,+w,term(+b,book,..)]]
\end{verbatim}

\item
De Re: same book (strict)
\begin{verbatim}
[+b]:[and,
       S,
       S:{term(+w,...)/term(+j,...), +w/+j,
          +b/term(+b,...)}]
\end{verbatim}

\item
De Re: different book (sloppy)
\begin{verbatim}
[and,[+b]:S,
     [+b1]:S:{term(+w,...)/term(+j,...), 
              +w/+j,+b1/+b}]
\end{verbatim}

\item
De Dicto: different book
\begin{verbatim}
[and,
 [wants,term(+w,...,bill)
        [+b]:[read,+w,term(+b,book,..)]],
 [wants,term(+w,...,bill)
        [+b]:[read,+w,term(+b,book,..)]:
        {term(+w,...)/term(+j,...), +w/+j,
         +b1/+b}]]
\end{verbatim}

\item
*De Dicto: same book
\begin{verbatim}
[and,[wants,term(+w,...,bill)
            [+b]:[read,+w,term(+b,book,..)]],
     [wants,term(+w,...,bill)
            [+b]:[read,+w,term(+b,book,..)]:
            {term(+w,...)/term(+j,...), +w/+j,
             +b/term(+b,...)}]]
\end{verbatim}
\end{itemize}
De re readings are obtained by giving the \term\ for {\em a book}
wide scope over the intentional predicate {\em want}.  This can be
done in two ways, so that it is either the same book that John and Bill
want to read or different books.  By contrast, there is only one way of 
obtaining a de dicto reading, and here there is no commitment to it being the
same book that John and Bill want to read. (The strict, de dicto reading is
uninterpretable, since it will leave an undischargeable index, \verb!+b!,
in the ellipsis)

\subsubsection{Antecedent Contained Ellipsis}

The following sentence might appear to present problems for our treatment
of ellipsis, in that the ellipsis is contained within the (object noun phrase
of the) antecedent.
\begin{quote}
{\it John greeted everyone that Bill did.}
\end{quote}
However, this merely serves to rule out the possibility of making a
sloppy substitution for the offending \term.
\begin{itemize}
\item
\begin{verbatim}
S=
[greet,term(+j,...,john),
       term(+e,X^[and,[person,X],
                    form(<type=vpellipsis>,
                         R^[R,term(+b,...,bill)],
                         _E)])]
\end{verbatim}

\item
Strict identity for {\it everyone}:
\begin{verbatim}
_E =
S:{term(+b,...)/term(+j,...), +b/+j,
   +e/term(+e,...)}

[+e]:
 [greet,term(+j,...,john),
        term(+e,X^[and,[person,X],
                    [greet,term(+b,...,bill),+e],
             ..)]
\end{verbatim}
\medskip

\item
Sloppy identity for {\it everyone}:
\begin{verbatim}
_E = 
 S:{term(+b,...)/term(+j,...), +b/+j,
    +e1/+e}
[greet,
 term(+b,...,bill),
 term(+e1,
      X^[and,[person,X],
         form(<type=vpellipsis>,
              R^[R,term(+b,...,bill)],
              [greet,term(+b,...,bill),
                     term(+e1,X^[and,[person,X],
                                   form(****)],..)],
                         )])]
\end{verbatim}
\end{itemize}
Here, giving a sloppy reading for ``everyone'' leads to an uninterpretable
QLF --- the ellipsis occurs in an undischarged form in its own resolution.

\subsubsection{Prepositional Phrase Ellipsis}

So far we have concentrated on ellipses where resolution essentially involves
replacing one \term\ by another.  Not all ellipses are like this, of course.
For example in,
\begin{quote}
John won in 1990. And in 1991.
\end{quote}
the ellipsis essentially involves replacing one prepositional phrase by 
another.
This type of ellipsis can readily be accommodated within our framework:
one substitutes for \forms\ and \form\ indices in the same way as
for \terms.

\subsubsection{Tense and Mood}

The current treatment of tense and mood in QLF forces a slight departure from
a strict substitutional approach to ellipsis.  This affects pairs of sentences
such as the following
\begin{quote}
John slept.  So will Mary.
\end{quote}
Because tense, aspect and mood information is all contained in the verb
\form\ category, it will not do simply to re-index the original verb
\form\ in resolving the ellipsis.  This would result in a past tense
resolution of the elliptical sentence.  

Instead, the verb \form\ categories on the antecedent and ellipsis are merged.
A new \form\ is constructed with the merged category and the 
body of the antecedent \form\, and the ellipsis resolution substitutes
this constructed \form\ for the original.

\subsection{Ellipsis and Higher-Order Unification}

Dalrymple, Shieber and Pereira (1991) (henceforth referred to as DSP)
suggest that the resolution of ellipsis can be cast as a matter of solving a
number of equations using higher order unification.  

\paragraph{Simple Example}
As a simple case
(which doesn't require the full power of higher order unification) consider
\begin{quote}
Dan likes golf.  So does George.
\end{quote}
{\it Dan} and {\it George} are parallel elements in the ellipsis and
it antecedent.  Resolving the ellipsis is a matter of recovering a property from
the antecedent which (i) when applied to the parallel elements in the
antecedent yields the antecedent itself, and (ii) when applied to the parallel
elements in the elllipsis yield the resolution of the ellipsis.

In simplified form, we can represent the `logical forms' of the sentences above
as
\begin{quote}
{\it like(dan,golf)}\\
{\it P(george)}
\end{quote}
where {\it P} is (a variable standing for) the property that has to be recovered.
Applying $P$ to the parallel elements in the antecedent, we get the following
equation
\begin{eqnarray}
P(dan) = like(dan,golf) \label{simple-eqn}
\end{eqnarray}
The solution to this equation is given by
\begin{quote}
$P = \lambda x. like(x,golf)$
\end{quote}
When applied to the parallel ellipsis elements, this solution yields
\begin{quote}
$\lambda x. like(x,golf)[george] = like(george,golf)$
\end{quote}
In more complex cases, solving equations like (\ref{simple-eqn}) may involve
using higher order unification.  

\paragraph{Strict and Sloppy}
The difference between strict and sloppy readings for
\begin{quote}
Dan likes his wife.  So does George
\end{quote}
(strict = George likes Dan's wife, sloppy = George likes George's wife) emerges
from different ways of solving the ellipsis equations:
\begin{quote}
{\it P(dan) = like(dan,wife-of(dan)}\\
 strict: $P = \lambda x. like(x,wife-of(dan))$\\
 sloppy: $P = \lambda x. like(x,wife-of(x))$
\end{quote}
As DSP note, there are two other solutions to the equation, where the primary
or parallel occurrence of $dan$ is not abstracted over.  They impose an additional
constraint on solutions, namely that primary occurrences must always be abstracted
over.

\paragraph{Interactions with Quantifier Scope}
DSP assume a semantic framework where `quantifier scope is generated through a
mechanism of discharging [quantifier] assumptions introduced in the course
of a derivation'.  To see roughly how this works, consider a sentence like
\begin{quote}
John greeted every person.
\end{quote}
The quantified noun phrase {\it every person} is given an interpretation
consisting of a quantifier assumption and a term meaning:
\begin{quote}
$\langle every,x,person(x)\rangle \vdash x$
\end{quote}
which is to be read as saying that the meaning of the noun phrase is $x$ under
the assumption to the left of the turnstile.  At some point in the derivation of
the meaning for the whole sentence, the quantifier assumption will have to
be discharged.  Where it gets discharged determines the scope of the quantifier.
Applying the meaning of the verb phrase to the meaning of the noun phrase (i.e.
$x$) and then the meaning of the subject {\it John}, yields an interpretation
that still rests on an undischarged scope assumption:
\begin{quote}
$\langle every,x,person(x)\rangle \vdash greet(j,x)$
\end{quote}
Discharging the assumption yields
\begin{quote}
$\vdash every(x,person(x), greet(j,x))$
\end{quote}
which no longer rests on any quantifier assumptions.

Turning to the interactions of scope and ellipsis, note that
\begin{quote}
John greeted every person when Bill did
\end{quote}
has two readings.  One is a distributive reading where each person gets 
simultaneously greeted  by John and by Bill.  The other is a collective reading
where John greets everyone (en masse) and Bill does the same.
These differences are accounted for by resolving the ellipsis at different
points during the introduction and discharging of quantifier assumptions.

First take the distributive case.  Here we solve the ellipsis equations before
the quantifier assumption has been discharged.  Thus the equation is
\begin{quote}
$P(john) = greet(john,x)$
\end{quote}
(the undischarged assumptions do not form part of the equation).
Applying the solution to this equation to the sentence as a whole
\begin{quote}
$\langle every,x,person(x)\rangle \vdash when(greet(john,x),P(bill))$
\end{quote}
we get
\begin{quote}
$\langle every,x,person(x)\rangle \vdash when(greet(john,x),greet(bill,x))$
\end{quote}
Once the quantifier assumption is discharged, we have
\begin{quote}
$\vdash every(x,person(x), when(greet(john,x),greet(bill,x)))$
\end{quote}
giving a distributive reading.

For the collective case, the quantifier assumption is first discharged and
then the ellipsis equation is solved.  The equation is thus
\begin{quote}
$P(john) = every(x,person(x),greet(john,x))$
\end{quote}
the solution to which is
\begin{quote}
$\lambda y. every(x,person(x),greet(y,x))$
\end{quote}
Applying this to the ellipsis we get
\begin{quote}
$\vdash when(every(x,person(x),greet(john,x)),every(x,person(x),greet(bill,x)))$
\end{quote}
i.e. a collective reading.

\subparagraph{Higher than Second-Order Matching}
The interactions between ellipsis and scope sometimes give rise to the need to
use more than second order matching in the solution of ellipsis equations.
Consider
\begin{quote}
Every student revised his paper, and then John did
\end{quote}
As before, we can resolve the ellipsis before or after discharging the scope
assumption (leading to a difference between Bill revising his own paper or
every students).  

If the quantifier is discharged first, we need to solve an equation that is
approximately represented as
\begin{quote}
$P(''every \; student'') = every(x,student(x),revise(x,paperof(x)))$
\end{quote}
the problem with this is that in place of $''every \; student''$ we cannot 
use the
conditional noun phrase interpretation 
\begin{quote}
$\langle every,x,student(x)\rangle \vdash x$
\end{quote}
since the conditional interpretation only makes sense in the course of a
derivation.  To circumvent this, the noun phrase is type raised by applying
it to an arbitrary property $S$, then discharging the quantifier assumption, and
then abstracting over $S$.  This yields an ellipsis equation
\begin{quote}
$P(\lambda S. every(x,student(x),S(x)) = every(x,student(x),revise(x,paperof(x)))$
\end{quote}
And this equation requires more than just second order matching to solve.

\section{Discussion}
Pereira (1990) observes, in the context of categorial semantics, that it is
productive to look at the way in which the meaning of a sentence is derived
by composing the meanings of its constituents, as well as at the end result of
the derivation.  Ellipsis resolution involves recomposing some of the
constituents of the antecedent to form a property that can be applied to
the ellipsis arguments.  Higher-order unification is required because the
ellipsis equations are stated using the end results of the meaning derivations,
and not the derivations themselves.  Thus, one has to unpick the original
derivation before putting some of the constituents back together again.

As we have seen, monotonic semantics in the shape
of QLF ellipsis resolution can achieve the same effects, but without
recourse to higher-order unification.  Nor is the precise interleaving of scoping
decisions in attempts to solve ellipsis equation crucial.  Both these
factors considerably simplify the treatment of ellipsis in QLF.  Nevertheless,
the underlying model of ellipsis in monotonic semantics and in DSP's
account is essentially the same.

The reason that the QLF treatment of ellipsis can get away with being simpler
is this:  To a limited extent, QLF resolutions record the steps that need
to be taken in the course of a semantic derivation, rather than just the
end results of the derivation.  This is especially so for quantifier
scope constraints.   The scope constraints simply say when and where during
the course of the semantic derivation quantifier assumptions are to be discharged.
Having this derivation information more immediately available makes the
unpicking and recomposing of derivations more straightforward.
QLFs may thus be construed as explicitly recording aspects of semantic
interpretation that are left implicit in an approach like Pereira's (1990)
categorial semantics, as suggested in Chapter~\ref{monosem}.

\chapter{Generation}
\label{generation}

The most important uses of the generator in CLARE
are (i) to allow users to confirm the results of reference resolution, and
(ii) to generate sentences from database assertions (or questions from
incomplete assertions). 
We had expected that its use for paraphrasing
QLF analyses for structural disambiguation would be also be useful.
In practice, it turns out to be difficult to generate paraphrases exhibiting
a choice of structure without introducing further ambiguity.

Our model for natural language generation involves the processes of
{\it synthesis}, i.e. producing a natural language string from a given
QLF, and {\it description}, i.e. producing a QLF from which synthesis
can proceed.  In section~\ref{ling-gen} we review the way that synthesis
is implemented by means of a version of the ``semantic head driven''
algorithm.  In section~\ref{genmono} we discuss how description
operates when generating paraphrases to make reference resolutions to QLFs
explicit. Finally, in section~\ref{genass} we discuss description of TRL 
facts, which is used in the generation of statements and questions
from the contents of the database.

\section{Linguistic Generation (Synthesis)}\label{ling-gen}

Since the CLE grammar formalism is declarative, it can be
used for \inx{generation} (synthesis) as well as analysis. In particular,
given that \inx{QLF} analyses are built up by \inx{unification} rules for 
syntax, semantics, and morphology, it is also possible to devise
algorithms for building a tree of constituents by unification by
applying these rules in the ``reverse'' direction. If the semantic rules had
been more in the style of traditional Montague semantics, generation from
structures that had undergone lambda reductions would have presented search
difficulties because the reductions would have to be applied in reverse during
the generation process. This turns out to be an important practical advantage
of unification-based semantics over the traditional approach. 

In the context of a system like CLARE for interaction with a knowledge base, 
generation is potentially useful to the aim of natural
cooperative dialogues in a number of ways.  These include the generation of
responses:  answers, asking the user to confirm consequences of their requests
(e.g., changes to the knowledge base), explanations of analysis failure, and so
on.  Generation is also useful for interactive disambiguation as it often
allows user inputs to be \inx{paraphrase}d in a way that makes explicit their
interpretation by the system.

The CLE component for generation from QLF employs an algorithm that is based on
the ``\inx{semantic head} driven'' (SHD) generation algorithm (Shieber et al. 1990).
We will briefly describe the original algorithm and then discuss some steps we
have taken in the CLE implementation to make generation efficient in the
context of a large grammar.

\subsection{The SHD Generation Algorithm}
\label{genalg}

The SHD algorithm is usually applied to a pair consisting of the semantic
structure being generated from (in our case a QLF), together with a \inx{category}
for
the phrase type we are attempting to generate.  We will refer to such a pair as
a {\it node}, so \inx{semantic rule}s in the CLE grammar formalism are lists of such
nodes.  At the start of the process, the category may be more or less
instantiated, it may simply be the start category for the grammar, or it may
be a
suitably instantiated sentence or noun phrase category.  In the CLE
implementation, the output (which is nondeterministic) is a phrase structure
tree for some sentence having the input QLF as its semantic analysis.

The SHD algorithm proceeds as follows.  The QLF for the node currently being
generated is matched against the QLF of the mother of a 
{\it nonchain rule}
that is, a rule in which the QLF for the mother is different from that of any of
its daughters.  The choice of rule is constrained further to rules for which
there is a sequence (or chain) of categories, $C_1,\dots,C_n$, where $C_1$ is
the category of the node being generated, $C_n$ is the mother node category of
the chosen rule, and each pair $(C_i,C_{i+1})$ corresponds to the mother and
daughter categories of a {\it chain rule}.  A chain rule is one in which some
daughter (the semantic head) has the same QLF as the mother node of that rule.
To complete the construction of the phrase structure tree, all that is needed
is to apply the algorithm recursively to the daughters of the chosen nonchain
rule and to the nonhead daughters of each chain rule in the sequence.

The effectiveness of the algorithm comes from the way it quickly leads to
a backbone of the target phrase structure tree being proposed according to
the input QLF and the semantic head relation for the grammar. For a more
detailed presentation of the algorithm and a discussion of its merits, the
reader is referred to Shieber et al. 1990.

\subsection{Applying Rules for Generation}

In the CLE implementation, the rules used in generation are those built by
unifying the categories in \inx{semantic rule}s with the categories in their
syntactic counterparts (syntactic rules with 
matching rule identifiers) and the categories in word sense 
derivation rules with those in corresponding \inx{morphology rule}s.
It is also possible to run the generator simply with
semantic rules and \inx{sense derivation rule}s,
and apply the constraints from the
syntactic rules in a second stage, but this leads to much increased
nondeterminism and a consequent increase in processing time.  Syntactic
constraints for lexical entries are also unified into sense entries when these
are being applied, as daughterless nonchain rules, during generation.

The more \inx{compositional} the semantic rules of the grammar, the more efficient
generation becomes with this algorithm.  In the context of the CLE semantic
rules, this corresponds to having the mother QLF in a rule built from daughter
QLFs rather than from the values of semantic features.  As a result, applying
the CLE grammar in the generation direction, rather than the analysis direction
for which it was originally designed, was more effective after some
re-organization of the rules that increased the compositionality of the 
grammar---this also made the grammar easier to understand.

Different uses to which a generator is employed require different trade-offs
between efficiency and the number of paraphrases that the generator needs to
produce for a given QLF.  To control the set of rules that the generator
applies, the rule compilation process is sensitive to declarations specifying
which rules are to be included, and the order of their application, in the
generation process.

\subsection{Rule and Lexicon Precompilation}

The CLE implementation of the SHD algorithm includes a number of control
structure and precompilation optimizations.  The first of these is
precompilation of the transitive closure of the \inx{semantic head}
relation for the
categories of the grammar to allow more efficient application of the chaining
step.  This precompilation proceeds in a similar fashion to the computation of
the left-corner linking relation used by the CLE parser in that pairs in the
relation are kept in their most general form with subsumed pairs being deleted.

Further preprocessing takes place by unifying the daughter element of
pairs in the semantic-head relation with the mother categories of 
\inx{nonchain rule}s producing composed rules of the form
\begin{quotenr}
\begin{verbatim}
gen_rule(Index,ChainCategory,MotherNode,DauNodes).
\end{verbatim}
\end{quotenr}
These rules are applied at run time by matching the mother QLF
and the chain category against the target QLF and category being generated.
It is further assumed that these two terms (the QLF and category) will be
instantiated when the rule is applied, so a combined index is derived from
them to improve the efficiency of rule look-up. This assumption
can be enforced by reordering the generation of daughter nodes dynamically, 
as explained in the next section.

As mentioned above, sense entries are
treated by the algorithm as daughterless
nonchain rules, so it is possible to include them in the precompilation
process.  However, this results in a very large set of \verb!gen_rule! entries
(several times the number of entries in the lexicon) with corresponding
increased storage and access overheads.  Our strategy is to only include
lexicon entries in the compilation process if they cannot be indexed
effectively by a unique (or almost unique) symbol, such as the \inx{word sense}
predicate symbols in the QLFs for most noun, verb, and adjective entries.  In
effect, the precompilation process decides which words are to be treated as
``function'' words whose entries are compiled together with grammar rules, and
which are treated as ``content'' words.  Applying the former will be constrained
during generation by the semantic-head relation, while the latter will only be
considered when generating from a QLF (sub)expression if the symbol under which
it is indexed is present in that QLF.  Since there is a constant number of
\inx{function word}s, and a constant number of content words indexed from the
symbols in a QLF, the time taken for generation is effectively independent of
the size of the lexicon.

Literals appearing in grammar rules (i.e., daughters that are
specified
as words rather than categories) are compiled directly into the run-time rules,
so no access to the lexicon is necessary for them; they are treated as
daughters whose generation has been completed at compile time.

\subsection{Control and Search}

The CLE implementation of the SHD algorithm includes an improvement to the
basic control structure:  generation of nodes with as yet uninstantiated QLFs
can be delayed until the QLFs become instantiated.
Generating from
nodes with instantiated QLFs reduces unguided \inx{search} and allows the
improved indexing mentioned above.  There are cases where the delay
is essential.  

An example of
this is generation of subject-auxiliary
inversion questions like {\it Will John sleep?} if these are treated by a rule 
of the form:
\begin{quotenr}
\begin{verbatim}
sem(s_aux_np_vp,inverted,
    (AuxQLF,s:[type=ynq,...]) -->
    [(AuxQLF,aux:[arglist=[(VpQLF,VpCat)],subjval=Np],
     (Np,np:[...]),
     (VpQLF,VpCat)]).
\end{verbatim}
\end{quotenr}
When this rule is used for analysis, the subject noun phrase meaning {\tt Np}
is passed to the verb phrase meaning {\tt VpQLF} via the sense entry for
the auxiliary.
During generation, however, the noun phrase meaning variable {\tt Np} 
will only become instantiated after the verb phrase node 
{\tt (VpQLF,VpCat)} has been generated. It is therefore necessary to delay
generation of the noun phrase node until this has happened, as otherwise
all possible noun phrases compatible with the noun phrase category will be
generated before their meanings are matched against the noun phrase 
meaning in the input QLF.

In this particular case, it is only necessary to delay generation of the 
subject noun phrase locally, i.e. until the other daughters of the rule
introducing it have been generated.  More generally though, it may be 
necessary to delay generation non-locally, so that uninstantiated
nodes may be threaded upwards until nodes higher up the tree have been 
generated.

The search strategy for generation in the CLE implementation is an iterative
deepening strategy whereby a series of depth-limited searches are performed to
successively higher limits (Korf 1985).  The depth of a solution corresponds
here roughly to the number of applications of rules or lexical entries during
the search, so a shallower solution usually corresponds to a simpler sentence.
The strategy thus tends to locate the simple solutions closer to the top of
the search space than the depth-first strategy would.  
This is particularly suitable
for generation because in many applications we are only interested in a single,
simple rendering of the semantic representation in natural language.  This is
in contrast to parsing where we are often interested in all possible
analyses, the reason for the asymmetry being that the semantic representation
language is less ambiguous a representation than natural language.  The
iterative deepening strategy also means that finite solutions will be found
eventually, even if the grammar allows infinite phrase structure derivations
for the input semantic representation.

In order to reuse intermediate results from one search path (or from a
previous depth iteration), whenever a possible phrase structure for a 
constituent is generated, this is stored together with the QLF-category
node associated with it, provided that it is not subsumed by an existing
intermediate result. When a daughter node is being generated it is
first matched against previously generated nodes, so that
the stored phrase structure can be reused. Again, the fact that we do not
want to exhaustively search for all possible solutions means that such
a simple caching technique can be effective in reducing the time spent
finding initial solutions.

\section{Generation for Paraphrasing Resolutions}
\label{genmono}

\subsection{Principles of Resolution Paraphrase}

In Chapter~\ref{clareinterp}, reference resolution was characterised in terms
of finding solutions to the following kind of translation equivalence,
\begin{verbatim}
 C U A |= (Q <-> Q') 
\end{verbatim}
(where \verb!C! represents the context, \verb!A! any additional assumptions,
\verb!Q! the unresolved QLF, and \verb!Q'! the resolved QLF). Reference
resolution involves finding a \verb!Q'! that is syntactically more 
instantiated  than \verb!Q!, but equivalent to it given \verb!C! and
additional assumptions \verb!A!.  A more concrete example of this schema
is
\begin{verbatim}
context(C) /\ most_salient(i1,C,X^prototype(X),zed9) ... |=
    disintegrate(term(i1,<det=the>,X^prototype(X),_,_))
    <->
    disintegrate(term(i1,<det=the>,X^prototype(X),exists,zed9)).
\end{verbatim}
which shows how the sentence {\em The prototype disintegrated} might be
resolved.

In order to paraphrase a resolution, the translational schema can be applied
in the reverse direction.  That is, given a resolved QLF \verb!Q!, one
needs to find an unresolved QLF \verb!Q''! that is equivalent to \verb!Q!
in a context \verb!C'! plus assumptions \verb!A'!, where preferably \verb!C'!
should be entailed by the more specific context \verb!C! used in reference
resolution.  That is
\begin{verbatim}
C' U A' |= (Q' <-> Q'') 
\end{verbatim}
Here there is no requirement that \verb!Q''! be syntactically more instantiated
then \verb!Q'!, or vice versa.   As a more concrete example, we might have
\begin{verbatim}
wider-context(C',C) /\... |=
    disintegrate(term(i1,<det=the>,X^prototype(X),exists,zed9))
    <->
    disintegrate(term(i1,<proper_name>,X^name_of(X,'Zed-9'),_,_)).
\end{verbatim}
This shows how {\em The prototype disintegrated} could be paraphrased
as {\em Zed-9 disintegrated}.  In the new context, \verb!C'!, we no longer
need assume that \verb!zed9! is the most salient prototype, though it does
need to be the case that the name for \verb!zed9! is {\em Zed-9} in 
\verb!C'! (and hence also in the original context \verb!C!).  The idea
behind using a wider, less specific context is that the paraphrase should
consequently be more specific and less context-dependent.  In general, the
context would be widened by weakening conditions in the discourse model
(e.g. what is salient / has recently been mentioned) rather than altering
the domain model itself.

To paraphrase more complex QLFs one would normally proceed bottom up, replacing
QLF sub-expressions by equivalent paraphrases before trying to paraphrase
the expression itself.  This has the benefit of largely preserving the
predicate-argument structure of the original QLF, which might otherwise be
lost.  The QLF sub-expressions paraphrased will be individual \terms, \forms\
and scope constraints.  

For \terms\ and \forms, deriving 
equivalent paraphrases is in many respects the reverse of making 
resolutions.   Resolution involves instantiating
unspecified \term\ or \form\ resolvents, given specified categories and
restrictions; paraphrase involves instantiating categories and restrictions
given specified resolvents.  Thus, in forming a description for
\term\ \verb!i1!, we might start with an underspecified \term
\begin{verbatim}
term(i1,_,_,exists,zed9)
\end{verbatim}
and instantiate the category and restriction to appropriate values.
This, in effect applies reference resolution methods in reverse.

\paragraph{Resolution Paraphrase in Practice}\label{Reverse-resolution}
It is not currently feasible to derive QLF paraphrases by proving equivalence
using resolution methods in the reverse direction, and simultaneously fixing 
a wider context. As a simpler compromise, a number of what are in effect 
special purpose reverse resolution methods are employed.  These take
a fully resolved \term\ or \form\ and produce either (i)
a new (unresolved) \term\
or \form\ that is equivalent to the original in a wider variety of contexts,
or otherwise (ii) a copy of the original.   A paraphrase QLF is built up
through replacing \terms\ and \forms\ by their paraphrases.  In the next 
sections, we describe some of the reverse resolution / paraphrase methods
in more detail.

\subsection{Paraphrasing \terms}

The following examples show the effects achieved by paraphrasing \terms.
(Generated sentences are shown in typewriter font.)
\begin{itemize}

\item Pronouns:\\
{\it Does Peter Piper work on CLAM-BAKE? \dots}\\
{\it Which projects does he work on?}\\
\verb!>>> Which projects does Peter Piper (the employee) work on?!

{\it How many payments were there on CLAM-BAKE? \ldots}\\
{\it Who were they to?}\\
\verb!>>> Who were they (the payments on CLAM-BAKE (the project)) to?!

\item Ambiguous Names:\\
{\it Show all payments for 1990.}\\
\verb!>>> Show all payments during 1990 (the year).!\\
or\\
\verb!>>> Show all payments on account 1990.!

\item Ambiguous Quantifiers:\\
{\it Show payments on CLAM-BAKE}\\
\verb!>>> Show all payments on CLAM-BAKE (the project).!\\
or\\
\verb!>>> Show some payments on CLAM-BAKE (the project).!

{\it Which employees have cars?}\\
\verb!>>> Which employees have some cars?!\\
or\\
\verb!>>> Which employees have all cars?!
\end{itemize}
Below, we explain how these paraphrases are brought about.

\subsubsection{Referring \terms}

When \terms\ refer to a specific entity, an appositive proper name 
\term\ is created in its place.  The restriction for the \term\ is formed as
follows. The inference engine is used to determine the name
of the entity.  The term in the domain model referring to the entity
will normally be sorted; that is, part of the term will include a property
(such as {\tt project\_Activity} or {\tt employee}) giving the basic sort
of the entity.  This is used to form an appositive modification to the
\term's restriction (in order to generate NPs like {\em Peter Piper (the
employee)}).

This method can also be used to produce \terms\ referring to objects from
scratch, and is indeed used when generating sentences from domain facts
(section~\ref{genass}).

\subsubsection{Contextually Restricted Quantification}

Some \terms\ are resolved to act as contextually restricted quantifiers,
e.g. pronouns that do not specific entities. Normally, the contextual
restriction corresponds to the full description (Chapter~\ref{clareinterp})
of an antecedent \term.  This description is used to form an
appositive modification to the original \term\ restriction.

\subsubsection{Bare Plurals}

Some determiners are resolved in non-trivial ways.  For example, bare plurals
may sometimes correspond either to universal or existential quantifiers.
Paraphrase can make the resolution explicit by altering the category of the
\term.

\subsection{Paraphrasing \forms}

Examples of \form\ paraphrases are:
\begin{itemize}
\item Compound Nominals:\\
{\it Show me the Scandy payments for last year}\\
\verb!>>> Show me the payments to Scandy for 1990.!\\
{\it Show me the CLAM-BAKE payments for last year}\\
\verb!>>> Show me the payments on CLAM-BAKE for 1990.!\\

\item Vague Prepositions and Possessives:\\
{\it Show payments for CLAM-BAKE}\\
\verb!>>> Show all payments on CLAM-BAKE (the project).!

{\it Show CLAM-BAKE's payments}\\
\verb!>>> Show all payments on CLAM-BAKE (the project).!

\item Ellipsis:\\
{\it How many projects does James Cagney work on?}\\
{\it Peter Piper?}\\
\verb!>>> How many projects does Peter Piper work on?!
\end{itemize}

\subsubsection{Vague Relations}

Compound nominals, possessives and some prepositions introduce \forms\ that
are typically resolved to specific prepositional senses.  When this is
the case, the category of the \form\ is altered so that it regenerates as
the appropriate preposition.

\subsubsection{Ellipsis}

Ellipsis resolutions are paraphrased essentially by making the necessary
substitutions marked on the ellipsis antecedent, and generating from that.

\section{Generation of Facts and Questions}
\label{gentrl}\label{genass}

\subsection{Generation from Assertions}

The description part of generation from a TRL assertion requires 
two steps to be carried out: (i) determining an appropriate set of linguistic
predications (i.e. applications of word
sense predicates) for the assertion, and (ii) producing a set
of QLFs which can be used for linguistic synthesis. 

Rules for the first of these steps are compiled from the
domain model equivalences expressing the relationship between
linguistic and database predicates. This compilation process
involves application of the equivalences in the ``reverse''
direction, i.e. starting from database predicates and ending
with word senses. (The initial stages of this
compilation process are the same as that required by
the interpretation process for domain-specific resolution of
vague relations --- see Chapter~\ref{clareinterp}.)

The second step could in principle proceed by the process
of monotonic description sketched out in Section~\ref{genmono}.
However, this process would be highly nondeterministic,
both in selecting possible restrictions and
possible categories for the QLFs being built up.
Our first attempt at a solution to the problem simplifies it in
two ways. Firstly, we only consider description from the most common
kind of TRL form, namely an existentially quantified conjunction
of atomic goals. Secondly, and more crucially, description is not
free, but rather constrained to follow one of a set of ``template''
rules. These are automatically produced by generalizing specific
training examples, using a variant of the so-called ``case-based
learning'' method. The learning procedure is 
described in Section~\ref{Case-based-learning}. It produces rules
of the approximate form
\begin{quote}\begin{verbatim}
qlf_for_trl(<QLF>) :- <Body>
\end{verbatim}\end{quote}
where \verb!<Body>! is a set of TRL goals. (We refer to these
as TRL description generation
rules, or simply TRL generation rules.) The goals in \verb!<Body>!
can be of several different types: they include 
the component conjuncts in the TRL form that is being generating from,
and conditions constraining {\tt term}s in \verb!<QLF>! to refer to
TRL constants. Examples
of learned TRL description rules are given in the next 
section.

The set of TRL description rules is not complete; a QLF can be produced 
only if it is of the same form as one of the training examples.
They are not necessarily guaranteed to be sound either, though 
we have not in practice observed any unsound rules to have been
inferred by the process. The status of the rules is thus that of
{\it generation heuristics}, induced by analysis of specific 
examples. To check the soundness of a rule, it necessary to make
sure that the original TRL can be recovered from the proposed QLF.

Loss of completeness is not necessarily important, since
there is no need to be able to produce {\it every} way of expressing
something. Indeed, everyday experience suggests that people generally
make use of an array of stock phrases when expressing themselves;
there is no obvious reason to believe that one has to have a general
mechanism for generation at this level. The unsoundness of the method
is more significant, since checking correctness is moderately 
expensive. One way to attack this problem might be to attempt to
infer rules using the
Explanation-Based Learning method (Hirsh 1987, Rayner 1988), the
application of reference resolution methods would be conditions
on the inferred rules.
In the rest of this chapter we describe in more detail TRL generation
as implemented in CLARE.

\subsection{Learning Case-based Description Rules}
\label{Case-based-learning}

The learning method used for acquiring TRL description rules is
extremely simple, and is perhaps best introduced by an example.
Suppose that the training example is the sentence
\begin{quote}
{\it John is a dog.}
\end{quote}
uttered in a context in which ``John'' can be used to refer to the entity
{\tt john1}.
This produces the QLF
\begin{quote}\begin{verbatim}
[dcl,
 form(verb(pres,no,no,no,y),
      A,
      B^
      [B,
       [be,
        A,
        term(proper_name(tpc),C,D^[name_of,D,John],E,F),
        G^[eq,G,term(q(H,a,sing),I,J^[dog_Animal,J],K,L)]]],
      M)]
\end{verbatim}\end{quote}
which in turn produces a TRL form which after simplification is
\begin{quote}\begin{verbatim}
dog_Animal(john1)
\end{verbatim}\end{quote}
The idea is to generalize the training sentence by looking for 
correspondences between the QLF and the TRL representations, 
and hypothesizing that they are instances of a general rule.
In our example, there are two such correspondences. The predicate
symbol {\tt dog\_\-Animal} occurs in both QLF and TRL; also,
the QLF contains the term
\begin{quote}
\begin{verbatim}
term(proper_name(tpc),C,D^[name_of,D,John],E,F)
\end{verbatim}
\end{quote}
which the according to the information in the domain model has the 
term {\tt john1} occurring in the TRL representation, as a referent. 
Looking up the linguistic properties of {\tt dog\_\-Animal} using the predicate
\verb!genpred/2!, we have
\begin{quote}\begin{verbatim}
genpred(dog_Animal,([dog_Animal,_],nbar(_,_)))
\end{verbatim}\end{quote}
-- that is, \verb!dog_Animal! is a one-place predicate and a
common noun (\verb!nbar!) sense. Our precise hypothesis with regard to 
replacing \verb!dog_Animal! is that this information constitutes a
sufficiently detailed description of its properties that any other
predicate with a similar entry would be processed in the same way.
So we are guessing, for example, that in a context where ``Mary'' can be used
to refer to the entity {\tt mary1}, the TRL 
\begin{quote}\begin{verbatim}
cat_Animal(mary1)
\end{verbatim}\end{quote}
could have been produced from the QLF
\begin{quote}\begin{verbatim}
[dcl,
 form(verb(pres,no,no,no,y),
      A,
      B^
      [B,
       [be,
        A,
        term(proper_name(tpc),C,D^[name_of,D,Mary],E,F),
        G^[eq,G,term(q(H,a,sing),I,J^[cat_Animal,J],K,L)]]],
      M)]
\end{verbatim}\end{quote}
our justification being that
\begin{quote}\begin{verbatim}
genpred(cat_Animal,([cat_Animal,_],nbar(_,_)))
\end{verbatim}\end{quote}
holds and that 
\begin{quote}
\begin{verbatim}
term(proper_name(tpc),C,D^[name_of,D,Mary],E,F)
\end{verbatim}
\end{quote}
resolves to {\tt mary1}.
In the general case, our heuristic rule will be to guess that
\begin{quote}\begin{verbatim}
[dcl,
 form(verb(pres,no,no,no,y),
      A,
      B^
      [B,
       [be,
        A,
        <Term>,
        G^[eq,G,term(q(H,a,sing),I,J^[<PredAtom>,J],K,L)]]],
      M)]
\end{verbatim}\end{quote}
is a QLF from which it would be possible to derive the TRL form
\begin{quote}\begin{verbatim}
<PredAtom>(<Ref>)
\end{verbatim}\end{quote}
under the assumptions that \verb!<PredAtom>! is a atom, \verb!<Term>!
is a term, the \verb!genpred! entry for \verb!<PredAtom>! is
\begin{quote}\begin{verbatim}
genpred(<PredAtom>,([<PredAtom>,_],nbar(_,_)))
\end{verbatim}\end{quote}
and \verb!<Ref>! is the referent of \verb!<Term>!.
This information is what is encoded in the induced rule,
\begin{verbatim}
trl_gen([dcl,
         form(verb(pres,no,no,no,y),A,
              B^[B,[be,
                    A,
                    Term,
                    C^[eq,C,term(q(_,a,sing),_,D^[Pred,D],_,_)]]],
              _)],
        [Pred,Ref],
        [entity_ref_term(Ref,Term),
         trlgoal([Pred,Ref],genpred(Pred,(E^[Pred,E],nbar(_,_))))])
\end{verbatim}
The rule is capable of producing QLFs for sentences of the form
\begin{quote}
\var{NP} is a/an \var{Noun}.
\end{quote}
where \var{NP} is any referring expression for which there
is a reverse resolution method (see section~\ref{Reverse-resolution}).
This includes NPs like ``John'', ``Mary (the employee)'', ``1990 (the year)''
and ``1/1/89 (the day)''.

\subsection{Description of Database Entities}
\label{Talk-about}

Generation from TRL is currently used in the database entity description
(or simply ``description'') module, which is linked to the semantics of 
English verbs like
``talk about (something)'' and ``tell (someone) about (something)''. 
Processing of a sentence such as
\begin{quote}
{\it Talk about Peter Piper.}
\end{quote}
produces a call to describe the conceptual object
\begin{quote}\begin{verbatim}
SF(c(x^[employee_Worker,x],Peter Piper))
\end{verbatim}\end{quote}
The description module is passed the conceptual object $X$ that is to be
described, and begins by finding all conceptual goals $C_i$
(see section~\ref{DBLDT}) which have the property that at least 
one argument position is filled by an occurrence of $X$. This 
provides an abstract ``conceptual-level'' summary of ``what the
database knows about $X$''. The main part of the processing is then
carried out by two sub-modules,
the {\it utterance synthesizer} and the {\it utterance filter}, 
which are called in alternation.
The synthesizer uses the TRL description rules to attempt to find new
utterances whose truth-conditions are implies by some $C_i$.
After each such utterance is
found, but before it is displayed, the utterance filter is called
into play to determine whether or not the new utterance should be
shown to the user, and if so whether any more utterances need be searched for.
We illustrate the behaviour of the description module by sketching
the course of processing for the example given immediately above.
The first $C_i$ found is the predication {\tt (C1)}
\begin{verbatim}
employee(SF(c(x^[company_Organization,x],CRI)),
         SF(c(x^[employee_Worker,x],Peter Piper)),
         m,
         y)                                                  (C1)
\end{verbatim}
which conveys the information that the employee Peter Piper is a employee
at the company CRI, is of sex {\tt m} (for ``Man''), and has a car
({\tt y} as opposed to {\tt n}).

The utterance synthesizer is called first. When passed a conceptual
goal $C_i$, the synthesizer begins by computing
the set of goals logically implied by the union of $C_i$ with 
the current linguistic domain theory {\it excluding} the database: the
result is then uses as input to the learned TRL description rules.
The efficiency of the process is increased by pre-compiling the
set of consequences of a generic representative of each class of
tuple, and caching the results in the form of ``TRL lemmas''.
The relevant lemmas will be of the form
\begin{verbatim}
Conditions -> (<C> -> <L>)
\end{verbatim}
where \verb!<C>! is a conceptual goal matching $C_i$, 
{\tt Conditions} is a conjunction of evaluable 
predicates, and \verb!<L>! is an atomic goal whose predicate
is a linguistic predicate
(cf sections~\ref{TRL-lemmas} and~\ref{Abductive-proof}).
Continuing with the example, synthesis from the goal {\tt (C1)}
proceeds as follows. First the TRL lemmas are used to find
the set of consequences of {\tt (C1)}.
When the concrete values for the relevant instances have been
instantiated, the result is a list which includes the goals
\begin{quote}\begin{verbatim}
car_Vehicle(sk12(SF(c(x^[employee_Worker,x],Peter Piper))))
have_3p(sk13(SF(c(x^[employee_Worker,x],Peter Piper))),
        SF(c(x^[employee_Worker,x],Peter Piper)),
        sk12(SF(c(x^[employee_Worker,x],Peter Piper)))),
male_Masculine(SF(c(x^[employee_Worker,x],Peter Piper))),
man_MalePerson(SF(c(x^[employee_Worker,x],Peter Piper))),
name_IdentifyingPhraseOf(Peter Piper,SF(c(x^[employee_Worker,x],
                         Peter Piper))),
sex_GenderOf(Male,SF(c(x^[employee_Worker,x],Peter Piper))),
person_HumanBeing(SF(c(x^[employee_Worker,x],Peter Piper))),
at_Locational(SF(c(x^[employee_Worker,x],Peter Piper)),
              SF(c(x^[company_Organization,x],CRI))),
employee_Worker(SF(c(x^[employee_Worker,x],Peter Piper))),
\end{verbatim}\end{quote}
Here, {\tt sk12(X)} and {\tt sk13(X)} are Skolem functions representing,
respectively, the car employee {\tt X} has (if such an object exists),
and the event of his or her having it. It is now possible
to attempt to apply learned description rules, some of which succeed.
For example, the first two ``consequence'' goals licence an application of
the rule generalized from
\begin{quote}
{\it John loves a cat.}
\end{quote}
with {\tt have\_3p} substituting {\tt love\_\-Like}, 
{\tt car\_\-Vehicle} substituting {\tt cat\_Animal}, and
{\tt Peter Piper} substituting {\tt John}.
Synthesis proceeds from the QLF produced, yielding the sentence
\begin{quote}
{\it Peter Piper (the employee) has a car.}
\end{quote}
Before the sentence is displayed, the utterance filter is called.
The filter keeps track of the combined propositional
content of the sentences so far uttered when describing the current tuple,
which is stored as a global variable $P_{old}$. (Initially, the value
of $P_{old}$ is $true$).
When the filter is passed a new candidate utterance with propositional
content $P$, it first ascertains whether or not $P$ is implied by 
$P_{old}$. If this is the case, it returns the information that the new
utterance is redundant and need not be displayed. If not, it displays
the new utterance, and updates the combined propositional content to
$P_{old}\wedge P$. It then checks again to find out whether $P_{old}\wedge P$
is equivalent to the current tuple. If so, no more utterances need to be
generated.

The expression representing the content is first transformed if necessary
by replacing Skolem functions with existentially quantified variables
and then translated into conceptual form. Thus
the propositional content of the first utterance is originally
\begin{verbatim}
and(car_Vehicle(sk12(SF(c(x^[employee_Worker,x],Peter Piper)))),
    have_3p(sk13(SF(c(x^[employee_Worker,x],Peter Piper))),
            SF(c(x^[employee_Worker,x],Peter Piper)),
            sk12(SF(c(x^[employee_Worker,x],Peter Piper)))))
\end{verbatim}
After replacement of Skolem functions by existentially quantified variables,
this becomes
\begin{verbatim}
exists([X,Y],
   and(car_Vehicle(X),
       have_3p(Y,
               SF(c(x^[employee_Worker,x],Peter Piper)),
               X)))                                           (C2)
\end{verbatim}
and after translation,
\begin{verbatim}
exists([X],
   employee(SF(c(x^[company_Organization,x],CRI)),
            SF(c(x^[employee_Worker,x],Peter Piper)),
            X,
            y)                                                (C3)
\end{verbatim}
Translating {\tt (C2)} into the canonical form {\tt (C3)}
makes it simple to ascertain that the 
tuple has so far not been completely described, by comparing {\tt (C3)} and 
the original tuple {\tt (C1)}.
 On the next cycle,
the compiled consequence goal
\begin{quote}\begin{verbatim}
man_MalePerson(SF(c(x^[employee_Worker,x],Peter Piper)))
\end{verbatim}\end{quote}
is used to generate the sentence
\begin{quote}
{\it Peter Piper (the employee) is a man.}
\end{quote}
The conjunction of {\tt (C3)} and the propositional content of the second
sentence now however translates to {\tt (C1)}, so the filter can return 
with the information that the tuple has been completely described.

For tuples whose structure is more complex, the filter leads to a 
very substantial reduction in the verbosity of the output produced
by the description module. For example, when generating from a 
{\tt project} tuple the output produced is something like the following
\begin{verbatim}
>>>>>>  CLAM-BAKE (the project)'s end date is 19/11/1992 (the day).
>>>>>>  CLAM-BAKE (the project)'s number is 8468.
>>>>>>  CLAM-BAKE (the project)'s start date is 20/11/1989 (the day).
\end{verbatim}
With the utterance filter disabled, the output becomes
\begin{verbatim}
>>>>>>  CLAM-BAKE (the project)'s end date is 19/11/1992 (the day).
>>>>>>  CLAM-BAKE (the project)'s number is 8468.
>>>>>>  CLAM-BAKE (the project)'s start date is 20/11/1989 (the day).
>>>>>>  CLAM-BAKE (the project)'s account is account 8468.
>>>>>>  CLAM-BAKE (the project)'s project number is 8468.
>>>>>>  CLAM-BAKE (the project)'s account number is 8468.
>>>>>>  19/11/1992 (the day) is an end date.
>>>>>>  8468 is a number.
>>>>>>  20/11/1989 (the day) is a start date.
>>>>>>  account 8468 is an account.
>>>>>>  CLAM-BAKE (the project) is a project.
>>>>>>  8468 is a project number.
>>>>>>  8468 is an account number.
>>>>>>  CLAM-BAKE (the project) began on 20/11/1989 (the day).
>>>>>>  CLAM-BAKE (the project) ended on 19/11/1992 (the day).
>>>>>>  CLAM-BAKE (the project) finished on 19/11/1992 (the day).
>>>>>>  CLAM-BAKE (the project) started on 20/11/1989 (the day).
>>>>>>  CLAM-BAKE (the project) began.
>>>>>>  CLAM-BAKE (the project) ended.
>>>>>>  CLAM-BAKE (the project) finished.
>>>>>>  CLAM-BAKE (the project) started.
\end{verbatim}
As can been seen, the filter's simple uniform mechanism manages to eliminate
several different kinds of excess verbosity, some of which are not
obviously redundant and require non-trivial inference to detect.

\subsection{Generating Questions from Incomplete Assertions}\label{Generating-Questions}

It is also possible, using the methods we have just
described, to generate questions. The module in CLARE 
responsible for doing this is 
invoked by the {\tt .ask} command: if some incomplete tuples have been
created by processing assertions (see section~\ref{Simplify-assertions}),
CLARE will respond to {\tt .ask} by entering a loop where it poses
questions to the user intended to elucidate the information necessary to
fill the unknown argument positions. 
This can involve generation of both Y-N and WH- questions. We begin by 
giving an example dialogue: utterances by CLARE and the user are 
prefixed by \verb!clare>! and \verb!user>! respectively, and comments are
in italics.
\begin{verbatim}
user>  ARLA is a project.
\end{verbatim}
{\it The user inputs a declarative statement. After processing it, CLARE has
an incomplete {\tt project} record.}
\begin{verbatim}
user>  .ask
\end{verbatim}
{\it The user invokes the question-asking module to fill the unspecified fields
in the record.}
\begin{verbatim}
clare> what is ARLA (the project)'s account?
\end{verbatim}
{\it CLARE asks a question to find out the filler of the 
``account'' field in the record, and hands control back to the user to get the
answer.}
\begin{verbatim}
user>  1234.
\end{verbatim}
{\it The user answers. The utterance is processed by CLARE in the normal way,
which involves syntactic and semantic analysis, followed by ellipsis resolution
in the context of the previous question.}
\begin{verbatim}
clare> account 1234 is ARLA (the project)'s account.
\end{verbatim}
{\it CLARE prints out a paraphrase showing the result of performing resolution,
to keep the user informed. It then translates the result in the usual way, merging
it with the previous one as described in section~\ref{Simplify-assertions}.}
\begin{verbatim}
clare> what is ARLA (the project)'s start date?
\end{verbatim}
{\it CLARE asks another question, this time to find out the project's start date.
Processing is as before.}
\begin{verbatim}
user>  1/4/93.
clare> 1/4/1993 (the day) is ARLA (the project)'s start date.
clare> what is ARLA (the project)'s end date?
\end{verbatim}
{\it Another question from CLARE; this time, the user replies with a sentence.
Processing is still the same as usual.}
\begin{verbatim}
user>  ARLA will end on 1/4/96.
\end{verbatim}
{\it When this sentence has been processed, CLARE has successfully filled all the
fields in the tuple.}

The range of questions currently handled is limited to Y-N questions whose propositional
content is an existentially quantified conjunction, and WH-questions whose
propositional content is a lambda-bound existentially quantified conjunction. For 
questions of these types, we can 
acquire case-based rules by a slight extension of the methods described in
section~\ref{Case-based-learning};
the only extra point is that it is necessary to store the lambda-bound variable
for a WH-question. Thus for example the rule induced from the training 
sentence
\begin{quote}
{\it Who loves Mary?}
\end{quote}
is
\begin{verbatim}
trl_gen([whq,
         form(verb(pres,no,no,no,y),
              A,
              B^[B,[C,A,term(q(tpc,wh,_),_,D^[personal,D],_,_),E]],_)],
        F^exists([G],
             and([personal,F],
                 [C,G,F,H]))),
        [entity_ref_term(H,E),
         personal(I),
         trlgoal([C,_,I,H],
                 genpred(C,(form(_,verb(_,_,_,_,_),
                                 J,K^[K,[C,J,_,_]],_),
                                 v(_,_))))],
        [I])
\end{verbatim}
where the fourth argument {\tt [I]} to {\tt trl\_gen} is the list of
lambda-bound variables in the conditions; note that the second argument,
the propositional content, is already a lambda-bound form. We now describe
how rules of this kind can be used.

Let us begin by supposing that we have a conceptual goal $C$, in which
some argument $X$ is so far uninstantiated. There are two obvious
strategies to use when trying to find a filler for $X$. The first is 
to attempt to frame a WH-question whose propositional content is equivalent
to $\lambda X.C$; the second is to substitute some plausible value $a$ for
$X$ in $C$, and then try to find a Y-N question whose content is equivalent to
$C[X/a]$. CLARE can use either strategy, choosing according to the nature
of the argument position. If it is an argument that has to be filled by
one of a known finite set of possible code values, it tries the second
strategy; otherwise, it uses the first one. 

For both strategies, the major difference compared to generating assertions is that
what is required is an utterance whose propositional content is {\it equivalent} to that
of the question, not {\it implied} by it. We will 
refer to the initial expression, either $\lambda X.C$ or $C[X/a]$ as the case
may be, as the {\it abstract question}; we can reduce the two cases to one
by considering abstract questions as forms bound by a set of lambdas, where the
set is empty for Y-N questions and non-empty for WH-questions. 

Suppose then that the abstract question is $\lambda \vec{X}.C$ where 
$\vec{X}$ is a vector of variables. Generation proceeds as follows. We
try to find a description rule $R$, generalized from a question, for which
the set of lambda-bound variables is $\vec{Y}$, the conditions are
$Conds$, the propositional content is $\lambda \vec{Y}.P_R$, 
and the following hold:
\begin{enumerate}
\item The cardinalities of $\vec{X}$ and $\vec{Y}$ are equal.
\item If $\vec{a}$ is a set of unique constants with the same cardinality as
$\vec{X}$ and $\vec{Y}$, then $Conds[\vec{Y}/\vec{a}]$ is implied by
$C[\vec{X}/\vec{a}]$. 
\item $P_R[\vec{Y}/\vec{a}]$ implies $C[\vec{X}/\vec{a}]$
\end{enumerate}
If these conditions are met, then the propositional content of $R$ is equivalent
to that of $\lambda \vec{X}.C$. The conditions are phased the way they are so as to
allow use of the techniques presented in section~\ref{Talk-about}; thus the
second condition is satisfied using the compiled TRL lemmas, and the third one
with essentially the same mechanism as that used by 
the ``utterance filter'' described at the end
of the last section.

\chapter{The Reasoning Engine}
\label{ReasoningEngine}

CLARE's inference engine is used by several other components.
It supports proofs from conjunctive contexts when performing
translation (cf. section~\ref{Translation-schemas}), and evaluates
translated queries; it is also employed to compile the lemmas 
used when generating facts and questions (chapter~\ref{genass}),
and for various other purposes during reference resolution
(chapter~\ref{clareinterp}). In each case,
the basic functionality is some version of the following.
We are provided with a Horn-clause theory $\Gamma$, a formula $F$ 
(which may contain free variables $\bf \vec{x}$),
and a criterion which determines
permissible abductive assumptions. The task is to 
find a substitution $\theta$ on $\bf \vec{x}$
and a set of permissible assumptions $A$ such that 
\begin{eqnarray}
\Gamma \Rightarrow (A \rightarrow \theta(F))\label{Basic-infer}
\end{eqnarray}
or terminate with the information that no such $\theta$ and $A$
exist. This chapter describes how the inference functionalities are
realized in CLARE as a type of Prolog meta-interpreter.
Since these have been
described many times in the literature (e.g. (Sterling and Shapiro, 1986)),
we will concentrate on the unusual features of the CLARE inference
engine. The expositional strategy used will be to describe a simplified
version of the inference engine, beginning with a 
the basic meta-interpreter for pure Horn-clause theories
shown in figure~\ref{Basic-interpreter} and adding features as we
go along to cover new functionalities.
\begin{figure}
\hrule
\begin{verbatim}
prove(and(L,R)) :-
  prove(L),
  prove(R).

prove(G) :-
  horn_clause_in_theory(G <- Body),
  prove(B).

prove(G) :- 
  unit_clause_in_theory(G).
\end{verbatim}   
\hrule
\caption{Basic Horn-clause interpreter\label{Basic-interpreter}}
\end{figure}

\section{Logical structure of the reasoning engine}\label{Reasoning-basic}

The basic form of the inference engine is that of
a slightly extended Prolog meta-interpreter. Superimposed
on top of it, there is an iterated-deepening A* search strategy.
We will only consider the logical aspects of
the inference engine in this section, 
postponing discussion of the search strategy 
until section~\ref{Inference-search}. There are two important
extensions over the standard Prolog meta-interpreter: these 
are to cover abductive proofs and proofs of universally quantified
implications. We take them in turn.

\subsection{Abductive proofs}\label{Abductive-proof}

We will use the term ``abductive proof'' in this section to mean
construction of a proof-tree in which some leaves are undischarged
assumptions. CLARE performs abductive proofs for two essentially
distinct reasons. The first has been referred to many times
already in chapters~\ref{translation} and~\ref{claredm}; when 
using logic to reason
about common-sense concepts, such as those expressed in natural 
language, it is frequently necessary to assume background knowledge
which is not part of the explicit content of the utterance's
logical representation. In this case, the abductively assumed 
conditions are normally shown to the user in order to check their
appropriateness. Abductive proofs are also used in order to
compile ``implicational lemmas'', which are utilized by several
other components of the system (see chapters~\ref{clareinterp} 
and ~\ref{generation}). In this mode of operation, the reasoning
engine is given a goal $F$ containing free variables $\vec{x}$
and a set of criteria for making abductive assumptions; the assumptions
may also contain free variables. If an abductive proof can be found with 
assumptions $A$ and a substitution $\theta$ for the $\vec{x}$,
this can be regarded as a strict (non-abductive) proof for the
Horn-clause
\begin{displaymath}
\forall \vec{y}.(A \rightarrow \theta(F))
\end{displaymath}
where $\vec{y}$ are the free variables in $\theta(\vec{x})$.
We will refer to formulas of this kind as ``implicational TRL lemmas'',
or just ``TRL lemmas''. As part of the process of compiling a linguistic
domain theory, all TRL lemmas of a particular type are generated.
At present,
the criterion used for limiting abductive assumptions restricts $A$
to being a set of which all but at most two members are 
{\it arithmetic relation goals}
(cf section~\ref{Effective-translation}). The remaining members of $A$ are
either a single atomic predication whose predicate is
a conceptual predicate, or a conjunction of two such predications
sharing a common variable and licenced by a {\tt db\_\-chain}
declaration (see the manual). 

Two minor changes 
are needed to extend a Prolog meta-interpreter to allow abductive
proofs: another argument is added to the main 
predicate, which passes around a difference-list holding the assumptions,
and an extra clause is added which allows ``proof'' of an atomic
goal by adding it to the assumption-list, if the abductive assumability
criteria permit this. The basic interpreter, modified in the way described,
is shown in figure~\ref{Abductive-interpreter}.
\begin{figure}
\hrule
\begin{verbatim}
prove(and(L,R),AssumptionsIn-AssumptionsOut) :-
  prove(L,AssumptionsIn-AssumptionsNext),
  prove(R,AssumptionsNext-AssumptionsOut).

prove(G,AssumptionsIn-AssumptionsOut) :-
  horn_clause_in_theory(G <- Body),
  prove(Body,AssumptionsIn-AssumptionsOut).

prove(G,AssumptionsIn-AssumptionsIn) :-
  unit_clause_in_theory(G).

prove(G,AssumptionsIn-[G|AssumptionsIn]) :-
  abductively_assumable(G).
\end{verbatim}   
\hrule
\caption{Abductive Horn-clause interpreter\label{Abductive-interpreter}}
\end{figure}

\subsection{Proving implications}\label{Prove-implications}

We now consider the task of extending the inference engine to 
be able to prove formulas of type
\begin{quote}
\begin{verbatim}
forall(Vars,impl(LHS,RHS))
\end{verbatim}
\end{quote}
where the variables in {\tt Vars} occur free in {\tt LHS} and {\tt RHS}.
Two well-known strategies exist for attempting to prove such expressions;
they can be proved either {\it intensionally} or {\it extensionally}.
Intensional proofs are carried out by substituting unique constants 
for the {\tt Vars}, assuming {\tt LHS}, and attempting to prove {\tt RHS};
extensional proofs by listing all values of {\tt Vars} for which a proof
of {\tt LHS} can be found, and for each one ascertaining that a proof
of {\tt RHS} exists. The CLARE inference engine can use either strategy.

To implement the intensional proof strategy it is necessary to add yet another
argument to the {\tt prove} predicate in the interpreter, to hold the
set of assumptions derived from the {\tt LHS}; it is tempting to try
and include them in the abductive assumption list, but in practice this
manoeuvre does not seem to simplify matters. 
The need to
take account of possible abductive assumptions also introduces
some complications in the extensional
strategy, since it is necessary to collect the
assumptions used for each separate proof of {\tt RHS}. The skeleton
interpreter, modified to allow proof of universally quantified 
implications by the intensional and extensional strategies, 
is presented in figure~\ref{Implic-interpreter}. Here the first
clause of form {\tt prove(forall...)} is for the intensional
strategy; {\tt replace\_\-by\_\-unique\_\-constants} is assumed to be a
predicate which replaces all the variables in its argument
by unique ground terms. The following clause is for the 
extensional strategy. Here 
{\tt safe\_bagof(X,G,Xs)} is like 
{\tt bagof(X,G,Xs)}, with the difference that it first binds any
free variables in G, and succeeds with {\tt Xs=[]} if there is no 
proof of {\tt G}; {\tt union\_\-n\_\-1(List,Union)} takes a list of lists
as its first argument and returns their union as the second one.

\begin{figure}
\hrule
{\tt (All clauses from the interpreter in figure\ref{Abductive-interpreter})}
\begin{verbatim}

prove(G,AssumptionsIn-AssumptionsIn,ImplAssumptions) :-
  member(G,ImplAssumptions).

prove(forall(Vars,impl(LHS,RHS)),AssIn-AssOut,ImplAss) :-
    replace_by_unique_constants(Vars),
    add_implicational_assumptions(LHS,ImplAss,ImplAss1),
    prove(R,AssIn-AssOut,ImplAss1).

prove(forall(Vars,impl(LHS,RHS)),AssIn-AssOut,ImplAss)
    safe_bagof(environment(Vars,AssOutLeft),
               prove(LHS,AssIn-AssOutLeft,ImplAss)
               LeftEnvironmentList),      
    prove_forall_cases(LeftEnvironmentList,Vars^RHS,
                       ImplAss,AssOutList),
    union_n_1(AssOutList,AssOut).

prove_forall_cases([],_Vars^_RHS,_ImplAss,[]).
prove_forall_cases([EnvFirst|EnvRest],Right,ImplAss,
                   [AssOutFirst|AssOutRest]) :-
    prove_forall_case(EnvFirst,Right,ImplAss,AssOutFirst),
    prove_forall_cases(EnvRest,Right,ImplAss,AssOutRest).

prove_forall_case(environment(VarsLeft,AssOutLeft),
                  Vars^RHS,ImplAss,AssOut) :-
    copy_term([Vars^RHS,ImplAss],[Vars1^RHS1,ImplAss1]),
    VarsLeft = Vars1,
    prove(RHS1,AssOutLeft-AssOut,ImplAss1).

add_implicational_assumptions(and(P,Q),ImplAssIn,ImplAssOut) :- !,
    add_implicational_assumptions(P,ImplAssIn,ImplAssNext),
    add_implicational_assumptions(Q,ImplAssNext,ImplAssOut).
add_implicational_assumptions(G,ImplAss,[G|ImplAss]) :- 
    atomic_goal(G), !.
add_implicational_assumptions(_Other,ImplAss,ImplAss).
\end{verbatim}   
\hrule
\caption{Abductive interpreter for Horn-clauses and implications
\label{Implic-interpreter}}
\end{figure}

\section{Search strategies for inference}\label{Inference-search}

In this section we consider the search strategy used in the
inference engine, which as already stated is an
iterated-deepening A* search strategy
(Korf, 1986, Stickel, 1986); that is, search proceeds by first defining a
cost limit $D$, and then performing a depth-first search where
each branch is cut off when the sum of the cost of goals already
processed, together with an estimate of the cost of solving pending
goals, exceeds $D$. The cost function used at present charges 
one unit for each rule application, and conservatively estimates
one unit for each pending goal; as explained in section~\ref{Assumptions},
it is possible to define extra costs for abductively assumed
goals. The motivation for using IDA*, rather
than simple depth-first search, is that it avoids getting stuck
in infinite recursions; it essentially performs the function
of simulating breadth-first search, and is thus theoretically complete.
However, experiment quickly shows that infinite recursion
still causes severe practical difficulties without further elaboration
of the method. 

There are basically three problems, which to some extent are 
related. Firstly, since the greater part of the
Horn-clause theory is derived from a set of equivalences 
(cf. section~\ref{Translation-schemas})
it is easy to wander aimlessly around in
the search-space, alternately using ``normal'' and ``backward''
Horn-clause versions of equivalences. 
Secondly, even when this does not happen,
it may be the case that there are multiple redundant proofs of
a goal. The third problem arises from inefficient ordering of
conjunctions. When trying to prove a conjunction, it can sometimes
be the case that there are many proofs of the first conjunct (which
take a correspondingly long time to generate), but 
that the second conjunct can quickly be discovered to have either one
or no proofs; in this case attempting to prove the conjuncts in 
a different order can pay dividends.
The non-obvious aspects of the search strategy,
which we will now describe,
stem from attempts to counteract these problems. We describe them 
one at a time.

\subsection{Avoiding infinitely recursive branches}

We currently use two different approaches to
help minimize time spent in infinitely recursive branches of the 
search-tree. The first approach is based on a standard loop-avoidance
method. It can be proved that it is sound to cut off search
if a goal is {\it identical} to one of its ancestors. The inference
engine uses this criterion, both in its simple form and in a ``look-ahead''
mode: when expanding a goal $G$ to a conjunction $B_1\wedge B_2 \ldots$, 
it checks before adding the $B_i$ to the stack of pending goals,
to make sure that none of them are identical to an ancestor.
Perhaps surprisingly (since this is a fairly expensive operation),
it turns out that performing the ``look-ahead'' check 
increases efficiency quite substantially.

It would be pleasant if the ``identity'' condition could be
relaxed a little, for example by replacing it with a test of
unification or subsumption with one of the ancestor goals.
Unfortunately, it is easy to prove that weakening the condition
is in general not correct. However,
a strong syntactic similarity between a goal and an ancestor is
often a good indicator that a loop has arisen, and can be taken as
a justification for exacting a ``penalty'', causing the 
cost limit to be reached more quickly. In the current version,
the penalty is applied if the goal is subsumed by an ancestor,
a strategy which we have found to work well in practice. The method is 
sound, since its only effect is to devote
less effort to search of the affected branches without cutting
them off completely, and appears rather more general than
the approach advocated in (Appelt and Hobbs 90), which involves recoding
of the meaning postulates.

The second approach to the ``looping'' problem builds on the assumption
that the equivalences should be read as translation rules; the
point is to exploit this when using them in their compiled
Horn-clause forms. Recall that 
\begin{eqnarray}
P_1\wedge P_2\wedge \ldots \leftrightarrow Q_1\wedge Q_2\wedge \ldots 
                                                     \label{Original-equiv}
\end{eqnarray}
produces one ``backward'' Horn-clause,
\begin{eqnarray}
Q_i \leftarrow P_1\wedge P_2\wedge \ldots \label{Backwards-HC}
\end{eqnarray}
for each $Q_i$. Now suppose that $Q_i$ occurs as a condition when 
trying to expand some other goal $G$. 
We demand that the rule~(\ref{Backwards-HC}) only be invoked to
prove $Q_i$ if this is consistent with the assumption that
$Q_i$ could be produced from the original rule by a chain of translations
the last one of which used~(\ref{Original-equiv}). (In other words,
(\ref{Backwards-HC}) can only be used to attempt to reconstruct
such a chain). What this means in terms of limiting search is that
the first goal in~(\ref{Backwards-HC}), $P_1$, must now be proved 
either by looking it up in the environment of assumptions, or by
invoking a further ``backward'' Horn-clause of which $P_1$ is the
head. The result is that search now proceeds only by chaining 
``backward'' rules, until either a goal can be found in the
environment or a dead-end is reached, a process that normally
terminates quickly. Without the restriction, any ``backward'' rule
invocation could be followed by a ``normal'' one, leading to aimless
meandering in the search-space that is only terminated by the
depth-bound being exceeded. For non-trivial sets of equivalences
the restriction on use of ``backward'' Horn-clauses seems necessary,
but also intuitively acceptable.

\subsection{Avoiding redundant search}

The second problem, that of avoiding redundant
recomputation, is currently only
applied to ground goals, that is goals which contain
no uninstantiated variables. We have observed in many cases
that, without some kind of restrictions,
a great deal of effort is wasted in producing multiple
proofs for goals of this kind. Since a proof of a ground goal
is basically just a ``yes'' or ``no'', it is tempting to
adjust the backtracking mechanism so as to
block attempts to find new proofs of ground goals that have already
been successfully proved. However, this is not quite right; 
the problem is that the ordering of the search-space may mean
that the first proof produced is an expensive one, which 
leaves the search process on the verge of exceeding the cost
limit, and prevents further progress. The correct realization of
the idea seems rather to be to block retrying of ground goals
only if the cost of the existing proof falls under a given
bound, which we have at present arbitrarily fixed as a third
of the total cost limit.

\subsection{Dynamic re-ordering of conjuncts}

The time taken to find a proof for a conjunctive goal (or to ascertain
that there is no proof) can often be strongly dependent on the order
in which the conjuncts are attempted. We have made no attempt to solve
this problem in its general form, but two special cases are
sufficiently important that the inference engine needs to take account
of them. Suppose first that the interpreter is about to attempt to
prove the conjunction $L\wedge R$.  If it is the case that there is no
proof of $R$, then time lost establishing this can be recovered by
omitting the attempt to prove $L$. Thus it can pay to perform
``look-ahead'' and search for goals which can quickly be shown to be
unachievable. Similarly, if $L$ contains free variables which also
occur in $R$ then the time needed to find a proof can be reduced if
the number of proofs for $R$ is substantially less than the number of
proofs for $L$, since instantiating free variables nearly always has
the function of reducing the number of possible proofs. In particular,
if it can quickly be shown that there is only {\it one} proof of $R$
then it is very likely that it will be correct to attempt it first.

It is possible to include declarations of the form
\begin{quote}\begin{verbatim}
quick_failure_test(<Goal>) :- <Conds>
\end{verbatim}\end{quote}
or
\begin{quote}\begin{verbatim}
quick_determinism_test(<Goal>) :- <Conds>
\end{verbatim}\end{quote}
where \verb!<Goal>! is an atomic goal and \verb!<Conds>! an arbitrary
Prolog form; the intended semantics are that there should be, respectively,
no proofs or exactly one proof of \verb!<Goal>! if \verb!<Conds>! hold.
For declarations of this kind to be useful, \verb!<Conds>! should be
quickly evaluable. In the present version of CLARE, instances of 
\verb!<Conds>! never do more than check instantiations of variables,
or attempt to unify them with other terms.
Tests for both cases are applied when a
Horn-clause is used to expand a goal {\tt G}; after the head of
the clause has been unified with {\tt G}, the goals in the body
are examined. If a {\tt quick\_\-failure\_\-test} succeeds then 
proof of the whole conjunction fails; if a {\tt quick\_\-determinism\_\-test}
succeeds on a subgoal, then that subgoal is evaluated first. There
are currently about 30 ``quick failure and determinism'' declarations,
all of them for the ``conversion'' predicates described in 
section~\ref{Equiv-dbconnect}.

\section{Finding finite proof procedures}\label{Qopt}

In the final section of this chapter we describe the module responsible
for searching for finite proof procedures 
(see also section~\ref{Effective-translation}). Since the inference
engine is basically a Prolog meta-interpreter, it is sensitive to
the ordering of conjuncts: it attempts to prove a goal of the form
{\tt and(P,Q)} by first proving {\tt P} and then proving {\tt Q}. If
{\tt P} and {\tt Q} share a free variable {\tt X}, the order can be important.
To repeat the example from~\ref{Effective-translation}, if $TRANS/3$ is a
database predicate then the strategy ``find an $X$ 
such that $X > 100$, then find values of
$Y$ such that $TRANS(john,X,Y)$'' is not a finite strategy; however, reversing
the order to make the strategy ``find $X$ and $Y$ such that 
$TRANS(john,X,Y)$, then determine whether $X > 100$ holds'' is finite.

The search for finite proof strategies is carried out by another extended
Prolog meta-interpreter, using an abstract interpretation method. 
The information needed to deal with the base case of attempting to 
prove an atomic goal is provided by declarations of the form
\begin{quote}
\begin{verbatim}
call_pattern(<Pred>(Arg1,Arg2,...Argn),InArgs,OutArgs)
\end{verbatim}
\end{quote}
where \verb!<Pred>! is a predicate and {\tt InArgs} and {\tt OutArgs}
are subsets of the set {\tt \{Arg1, ..., Argn\}}.
The intended semantics are that there are finitely many proofs of
the goal 
\begin{quote}
\begin{verbatim}
<Pred>(Arg1,Arg2,...Argn)
\end{verbatim}
\end{quote}
if the variables in {\tt InArgs} are already instantiated, 
and that each of them will 
result in the arguments {\tt OutArgs} becoming instantiated.
The basic structure of the interpreter is shown in 
figure~\ref{Qopt-interpreter}; only the clauses for conjunctive
and atomic goals are displayed. 
\begin{figure}
\hrule
\begin{verbatim}
rearrange(OldForm,NewForm) :-
    copy_term(OldForm,EvalForm),
    rearrange0(OldForm,NewForm,EvalForm,[]-[],[]-_), !.

rearrange0(Old,New,Eval,FrozenIn-FrozenOut,DoneIn-DoneOut) :-
    rearrange1(Old,New0,Eval,FrozenIn-FrozenNext,DoneIn-DoneNext),
    try_to_thaw(New0,New,FrozenNext-FrozenOut,DoneNext-DoneOut).

rearrange1(and(LOld,ROld),and(LNew,RNew),and(L,R),
           FrozenIn-FrozenOut,DoneIn-DoneOut) :- !,
    rearrange0(LOld,LNew,L,FrozenIn-FrozenNext,DoneIn-DoneNext),
    rearrange0(ROld,RNew,R,FrozenNext-FrozenOut,DoneNext-DoneOut).
rearrange1(OldG,OldG,G,Frozen-Frozen,Done-[goal_executed|Done]) :-
    abstract_call(G),!.
rearrange1(OldG,true,G,Frozen-[frozen(G,OldG)|Frozen],Done-Done).

try_to_thaw(Form,Form,[]-[],Done-Done) :- !.
try_to_thaw(Form,OutForm,
            [frozen(FrozenEval,FrozenOld)|FrozenNext]-FrozenOut,
            DoneIn-DoneOut) :-
    rearrange1(FrozenOld,FrozenNew,FrozenEval,[]-[],[]-[_|_]),!,
    try_to_thaw(and(Form,FrozenNew),OutForm,FrozenNext-FrozenOut,
                [goal_executed|DoneIn]-DoneOut).
try_to_thaw(Form,OutForm,
            [FrozenFirst|FrozenNext]-[FrozenFirst|FrozenOutRest],
            DoneIn-DoneOut) :-
    try_to_thaw(Form,OutForm,FrozenNext-FrozenOutRest,DoneIn-DoneOut).

abstract_call(X=Y) :-
     X = Y, !.
abstract_call(_X=_Y) :- !.
abstract_call(G) :-
     call_pattern(G,InArgs,OutArgs),
     is_instantiated(InArgs),
     make_instantiated(OutArgs).     
\end{verbatim}   
\hrule
\caption{Interpreter for finding finite proof strategies
\label{Qopt-interpreter}}
\end{figure}
In the implemented interpreter,
there is a clause for each logical operator of the TRL language.
The central idea is to rearrange
the expression by using an abstract version of the Prolog
``freeze'' or delayed-goal primitive. 
The arguments of the main predicates
{\tt rearrange0} and {\tt rearrange1} are
\begin{quote}
\begin{verbatim}
rearrange0/1(Original,Rearranged,Eval,Frozen,Done)
\end{verbatim}
\end{quote}
where
\begin{description}
\item[{\tt Original}] is the original expression.
\item[{\tt Rearranged}] is the rearranged expression.
\item[{\tt Eval}] is a copy of {\tt Original} used to keep track of 
      the variables currently instantiated in the abstract interpretation.
\item[{\tt Frozen}] is a difference-list of delayed goals, i.e. goals which
     were insufficiently instantiated to be executed when first called.
\item[{\tt Done}] is a difference-list that keeps track of progress made
     in evaluating goals, to prevent infinite postponement of goal 
     execution.
\end{description}
the two predicates call each other recursively, with {\tt rearrange0}
calling {\tt rearrange1} and then trying to ``thaw'' any frozen goals
that may have been created.
Note the special treatment of equality goals: if {\tt Eval} is
an equality, its arguments are unified if possible.

\chapter{The Preference Mechanism}
\label{clarepref}

Linguistic coverage of the CLE subsystem of CLARE has been increasing
gradually as the system develops, both in terms of grammatical coverage
and in the coverage of referential terms handled during interpretation. 
While this has the obvious advantage of being able to handle new types
of sentences, it also increases the number of analyses and interpretations 
produced for a sentence. If these alternatives are regarded as equally 
likely, then for sentences that were already treated by the narrower 
coverage the chance of selecting the correct interpretation of a sentence 
can decrease with wider coverage. Clearly this problem can be overcome
by providing a mechanism that gives an accurate indication of when
to prefer one alternative over another. 
This chapter describes preference metrics and mechanisms implemented in 
CLARE (for related approaches, see McCord 1989 or Hobbs and Bear 1990).
We start by describing the role of the preference mechanism
in the treatment of alternatives in the current architecture, then the 
general approach to computing preferences. The last three sections describe 
the actual preferences used by the system.

\section{Preference and the CLE Architecture}

The preference mechanism applies to unresolved semantic analyses as
well as to resolved and scoped interpretations. More specifically,
the mechanism is currently applied at the following three points
in the course of processing input sentences:
\begin{itemize}

\item the derivation of a set of unresolved QLFs by the application
of semantic rules and sense entries to the packed syntactic analysis
records;

\item the derivation of resolved QLFs by the application of reference
resolution rules;

\item the application of the scoping algorithm to resolved QLFs yielding
scoped resolved QLFs.

\end{itemize}

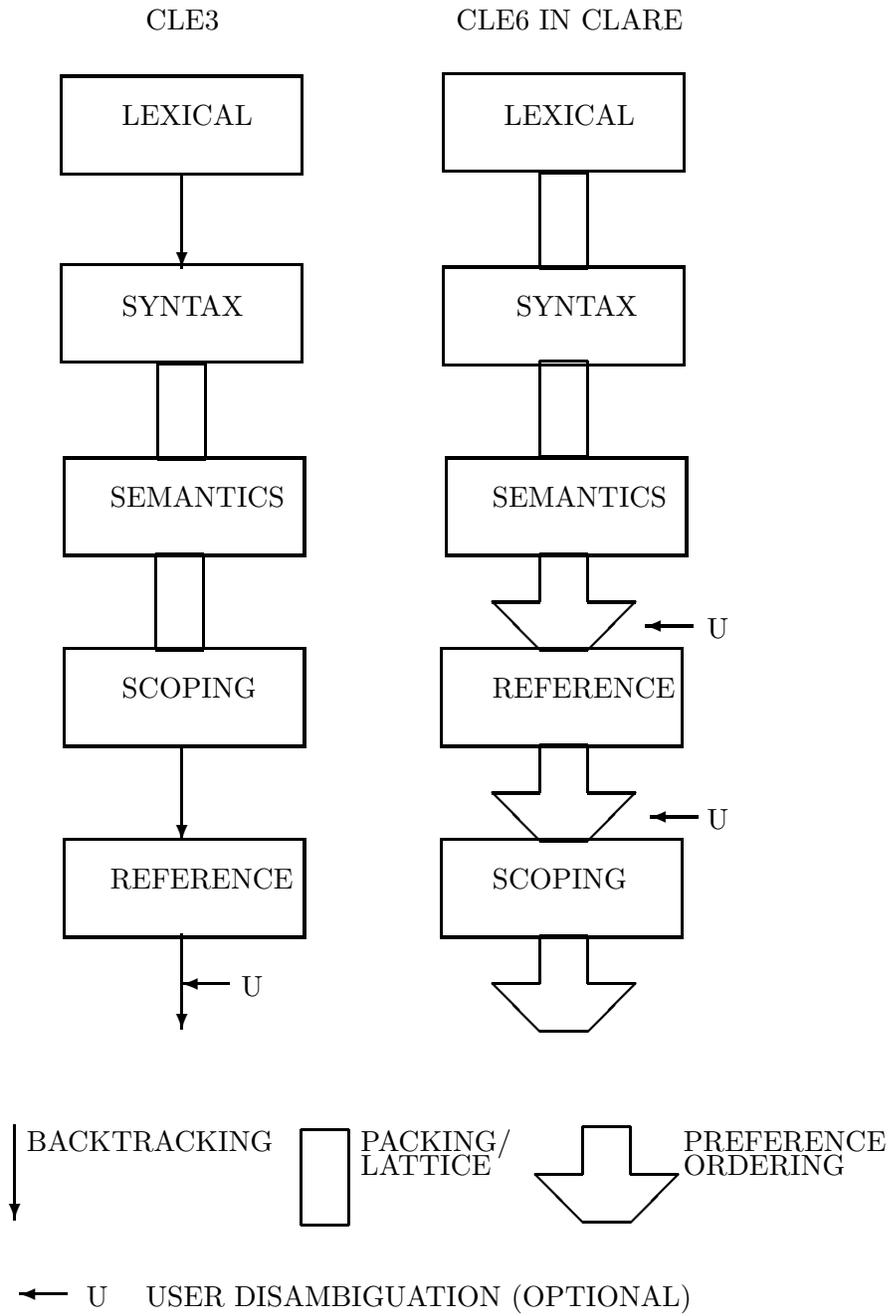
\begin{figure}
\setlength{\unitlength}{0.0125in}%
\begin{center}
\begin{picture}(290,547)(60,285)
\thicklines
\put( 82,290){\vector(-1, 0){ 20}}
\put(344,570){\vector(-1, 0){ 20}}
\put(346,490){\vector(-1, 0){ 20}}
\put(150,420){\vector(-1, 0){ 20}}
\put( 60,361){\vector( 0,-1){ 40}}
\put(298,360){\line( 0,-1){ 20}}
\put(298,340){\line(-1, 0){ 20}}
\put(278,340){\line( 1,-1){ 20}}
\put(298,320){\line( 1, 0){ 20}}
\put(318,320){\line( 1, 1){ 20}}
\put(338,340){\line(-1, 0){ 20}}
\put(318,340){\line( 0, 1){ 20}}
\put(318,360){\line(-1, 0){ 20}}
\put(280,760){\line( 0,-1){ 40}}
\put(280,720){\line( 1, 0){ 20}}
\put(300,720){\line( 0, 1){ 40}}
\put(300,760){\line(-1, 0){ 20}}
\put(280,681){\line( 0,-1){ 40}}
\put(280,641){\line( 1, 0){ 20}}
\put(300,641){\line( 0, 1){ 40}}
\put(300,681){\line(-1, 0){ 20}}
\put(120,680){\line( 0,-1){ 40}}
\put(120,640){\line( 1, 0){ 20}}
\put(140,640){\line( 0, 1){ 40}}
\put(140,680){\line(-1, 0){ 20}}
\put(119,600){\line( 0,-1){ 40}}
\put(119,560){\line( 1, 0){ 20}}
\put(139,560){\line( 0, 1){ 40}}
\put(139,600){\line(-1, 0){ 20}}
\put(180,359){\line( 0,-1){ 40}}
\put(180,319){\line( 1, 0){ 20}}
\put(200,319){\line( 0, 1){ 40}}
\put(200,359){\line(-1, 0){ 20}}
\put(130,441){\vector( 0,-1){ 40}}
\put(280,440){\line( 0,-1){ 20}}
\put(280,420){\line(-1, 0){ 20}}
\put(260,420){\line( 1,-1){ 20}}
\put(280,400){\line( 1, 0){ 20}}
\put(300,400){\line( 1, 1){ 20}}
\put(320,420){\line(-1, 0){ 20}}
\put(300,420){\line( 0, 1){ 20}}
\put(300,440){\line(-1, 0){ 20}}
\put(280,600){\line( 0,-1){ 20}}
\put(280,580){\line(-1, 0){ 20}}
\put(260,580){\line( 1,-1){ 20}}
\put(280,560){\line( 1, 0){ 20}}
\put(300,560){\line( 1, 1){ 20}}
\put(320,580){\line(-1, 0){ 20}}
\put(300,580){\line( 0, 1){ 20}}
\put(300,600){\line(-1, 0){ 20}}
\put(280,520){\line( 0,-1){ 20}}
\put(280,500){\line(-1, 0){ 20}}
\put(260,500){\line( 1,-1){ 20}}
\put(280,480){\line( 1, 0){ 20}}
\put(300,480){\line( 1, 1){ 20}}
\put(320,500){\line(-1, 0){ 20}}
\put(300,500){\line( 0, 1){ 20}}
\put(300,520){\line(-1, 0){ 20}}
\put(239,440){\framebox(100,40){}}
\put(239,520){\framebox(100,40){}}
\put(241,600){\framebox(100,40){}}
\put(240,680){\framebox(100,40){}}
\put(240,761){\framebox(100,40){}}
\put(130,520){\vector( 0,-1){ 40}}
\put(130,760){\vector( 0,-1){ 40}}
\put( 81,440){\framebox(100,40){}}
\put( 81,520){\framebox(100,40){}}
\put( 81,600){\framebox(100,40){}}
\put( 80,681){\framebox(100,40){}}
\put( 80,760){\framebox(100,40){}}
\put(115,285){\makebox(0,0)[lb]{\raisebox{0pt}[0pt][0pt]{\small USER DISAMBIGUATION (OPTIONAL)}}}
\put( 90,285){\makebox(0,0)[lb]{\raisebox{0pt}[0pt][0pt]{\small U}}}
\put(350,565){\makebox(0,0)[lb]{\raisebox{0pt}[0pt][0pt]{\small U}}}
\put(350,485){\makebox(0,0)[lb]{\raisebox{0pt}[0pt][0pt]{\small U}}}
\put(155,415){\makebox(0,0)[lb]{\raisebox{0pt}[0pt][0pt]{\small U}}}
\put(245,820){\makebox(0,0)[lb]{\raisebox{0pt}[0pt][0pt]{\small CLE6 IN CLARE}}}
\put(115,820){\makebox(0,0)[lb]{\raisebox{0pt}[0pt][0pt]{\small CLE3}}}
\put(340,340){\makebox(0,0)[lb]{\raisebox{0pt}[0pt][0pt]{\small ORDERING}}}
\put(340,350){\makebox(0,0)[lb]{\raisebox{0pt}[0pt][0pt]{\small PREFERENCE}}}
\put(205,340){\makebox(0,0)[lb]{\raisebox{0pt}[0pt][0pt]{\small LATTICE}}}
\put(205,350){\makebox(0,0)[lb]{\raisebox{0pt}[0pt][0pt]{\small PACKING/}}}
\put( 65,350){\makebox(0,0)[lb]{\raisebox{0pt}[0pt][0pt]{\small BACKTRACKING}}}
\put(105,780){\makebox(0,0)[lb]{\raisebox{0pt}[0pt][0pt]{\small LEXICAL}}}
\put(105,540){\makebox(0,0)[lb]{\raisebox{0pt}[0pt][0pt]{\small SCOPING}}}
\put(265,780){\makebox(0,0)[lb]{\raisebox{0pt}[0pt][0pt]{\small LEXICAL}}}
\put(270,700){\makebox(0,0)[lb]{\raisebox{0pt}[0pt][0pt]{\small SYNTAX}}}
\put(260,620){\makebox(0,0)[lb]{\raisebox{0pt}[0pt][0pt]{\small SEMANTICS}}}
\put(260,460){\makebox(0,0)[lb]{\raisebox{0pt}[0pt][0pt]{\small SCOPING}}}
\put(260,540){\makebox(0,0)[lb]{\raisebox{0pt}[0pt][0pt]{\small REFERENCE}}}
\put(100,620){\makebox(0,0)[lb]{\raisebox{0pt}[0pt][0pt]{\small SEMANTICS}}}
\put(100,460){\makebox(0,0)[lb]{\raisebox{0pt}[0pt][0pt]{\small REFERENCE}}}
\put(105,700){\makebox(0,0)[lb]{\raisebox{0pt}[0pt][0pt]{\small SYNTAX}}}
\end{picture}
\end{center}
\caption{Handling of alternatives by main CLE phases}
\label{prefarch}
\end{figure}

Note that reference resolution takes place before scoping in the
revised architecture of the CLE subsystem employed in CLARE.
This change was made possible by the implementation of a more
monotonic model for semantic interpretation as discussed in
detail in Chapter~\ref{clareinterp}. At each of the points listed,
the preference mechanism attempts to ensure that alternatives
are passed to the next phase of processing starting with the
most likely alternative, the next most likely alternative, and so on.

As in earlier versions of the CLE, all syntactic
analyses for the sentence are represented implicitly in the packed
syntax records, but unlike earlier versions, explicit QLFs are
derived directly from the syntax records rather than being
first coded as packed semantic records. This change is
related to the following considerations: (i) direct computation of
QLFs allows a restriction on the form of semantic rules
(daughter QLFs being restricted to variables) to be lifted, 
making the rule set more amenable to efficient generation,
(ii) since preferences on QLFs often depend on
non-local structural properties, it is difficult to compute
preference measures for a set of QLFs represented as 
records with local ambiguity packing, and (iii) prepositions
are now given single vague-relation meanings in unresolved QLF 
(the relations are refined during reference resolution),
so much of the
ambiguity which motivated packed semantic records has been
removed from this level of analysis.

The overall changes in the handling of alternatives
in CLARE compared with the CLE in NATTIE are
summarized in the Figure~\ref{prefarch}.

The preference mechanism improves user interaction since 
alternatives can be presented for confirmation starting with
the most likely alternative analysis or interpretation according 
to the preference measures. It should therefore reduce the
amount of interaction that is necessary, or even allow the user
to run the system in fully automatic mode.

Figure~\ref{prefarch} also shows
that user interaction is no longer at the end of the whole
interpretation process, when a resolved LF was presented to the
user, but instead takes place after semantic analysis and
reference resolution. These two points were chosen for
interactive disambiguation because they involve choices
that can often be displayed in an informative way to non-expert
users (QLFs and LFs are no longer displayed in ``user mode'').
After semantic analysis, word sense choices and meaningful
phrasal analysis (i.e. identification of phrases corresponding
to objects or basic assertions) are presented. After
resolution, new entities and relations identified in a possible
paraphrase using the referents (examples are presented in 
Section~\ref{genfromrqlf}).

\section{User Interaction for QLF Selection}\label{genfromrqlf}

After applying its own preference criteria, the system by default
presents the user with information about the various QLFs in order
of preference. The first QLF the user accepts is the one used for
further processing.

Rather than print the QLF itself, which would be useful only to
specialists, the system gives a bracketing of the sentence, showing
the noun phrases it is analysed as containing according to that QLF.
It also prints paraphrases of word senses, contrasting them with
alternatives that may be defined in the lexicon.

Thus inputting the sentence {\it John met the man in the bank} causes
the following to be printed:
\begin{quote}
\begin{verbatim}
   Complete sentence with bracketing:

     "{John} met {{the man} in {the bank}}."

   Word senses (unordered):

      meet: encounter (rather than "be adjacent")
      man: male person
      bank: company (rather than "building" or "edge")

Confirm this analysis? (y/n/c/p/?): 
\end{verbatim}
\end{quote}
The user is being told that according to this analysis,
\begin{itemize}
\item the input is being interpreted as a complete sentence rather
than an elliptical one; an ambiguity of this type would occur for the
input {\it the men met in the bank}, which can either be a sentence or a
noun phrase meaning {\it the men who were met in the bank};
\item {\it the man in the bank} is a constituent; 
i.e.\ {\it in the bank} attaches to {\it man} rather than {\it met}; 
\item {\it meet} is being interpreted as ``encounter'' rather than 
``be adjacent'' -- a sense that would be appropriate for {\it the two
walls meet at the corner of the property};
\item {\it man} is being interpreted as ``male person'' -- there is no
alternative to this;
\item {\it bank} is being interpreted as a kind of company rather than
as a building or as the edge of something such as a river.
\end{itemize}

The user may accept the analysis by typing \verb!y! or reject it with
\verb!n!. He or she may also have the QLF itself printed by typing
\verb!p! and then make a decision. However, it is also possible to 
reject it and specify the reasons why it is unsatisfactory by typing
\verb!c! (for ``constraints''). This facility can greatly reduce the
amount of interaction required when there are a large number of QLFs
which correspond to the various combinations of decisions on several
different ambiguities.

Let us suppose that the user is unhappy with the QLF because
of the sense of {\it bank} chosen, but wants to confirm the bracketing.
The response to \verb!c! is a menu:

\begin{quote}
\begin{verbatim}
Possible constraints:
    1 This bracketing 
    2 Complete sentence (rather than phrase) 
    3 "meet" meaning "encounter"  
    4 "bank" meaning "company"  
Enter some signed numbers (+n to insist, -n to disallow) 
separated by spaces: 
\end{verbatim}
\end{quote}
If the user gives the response ``\verb!+1 -4!'', then QLFs whose
bracketings are not as given above will be rejected without the user
seeing them, as will QLFs where ``bank'' means ``company''. The next
QLF that matches the user's constraints is therefore described as
follows:
\begin{quote}
\begin{verbatim}
   Complete sentence with bracketing:

     "{John} met {{the man} in {the bank}}."

   Word senses (unordered):

      meet: encounter (rather than "be adjacent")
      man: male person
      bank: edge (rather than "building" or "company")

Confirm this analysis? (y/n/c/p/?): 
\end{verbatim}
\end{quote}
This time, the user might simply time \verb!n!, to reject the QLF. If
so, a further QLF, with {\it bank} this time meaning ``building'' but
with the same bracketing, is tried.

In this way, users without an understanding of the QLF language can
make intelligent interpretation choices. They can also avoid a lot of
unnecessary interaction; in the above example, only three QLFs were
presented out of a total of six found by the system. The other three
are automatically rejected because of the bracketing constraint.

\section{Preference Measures}

At each of the points at which the preference mechanism is applied,
a numerical `preference measure' is computed for each of
the alternative representations being considered at that point,
and the alternatives are sorted into an order respecting
the preference measure. There are system parameters (\verb!maxqlfs!, etc.)
for controlling the maximum number of alternatives considered
at each stage, and a minimum preference measure threshold below which 
alternatives are discarded immediately. Different preference measures are 
used for the different points at which the mechanism is applied: 
the measures are meant to reflect the contribution of new choices 
made during interpretation that have not yet been evaluated by 
the mechanism.

Most (unscaled) scoring functions are simply
integers corresponding to counts of particular constructs
in the representation, for example the number of unresolved
ellipsis \verb!form!s in QLF. Others, such as the syntax rule score
and the semantic collocation metrics are real numbers derived from
automatic training processes.
Some of the preferences now implemented as scoring functions
with relatively high negative scaling factors
were previously regarded as inviolable constraints applied,
in particular, to scopings and resolutions.
We now turn to the scoring functions used, respectively, to compute
preference measures for QLFs, resolved QLFs, and scoped resolved 
QLFs respectively.

A preference measure is itself a weighted sum of a set of 
scoring functions declared by assertions of the form
\begin{quote}\begin{verbatim}
scoring_function(Identifier,Fi,Ci).
\end{verbatim}\end{quote}
Here \verb!Id! is an identifier for the preference measure
(e.g. \verb!qlf!, \verb!scope!), \verb!Ci! is a scaling factor 
(positive or negative) and \verb!Fi! is the name of the scoring
function. Examples are:
\begin{quote}\begin{verbatim}
scoring_function(qlf,qlf_left_nesting,-3)
scoring_function(qlf,qlf_sense_weights,10).
\end{verbatim}\end{quote}
The scoring function takes a complete representation
(QLF or resolved QLF) as input, returning a numerical 
value \verb!Si!, the preference measure being computed by summing 
over the product of scaled scores for all scoring 
functions:
\begin{quote}\begin{verbatim}
C1S1+...+CmSm.
\end{verbatim}\end{quote}

\section{Scaling Factors}

When the preference mechanism was first used, an initial set
of scaling factors was chosen by hand according to knowledge of 
the how the particular raw preference metrics were computed and 
introspection about the `strength' of the metrics as indicators
of preference or dispreference. These initial scaling factors
were subsequently revised according to their observed behaviour
in ranking analyses in the CLARE PRM application. This tuning
process was time consuming but it did eventually lead
to reasonably well behaved rankings for this domain and 
application. It is difficult to
quantify these remarks because, for this application, there was
no independently collected corpus of examples to use for evaluation.

However, there are a number of disadvantages to manual tuning
of scaling factors. These include the effort spent in maintaining
the parameters, this effort being greater for those with less
knowledge of how the raw preference functions are computed since
this increases the effort for trial-and-error tuning. A point of
diminishing returns is also reached after which further attempts
at improvement through hand-tuning often turn out to be
counter-productive. Another problem was that it became difficult
to detect preference metrics that were ineffective, or simply
wrong, if they were given sufficiently low scaling factors.
Probably a more serious problem is that the contributions of different
preference factors to selecting the most plausible analyses seem
to vary from one sublanguage to another.

These disadvantages point to the need for automatic
procedures to determine scaling factors that optimize preference
metric rankings for a particular sublanguage. This requires
a corpus of sentences that are representative of the sublanguage
and some independent measure of the correctness or plausibility of analyses
of these sentences. (With a sufficiently large corpus, it may
be possible to start with some initial plausibility measure and
refine it through re-estimation, although we have not attempted
this.) One way of gathering the correctness information is to
retain the choices made during interactive disambiguation by 
operators of information system interfaces or machine aided translation
systems. This could result in systems whose automatic preference
choices would improve gradually over time.

Work on optimizing the scaling factors for the CLARE preference
mechanism has been carried out under a project for building a speech
translation system based on CLARE's linguistic components 
(Section~\ref{slt}).
These experiments made use of
the Air Travel Information System (ATIS) corpus of transcribed
speech sentences. We chose this application for these experiments because
there existed a hand-parsed sub-collection of this corpus built as
part of the University of Pennsylvania ``Treebank'' project. 
Further skeletal trees induced by QLF analyses, and verified manually,
were collected at Cambridge. We then defined a penalty function on QLFs
which reflected the extent to which a QLF analysis for a sentence differed
from the treebank analysis. It was then possible to determine an
optimal set of scaling factors, with respect to this penalty function,
using the least squares minimization: this turns 
into the straightforward analytic procedure to find the unique solution of
a set of linear simultaneous equations. The experiments are described
in Alshawi and Carter 1992. In summary, the results from blind testing
indicated that despite the crude penalty function used, the method yielded 
a set of scaling factors that is at least as good as the set derived by 
labour intensive hand tuning.

\section{QLF Preferences}
\label{prf-qlf}

The scoring functions contributing to the preference
measure for QLFs can be divided into two categories, those expressing 
structural preferences and those expressing semantic preferences.
We will present examples relevant to some of the fifteen or so
QLF scoring functions currently in use.

\subsection{Structural Preferences}

QLF structural preferences are often the same as, or very similar to,
those proposed in previous work for selecting between possible
syntactic analyses (Hobbs and Bear 1990 discuss principles underlying
parsing preferences). Although the QLF representation is
semantic in nature, it is possible to use such syntactic preference 
criteria because of the presence of linguistic categories in QLF
and because much of the phrasal structure of sentences is implicit
in QLFs. Structural QLF preferences include the following:

\begin{itemize}

\item Prefer subcategorization to adjuncts. For example,
the reading of {\it John laughed at the school} in which
{\it laugh at} is treated as a verb-particle construction
is preferred to the reading in which {\it at the school} 
is an optional modifier.

\item Prefer low attachment. For example, this prefers
the reading of {\it John saw a cat in a tree in a garden}
in which the cat is in a tree that is in a garden.

\item Disprefer demonstrative readings of {\it that}
and {\it there}. This applies to examples like
{\it I know that sugar is expensive} and
{\it  What cars are there?}

\item Prefer QLFs with structurally balanced conjunctions,
for example the more likely reading of {\it They sell
motorcycles in France and caravans in Germany}. (There is
also a sort parallelism preference for conjunctions.)

\end{itemize}

\subsection{Semantic Preferences}

QLF semantic preferences are used to downgrade vague analyses,
such as those involving \verb!form!s, or to prefer structures
which lexical semantics makes more plausible. Semantic QLF
preferences include the following:

\begin{itemize}

\item Prefer QLFs which include more frequently observed
semantic collocations. This type of metric is considered in
more detail in the section below on semantic head collocations.

\item Disprefer QLFs with \verb!form!s corresponding
to NBAR ellipsis. For example, this disprefers analysing
{\it Mary sold some tickets} as elliptical for, say, {\it Mary
sold tickets to some customers}.

\item Disprefer QLFs with \verb!form!s arising from verb phrase
ellipsis. For example, this disprefers analysing {\it Mary
is in Cambridge} as elliptical for, say, {\it Mary is working in
Cambridge}.

\item Disprefer readings with properties as arguments. The progressive
reading of {\it It is eating} is thus preferred over the equative
reading.

\item Prefer QLFs which include word senses with higher probabilities
indicated by \verb!weight! declarations in the lexicon. These preferences
are often application specific. For example, in the PRM supplementary
lexicon, the
`money' sense of {\it pound} is strongly preferred to the `weight measure'
sense.

\end{itemize}

\subsection{Semantic Head Collocations}

Semantic collocation metrics, especially those determined statistically have
recently attracted a lot of attention in computational linguistics
(Calzolari and Bindi 1990, Church and Hanks 1990, Hindle and Rooth 1990,
Sekine et al. 1992).
They are typically derived by observing the occurrences of tuples 
(usually pairs or triples) that summarize relationships present 
in a semantic analysis of a text, or relate lexical heads of phrases.

In our case, we use semantic collocations extracted from QLF
expressions in the form of \verb!(H1,R,H2)! triples where \verb!H1! 
and \verb!H2! are the head predicates of phrases in a sentence 
and \verb!R! indicates the semantic relationship (e.g. a preposition
or an argument position corresponding to a case relation) between
the two phrases in the proposed analysis.
Semantic collocation metrics are especially relevant to choosing between
analyses with different prepositional attachments and different
word senses; the can be viewed as a statistical analogue of selectional
restrictions based on concept hierarchies.

Data collection for the semantic collocation metrics proceeds by
deriving a set of triples from each QLF analysis of the sentences
in the training set. This is followed by statistical analysis to
produce the following functions of each triple in the observed
triple population:
\begin{itemize}

\item mutual information: this relates the probability
of the triple assuming independence between its three fields
(\verb!P1*P2*P3!) with the probability \verb!A! estimated 
from actual observations of triples derived from analyses 
ranked highest (or joint highest) by the tree bank goodness metric.
More specifically we use \verb!ln(A)-ln(P1)-ln(P2)-ln(P2)!.

\item chi-squared: compares expected frequency \verb!E! of a
triple with the square of the difference between the observed 
frequency \verb!F! of the triple and the expected frequency.
Here the observed frequency is in analyses ranked highest 
(or joint highest) by the tree bank goodness metric and the 
``expected'' frequency assumes independence between triple fields. 
More specifically we use \verb!|F-E|*(F-E)/E!
(This variant of chi-squared in which the numerator is
signed is used so that the function is monotonic making it
more suitable in preference metrics.) 

\item mean distance: the average of the tree bank goodness
metric for all QLF analyses (not necessarily highest ranked
ones) which includes the semantic collocation corresponding
to the triple.

\end{itemize}

Computation of the mutual information and chi-squared functions
for triples involves the simple smoothing technique of adding
0.5 to actual counts. 
From these three functions on triples we define three semantic
collocation preference metrics applied to QLF, in each case by 
summing over the result of applying the function to each triple 
derived from a QLF. In the experiments described in Alshawi and
Carter 1992 we compared the effectiveness of these semantic
collocation metrics. Chi-squared performed best, with mean distance
a close second and mutual information performing significantly less
well than the other two, despite our impression that it is the
most commonly used collocation metric.

\section{Resolved QLF Preferences}

Resolved QLF preferences can be split into two broad types: those that depend
on the precise resolution method used, and those that depend solely on the
form of the finally resolved QLF.  This division arises because more than one
resolution may be applicable to an unresolved expression, but one cannot
tell which method was used just by looking at the form of the resolved expression.
For example, a domain specific and a non-specific resolution method may 
give rise to exactly the same resolution, but greater weight should be placed
on the domain specific case.

To impose a preference ranking over resolution methods, {\tt rule\_score}
declarations are given.  These take the form
\begin{quote}\begin{verbatim}
rule_score(Method,Category,Weight).
\end{verbatim}\end{quote}
\verb!Method! names a resolution method, \verb!Category! specifies a category
of unresolved expression to which the method can be applied, and
\verb!Weight! assigns a preference weighting to the use of the method on that
category. (Currently, the category is left unspecified in all declarations).
During resolution, a score is accumulated as resolution methods are applied.
The overall score is then weighted by the
\begin{quote}\begin{verbatim}
scoring_function(ref(_),ref_rules_score,Weight).
\end{verbatim}\end{quote}
declaration.

Other preference metrics operate in the more conventional way by 
considering just the resolved QLF.  These preferences include
\begin{itemize}
\item
Checking for violations of reflexive and non-reflexive pronoun constraints.
Violations are heavily penalised.

\item
Penalising duplicate resolved QLFs.  Duplicates may sometimes arise through
distinct resolution methods leading to identical results.

\item
Penalising sortal violations.

\item
Dispreferring non-parallel substitutions in ellipsis, e.g. where a non-quantified
term is substituted for a quantified one or vice versa.

\item
Scoring resolutions of determiners to different collective and distributive
quantifiers.   {\tt det\_to\_quant} rules are given scores which contribute
to the ranking, and certain QLF constructions can be identified as demanding
collective (or distributive) quantifiers.
\end{itemize}

The number of resolutions made in the first place, and the number returned after
preference ranking are controlled by setting a resolution limit and a ranked
resolution limit respectively. Also, resolution is carried out in two
stages, which lead to different metrics being used. The first bout of resolutions
deals with `major' resolutions, e.g. resolution of pronouns, vague relations,
ellipsis, and so forth.  These resolutions are fairly easy to paraphrase back to
the user for confirmation.  The second stage involves resolving determiners
to quantifiers and deciding on distributive or collective readings, which are
not so easily paraphrasable.  Separate limits are defined for the two stages.

\section{Scoping Preferences}

Preferences defined for scoped QLFs include
\begin{itemize}
\item
Penalising redundant scope permutations, e.g. ignoring one of $\exists X \exists
Y$ and $\exists Y \exists X$
\item
Heavily dispreferring violations of domain specific functional information
(see Section~\ref{scp-func}).
\item
Scoring how far quantifiers have been `raised' through the QLF
\item
Making pairwise comparisons of quantifier -- quantifier scopes, and 
quantifier -- operator scopes.  These call on subsidiary definitions
stating four way preferences about the relative scopes of specific
quantifiers and operators.
\end{itemize}

\chapter{Lexical and Syntactic Analysis and Failure Recovery}
\label{clarelexproc}

\section{Introduction}

In many language processing systems, uncertainty in the boundaries of
linguistic units, at various levels, means that data are represented
not as a well-defined sequence of units but as some kind of lattice.
It is common for speech recognizers to maintain a lattice of
overlapping word hypotheses from which one or more plausible complete
paths are subsequently selected. Syntactic parsing, in the CLARE and
many other systems, makes use of a chart or well-formed substring
table because the correct bracketing of a sentence cannot (easily) be
calculated deterministically.  And lattices are also often used in the
task of converting Japanese text typed in kana (syllabic symbols) to
kanji; the lack of interword spacing in written Japanese and the
complex morphology of the language mean that lexical items and their
boundaries cannot be reliably identified without applying syntactic
and semantic knowledge (Abe {\it et al}, 1986).

In contrast, however, most systems accepting typed input have
attempted to group their input deterministically into a sequence of
words and punctuation symbols to be handed to the parser, possibly
performing word-by-word morphological analysis on the way. Such an
approach is sometimes adopted even when typographically complex inputs
are handled; see, for example, Futrelle {\it et al}, 1991.

In this chapter we argue that to deal with a range of phenomena that
occur in written English, a lattice representation must also be used
in front end (pre-parsing) analysis.  We describe how the CLARE system
uses such a representation, and show how this allows the system to
deal straightforwardly with combinations of phenomena that would be
difficult or impossible to process correctly under a sequence
representation. The chapter also describes changes to the version of
the CLE parser described in section 7.1 of Alshawi {\it et al} (1992),
which was current near the beginning of the CLARE project.

For typed input, spaces do not necessarily correspond to boundaries
between lexical items, both for linguistic reasons and because of the
possibility of typographic errors.  This means that a lattice
representation, not a simple sequence, should be used throughout front
end analysis. As evidence for the performance of the approach taken, we
describe an evaluation of CLARE's ability to deal with typing and
spelling errors.  Such errors are especially common in interactive
use, for which CLARE is designed, and the correction of as many of
them as possible can make an appreciable difference to the usability
of a system.

The word identity and word boundary ambiguities encountered in the
interpretation of errorful input often require the application of
syntactic and semantic knowledge on a phrasal or even sentential
scale. Such knowledge may be applied as soon as the problem is
encountered; however, this brings major problems with it, such as the
need for adequate lookahead, and the difficulties of engineering large
systems where the processing levels are tightly coupled. To avoid
these difficulties, CLARE adopts a staged architecture, in which
indeterminacy is preserved until the knowledge needed to resolve it is
ready to be applied. An appropriate representation is of course the
key to doing this efficiently.

When an input is outside the system's grammatical coverage or is
judged sortally incoherent, it is much more difficult to make a
correction than when a spelling error occurs. This is because the
space of possibilities is much larger and less well-defined; even if
one could enumerate the possible corrections, choosing between them
would be very difficult, and there would often be so many of them that
no motivated choice could be made in a reasonable time. CLARE
therefore reacts to grammatical and sortal problems not by attempting
a recovery but by offering the user an analysis of them which is
intended to provide guidance for rephrasing the input or, during
development, modifying linguistic data.

\section{Lattices for Lexical Processing}
\label{latforlex}

In general, typing errors are not just a matter of one intended input
token being miskeyed as another one. Spaces between tokens may be
deleted (so that two or more intended words appear as one) or inserted
(so that one word appears as two or more). Multiple errors, involving
both spaces and other characters, may be combined in the same intended
or actual token. A reliable spelling corrector must allow for all
these possibilities, which must, in addition, be distinguished from
the use of correctly-typed words that happen to fall outside the
system's lexicon.

However, even in the absence of ``noise'' of this kind, spaces do not
always correspond to lexical item boundaries, at least if lexical
items are defined in a way that is most convenient for grammatical
purposes. For example, ``special'' forms such as telephone numbers or
e-mail addresses, which are common in many domains, may contain
spaces. In CLARE, these are analysed using regular expressions (cf
Grosz {\it et al}, 1987), which may include space characters. When
such an expression is realised, an analysis of it, connecting
non-adjacent vertices if it contains spaces, is added to the lattice.

The complexities of punctuation are another source of uncertainty:
many punctuation symbols have several uses, not all of which
necessarily lead to the same way of segmenting the input. For example,
periods may indicate either the end of a sentence or an abbreviation,
and slashes may be simple word-internal characters (e.g.\ {\it
X11/NeWS}) or function lexically as disjunctions, as in
\begin{quote}\
{\it I'm looking for suggestions for vendors to deal with/avoid.}
\footnote{These two examples are taken from the Sun-spots
electronic bulletin board on Sun systems administration.}
\end{quote}
Here, the character string ``{\it with/avoid}'', although it contains
no spaces, represents three lexical items that do not even form a
syntactic constituent.

And, particularly in running text, a number of different punctuation
symbols must be dealt with.  While some of these phenomena can be
handled fairly effectively on their own without resorting to a lattice
representation, they tend to interact in ways that make apparent the
inadequacy of modelling the input as a deterministic sequence of
tokens.

An example that shows the need for a lattice representation of
possible words is as follows:
\begin{quote}{\tt
Does anyone out there in net.land know of/sell/use a Centronics
parallel (redundent I know) output board for VME bus?}
\end{quote}
To prepare this sentence for parsing, at least four questions would
need to be answered:
\begin{itemize}
\item If \verb!net.land! is not in the lexicon, should it
be interpreted as \verb!netland! or \verb!net land!? (Note that the
dot cannot be an end-of-sentence marker here).
\item Is \verb!of/sell/use! to be interpreted somehow as a single lexical
item, or do the slashes have a wider syntactic function? If the
latter, how?  (Other uses of the slash character common in the
bulletin board are \verb!X11/NeWS! and \verb!Sun 4/330GX!; in both
cases, the slash has no special status inside the lexical item.)
\item Do the bracket characters function as parentheses? There are
other uses, as in another sentence from the same bulletin board:
\begin{quote}{\tt
PS: Dear wnl, outlaw those insipid ":-)" smiley faces now!
}\end{quote}
\item What is the correct version of the misspelled word 
\verb!redundent!? Misspellings sometimes involve spaces being
deleted, as in tokens like \verb!soundsthe!, or \verb!witha!. In the
latter case, whether the intended input is ``with a'' or ``with''
cannot be determined without knowledge of syntax, which is unlikely to
be easily applicable before (candidate) words have been identified.
\end{itemize}

The specific task considered in this chapter is the process of mapping
single sentences from character strings to quasi logical form
representations. Two kinds of issue are therefore not discussed here.
These are the problem of segmenting a text into sentences and dealing
with any markup instructions, which is logically prior to producing
character strings; and possible context-dependence of the lexical
phenomena discussed, which would need to be dealt with after the
creation of QLFs.

Many of the aspects of language under discussion here are especially
liable to vary between applications. Since CLARE is intended to be
tailorable to different domains and tasks without the intervention of
expert linguists, it is important that non-linguist application
experts be able to write the appropriate rules and lexical entries.
This constraint, too, is reflected in the solutions to be discussed.

\section{CLARE's Lexical Processing Stages}
\label{stages}

In the analysis direction, CLARE's lexical processing stages are as
follows.

\begin{enumerate}
\item[] {\bf Clustering}. A sentence is deterministically divided into a
sequence of {\it clusters} separated by white space. An example of a
cluster is ``\verb!books,!''  (including the comma).
\item[] {\bf Tokenization}. Each cluster is divided into one or more {\it
tokens}; for example,
``\verb!books!'' and ``\verb!,!''. A token is anything that the parser
treats as a terminal symbol; the term therefore includes words,
special forms and punctuation symbols. Because tokenization is
non-deterministic, a lattice representation is adopted from this
stage onwards.
\item[] {\bf Segmentation}. Each token, or sometimes a sequence of several
tokens, is analysed nondeterministically as a sequence of one or more
{\it segments}. For words, these segments are morphemes; for example,
``\verb!book!'' and ``\verb!es!''.
\item[] {\bf Recovery}. The failure to analyse a token at stage 3 usually
means that another division of the cluster it originated in (in stage
2) was the appropriate one. Such tokens are simply pruned away.
However, if there is {\it no} path through the lattice that avoids
unanalysable tokens, then some of those tokens are handed to various
{\it failure recovery} modules. These may create new tokens or new
definitions for the existing tokens.
\item[] {\bf Idiom and Abbreviation Recognition}. Some words or phrases may
have been defined as idioms or abbreviations that are equivalent to
other phrases. If so, segmented versions of the latter are
added to the lattice.
\item[] {\bf Word Parsing}. The segments making up each token that is a word 
undergo word parsing with respect to the system's morphological
rules to yield one or more syntactic and semantic analyses. Thus
``\verb![book,es]!'' is analysed as a plural noun; if ``book'' were to
be defined as a verb, a third person singular present verb analysis
would also result. Non-word tokens undergo analogous processes. Word
parsing is not further described here as it is unchanged from earlier
versions of the system.
\item[] {\bf Phrase Parsing}. Before sentence parsing proper (for which see
the final Nattie report), a set of grammar rules known as {\it phrase
rules} are applied to the word analyses.
\end{enumerate}

In synthesis, analogous process are applied. The generator (see
Chapter \ref{generation}) creates a parse tree, and the reverse of the
segmentation, tokenization and clustering phases are then applied. The
same rules are used as for analysis. In the following sections, the
stages of lexical processing are described in the analysis order.

\section{Clustering}

The clustering phase is the simplest of the lexical processing phases.
In analysis, the input characters are separated at white spaces into
atoms known as {\it clusters} that contain no white space. Multiple
contiguous white space characters are treated just like a single one.

Recognition of the end of a sentence in general may require the
application of syntactic and semantic knowledge, because there are
situations in which a cluster-final character such as a period or an
exclamation mark does not mark the end of a sentence. However, in the
interests of practicability, the rules in CLARE, which should be
well suited to interactive input, are as follows.
\begin{itemize}
\item A sentence can only end at the end of a line of input. That is, a
new sentence cannot start on the same line.
\item If the last cluster on a line ends in a sentence-terminating final
punctuation symbol it is taken as an end of sentence. That is, a
sentence can be input over multiple lines as long as the last cluster
on a line is not one that can finish a sentence.
\end{itemize}

The cluster \verb!##! acts as a signal to abandon the
sentence currently being input. It is useful if the user notices an
error only after continuing a sentence on a new line.

Declustering, the trivial counterpart of clustering in synthesis,
consists merely of outputting the clusters generated by detokenization
(see below) with space characters between them.

\section{Tokenization}
\label{lex-tok}

In earlier versions of the system, tokenization was hard-wired and
deterministic. For example, a lexical item that happened to end in a
colon could not be recognized because the colon would be stripped off
before lexical access took place. In CLARE, however, tokenization is
performed by the same code as word segmentation, using rules of the
same format. All rules are treated as optional; thus the cluster
``\verb!books,!'' will yield one tokenization consisting of
``\verb!books!''  and a comma, and one consisting just of
``\verb!books,!''  itself. (The latter will normally, of course, fail
segmentation). The symbol ``\verb!,!'', like, say, ``\verb!ed!'', is
defined as a suffix, but one that is treated by the grammar as a
separate token rather than a bound morpheme. 

Rules for punctuation characters tend to be very simple because no
spelling changes are involved.  However, the possessive ending
``\verb!'s!'' is treated as a separate lexical item, not a bound
morpheme, in the CLARE grammar to allow the correct analysis of
phrases such as ``the man in the corner's wife'', and spelling changes
can be involved here: the possessive of ``cheques'', for example, is
\verb!cheques'! and not \verb!cheques's!.

As well as prefix and suffix symbols, a new class, that of infixes,
has been defined. This is to allow symbols such as hyphens and
slashes, which typically appear within clusters, to be treated
syntactically. In the example given in section \ref{latforlex}
the cluster sequence ``\verb!know of/sell/use!'' would best be
analysed as something like ``know of or sell or use'', with the
slashes acting independent lexical items.

The CLARE system is, by default, sensitive to the casing of its
input; thus if ``Bill'' (the name) and ``bill'' (a noun or verb) are
both defined, an occurrence of \verb!bill! will access only the
latter.  Casing errors are dealt with as a simple case of spelling
correction (see below). However, if the token (or tokens) at the
beginning of the lattice is capitalized, an uncapitalized version of
it is added to ensure correct behaviour for sentences like ``Bill
six hours to the WHIZ project''.

In the detokenization phase of synthesis, a token sequence is turned
into a cluster sequence. At each token boundary, the system must
decide whether to join the tokens together or to start a new cluster.
If token \verb!T1! is followed by token \verb!T2!, a new cluster must
be started at \verb!T2! if \verb!T1! can only be a root or a suffix
and \verb!T2! can only be a root or a prefix.\footnote{The terms
``root'', ``prefix'' and ``suffix'' apply at the cluster level here;
thus an English word would be a ``root'' even if inflected, a comma
would be a typical ``suffix'', and an opening bracket a typical
``prefix''.} A new cluster is also started if spelling changes prevent
the creation of a cluster including both \verb!T1! and \verb!T2!
(though such spelling changes are very rare).

\section{Segmentation}

A fairly simple affix-stripping approach to
word segmentation is adopted in CLARE because inflectional spelling
changes in English tend not to be complex enough to warrant more
powerful, and potentially less efficient, treatments such as two-level
morphology (Koskenniemi, 1983). Derivational relationships, such as
that between {\it destruction} and {\it destroy}, usually involve
semantic peculiarities as well, necessitating the definition of words
like {\it destruction} in the lexicon in their own right.

Segmentation rules are applied recursively to cover cases like
``workers'' (analysed as \verb!work+er+es!) without a special rule for
the combination \verb!er+es!. This is useful not only in token
segmentation but also in cluster tokenization; the cluster
\verb!of/sell/use! referred to above requires two applications of an
infix rule to be correctly tokenized.

\subsection{Special Forms}

As in earlier releases, special forms may be defined in CLARE; see
sections 6.2 and C.3.2 in the Nattie final report for general
principles.  In CLARE, however, special forms may include spaces.
For example, a car registration number such as \verb!XYZ 987! could be
recognized as a single special form, with both halves also perhaps
being treated as (unrelated) special forms in their own right.  It is
difficult to see how this could easily be done if a deterministic word
sequence had to be created before parsing. When a special form is
recognized, an edge is added to the lattice of possible segmentations;
a special form with a space corresponds to an edge connecting
non-adjacent vertices.

The passing of information from surface string to logical form has
also changed somewhat.  The following special form definition
illustrates the use of regular expressions and the way information in
the items they match is used to create structures in the logical form.
It handles dates in the form of three numbers separated by colons or
slashes. As before, the pattern matching is based on regular expressions.
\begin{quote}\begin{small}\begin{verbatim}
special_form('\([0-9]+\)\([/:]\)\([0-9]+\)\2\([0-9]+\)',
             a_term(time(date),
                    V,
                    [and,[day_num,V,reg(1)],
                         [and,[month_num,V,reg(3)],
                              [year_num,V,reg(4)]]]),
             parallel('Monday',monday_DayName)).
\end{verbatim}\end{small}\end{quote}
The sequences \verb!\(! and \verb!\)! in the first line of the entry
cause the characters matched to be assigned to a register. Thus for
a date of the form \verb!26/9/91!, the registers would be given the
contents \verb!26!, \verb!/!, \verb!9! and \verb!91!. The characters
\verb!\2! in the first line refer to the contents of register
two, which is set when the earlier pattern \verb![/:]! is matched
(with either a slash or a colon); this enforces the constraint that
the same separator must be used in both positions (\verb!26/9:91!, for
example, is not valid).  The first, third and fourth register values
are substituted for the \verb!reg(!$N$\verb!)!  structures in the
\verb!a_term! structure to create the logical constant
\begin{quote}\begin{small}\begin{verbatim}
a_term(time(date),
       V,
       [and,[day_num,V,26],
            [and,[month_num,V,9],
                 [year_num,V,91]]]).
\end{verbatim}\end{small}\end{quote}
This constant later forms part of any successfully constructed QLFs
for the sentence in which the date occurs.  

In synthesis, information flows in the reverse direction, from the
logical constant to the registers, which are matched with the regular
expression to compute the appropriate character sequence (in this
case, there would be two possibilities, one for each choice of
separator).

The final line of the entry,
\verb!parallel('Monday',monday_DayName))!, describes the syntactic
behaviour of the items being defined. On encountering it, the system
looks (at the word parsing stage, in fact) for a definition of the
word \verb!Monday! involving the constant \verb!monday_DayName!, and
creates syntactic and semantic entries for the special form by analogy
with it. This allows the application developer to specify the
behaviours of special forms in sentences by reference to existing
words which have earlier been defined using VEX.

Other options, allowing more flexibility, exist for the use of more
linguistically knowledgeable developers. Instead of a
``\verb!parallel(...)!'' structure, an atomic ``prototype'' may be
given; this is, as in earlier releases, expected to have its own
definitions. Alternatively, if the required syntactic and semantic
behaviour follows a known paradigm, a structure of the form
\verb!paradigm(Para,Proto)! may be given; this causes the 
prototype \verb!Proto! to be defined using the paradigm name
\verb!Para!.

\section{Lexical Failure Recovery}

When a token identified by tokenization cannot be segmented as a
normal word or a special form, and when there is no alternative path
through the lattice that avoids such tokens, CLARE has two basic
strategies available. One is to assume that the token is as intended
but the lexicon is incomplete, and attempt to derive an entry by
accessing an external database such the OUP (MRC) lexicon; eliciting
information from the user (via VEX); or simply by guesswork.  The
mechanisms for this are as described in Alshawi, 1992.
The other strategy is that the word is not as intended, i.e.\ that a
spelling or typing error has occurred.  All of these possibilities
correspond to switches that may be turned on and off by users or
application developers.

CLARE's token segmentation phase attempts to find analyses
for all the single tokens in the lattice, and for any special forms,
which may include spaces and therefore span multiple tokens.  Next, a
series of {\it recovery methods}, which may be augmented or reordered
by the application developer, are applied. {\it Global} methods apply
to the lattice as a whole, and are intended to modify its contents or
create required lexicon entries on a scale larger than the individual
token.  {\it Local} methods apply only to single still-unanalysed
tokens, and may either supply analyses for them or alter them to other
tokens. The default methods, all of which may be switched on or off
using system commands, supply facilities for inferring entries through
access to an external machine-readable dictionary; for defining
sequences of capitalized tokens as proper names; for spelling
correction (described in detail in the next section); and for
interacting with the user who may suggest a replacement word or phrase
or enter the VEX lexical acquisition subsystem (Carter, 1989) to
create the required entries.

After a method has been applied, the lattice is, if possible, pruned:
edges labelled by unanalysed tokens are provisionally removed, as are
other edges and vertices that then do not lie on a complete path. If
pruning succeeds (i.e.\ if at least one problem-free path remains)
then token analysis is deemed to have succeeded, and unanalysed tokens
(such as {\it with/avoid}) are forgotten; any remaining global methods
are invoked, because they may provide analyses for token {\it
sequences}, but remaining local ones are not.  If full pruning does
not succeed, any subpath in the lattice containing more unrecognized
tokens than an alternative subpath is eliminated. Subpaths containing
tokens with with non-alphabetic characters are penalized more heavily;
this ensures that if the cluster ``\verb!boooks,!'' is input, the
token sequence ``\verb!boooks ,!''  (in which ``\verb!boooks!''  is an
unrecognized token and ``\verb!,!''  is a comma) is preferred to the
single token ``\verb!boooks,!'' (where the comma is part of the
putative lexical item). The next method is then applied.\footnote{In
fact, for completeness, CLARE allows the application of two or more
methods in tandem and will combines the results without any
intermediate pruning. This option would be useful if, in a given
application, two sources of knowledge were deemed to be about equally
reliable in their predictions.}

One major advantage of the simplicity of the affix-stripping mechanism
is that spelling correction can be interleaved directly with it. Root
forms in the lexicon are represented in a discrimination net for
efficient access (cf Emirkanian and Bouchard, 1988). When the spelling
corrector is called to suggest possible corrections for a word, the
number of simple errors (of deletion, insertion, substitution and
transposition; e.g.~Pollock and Zamora, 1984) to assume is given.
Normal segmentation is just the special case of this with the number
of errors set to zero. The mechanism nondeterministically removes
affixes from each end of the word, postulating errors if appropriate,
and then looks up the resulting string in the discrimination net,
again considering the possibility of error.\footnote{This is the
reverse of Veronis' (1988) algorithm, where roots are matched before
affixes. However, it seems easier and more efficient to match affixes
first, because then the hypothesized root can be looked up without
having to allow for any spelling changes; and if both prefixes and
suffixes are to be handled, as they are in CLARE, there is no obvious
single starting point for searching for the root first.}

Interleaving correction with segmentation like this promotes
efficiency in the following way. As in most other correctors, only up
to two simple errors are considered along a given search path.
Therefore, either the affix-stripping phase or the lookup phase is
fairly quick and produces a fairly small number of results, and so the
two do not combine to slow processing down. Another beneficial
consequence of the interleaving is that no special treatment is
required for the otherwise awkward case where errors overlap morpheme
boundaries; thus {\it desigend} is corrected to {\it designed} as
easily as {\it deisgned} or {\it designde} are.

If one or more possible corrections to a token are found, they may
either be presented to the user for selection or approval, or, if the
number of them does not exceed a preset threshold, all be preserved as
alternatives for disambiguation at the later syntactic or semantic
stages.  The lattice representation allows multiple-word corrections
to be preserved along with single-word ones.

It is generally recognized that spelling errors in typed input are of
two kinds: competence errors, where the user does not know, or has
forgotten, how to spell a word; and performance errors, where the
wrong sequence of keys is hit. CLARE's correction mechanism is
oriented towards the latter. Other work (e.g.\ Veronis, 1988,
Emirkanian and Bouchard, 1988, van Berkel and De Smedt, 1988)
emphasizes the former, often on the grounds that competence errors are
both harder for the user to correct and tend to make a worse
impression on a human reader. However, Emirkanian and Bouchard
identify the many-to-one nature of French spelling-sound
correspondence as responsible for the predominance of such errors in
that language, which they say does not hold in English; and material
typed to CLARE tends to be processed further (for database access,
translation, etc) rather than reproduced for potentially embarrassing
human consumption.  A performance-error approach also has the
practical advantage of not depending on extensive linguistic
knowledge; and many competence errors can be detected by a performance
approach, especially if some straightforward adjustments (e.g.\ to
prefer doubling to other kinds of letter insertion) are made to the
algorithm.

As well as coping quite easily with morpheme boundaries, CLARE's
algorithm can also handle the insertion or deletion of word boundary
spaces. For the token {\it witha}, CLARE postulates both {\it with}
and {\it with a} as corrections, and (depending on the current switch
settings) both may go into the lattice. The choice will only finally
be made when a QLF is selected on sortal and other grounds after
parsing and semantic analysis. For the token pair {\it nev er}, CLARE
postulates the single correction {\it never}, because this involves
assuming only one simple error (the insertion of a space) rather than
two or more to ``correct'' each token individually. Multiple
overlapping possibilities can also be handled; the input {\it Th m n
worked} causes CLARE to transform the initial lattice
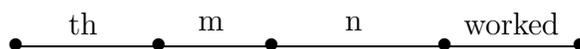
\setlength{\unitlength}{1mm}

\begin{center}
\begin{picture}(75,10)
\put(0,0.2){$\bullet$}
\put(1,1){\line(1,0){75}}
\put(10,4){\makebox(0,0)[c]{th}}
\put(19,0.2){$\bullet$}
\put(27,4){\makebox(0,0)[c]{m}}
\put(34,0.2){$\bullet$}
\put(46,4){\makebox(0,0)[c]{n}}
\put(57,0.2){$\bullet$}
\put(67,4){\makebox(0,0)[c]{worked}}
\put(75,0.2){$\bullet$}
\end{picture}
\end{center}
\noindent
into a corrected lattice containing analyses of the words shown
here:\newline
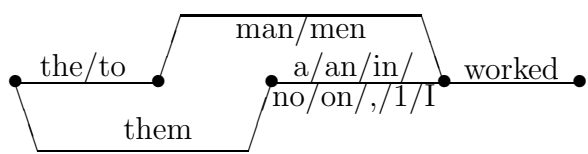
\begin{center}
\begin{picture}(75,25)
\put(0,10.2){$\bullet$}
\put(1,11){\line(1,-3){3}}
\put(4,2){\line(1,0){28}}
\put(20,5){\makebox(0,0)[c]{them}}
\put(32,2){\line(1,3){3}}
\put(1,11){\line(1,0){19}}
\put(10,13){\makebox(0,0)[c]{the/to}}
\put(19,10.2){$\bullet$}
\put(20,11){\line(1,3){3}}
\put(23,20){\line(1,0){32}}
\put(39,18){\makebox(0,0)[c]{man/men}}
\put(55,20){\line(1,-3){3}}
\put(34,10.2){$\bullet$}
\put(35,11){\line(1,0){23}}
\put(46,13){\makebox(0,0)[c]{a/an/in/}}
\put(46,9){\makebox(0,0)[c]{no/on/,/1/I}}
\put(57,10.2){$\bullet$}
\put(58,11){\line(1,0){18}}
\put(67,13){\makebox(0,0)[c]{worked}}
\put(75,10.2){$\bullet$}
\end{picture}
\end{center}
\noindent
The edges labelled ``them'' and ``man/men'' are constructed first by
the ``global'' spelling correction method, which looks for possible
corrections across token boundaries. The edge for the token ``m'' is
then removed because, given that it connects only to errorful tokens
on both sides, it cannot form part of any potentially optimal path
through the lattice.  Corrections are, however, sought for ``th'' and
``n'' as single tokens when the local spelling correction method is
invoked. The corrected lattice then undergoes syntactic and semantic
processing, and QLFs for the sequences ``the man worked'' and ``the
men worked'', but not for any sequence starting with ``them'' or
``to'', are produced.

\section{Idiom and Abbreviation Recognition}

However wide the coverage of a grammar, it is inevitable that some
constructions will be encountered that go beyond it.  This is
particularly true when tailoring to applications is considered.  There
will also be constructions that are within syntactic coverage but
whose semantics differ from what the semantic rules would predict.
Such constructions are, at least from the standpoint of the system's
inevitably incomplete grammar, idiomatic. They present a problem when
an application must be developed without help from a linguist
competent to extend the grammar in the normal way.

CLARE therefore allows application developers and other users to write
lexical entries that define phrases as equivalent to other words or
phrases.  Flexibility is allowed in affixation but not in the sequence
of roots; this is in contrast to a similar facility in the TEAM system
(Grosz et al, 1987), 
where such phrases were defined using regular expressions. Thus in
CLARE, a user can specify that occurrences of the phrase {\it kick
the bucket} should, as one analysis possibility, be treated as if the
word {\it die} had been input instead, and that any affixes on {\it
kick} should be transferred to {\it die}. This allows idiomatic
analysis of {\it John kicked the bucket}, but not of {\it John kicked
the proverbial bucket} or {\it The bucket was kicked by John}. Such
constraints on the applicability of entries can be quite easily
understood by non-linguists, who may create the entries through
interacting with the VEX lexical acquisition subsystem; this might not
be true if regular expressions were used. In our example, the user
need only supply the underlined material. He or she does, however,
need to choose a target expression that the system can interpret
unambiguously in the contexts where the idiom is likely to occur,
because if ``die'' has multiple senses, the system has no way of
knowing which one means the same as ``kick the bucket''.
\begin{small}\begin{quote}{\tt
Enter the word or word group to be defined: \underline{kick the bucket} \\
Is this an adjective, noun, verb or idiom in your domain? \\
Enter one or more of a, n, v or i: \underline{i} \\
 \\
I am now defining "kick the bucket" as an idiom. \\
 \\
Enter a phrase with an equivalent meaning: \underline{die}
 \\
Where should any affix on "kick" be attached? \\
\hspace{5mm}1 Don't allow any affix. \\
\hspace{5mm}2 Ignore any affix. \\
\hspace{5mm}3 Attach it to "die". \\
Enter a number between 1 and 3: \underline{3} \\
 \\
Where should any affix on "bucket" be attached? \\
\hspace{5mm}1 Don't allow any affix. \\
\hspace{5mm}2 Ignore any affix. \\
\hspace{5mm}3 Attach it to "die". \\
Enter a number between 1 and 3: \underline{1}}
\end{quote}\end{small}

Phrases treated as idioms in this way do not have to be idioms in the
usual linguistic sense. Abbreviations and variant spellings can also
be treated with this mechanism, defining, say, {\it i.e.} as {\it that
is} and {\it color} as {\it colour}.  In the true idiom case, however,
provision must be made for literal as well as idiomatic
interpretations.  It is not correct to replace an occurrence of {\it
kicking the bucket} with the single word {\it dying}; both must be
preserved. The easiest way to achieve the desired effect is to
represent both possibilities in parallel in a word lattice.

Users may also define ``special form expansions'' in which one word
sequence, specified using regular expressions, is defined as
equivalent to another. For example, the entry
\begin{quote}
\begin{verbatim}
special_form_expansion('\$\([0-9]+\)',['\1',dollars]).
\end{verbatim}
\end{quote}
causes a token such as \verb!$523! to be handled as if it were
\verb!523 dollars!, which is more convenient for syntactic reasons.

Idioms and special form expansions are ignored in generation, because
in the generation direction they would simply replace one word string
by another, perhaps more opaque, one, adding nothing to the coverage
of the generator. That is, if it is possible to derive a sentence from
a QLF using idiom definitions, it is also possible to do so without
using them.

\section{Phrase Parsing}\label{Phrasal-Parsing}

Just before sentence parsing proper, the {\it phrase parsing} phase
applies a designated set of syntactic rules to the lattice of
categories created by word parsing. These rules have exactly the same
format as other syntactic rules. However, they are applied in a simple
left-corner, bottom-up manner, and may not therefore involve the
identification of syntactic ``gaps'' in the input (although, in
contrast to earlier versions of CLARE, they may involve recursion). In
the current grammar, phrase rules match such sequences as
\begin{quote}
how \var{adjective} a \\
the \var{adjective}est \\
at least \var{number} \\
\end{quote}
The distinction between normal and phrasal grammar rules promotes
efficiency in two ways. Firstly, the simpler mechanism used to apply
phrase rules is inherently more efficient. Secondly, any factorization
of the grammar will reduce the search space; phrasal rules obviously
cannot apply to the output of the main parsing phase.

When phrase parsing begins, pseudo-edges for the start and end of the
sentence are added to the lattice. These allow a more general
treatment of words like ``however'', which, when used
sentence-internally as sentential adverbs, appear between a pair of
commas, one of which can be viewed as being deleted when the adverb
begins or ends the sentence.

The phrase rule mechanism may initially seem to provide similar
functionality to the idiom mechanism. However, the two differ in
important ways:
\begin{itemize}
\item A phrase rule, but not an idiom definition, specifies a pattern that 
can include syntactic categories as well as words, and may be
recursively applied. For this reason, non-linguists are not
recommended to write phrase rules. The idiom mechanism was introduced
precisely to allow non-linguists to extend the coverage of the system.
\item An idiom definition, but not a phrase rule, can map its
input onto a sequence rather than a single symbol. Indeed, the members
of the sequence may form part of quite different constituents in the
ultimate parse.
\item Phrase rules, like other syntactic rules, are essential for
successful generation of a word string from some QLFs. Idiom
definitions are ignored in generation, as stated above.
\end{itemize}

\section{Examples of Lattice Processing}

In CLARE, morphological, syntactic and semantic processing all act on
the lattice of word candidates identified by the processes just
described, and resulting word-identity ambiguities are resolved as a
by-product of the selection (by the user and/or the system) of the
appropriate QLF for the sentence, which will correspond to a
particular path through the lattice. Thus final decisions are delayed
until sufficient knowledge is available to make them.

A simple example of this delaying effect, involving spelling
correction, is:
\begin{quote}\begin{verbatim}
Show methe project nambers.
\end{verbatim}\end{quote}
where the intended input, typical of those occurring in resource
management applications such as the PRM system developed under CLARE,
is {\it Show me the project numbers}. The word lattice, after spelling
correction is applied to \verb!methe!  and \verb!nambers!, consists of
segmentations of the following word candidates:
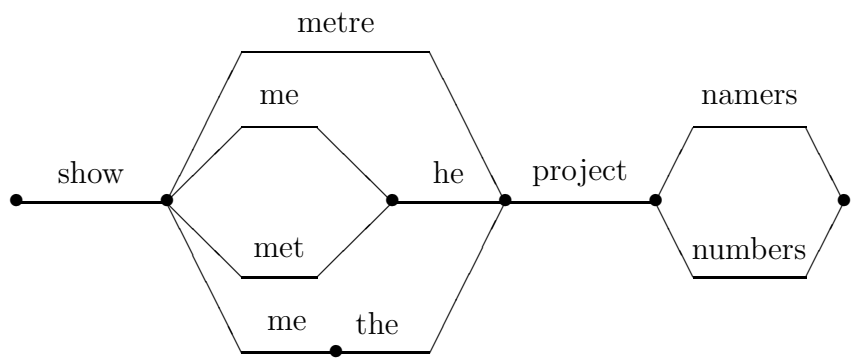
\begin{center}
\begin{picture}(100,50)
\put(0,24.2){$\bullet$}
\put(1,25){\makebox(20,8)[c]{show}}
\put(1,25){\line(1,0){20}}
\put(20,24.2){$\bullet$}
\put(21,25){\line(1,2){10}}
\put(31,45){\makebox(25,8)[c]{metre}}
\put(31,45){\line(1,0){25}}
\put(56,45){\line(1,-2){10}}
\put(56,5){\line(1,2){10}}
\put(21,25){\line(1,-2){10}}
\put(31,5){\makebox(12,8)[c]{me}}
\put(41.5,5){\makebox(15,8)[c]{the}}
\put(31,5){\line(1,0){25}}
\put(42.5,4,2){$\bullet$}
\put(51,25){\makebox(15,8)[c]{he}}
\put(51,25){\line(1,0){15}}
\put(50,24.2){$\bullet$}
\put(21,25){\line(1,1){10}}
\put(31,35){\makebox(10,8)[c]{me}}
\put(31,35){\line(1,0){10}}
\put(41,35){\line(1,-1){10}}
\put(21,25){\line(1,-1){10}}
\put(31,15){\makebox(10,8)[c]{met}}
\put(31,15){\line(1,0){10}}
\put(41,15){\line(1,1){10}}
\put(65,24.2){$\bullet$}
\put(66,25){\makebox(20,8)[c]{project}}
\put(66,25){\line(1,0){20}}
\put(85,24.2){$\bullet$}
\put(86,25){\line(1,2){5}}
\put(86,25){\line(1,-2){5}}
\put(91,35){\makebox(15,8)[c]{namers}}
\put(91,35){\line(1,0){15}}
\put(91,15){\makebox(15,8)[c]{numbers}}
\put(91,15){\line(1,0){15}}
\put(106,35){\line(1,-2){5}}
\put(106,15){\line(1,2){5}}
\put(110,24.2){$\bullet$}
\end{picture}
\end{center}
When this lattice is parsed, four of the eight possible strings (those
involving the sequences ``met he'' and ``metre'' and finishing with
either ``namers'' or ``numbers'') are effectively ruled out, because
no complete parse of the sentence can be based on those sequences.
Semantic processing leads to QLFs being produced first for {\it Show
met project numbers} and second for {\it Show me the project namers},
but the latter is rejected by users (or, potentially, by the system
itself) because the phrase {\it project namers} is uninterpretable in
the PRM domain. The first QLF eventually causes an appropriate list of
project numbers to be displayed. All this is done without the user
ever being explicitly queried about the required spelling changes.

Theoretically, an alternative to the use of a lattice would be to
create a list of the possible complete word sequences for the
sentence.  However, this would involve much repeated work and be very
inefficient.  Currently in CLARE, the time taken for parsing and
semantic analysis depends on the size of the lattice, and not directly
on the number of distinct word sequences it represents. For sentences
with many word identity ambiguities, this can be a major advantage.
Indeed, it is possible for even fairly short sentences containing
instances of the lexical phenomena discussed in this chapter,
especially spelling errors, to give rise to lattices containing tens
of edges and representing hundreds of possible sequences of
morphologically analyses of words.

The above example showed interaction between two lexical processing
problems of the same type. However, interacting problems of different
types can be handled as easily. Suppose that the phrase {\it what
number} is defined as an idiom meaning {\it how many}, and the slash
character as, potentially, a word separating punctuation symbol
defined in the lexicon as a conjunction.\footnote{This example is in
fact beyond the coverage by the CLARE/PRM system as delivered, but
is included for illustrative reasons.} Then the input
\begin{quote}\begin{verbatim}
Tell mewht number of our bills/ivoices have not been paid.
\end{verbatim}\end{quote}
where \verb!mewht! is a mistyping of \verb!me what! and \verb!ivoices!
of \verb!invoices!, causes various spelling corrections to be
hypothesized, triggering recognition of the idiom. After spelling
correction, the word lattice (with some uninteresting parts omitted)
therefore becomes\footnote{The dashed edge, for
\verb!bills/invoices!, would only exist if for some reason that string
were defined in the lexicon or otherwise analysable as a single word,
and were therefore available as a correction for
\verb!bills/ivoices!.}:

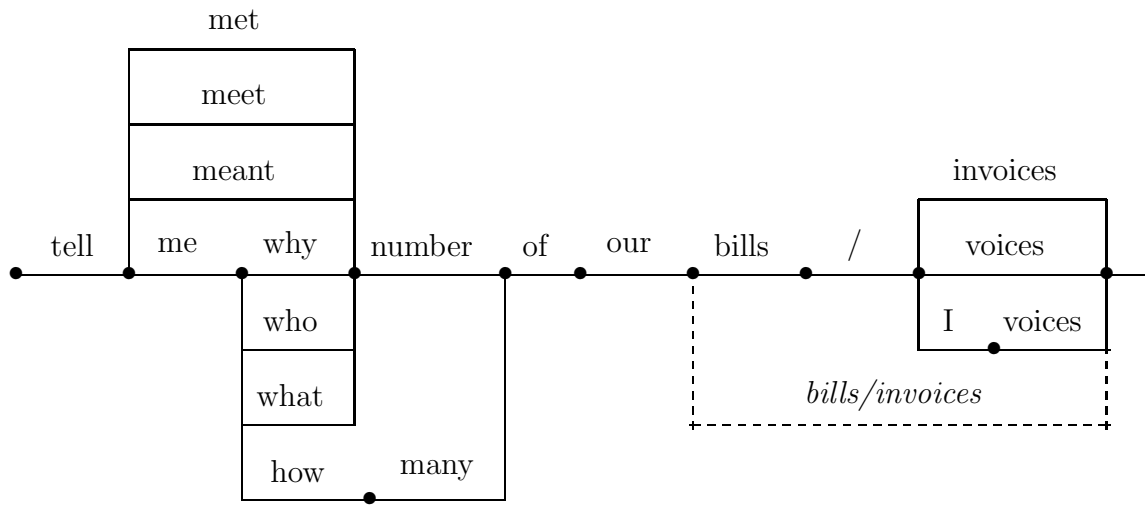
\begin{center}
\begin{picture}(150,80)
\put(0,34.2){$\bullet$}
\put(1,35){\makebox(15,8)[c]{tell}}
\put(1,35){\line(1,0){150}}
\put(16,35){\line(0,1){30}}
\put(15,45){\makebox(30,8)[c]{meant}}
\put(16,45){\line(1,0){30}}
\put(15,55){\makebox(30,8)[c]{meet}}
\put(16,55){\line(1,0){30}}
\put(15,65){\makebox(30,8)[c]{met}}
\put(16,65){\line(1,0){30}}
\put(46,35){\line(0,1){30}}
\put(31,35){\line(0,-1){30}}
\put(30,25){\makebox(15,8)[c]{who}}
\put(31,25){\line(1,0){15}}
\put(30,15){\makebox(15,8)[c]{what}}
\put(31,15){\line(1,0){15}}
\put(30,5){\makebox(17,8)[c]{how}}
\put(31,5){\line(1,0){35}}
\put(47,4.2){$\bullet$}
\put(48,5){\makebox(18,8)[c]{many}}
\put(46,35){\line(0,-1){20}}
\put(66,35){\line(0,-1){30}}
\put(15,34.2){$\bullet$}
\put(15,35){\makebox(15,8)[c]{me}}
\put(30,34.2){$\bullet$}
\put(30,35){\makebox(15,8)[c]{why}}
\put(45,34.2){$\bullet$}
\put(45,35){\makebox(20,8)[c]{number}}
\put(65,34.2){$\bullet$}
\put(65,35){\makebox(10,8)[c]{of}}
\put(75,34.2){$\bullet$}
\put(75,35){\makebox(15,8)[c]{our}}
\put(90,34.2){$\bullet$}
\put(90,35){\makebox(15,8)[c]{bills}}
\put(105,34.2){$\bullet$}
\put(105,35){\makebox(15,8)[c]{/}}
\put(120,34.2){$\bullet$}
\put(120,35){\makebox(25,8)[c]{voices}}
\put(145,34.2){$\bullet$}
\put(91,15){\dashbox(0,20){}}
\put(91,15){\dashbox(55,0){}}
\put(146,15){\dashbox(0,10){}}
\put(90,15){\makebox(55,8){\it bills/invoices}}
\put(121,25){\line(0,1){20}}
\put(146,25){\line(0,1){20}}
\put(121,45){\line(1,0){25}}
\put(120,45){\makebox(25,8)[c]{invoices}}
\put(121,25){\line(1,0){25}}
\put(130,24.2){$\bullet$}
\put(120,25){\makebox(10,8)[c]{I}}
\put(130,25){\makebox(15,8)[c]{voices}}
\end{picture}
\end{center}

The correct word string can again be identified by subsequent
processing; only the initial path ``tell me how many...'' supports
any complete syntactic analyses, and the choice between ``invoices''
and ``voices'' could be made at QLF level (on domain-specific word
frequency grounds, for example).

A lattice structure also means that the decision on whether to attempt
correction or recovery for unrecognized tokens can be made sensitive
to context. If, say, CLARE were to be used to access a database of
vehicles, a special form entry might be created to allow recognition
of registration numbers of the form \verb!lll nnn! where each
\verb!l! is a letter and each \verb!n! a number. Consider the queries
\begin{verbatim}
  What is the value of BMQ 587?
  What is the value of the BMQ?
\end{verbatim}
Suppose \verb!BMW!, but not \verb!BMQ!, is defined in the
lexicon as a count noun denoting a type of car. In processing the
first query, \verb!BMQ 587! is recognized as a registration number,
and we do not want the system to attempt to correct the spelling of
\verb!BMQ!. But in the second, no registration number is recognized,
and possibility of correcting \verb!BMQ! to \verb!BMW! should be
considered.

When faced with alternative paths through the lattice, such as would
arise in the \verb!BMQ 587! example, CLARE therefore prunes any path
containing $N$ unanalysable words as long as there is an alternative
path with less than $N$. The lattices for the two examples before
correction, with italics denoting unanalysability, are as follows:

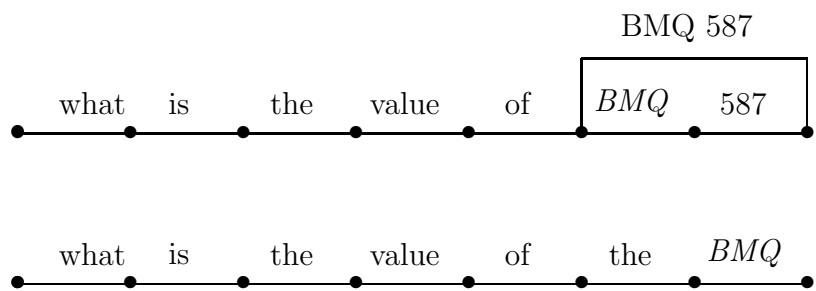
\begin{center}
\begin{picture}(105,40)
\put(0,24.2){$\bullet$}\
\put(1,25){\makebox(20,8)[c]{what}}
\put(15,25){\makebox(15,8)[c]{is}}
\put(30,25){\makebox(15,8)[c]{the}}
\put(45,25){\makebox(15,8)[c]{value}}
\put(60,25){\makebox(15,8)[c]{of}}
\put(75,25){\makebox(15,8)[c]{\it BMQ}}
\put(90,25){\makebox(15,8)[c]{587}}
\put(1,25){\line(1,0){105}}
\put(0,24.2){$\bullet$}
\put(15,24.2){$\bullet$}
\put(30,24.2){$\bullet$}
\put(45,24.2){$\bullet$}
\put(60,24.2){$\bullet$}
\put(75,24.2){$\bullet$}
\put(90,24.2){$\bullet$}
\put(105,24.2){$\bullet$}
\put(76,25){\line(0,1){10}}
\put(106,25){\line(0,1){10}}
\put(76,35){\line(1,0){30}}
\put(75,35){\makebox(30,8)[c]{BMQ 587}}
\put(0,4.2){$\bullet$}\
\put(1,5){\makebox(20,8)[c]{what}}
\put(15,5){\makebox(15,8)[c]{is}}
\put(30,5){\makebox(15,8)[c]{the}}
\put(45,5){\makebox(15,8)[c]{value}}
\put(60,5){\makebox(15,8)[c]{of}}
\put(75,5){\makebox(15,8)[c]{the}}
\put(90,5){\makebox(15,8)[c]{\it BMQ}}
\put(1,5){\line(1,0){105}}
\put(0,4.2){$\bullet$}
\put(15,4.2){$\bullet$}
\put(30,4.2){$\bullet$}
\put(45,4.2){$\bullet$}
\put(60,4.2){$\bullet$}
\put(75,4.2){$\bullet$}
\put(90,4.2){$\bullet$}
\put(105,4.2){$\bullet$}
\end{picture}
\end{center}

In first lattice, therefore, the lower path, where \verb!BMQ! and
\verb!587! are separate items, is pruned without any correction being
considered, because it contains an unanalysable word and the
alternative upper path, with \verb!BMQ 587! as a single special form,
does not. In the second lattice, in contrast, there is no alternative
path to compete with that along which \verb!BMQ! is unanalysable.  The
\verb!BMQ! is thus not pruned, and spelling correction and/or user
intervention are invoked, depending on the system's switch settings.

\section{An Evaluation}

To assess the usefulness of syntactico-semantic constraints in CLARE's
spelling correction, the following experiment, intended to simulate
performance (typographic) errors, was carried out.  Five hundred
sentences, of up to ten words in length, falling within CLARE's
current core lexical (1600 root forms) and grammatical coverage were
taken at random from the LOB corpus. These sentences were passed,
character by character, through a channel which transmitted a
character without alteration with probability 0.99, and with
probability 0.01 introduced a simple error. The relative
probabilities of the four different kinds of error were deduced from
Table X of Pollock and Zamora, 1984; where a new character had to be
inserted or substituted, it was selected at random from the original
sentence set.  This process produced a total of 102 sentences that
differed from their originals. Some examples are:
\begin{quote}
\begin{verbatim}
Andb I agree with it.
Do we or do we not wa nt registration?
She was very goood aboutit.
Les sthan 2,000 now remain.
\end{verbatim}
\end{quote}
The average length was 6.46 words, and there were 123 corrupted tokens
in all, some containing more than one simple error.  Because longer
sentences were more likely to be changed, the average length of a
changed sentence was some 15\% more than that of an original one.

The corrupted sentence set was then processed by CLARE with only the
spelling correction recovery method in force and with no user
intervention. Up to two simple errors were considered per token. No
domain-specific or context-dependent knowledge was used.

Of the 123 corrupted tokens, ten were corrupted into other known
words, and so no correction was attempted. Parsing failed in nine of
these cases; in the tenth, the corrupted word made as much sense as
the original out of discourse context. In three further cases, the
original token was not suggested as a correction; one was a special
form, and for the other two, alternative corrections involved fewer
simple errors.  The corrections for two other tokens were not used
because a corruption into a known word elsewhere in the same sentence
caused parsing to fail.

Only one correction (the right one) was suggested for 59 of the
remaining 108 tokens. Multiple-token correction, involving the
manipulation of space characters, took place in 24 of these cases.

This left 49 tokens for which more than one correction was suggested,
requiring syntactic and semantic processing for further
disambiguation.  The average number of corrections suggested for these
49 was 4.57.  However, only an average of 1.69 candidates (including,
because of the way the corpus was selected, all the right ones)
appeared in QLFs satisfying selectional restrictions; thus only 19\%
of the wrong candidates found their way into any QLF.  If, in the
absence of frequency information, we take all candidates as equally
likely, then syntactic and semantic processing reduced the average
entropy from 1.92 to 0.54, removing 72\% of the uncertainty (see
Carter, 1987, for a discussion of why entropy is the best measure to
use in contexts like this).

When many QLFs are produced for a sentence, CLARE orders them
according to a set of scoring functions encoding syntactic and
semantic preferences. For the 49 multiple-candidate tokens,
removing all but the best-scoring QLF(s) eliminated 7 (21\%) of the 34
wrong candidates surviving to the QLF stage; however, it also
eliminated 5 (10\%) of the right candidates.  It is expected that
future development of the scoring functions will further improve these
figures, which are summarized in Table \ref{ctab}.
\begin{table}
\begin{center}
\begin{tabular}{|l|l|l|l|} \hline
Stage & Right & Wrong & Average \\ 
      & cand's & cand's & number \\ \hline
Suggested &  49 & 175 & 4.57 \\ \hline
In any QLF & 49 & 34 &  1.69 \\ \hline
In best-scoring QLF(s) & 44  & 27 & 1.45 \\ \hline
\end{tabular}
\end{center}
\caption{Correction candidates for the 49 multiple-candidate tokens}
\label{ctab}
\end{table}

The times taken to parse lattices containing multiple spelling
candidates reflect the characteristics of CLARE's parser, which uses a
backtracking, left-corner algorithm and stores well-formed
constituents so as to avoid repeating work where possible.  In
general, when a problem token appears late in the sentence and/or when
several candidate corrections are syntactically plausible, the lattice
approach is several times faster than processing the alternative
strings separately (which tends to be very time-consuming). When the
problem token occurs early and has only one plausible correction, the
two methods are about the same speed.

For example, in one case, a corrupted token with 13 candidate
corrections occurred in sixth position in an eight-word sentence.
Parsing the resulting lattice was three times faster than parsing each
alternative full string separately. The lattice representation avoided
repetition of work on the first six words. However, in another case,
where the corrupted token occurred second in an eight-word sentence,
and had six candidates, only one of which was syntactically plausible,
the lattice representation was no faster, as the incorrect candidates
in five of the strings led to the parse being abandoned early.

An analogous experiment was carried out with 500 sentences from the
same corpus which CLARE could {\it not} parse. 131 of the sentences,
with average length 7.39 words, suffered the introduction of errors.
Of these, only seven (5\%) received a parse. Four of the seven
received no sortally valid QLFs, leaving only three (2\%) ``false
positives''. This low figure is consistent with the results from the
originally parseable sentence set; nine out of the ten corruptions into
known words in that experiment led to parse failure, and only 19\% of
wrong suggested candidates led to a sortally valid QLF.  If, as those
figures suggest, the replacement of one word by another only rarely
maps one sentence inside coverage to another, then a corresponding
replacement on a sentence {\it outside} coverage should yield
something within coverage even more rarely, and this does appear to be
the case.

These experimental results suggest that general syntactic and semantic
information is an effective source of constraint for correcting typing
errors, and that a conceptually fairly simple staged architecture,
where word identity and word boundary ambiguities are only resolved
when the relevant knowledge is ready to be applied, can be acceptably
efficient. The lattice representation also allows the system to deal
cleanly with word boundary uncertainty not caused by noise in the
input.   
 
A fairly small vocabulary was used in the experiment. However, these
words were originally selected on the basis of frequency of
occurrence, so that expanding the lexicon would involve introducing
proportionately fewer short words than longer ones.  Mistyped short
words tend to be the ones with many correction candidates, so the
complexity of the problem should grow less fast than might be expected
with vocabulary size.  Furthermore, more use could be made of
statistical information: relative frequency of occurrence could be
used as a criterion for pruning relatively unlikely correction
candidates, as could more sophisticated statistics in the suggestion
algorithm, along the lines of Kernighan {\it et al} (1990).
Phonological knowledge, to allow competence errors to be tackled more
directly, would provide another useful source of constraint.

\section{Responding to Parse Failures}

By default, CLARE responds to a failure in syntactic parsing by
searching for and displaying the substrings of the input that can be
parsed and interpreted as sentences, noun phrases, or other
constituents that the application developer may specify.
An example is the input
\begin{quote}\begin{verbatim}
Can you tell me what colleges are pretty much within
walking distance of the centre of the city?
\end{verbatim}\end{quote}
to which CLARE responds
\begin{quote}\begin{verbatim}
... No parses found.

Minimal sequence(s) of words or phrases have length 4:
can you tell me which colleges are pretty | much within walking |
distance | of the centre of the city
\end{verbatim}\end{quote}
Additional information may also be requested. In this case, it is as
follows.
\begin{quote}\begin{verbatim}
Maximal entity descriptions:
  ---  you
  ---  me
  ---  which colleges are
  ---  much within walking
  ---  the centre of the city

Maximal embedded sentence:
  ---  can you tell me which colleges are pretty

Shortest sequence(s) of words and complete constituents:

    S:8 NP:3 distance PP:6
\end{verbatim}\end{quote}
The parsed substrings are intended to give the user some idea of where
the problem parts of the sentence are. In this example, the user
should notice that the beginning and end of the sentence were both
parsed successfully, but that no significant progress was made on the
middle part, ``pretty much within walking distance of''.
This should give the user some guidance in rephrasing his request,
e.g.:
\begin{quote}\begin{verbatim}
Can you tell me which colleges are near the centre of the city?
\end{verbatim}\end{quote}
for which parsing and semantic analysis succeed.

Some caution needs to be exercised in interpreting partial parses,
however. The embedded sentence in the example, ``Can you tell me which
colleges are pretty?'' is not in fact a constituent of the intended
interpretation of either the original input or the subsequent,
simplified version; it involves treating ``pretty'' in its meaning of
``attractive'' rather than as a modifier of what follows.
Nevertheless, this bracketing does give a fairly good idea of what a
relevant part of the system's coverage is, indicating, for example,
that sentences of the form ``Can you ...'', and occurrences of
``tell'' with a dative and a ``wh-'' sentential complement, are
covered.

For the sake of brevity, constituents spanning a strict subpart of the
text scanned by constituents of the same type are not reported. Thus
although the phrase ``Tell me which colleges are'' is identified as a
sentence, it is not included in the list of sentences found.

Sentences (major category \verb!s! in the grammar) and noun phrases or
``entity descriptions'' (major category \verb!np!) were selected for
display because they seem to correspond to easily identifiable
concepts more closely than some other categories do. While naive users
can be expected to understand the word ``sentence'' and to assign some
meaning to the phrase ``entity description'', it is hard to find ways
of expressing, say, categories \verb!vp! (verb phrase) or \verb!nbar!
in any non-technical way that will not sometimes be misleading.
Also, the categories \verb!s! and \verb!np! may appear as complete
utterances rather more often than some others, thus making them easier
for the user to interpret as sentence fragments. However, the types of
constituent reported may be varied if desired.

The initial (failed) parse of a sentence does not necessarily create
entries in the well-formed substring table for every possible
constituent relevant to partial parsing. This is because during
parsing, contextual expectations, derived from the phrases already
identified, are applied. To find the complete set of required
constituents, it is therefore necessary to parse the sentence starting
from every word boundary within it, searching for the desired
constituents.

The additional processing seems typically to double the total amount
of time needed for parsing, as compared with a successfully parsed
sentence of similar length, though there is a fair amount of
variation.

If semantic analysis is switched on when a partial parse is produced,
one or more partial semantic analyses will also be sought, based on
the minimal parsing sequence(s). Only certain constituents (certain
types of \verb!np!, \verb!s!, \verb!pp!, \verb!adjp! and \verb!advp!)
are used as the starting points for such top-down analysis. The QLF
returned is a \verb!form! involving the application of an unspecified
predicate to the list of arguments formed by analysing each
constituent in the sequence.

\section{Responding to Sortal Restriction Failures}

CLE-3 (Alshawi, 1992) contained a facility for examining logical forms
after all analyses of a sentence had failed sortal restrictions, and
showing how sortal restriction violations arose. This facility is
useful to people familiar with the contents of logical forms and with
the sort hierarchy currently in force; however, it is of very limited
use to non-experts. Therefore, while it is retained in CLARE, an
additional facility is provided that gives English explanations of
sortal failures.

An example is:
\begin{quote}\begin{verbatim}
Did the house decide to consider the plan?
\end{verbatim}\end{quote}
With only the core lexicon loaded, all semantic analyses fail sortal
checks. The CLARE system gives the following output:
\begin{quote}\begin{verbatim}
No sorted semantic analyses found.

Examining ill-sorted analysis 1 (yes/no question)...
Examining ill-sorted analysis 2 (yes/no question)...
The (best) analyses were both ill-sorted for the following reasons
   Analyses 1 and 2: 
    A house cannot consider anything.
consider_ThinkAbout ---> animate           => agent;
house_Building      ---> passive_nonliving => nonagent.

   Analyses 1 and 2: 
    A house cannot decide to do anything.
decide_ChooseTo ---> animate           => agent;
house_Building  ---> passive_nonliving => nonagent.
\end{verbatim}\end{quote}
The location of the first violation is indicated by the sentence ``A
house cannot consider anything'', and its nature is given on the
following two lines. It should be clear that, according to the lexical
entries currently loaded, only animate beings are able to consider
things, and a house (only understood in the sense of ``building'') is
not animate.  Similarly, according to these entries, houses are not
able to make decisions.

Sometimes, all interpretations fail sorts, but for different reasons;
or there is more than one reason why a given reading fails. The system
attempts to summarize in such situations, grouping one reason together
under multiple readings when appropriate.

To gather the information necessary to provide these explanations, the
sortal restriction checker behaves differently when the relevant
system switch is turned on. Each variable in a logical form is
associated not just with a sort term (Alshawi, 1992, p174) but also
with a list of structures indicating what restrictions have applied to
it. For example, when the following QLF for ``The man left''
\begin{quote}\begin{verbatim}
[dcl,
 form(l([the,man,left]),verb(past,no,no,no,y),A,
      B^
      [B,
       [leave_Depart,A,
        term(l([the,man]),ref(def,the,sing,l([])),_,
             C^[man_MalePerson,C],_,_)]],
      _)]
\end{verbatim}\end{quote}
is checked, the variable \verb!C! is (during checking only) instantiated
to
\begin{verbatim}
  (D;
   ss(s(...,animate(human(human_being),male),...),
      [...
       arg(man_MalePerson,1),
       ...,
       arg(leave_Depart,2)|T]))
\end{verbatim}
Here, ``\verb!ss!'' stands for ``sourced sort'', indicating that both
a sort term, \verb!s(...)!, and a list of sources of that sort are
represented. The source list here
indicates that the variable has had to satisfy the sortal constraints
the first (and only) argument of the
predicate \verb!man_MalePerson!, and the second (agent) argument of
\verb!leave_Depart!. The list finishes with a variable tail, \verb!T!,
to allow it to be extended by unification during sort checking.

When two \verb!ss! terms are combined, the source list of each one is
extended by the other. If the sort terms themselves unify, nothing
further is done. Otherwise, the sort checker does not immediately
fail, as it would if explanations were not required. Instead, the sort
terms are left unchanged, but the (combined) source list is added to
an auxiliary {\it failure description list} (FDL). If, at the end of
checking a logical form, the FDL is empty, the
logical form is sortally consistent.  Otherwise, it is sortally
inconsistent; the sort checker fails, and the FDL is asserted in the
Prolog database in anticipation of {\it all} (quasi) logical forms for
the current sentence failing, and explanations being required.

If and when all readings do fail, and explanations are required, the
FDL for each reading is retrieved. Every source list in every FDL must
contain evidence of at least one {\it pairwise} sortal violation: that
is, there must be two source records that specify conflicting
sorts.\footnote{Non-pairwise conflicts are not possible in the absence
of pairwise ones, because sort terms never contain repeated variables.
If $T_1, \ldots, T_n$ are Prolog terms without repeated variables, and
if $T_i$ and $T_j$ unify for all $i$ and $j$, then unification of all
of $T_1, \ldots, T_n$ is guaranteed to succeed}.
The offending pair in each list is extracted, and the system attempts
to construct a QLF which, when handed to the generator, will give an
appropriate English sentence. For instance, when sortal checking fails
for the sentence ``The house slept'', it is known that there must be
a clash between two of the sorts referred to in the list
\begin{verbatim}
[...,
 arg(sleep_BeNaturallyUnconscious,2),
 ...
 arg(house_Building,1)|_]
\end{verbatim}
It is easily verified that the clash is between the two items given here.
The system then inspects the lexical entries from which the
offending predicates are derived in order to construct a sensible QLF.
A constant or a \verb!qterm!, which should be realised as a noun
phrase, is created from one of the source terms, if possible. Then the
main part of the QLF, into which the \verb!qterm! will slot, is
created from the other. The sentence patterns aimed for, with phrases
for the clashing terms shown in angle brackets include:
\begin{quote}
Nothing can \anglevar{Verb} ... \anglevar{NounPhrase} .... \\
\anglevar{NounPhrase} cannot \anglevar{VerbPhrase} \\
\anglevar{NounPhrase} cannot be \anglevar{AdjectivePhrase} \\
\anglevar{NounPhrase} cannot be \anglevar{NounPhrase} \\
Nothing can be \anglevar{Preposition} \anglevar{NounPhrase}\\
Nothing can be \anglevar{Adjective+Preposition} \anglevar{NounPhrase} \\
Nothing can \anglevar{Verb} \anglevar{Preposition} anything.
\end{quote}
The extra material indicated by dots in the first pattern, and the
contents of the verb phrase in the second pattern, depend on the
paradigm with which the predicate in question is defined. For an
intransitive verb, the verb phrase will be a single word:
\begin{verbatim}
>> The house slept.
 ...
A house can not sleep.
\end{verbatim}
But for a ditransitive verb or one with a sentential complement, it will
be more complex:
\begin{verbatim}
>> John sent Mary an idea.
 ...
Nothing can send anything an idea.
>> The house wants to leave.
 ...
A house cannot want to do anything.
\end{verbatim}
When a QLF cannot be constructed -- for example, when one of the
predicates is for a content word but is not derived from a paradigm
definition -- or when no QLF leads to a sentence being generated, the
system simply states that the two sources clash. This will be
comprehensible to a system developer but not to a naive user.

\section{Summary and Conclusions}

This chapter has described the way CLARE reacts to material that is
outside lexical, grammatical or sortal coverage. 

For the lexical stage, we have demonstrated that no simple
deterministic scheme can be adequate for word identification in a
system that aims to process real-world texts fully. The solution to
the problem adopted in the CLARE system is based on a lattice
representation, allowing conceptually clean treatments of individual
lexical phenomena to interact smoothly with one another. The
treatments selected are also aimed at allowing non-linguist
application developers to tailor the system to their needs at a level
beyond that of defining individual new words in the lexicon.

The current state of development provides a firm basis for development
to proceed on the syntactic and semantic coverage of the phenomena
whose lexical dimensions have been dealt with here. This can be quite
a major task; for example, it is rather easier to develop algorithms
to identify the different possible uses of bracket characters than to
write syntactic and semantic rules to account for all the uses of
parentheses in English.

More important, perhaps, is the evaluation of the claims made here
that non-linguists will be able to extend the system's coverage beyond
the lexical level by defining their own special forms and by using VEX
to define multiple-word idioms as well as single words. As CLARE is
tailored to prototype applications at various sites, the feasibility
of such an approach should become clearer.

CLARE's analysis of grammatical and sortal failure is aimed at
describing its view of the input so that the user can rephrase his or
her input appropriately. Its descriptions, which use bracketing and
English statements respectively, are intended to be useful to
non-linguists, both users and developers of applications.

\chapter{Coverage Extensions and Evaluations}
\label{clarecoverage}

This chapter describes some of the extensions to linguistic coverage made
during the CLARE project, and explains how documentation associated
with linguistic rules can be accessed to explore this coverage in more
detail. It then goes on to describe extensive tests carried out on the
delivered CLARE system to evaluate its coverage under different
conditions and in comparison to earlier versions of the system.
The chapter concludes with a description of some experimental work
with word tagging to enhance parsing efficiency.

\section{Additions to Coverage}

This section describes the most important of the changes or additions
that have been made to the syntactic and semantic analysis capabilities
of the CLE as developed under CLARE. Many smaller changes are not mentioned,
(for example, extensions to some new types of lexical item or idiosyncratic
construction),
even though  their impact on corpus coverage figures might actually be
rather greater.

\subsection{Ellipsis}

As described in \ref{ellipana}, considerable attention has been paid to 
elliptical sentences. The following constructions illustrate some
of the types of elliptical construction which CLARE is intended
to be able to analyse:

\begin{quotenr}
What about Smith?\\
How many?\\
Does Jones?\\
and Bill doesn't want to?\\
In Cambridge.\\
3 or 4.\\
\end{quotenr}

A variety of means are used to analyse such occurrences. Some are
regarded as syntactically incomplete sentences. Others are regarded
as syntactically complete phrases. Still others are regarded as
semi-idiomatic forms requiring special rules. All of them are interpreted
with respect to the current model of the context, in particular the immediately
preceding linguistic context.

It is worth noting that much of the attention in the linguistic literature
has focused on cases of ellipsis where the antecedent is to be found within
the same sentence. Examples of this include so called `gapping' and
`VP-deletion':

\begin{quotenr}
John likes fish, and Bill chips.\\
Joe likes Mary, and Bill does too.
\end{quotenr}

Our treatment of ellipsis will readily extend to examples
like this. However, the current version of the system has
some of the relevant rules placed in an unused rule group:
we found that in the corpora used for evaluation, and in the
sample applications we were building, the discourse versions
of these types of ellipsis were overwhelmingly more frequent.
The slight lack of coverage was more than compensated for by the
elimination of the spurious analyses all treatments of ellipsis
give rise to.

\subsection{Comparatives}

Compared to CLE, the grammar in CLARE intends to give a wider coverage
of comparatives. Two kinds of comparatives are distinguished: `compositional'
comparatives, where the meaning can be arrived at purely on the
basis of the sentence itself, and those involving some kind of ellipsis.
Examples of the two kinds are as follows:

\begin{quotenr}
Mary owns more dogs than Bill owns.\\
Mare owns more dogs than Bill owns cats.\\
Mary swims faster than Bill swims.\\
Mary owns a larger dog than Bill owns.\\
Mary looks older than Bill looks.\\

\medskip

Mary owns more dogs than Bill does.\\
Mary owns more dogs than Bill.\\
Mary swims faster than Bill does.\\
Mary swims faster than Bill.\\
Mary owns a larger dog than Bill.\\
Mary looks older than Bill does.\\
Mary looks older than Bill.
\end{quotenr}

The first group of sentences can all be analysed as containing
the appropriate type of gap in the part of the sentence
following the comparative marker (\verb!more/less-than!, \verb!as-as!).
The usual type of gap threading approach can be used to build
up the correct QLF. In the second group of sentences, however,
it is more accurate to regard what follows the comparative marker
as an elliptical sentence, and to build the appropriate QLF using
essentially the same technique as for ellipsis.

\subsection{NP postmodification}

During CLARE the CLE treatment of postmodification of NPs has changed,
although no external differences should be apparent to the user.
In the original treatment, a phrase like `every college in Cambridge',
or `every college that was designed by a bishop' was regarded as
having the following structure:
\begin{quote}
[every [college in Cambridge]]\\ \mbox{}
[every [college designed by a bishop]]
\end{quote}
This is the simplest structure that is consistent with the syntactic
properties found and which allows for the correct semantic
representation to be build up compositionally. However, it has several
disadvantages: firstly, it will make it difficult to extend this
analysis for other types of NP postmodification like
\begin{quote}
 [everyone in the room]\\ \mbox{}
 [the college, which was designed by a bishop, ....]
\end{quote}
since it is impossible to impose an analogous syntactic structure on
these (although they are not semantically identical). The most
important disadvantage, though, is that the analysis does not have
good properties from the point of view of robust parsing: for example,
in the case of parse failure, it would be nice to be able to first of
all identify all the `minimal' phrases that can be found, and then in
a later pass assemble these minimal phrases in semantically plausible
ways. But on this analysis, in order to combine `every college' with
`in Cambridge' we would first need to undo the phrase `every college'.

With a view to making it possible to make the various extensions
described, and to further work on robustness, we have changed the
analysis of these constructions so that they are of the form:
\begin{quote}
 [[every college][in Cambridge]]
\end{quote}
etc. The resulting semantic rules are more complex than before, but we
hope that this slight disadvantage will be outweighed by future
benefits.

\subsection{``Big PPs''}

An alteration in constituent structure similar in spirit to that described
in the previous section has also been made to the handling of prepositional
phrases. Previously, an NP or VP postmodified by a sequence of PPs
was analyzed in the traditional way, as illustrated by the 
following two examples:
\begin{quote}
[[people [in England]] [during 1991]]\\ \mbox{}
[He [[arrived [in England]] [during 1991]]]
\end{quote}
the NP or VP contained a succession of nested smaller constituents of
the same type, one for each modifier. It turns out, however, that there
are concrete advantages to be gained from adopting an alternate representation
where the sequence of PPs is regarded as a ``big PP'', a 
compound prepositional phrase in its own right. Thus the new analysis 
of the examples above is 
\begin{quote}
[people [[in England] [during 1991]]]\\ \mbox{}
[He [arrived [[in England] [during 1991]]]]
\end{quote}
The ``big PP'' rules are supplemented by a rule that allows a temporal NP
such as `today' or `this Tuesday' also to be treated as a PP. This implies
that expressions like
\begin{quote}
in London this week\\ \mbox{}
last year during the election campaign
\end{quote}
are also regarded as PPs.
These and other ``big PPs'' can thus appear on their own as elliptical
phrases, or be preposed to form expressions like
\begin{quote}
In London this week, several people said the same thing.
\end{quote}
These increases in coverage turn out to be quite significant in 
interactive query domains like the SLT translation project (section~\ref{slt}).
Another point is that introduction of the ``big PP' allows simple
definition of preference methods (Chapter~\ref{clarepref}) which
reward plausible combinations of PPs, in particular the very common
pair {\it from ... to}.

\subsection{Adverbial phrases}

The coverage of adverbial phrases has been substantially broadened in CLARE.
There are two main extensions. Firstly, a number of rules have been added
to cover various cases where sentences and verb-phrases function as adverbs,
as for example in the following:
\begin{quotenr}
John left England [to find a job].\\
John left [singing a song].\\
John slept [, I think].\\
\end{quotenr}
Secondly, the treatent of punctuation with regard to adverbial modification
has been properly systematized, by allowing commas to be used to demarcate
the boundaries of the phrase; in accordance with normal English usage,
commas can appear both before and after in medial position, only after in 
initial, and only before in final. This is illustrated in the next group
of examples:
\begin{quotenr}
[Quickly ,] John left.\\
John [, quickly ,] left.\\
John left [, quickly].\\
John [, Mary told me ,] has left.\\
John has left [, Mary told me].
\end{quotenr}
The implementation of these rules relies on the notional {\tt start\_\-marker}
and {\tt end\_\-marker} categories introduced at the beginning and end of
the utterance during lexical processing (see section~\ref{Phrasal-Parsing}).

\subsection{Idioms}

It is now possible to define multi-word lexical entries.  Thus an
entry like

lex([not,many],....)

is possible and will allow the sequence of words `not many' to be
analysed as a single item with the categories given in the entry.
Previously such combinations had to be recognized by putting in
special syntax rules, one for each such combination.  The addition of
the new mechanism has allowed us to simplify the grammar by removing
these rules.

The same mechanism is used by VEX in defining simple idioms, as
documented in the manual (see section 7.2.5 there).  A lexical entry
of the form

\index{{\tt idiom} lexical entries}

\begin{quote} {\tt
\begin{tabbing}

idiom(\= [\var{iword_1},...,\var{iword_i}], \\
      \> [\var{mword_1},...,\var{mword_j}], \\
      \> [\var{pair_1},...,\var{pair_k}])
\end{tabbing}}\end{quote}
defines the phrase \var{iword_1},...,\var{iword_i} (typically an idiom or an
abbreviation) as possibly equivalent to the phrase
\var{mword_1},...,\var{mword_j}
(its ``meaning'').

Some common fixed phrases and idioms have been added, although we are
still in the process of discovering from corpora which are the most
frequent and useful of such phrases.

\subsection{Dates, times and numbers}

The PRM application, as do many others, demands a comprehensive
treatment of dates. We have added the necessary special forms, entries
and rules to be able recognize dates in any of these common formats:

\begin{quote}
27/11/1991. \\
27 Nov 1991. \\
27th Nov 1991. \\
27th of Nov 1991. \\
27 Nov, 1991. \\
27th Nov, 1991. \\
27th of Nov, 1991. \\
Nov 27, 1991.
Nov 27  1991. \\
Nov 27th, 1991. \\
Nov 27th 1991. \\
Nov the 27th,  1991. \\
Nov the 27th  1991. \\
27 November 1991. \\
27th November 1991. \\
27th of November 1991. \\
27 November, 1991. \\
27th November, 1991. \\
27th of November, 1991. \\
November 27, 1991. \\
November 27  1991. \\
November 27th, 1991. \\
November 27th 1991. \\
November the 27th,  1991. \\
November the 27th  1991.
\end{quote}

If the year is omitted, it will be inferred to be the current year.
The year can also be abbreviated, for example, to 91.

There is also a fairly comprehensive coverage of the ``spelled-out''
forms of times and numbers. These have been added primarily for use
in the SLT translation project described in section~\ref{slt}, but
should be of general utility in domains where spoken-language
input is used. Examples follow of the types of construction for
which coverage exists.
\begin{quote}
six.\\
six o'clock.\\
six a m.\\
six o'clock a m.\\
six thirty.\\
six thirty a m.\\
six thirty five.\\
six oh five.\\
twenty two fifteen.\\
fourteen hundred hours.\\
thirty eight.\\
eight hundred.\\
eight hundred and ten.\\
nineteen ninety.\\
eight three. {\it (Spelled-out two-digit number)}\\
eight three five. {\it (Spelled-out three-digit number)}\\
eight three five seven. {\it (Spelled-out four-digit number)}
\end{quote}
The main problem with including constructions of this type in the grammar
is that single-digit phrases like ``two'' acquire numerous interpretions,
which without extra grammatical constraints can lead to an explosion in
the number of analyses produced for complex phrases. For example,
\begin{quotenr}
Six dogs.
\end{quotenr}
could potentially mean ``a half-dozen dogs'', ``six o'clock dogs'' 
(like ``six o'clock news''), or ``dogs with code six'' (like ``eight ball'').
In order to combat this tendency, analyses like the second and third ones
above in which a number is used as a pre-nominal modifier are currently
blocked by use of special features.

\section{Grammar Documentation Facilities}

CLARE incorporates a facility for documenting the intended behaviour
of individual grammatical rules. In addition, documentation fields for
each rule provide an index to relevant sentences in a small test
corpus which is designed to illustrate and provide a check on the
basic coverage of the grammar.

A \verb!doc! field within a grammar rule is used to associate
various types of information with the rule. The form of the field is
\begin{quote}
\begin{verbatim}
doc(FragmentString,CorpusIDList,CommentString)
\end{verbatim}
\end{quote}
Each documentation record contains a fragment of the
language covered by the rule, a number of corpus sentence identifiers,
and a brief comment describing the intended behaviour of the rule.

Rule name identifiers are the same as those used in the relevant
grammar file. The CLE command \verb!.dr! (see the software manual)
displays the documentation for a particular rule given its identifier.
The language fragment is a string consisting of a sentence parseable by
the CLE using the core lexicon. The part of this sentence covered by
the relevant rule is enclosed in square brackets.  A list of corpus
sentence identifiers is provided which indicate which sentences in the
test corpus are relevant to test or illustrate the behaviour of the
rule. The comment string provides a brief description of the intended
behaviour and function of the rule. For example, calling \verb!.dr! on
the rule \verb!syn s_np_vp_Normal! produces the following output:
\begin{verbatim}
(describe rule specified by `s_np_vp_Normal')

syn rule s_np_vp_Normal:
  Describes fragment:
    [Wren slept]
  Corpus sentences:
    d1: Wren designed a library.
    d3: The man slept.
    d4: I wonder who you like.
    e1: Wren has.
    e3: Who has.
    q2: What has Wren designed?
    q4: Who has designed a library?
  Comment:
    Covers most types of finite and subjunctive, uninverted clause
\end{verbatim}
The bracketed language fragment is shown and the test corpus sentences
indexed by the identifiers are listed, and the brief explanation of
the rule is shown.

The test corpus consists of a set of sentences parseable by the CLE
using the core lexicon. Each sentence has an identifier consisting of
one or two (minimally mnemonic) letters followed by a number (a =
adjectives, c = coordination, cp = comparatives, d = declaratives /
verb phrases, e = ellipsis, m = measure phrases / dates, n = noun
phrases, p = prepositional phrases, q = questions, i = imperatives, r
= relative clauses). The CLE command for testing a rule on the corpus
is \verb!.tr [<ruletype>] <rulename>!. This command invokes the CLE on
the language fragment in the documentation field and on the test corpus
sentences indexed there.

To date, many of the syntactic rules in the current grammar have been
documented and a test corpus constructed illustrating their coverage.
The description of the rules has been kept as informal and consistent
as possible.  Unfamiliar terms should be explained in a compatible
fashion in Quirk {\it et al} (1985).

\section{Conditions of Coverage Evaluation}

In order to evaluate progress during the CLARE project, CLARE-0 (the
version of the CLE at the end of the Nattie project), CLARE-1, CLARE-2
and CLARE-3 (the final release) were each run on parts of the LOB
(Lancaster-Oslo-Bergen) corpus, a mixture of many different kinds of
printed material. This corpus was selected as being the closest
available approximation to a general or ``unbiased'' set of sentences.
In addition to the LOB data, CLARE-2 was run on four other text sets;
this process is described in section \ref{covtexttype} below.

Because of the large amount of data involved, a client/server control
regime was adopted, with one client process distributing sentences to
CLARE server processes on several different machines (using the TCP
facilities provided with Quintus Prolog 3) and automatically handling
Prolog or UNIX error conditions by restarting servers when
necessary.\footnote{The code implementing this is not an official part
of the final CLARE release, but is included with it in files named
{\tt tcp*.pl}.}

The sentences selected for evaluation were the members of the text
sets that contained only ``legible'' characters from CLARE's point
of view: that is, alphanumerics, hyphens, commas and, in
sentence-final position only, full stops, question marks and
exclamation marks. Although the mechanism of CLARE is capable of
processing sentences containing other characters (see Chapter
\ref{clarelexproc}), its lexicon and grammar have not been
significantly extended in this direction. The ``legible'' sets were
further restricted by including only sentences of twenty words or
fewer, because longer sentences stand little chance of being
successfully processed by the system.

\subsection{Limiting the vocabulary}

For some of the experiments, a further subset of sentences was taken:
those consisting only of words within the coverage of the CLARE-2 core
and PRM lexicons. For the LOB corpus, over 78\% of the word
occurrences were within coverage. However, this meant that only around
9\% of the sentences consisted only of words that were within
coverage.

A plot of the distribution of sentence lengths in the LOB corpus is
given in figure \ref{loblenfig}. For each length between 1 and
100, the graph shows the number of sentences of that length in the
whole LOB corpus, in the ``legible'' subset, and in the
vocabulary-limited subset.

\setlength{\unitlength}{1mm}
\thicklines
\begin{figure}
\begin{picture}(130,100)
\put(10,5){\vector(1,0){120}}
\put(10,5){\vector(0,1){95}}
\put(0,  4.0){  0
}\put(  9.0,  5.0){\line(1,0){2}}
\put(0, 23.0){300
}\put(  9.0, 24.0){\line(1,0){2}}
\put(0, 42.0){600
}\put(  9.0, 43.0){\line(1,0){2}}
\put(0, 61.0){900
}\put(  9.0, 62.0){\line(1,0){2}}
\put(0, 80.0){1200
}\put(  9.0, 81.0){\line(1,0){2}}
\put(0, 99.0){1500
}\put(  9.0,100.0){\line(1,0){2}}
\put( 65.0, 95.0){$\cdot$ All sentences
}
\put( 65.0, 90.0){{\tiny $+$} Sentences with ``nice'' characters
}
\put( 65.0, 85.0){{\tiny $\odot$} Sentences within core+PRM vocabulary 
}
\put( 11.2, 37.6){$\cdot$}
\put( 11.3, 38.6){$\cdot$}
\put( 11.3, 39.6){$\cdot$}
\put( 11.4, 40.6){$\cdot$}
\put( 11.4, 41.6){$\cdot$}
\put( 11.5, 42.6){$\cdot$}
\put( 11.5, 43.6){$\cdot$}
\put( 11.6, 44.6){$\cdot$}
\put( 11.6, 45.6){$\cdot$}
\put( 11.6, 46.6){$\cdot$}
\put( 11.7, 47.6){$\cdot$}
\put( 11.7, 48.6){$\cdot$}
\put( 11.8, 49.6){$\cdot$}
\put( 11.8, 50.6){$\cdot$}
\put( 11.9, 51.6){$\cdot$}
\put( 11.9, 52.6){$\cdot$}
\put( 12.0, 53.6){$\cdot$}
\put( 12.0, 54.6){$\cdot$}
\put( 12.1, 55.6){$\cdot$}
\put( 12.1, 56.6){$\cdot$}
\put( 12.1, 57.6){$\cdot$}
\put( 12.2, 58.6){$\cdot$}
\put( 12.2, 59.6){$\cdot$}
\put( 12.3, 60.6){$\cdot$}
\put( 12.3, 61.6){$\cdot$}
\put( 12.4, 62.6){$\cdot$}
\put( 12.4, 63.6){$\cdot$}
\put( 11.3, 30.6){{\tiny $+$}}
\put( 11.3, 31.6){{\tiny $+$}}
\put( 11.4, 32.6){{\tiny $+$}}
\put( 11.4, 33.6){{\tiny $+$}}
\put( 11.5, 34.6){{\tiny $+$}}
\put( 11.5, 35.6){{\tiny $+$}}
\put( 11.6, 36.6){{\tiny $+$}}
\put( 11.6, 37.6){{\tiny $+$}}
\put( 11.7, 38.6){{\tiny $+$}}
\put( 11.7, 39.6){{\tiny $+$}}
\put( 11.8, 40.6){{\tiny $+$}}
\put( 11.8, 41.6){{\tiny $+$}}
\put( 11.9, 42.6){{\tiny $+$}}
\put( 11.9, 43.6){{\tiny $+$}}
\put( 12.0, 44.6){{\tiny $+$}}
\put( 12.0, 45.6){{\tiny $+$}}
\put( 12.1, 46.6){{\tiny $+$}}
\put( 12.1, 47.6){{\tiny $+$}}
\put( 12.2, 48.6){{\tiny $+$}}
\put( 12.2, 49.6){{\tiny $+$}}
\put( 12.3, 50.6){{\tiny $+$}}
\put( 12.3, 51.6){{\tiny $+$}}
\put( 12.4, 52.6){{\tiny $+$}}
\put( 11.5, 14.6){{\tiny $\odot$}}
\put( 11.9, 15.6){{\tiny $\odot$}}
\put( 12.2, 16.5){{\tiny $\odot$}}
\put( 12.5, 17.4){{\tiny $\odot$}}
\put( 12.5, 64.3){$\cdot$}
\put( 12.6, 65.3){$\cdot$}
\put( 12.6, 66.3){$\cdot$}
\put( 12.7, 67.3){$\cdot$}
\put( 12.8, 68.3){$\cdot$}
\put( 12.9, 69.3){$\cdot$}
\put( 13.0, 70.3){$\cdot$}
\put( 13.0, 71.3){$\cdot$}
\put( 13.1, 72.3){$\cdot$}
\put( 13.2, 73.3){$\cdot$}
\put( 13.3, 74.3){$\cdot$}
\put( 13.4, 75.3){$\cdot$}
\put( 13.4, 76.3){$\cdot$}
\put( 13.5, 77.3){$\cdot$}
\put( 13.6, 78.3){$\cdot$}
\put( 12.5, 53.7){{\tiny $+$}}
\put( 12.7, 54.7){{\tiny $+$}}
\put( 12.8, 55.7){{\tiny $+$}}
\put( 12.9, 56.7){{\tiny $+$}}
\put( 13.0, 57.7){{\tiny $+$}}
\put( 13.2, 58.7){{\tiny $+$}}
\put( 13.3, 59.7){{\tiny $+$}}
\put( 13.4, 60.7){{\tiny $+$}}
\put( 13.6, 61.7){{\tiny $+$}}
\put( 12.7, 18.0){{\tiny $\odot$}}
\put( 12.9, 19.0){{\tiny $\odot$}}
\put( 13.2, 19.9){{\tiny $\odot$}}
\put( 13.4, 20.9){{\tiny $\odot$}}
\put( 13.7, 21.9){{\tiny $\odot$}}
\put( 13.9, 79.2){$\cdot$}
\put( 14.2, 80.2){$\cdot$}
\put( 14.5, 81.1){$\cdot$}
\put( 14.8, 82.1){$\cdot$}
\put( 14.3, 61.2){{\tiny $+$}}
\put( 15.0, 60.6){{\tiny $+$}}
\put( 14.6, 21.6){{\tiny $\odot$}}
\put( 15.0, 83.0){$\cdot$}
\put( 15.3, 84.0){$\cdot$}
\put( 15.5, 85.0){$\cdot$}
\put( 15.7, 86.0){$\cdot$}
\put( 16.0, 86.9){$\cdot$}
\put( 15.2, 61.7){{\tiny $+$}}
\put( 15.6, 62.6){{\tiny $+$}}
\put( 16.0, 63.6){{\tiny $+$}}
\put( 15.5, 22.3){{\tiny $\odot$}}
\put( 16.2, 23.0){{\tiny $\odot$}}
\put( 13.0,  0.0){  5}
\put( 16.0,  4.0){\line(0,1){2}}
\put( 16.3, 88.0){$\cdot$}
\put( 16.6, 89.0){$\cdot$}
\put( 16.9, 89.9){$\cdot$}
\put( 17.2, 90.9){$\cdot$}
\put( 16.5, 64.4){{\tiny $+$}}
\put( 17.1, 65.3){{\tiny $+$}}
\put( 16.3, 21.8){{\tiny $\odot$}}
\put( 16.6, 20.9){{\tiny $\odot$}}
\put( 16.8, 19.9){{\tiny $\odot$}}
\put( 17.1, 19.0){{\tiny $\odot$}}
\put( 17.6, 91.7){$\cdot$}
\put( 17.9, 92.6){$\cdot$}
\put( 18.3, 93.6){$\cdot$}
\put( 17.8, 66.3){{\tiny $+$}}
\put( 18.5, 67.0){{\tiny $+$}}
\put( 17.7, 17.8){{\tiny $\odot$}}
\put( 18.3, 17.0){{\tiny $\odot$}}
\put( 19.1, 94.5){$\cdot$}
\put( 19.9, 95.2){$\cdot$}
\put( 19.3, 67.4){{\tiny $+$}}
\put( 19.0, 16.0){{\tiny $\odot$}}
\put( 19.7, 15.2){{\tiny $\odot$}}
\put( 20.1, 94.1){$\cdot$}
\put( 20.6, 93.2){$\cdot$}
\put( 20.1, 66.7){{\tiny $+$}}
\put( 20.7, 65.9){{\tiny $+$}}
\put( 20.2, 14.5){{\tiny $\odot$}}
\put( 20.8, 13.7){{\tiny $\odot$}}
\put( 21.1, 93.9){$\cdot$}
\put( 21.4, 94.8){$\cdot$}
\put( 21.7, 95.8){$\cdot$}
\put( 22.0, 96.7){$\cdot$}
\put( 21.1, 66.6){{\tiny $+$}}
\put( 21.3, 67.6){{\tiny $+$}}
\put( 21.6, 68.6){{\tiny $+$}}
\put( 21.8, 69.5){{\tiny $+$}}
\put( 22.1, 70.5){{\tiny $+$}}
\put( 21.3, 12.8){{\tiny $\odot$}}
\put( 21.8, 11.9){{\tiny $\odot$}}
\put( 22.2, 11.0){{\tiny $\odot$}}
\put( 19.0,  0.0){ 10}
\put( 22.0,  4.0){\line(0,1){2}}
\put( 23.0, 96.8){$\cdot$}
\put( 22.8, 69.6){{\tiny $+$}}
\put( 22.6, 10.7){{\tiny $\odot$}}
\put( 23.3,  9.9){{\tiny $\odot$}}
\put( 23.7, 96.0){$\cdot$}
\put( 24.2, 95.1){$\cdot$}
\put( 24.1, 69.8){{\tiny $+$}}
\put( 24.2,  9.8){{\tiny $\odot$}}
\put( 25.4, 94.7){$\cdot$}
\put( 24.8, 68.9){{\tiny $+$}}
\put( 25.2, 68.0){{\tiny $+$}}
\put( 25.7, 67.1){{\tiny $+$}}
\put( 25.3,  9.4){{\tiny $\odot$}}
\put( 26.0, 95.6){$\cdot$}
\put( 26.5, 96.5){$\cdot$}
\put( 26.9, 97.4){$\cdot$}
\put( 26.0, 68.2){{\tiny $+$}}
\put( 26.4, 69.1){{\tiny $+$}}
\put( 26.8, 70.0){{\tiny $+$}}
\put( 26.2,  8.5){{\tiny $\odot$}}
\put( 26.8,  7.7){{\tiny $\odot$}}
\put( 27.0, 96.2){$\cdot$}
\put( 27.2, 95.3){$\cdot$}
\put( 27.4, 94.3){$\cdot$}
\put( 27.6, 93.3){$\cdot$}
\put( 27.8, 92.3){$\cdot$}
\put( 28.0, 91.3){$\cdot$}
\put( 27.1, 69.1){{\tiny $+$}}
\put( 27.3, 68.2){{\tiny $+$}}
\put( 27.6, 67.2){{\tiny $+$}}
\put( 27.8, 66.2){{\tiny $+$}}
\put( 28.1, 65.3){{\tiny $+$}}
\put( 27.8,  7.4){{\tiny $\odot$}}
\put( 25.0,  0.0){ 15}
\put( 28.0,  4.0){\line(0,1){2}}
\put( 28.2, 92.4){$\cdot$}
\put( 28.3, 93.4){$\cdot$}
\put( 28.5, 94.4){$\cdot$}
\put( 28.7, 95.4){$\cdot$}
\put( 28.8, 96.4){$\cdot$}
\put( 29.0, 97.4){$\cdot$}
\put( 29.2, 98.4){$\cdot$}
\put( 28.4, 66.5){{\tiny $+$}}
\put( 28.8, 67.4){{\tiny $+$}}
\put( 29.2, 68.3){{\tiny $+$}}
\put( 29.0,  7.1){{\tiny $\odot$}}
\put( 29.5, 97.7){$\cdot$}
\put( 29.8, 96.7){$\cdot$}
\put( 30.1, 95.7){$\cdot$}
\put( 30.4, 94.8){$\cdot$}
\put( 29.7, 67.5){{\tiny $+$}}
\put( 30.1, 66.6){{\tiny $+$}}
\put( 30.6, 65.7){{\tiny $+$}}
\put( 30.2,  7.1){{\tiny $\odot$}}
\put( 31.3, 94.2){$\cdot$}
\put( 30.9, 65.2){{\tiny $+$}}
\put( 31.4, 64.3){{\tiny $+$}}
\put( 31.8, 63.4){{\tiny $+$}}
\put( 31.2,  6.6){{\tiny $\odot$}}
\put( 31.8, 93.1){$\cdot$}
\put( 32.0, 92.1){$\cdot$}
\put( 32.2, 91.1){$\cdot$}
\put( 32.5, 90.1){$\cdot$}
\put( 32.7, 89.2){$\cdot$}
\put( 32.9, 88.2){$\cdot$}
\put( 32.2, 63.0){{\tiny $+$}}
\put( 32.7, 62.2){{\tiny $+$}}
\put( 32.6,  6.3){{\tiny $\odot$}}
\put( 33.0, 87.6){$\cdot$}
\put( 33.3, 86.6){$\cdot$}
\put( 33.5, 85.6){$\cdot$}
\put( 33.8, 84.7){$\cdot$}
\put( 34.0, 83.7){$\cdot$}
\put( 33.1, 61.1){{\tiny $+$}}
\put( 33.4, 60.2){{\tiny $+$}}
\put( 33.7, 59.2){{\tiny $+$}}
\put( 34.0, 58.3){{\tiny $+$}}
\put( 33.8,  6.1){{\tiny $\odot$}}
\put( 31.0,  0.0){ 20}
\put( 34.0,  4.0){\line(0,1){2}}
\put( 34.9, 83.3){$\cdot$}
\put( 34.3, 57.4){{\tiny $+$}}
\put( 34.6, 56.5){{\tiny $+$}}
\put( 34.9, 55.5){{\tiny $+$}}
\put( 35.2, 54.6){{\tiny $+$}}
\put( 35.0,  5.8){{\tiny $\odot$}}
\put( 35.5, 82.1){$\cdot$}
\put( 35.7, 81.2){$\cdot$}
\put( 36.0, 80.2){$\cdot$}
\put( 36.2, 79.2){$\cdot$}
\put( 36.5, 78.3){$\cdot$}
\put( 35.5, 53.5){{\tiny $+$}}
\put( 35.8, 52.5){{\tiny $+$}}
\put( 36.1, 51.6){{\tiny $+$}}
\put( 36.3, 50.6){{\tiny $+$}}
\put( 36.2,  5.7){{\tiny $\odot$}}
\put( 36.6, 77.6){$\cdot$}
\put( 36.7, 76.6){$\cdot$}
\put( 36.9, 75.6){$\cdot$}
\put( 37.1, 74.7){$\cdot$}
\put( 37.3, 73.7){$\cdot$}
\put( 37.4, 72.7){$\cdot$}
\put( 37.6, 71.7){$\cdot$}
\put( 37.3, 49.9){{\tiny $+$}}
\put( 37.3,  6.0){{\tiny $\odot$}}
\put( 38.1, 72.6){$\cdot$}
\put( 38.5, 73.5){$\cdot$}
\put( 39.0, 74.4){$\cdot$}
\put( 38.1, 48.9){{\tiny $+$}}
\put( 38.6, 48.0){{\tiny $+$}}
\put( 38.6,  5.9){{\tiny $\odot$}}
\put( 39.0, 73.1){$\cdot$}
\put( 39.2, 72.1){$\cdot$}
\put( 39.3, 71.1){$\cdot$}
\put( 39.5, 70.1){$\cdot$}
\put( 39.7, 69.1){$\cdot$}
\put( 39.9, 68.1){$\cdot$}
\put( 40.1, 67.2){$\cdot$}
\put( 39.1, 46.7){{\tiny $+$}}
\put( 39.3, 45.7){{\tiny $+$}}
\put( 39.6, 44.7){{\tiny $+$}}
\put( 39.8, 43.8){{\tiny $+$}}
\put( 40.1, 42.8){{\tiny $+$}}
\put( 39.8,  5.6){{\tiny $\odot$}}
\put( 37.0,  0.0){ 25}
\put( 40.0,  4.0){\line(0,1){2}}
\put( 40.2, 66.6){$\cdot$}
\put( 40.5, 65.6){$\cdot$}
\put( 40.7, 64.7){$\cdot$}
\put( 41.0, 63.7){$\cdot$}
\put( 41.2, 62.7){$\cdot$}
\put( 40.5, 42.3){{\tiny $+$}}
\put( 41.0, 41.4){{\tiny $+$}}
\put( 41.0,  5.5){{\tiny $\odot$}}
\put( 41.8, 61.9){$\cdot$}
\put( 42.4, 61.1){$\cdot$}
\put( 41.6, 40.1){{\tiny $+$}}
\put( 42.0, 39.1){{\tiny $+$}}
\put( 42.4, 38.2){{\tiny $+$}}
\put( 42.2,  5.4){{\tiny $\odot$}}
\put( 42.7, 60.2){$\cdot$}
\put( 43.0, 59.2){$\cdot$}
\put( 43.3, 58.3){$\cdot$}
\put( 43.7, 57.3){$\cdot$}
\put( 42.8, 37.2){{\tiny $+$}}
\put( 43.3, 36.3){{\tiny $+$}}
\put( 43.7, 35.4){{\tiny $+$}}
\put( 43.4,  5.3){{\tiny $\odot$}}
\put( 44.3, 56.8){$\cdot$}
\put( 44.9, 56.0){$\cdot$}
\put( 44.2, 34.8){{\tiny $+$}}
\put( 44.7, 34.0){{\tiny $+$}}
\put( 44.6,  5.1){{\tiny $\odot$}}
\put( 45.1, 55.2){$\cdot$}
\put( 45.3, 54.2){$\cdot$}
\put( 45.6, 53.3){$\cdot$}
\put( 45.9, 52.3){$\cdot$}
\put( 46.1, 51.4){$\cdot$}
\put( 45.3, 33.1){{\tiny $+$}}
\put( 45.9, 32.3){{\tiny $+$}}
\put( 43.0,  0.0){ 30}
\put( 46.0,  4.0){\line(0,1){2}}
\put( 46.3, 50.9){$\cdot$}
\put( 46.7, 49.9){$\cdot$}
\put( 47.0, 49.0){$\cdot$}
\put( 46.4, 31.1){{\tiny $+$}}
\put( 46.8, 30.2){{\tiny $+$}}
\put( 47.2, 29.3){{\tiny $+$}}
\put( 47.7, 47.7){$\cdot$}
\put( 48.2, 46.8){$\cdot$}
\put( 48.6, 45.9){$\cdot$}
\put( 47.7, 28.5){{\tiny $+$}}
\put( 48.2, 27.6){{\tiny $+$}}
\put( 48.8, 45.4){$\cdot$}
\put( 49.2, 44.5){$\cdot$}
\put( 49.6, 43.6){$\cdot$}
\put( 48.8, 26.4){{\tiny $+$}}
\put( 49.1, 25.4){{\tiny $+$}}
\put( 49.5, 24.5){{\tiny $+$}}
\put( 50.4, 44.3){$\cdot$}
\put( 51.1, 44.9){$\cdot$}
\put( 50.2, 25.0){{\tiny $+$}}
\put( 50.9, 25.7){{\tiny $+$}}
\put( 51.0, 43.7){$\cdot$}
\put( 51.2, 42.7){$\cdot$}
\put( 51.4, 41.7){$\cdot$}
\put( 51.6, 40.7){$\cdot$}
\put( 51.8, 39.7){$\cdot$}
\put( 52.0, 38.8){$\cdot$}
\put( 51.1, 24.7){{\tiny $+$}}
\put( 51.5, 23.8){{\tiny $+$}}
\put( 51.8, 22.8){{\tiny $+$}}
\put( 49.0,  0.0){ 35}
\put( 52.0,  4.0){\line(0,1){2}}
\put( 52.3, 37.8){$\cdot$}
\put( 52.7, 36.9){$\cdot$}
\put( 53.0, 35.9){$\cdot$}
\put( 53.4, 35.0){$\cdot$}
\put( 52.5, 21.6){{\tiny $+$}}
\put( 53.1, 20.7){{\tiny $+$}}
\put( 54.0, 34.8){$\cdot$}
\put( 54.7, 34.2){$\cdot$}
\put( 54.2, 20.5){{\tiny $+$}}
\put( 54.8, 33.5){$\cdot$}
\put( 55.2, 32.6){$\cdot$}
\put( 55.6, 31.7){$\cdot$}
\put( 54.8, 19.5){{\tiny $+$}}
\put( 55.1, 18.6){{\tiny $+$}}
\put( 55.5, 17.6){{\tiny $+$}}
\put( 55.9, 30.7){$\cdot$}
\put( 56.3, 29.8){$\cdot$}
\put( 56.6, 28.9){$\cdot$}
\put( 56.4, 16.7){{\tiny $+$}}
\put( 57.5, 29.2){$\cdot$}
\put( 58.2, 29.9){$\cdot$}
\put( 57.8, 16.3){{\tiny $+$}}
\put( 55.0,  0.0){ 40}
\put( 58.0,  4.0){\line(0,1){2}}
\put( 58.6, 28.9){$\cdot$}
\put( 59.2, 28.1){$\cdot$}
\put( 58.7, 15.5){{\tiny $+$}}
\put( 59.4, 14.8){{\tiny $+$}}
\put( 59.4, 27.1){$\cdot$}
\put( 59.7, 26.1){$\cdot$}
\put( 59.9, 25.1){$\cdot$}
\put( 60.1, 24.2){$\cdot$}
\put( 60.3, 23.2){$\cdot$}
\put( 59.6, 14.2){{\tiny $+$}}
\put( 60.1, 13.3){{\tiny $+$}}
\put( 60.5, 12.4){{\tiny $+$}}
\put( 61.4, 22.9){$\cdot$}
\put( 61.3, 13.1){{\tiny $+$}}
\put( 62.0, 22.0){$\cdot$}
\put( 62.5, 21.1){$\cdot$}
\put( 62.9, 20.2){$\cdot$}
\put( 62.1, 12.4){{\tiny $+$}}
\put( 62.6, 11.5){{\tiny $+$}}
\put( 63.8, 20.3){$\cdot$}
\put( 63.6, 10.6){{\tiny $+$}}
\put( 61.0,  0.0){ 45}
\put( 64.0,  4.0){\line(0,1){2}}
\put( 64.5, 19.5){$\cdot$}
\put( 65.1, 18.6){$\cdot$}
\put( 65.0, 10.4){{\tiny $+$}}
\put( 66.2, 18.3){$\cdot$}
\put( 66.2, 10.1){{\tiny $+$}}
\put( 66.9, 17.4){$\cdot$}
\put( 67.4, 16.5){$\cdot$}
\put( 67.3,  9.6){{\tiny $+$}}
\put( 68.5, 16.7){$\cdot$}
\put( 68.6,  9.3){{\tiny $+$}}
\put( 69.8, 16.5){$\cdot$}
\put( 69.8,  9.2){{\tiny $+$}}
\put( 67.0,  0.0){ 50}
\put( 70.0,  4.0){\line(0,1){2}}
\put( 70.5, 15.6){$\cdot$}
\put( 71.1, 14.7){$\cdot$}
\put( 70.8,  8.5){{\tiny $+$}}
\put( 71.5,  7.9){{\tiny $+$}}
\put( 71.8, 13.8){$\cdot$}
\put( 72.4, 12.9){$\cdot$}
\put( 72.1,  7.8){{\tiny $+$}}
\put( 73.4, 12.7){$\cdot$}
\put( 73.4,  7.6){{\tiny $+$}}
\put( 74.6, 12.5){$\cdot$}
\put( 74.5,  7.1){{\tiny $+$}}
\put( 75.7, 12.0){$\cdot$}
\put( 75.8,  6.9){{\tiny $+$}}
\put( 73.0,  0.0){ 55}
\put( 76.0,  4.0){\line(0,1){2}}
\put( 76.6, 11.0){$\cdot$}
\put( 77.2, 10.2){$\cdot$}
\put( 76.8,  6.2){{\tiny $+$}}
\put( 77.6,  5.6){{\tiny $+$}}
\put( 78.0, 10.8){$\cdot$}
\put( 78.2,  6.0){{\tiny $+$}}
\put( 79.2, 10.4){$\cdot$}
\put( 80.0,  9.8){$\cdot$}
\put( 79.4,  6.4){{\tiny $+$}}
\put( 80.6,  9.9){$\cdot$}
\put( 80.5,  6.1){{\tiny $+$}}
\put( 81.7,  9.4){$\cdot$}
\put( 81.8,  6.0){{\tiny $+$}}
\put( 79.0,  0.0){ 60}
\put( 82.0,  4.0){\line(0,1){2}}
\put( 82.7,  8.6){$\cdot$}
\put( 83.5,  8.0){$\cdot$}
\put( 83.0,  5.8){{\tiny $+$}}
\put( 84.1,  8.6){$\cdot$}
\put( 84.2,  5.8){{\tiny $+$}}
\put( 85.4,  8.5){$\cdot$}
\put( 85.4,  5.8){{\tiny $+$}}
\put( 86.5,  8.8){$\cdot$}
\put( 86.6,  5.9){{\tiny $+$}}
\put( 87.8,  8.8){$\cdot$}
\put( 87.8,  6.1){{\tiny $+$}}
\put( 85.0,  0.0){ 65}
\put( 88.0,  4.0){\line(0,1){2}}
\put( 88.7,  8.0){$\cdot$}
\put( 89.4,  7.3){$\cdot$}
\put( 88.9,  5.7){{\tiny $+$}}
\put( 90.2,  7.3){$\cdot$}
\put( 90.2,  5.5){{\tiny $+$}}
\put( 91.3,  7.7){$\cdot$}
\put( 91.4,  5.4){{\tiny $+$}}
\put( 92.4,  7.3){$\cdot$}
\put( 92.6,  5.2){{\tiny $+$}}
\put( 93.8,  6.9){$\cdot$}
\put( 93.8,  5.2){{\tiny $+$}}
\put( 91.0,  0.0){ 70}
\put( 94.0,  4.0){\line(0,1){2}}
\put( 95.0,  6.8){$\cdot$}
\put( 95.0,  5.3){{\tiny $+$}}
\put( 96.2,  6.6){$\cdot$}
\put( 96.2,  5.1){{\tiny $+$}}
\put( 97.4,  6.3){$\cdot$}
\put( 97.4,  5.1){{\tiny $+$}}
\put( 98.4,  6.8){$\cdot$}
\put( 98.6,  5.2){{\tiny $+$}}
\put( 99.8,  7.1){$\cdot$}
\put( 99.8,  5.3){{\tiny $+$}}
\put( 97.0,  0.0){ 75}
\put(100.0,  4.0){\line(0,1){2}}
\put(101.0,  6.8){$\cdot$}
\put(101.0,  5.2){{\tiny $+$}}
\put(102.1,  6.3){$\cdot$}
\put(102.2,  5.1){{\tiny $+$}}
\put(103.3,  6.7){$\cdot$}
\put(103.4,  5.2){{\tiny $+$}}
\put(104.4,  6.3){$\cdot$}
\put(104.6,  5.0){{\tiny $+$}}
\put(105.8,  5.8){$\cdot$}
\put(105.8,  5.0){{\tiny $+$}}
\put(103.0,  0.0){ 80}
\put(106.0,  4.0){\line(0,1){2}}
\put(107.0,  5.9){$\cdot$}
\put(107.0,  5.1){{\tiny $+$}}
\put(108.2,  5.9){$\cdot$}
\put(108.2,  5.1){{\tiny $+$}}
\put(109.4,  5.7){$\cdot$}
\put(109.4,  5.0){{\tiny $+$}}
\put(110.6,  5.9){$\cdot$}
\put(110.6,  5.1){{\tiny $+$}}
\put(111.8,  5.9){$\cdot$}
\put(111.8,  5.1){{\tiny $+$}}
\put(109.0,  0.0){ 85}
\put(112.0,  4.0){\line(0,1){2}}
\put(112.9,  5.5){$\cdot$}
\put(113.0,  5.1){{\tiny $+$}}
\put(114.1,  5.7){$\cdot$}
\put(114.2,  5.0){{\tiny $+$}}
\put(115.4,  5.7){$\cdot$}
\put(115.4,  5.2){{\tiny $+$}}
\put(116.6,  5.6){$\cdot$}
\put(116.6,  5.1){{\tiny $+$}}
\put(117.8,  5.4){$\cdot$}
\put(117.8,  5.1){{\tiny $+$}}
\put(115.0,  0.0){ 90}
\put(118.0,  4.0){\line(0,1){2}}
\put(119.0,  5.6){$\cdot$}
\put(119.0,  5.2){{\tiny $+$}}
\put(120.2,  5.6){$\cdot$}
\put(120.2,  5.0){{\tiny $+$}}
\put(121.4,  5.6){$\cdot$}
\put(121.4,  5.1){{\tiny $+$}}
\put(122.6,  5.4){$\cdot$}
\put(122.6,  5.0){{\tiny $+$}}
\put(123.8,  5.5){$\cdot$}
\put(123.8,  5.0){{\tiny $+$}}
\put(121.0,  0.0){ 95}
\put(124.0,  4.0){\line(0,1){2}}
\put(125.0,  5.5){$\cdot$}
\put(125.0,  5.0){{\tiny $+$}}
\put(126.2,  5.4){$\cdot$}
\put(126.2,  5.0){{\tiny $+$}}
\put(127.4,  5.3){$\cdot$}
\put(127.4,  5.0){{\tiny $+$}}
\put(128.6,  5.3){$\cdot$}
\put(128.6,  5.0){{\tiny $+$}}
\put(129.8,  5.3){$\cdot$}
\put(129.8,  5.1){{\tiny $+$}}
\put(127.0,  0.0){100}
\put(130.0,  4.0){\line(0,1){2}}
\end{picture}
\caption{Sentence length distribution for LOB corpus}
\label{loblenfig}
\end{figure}
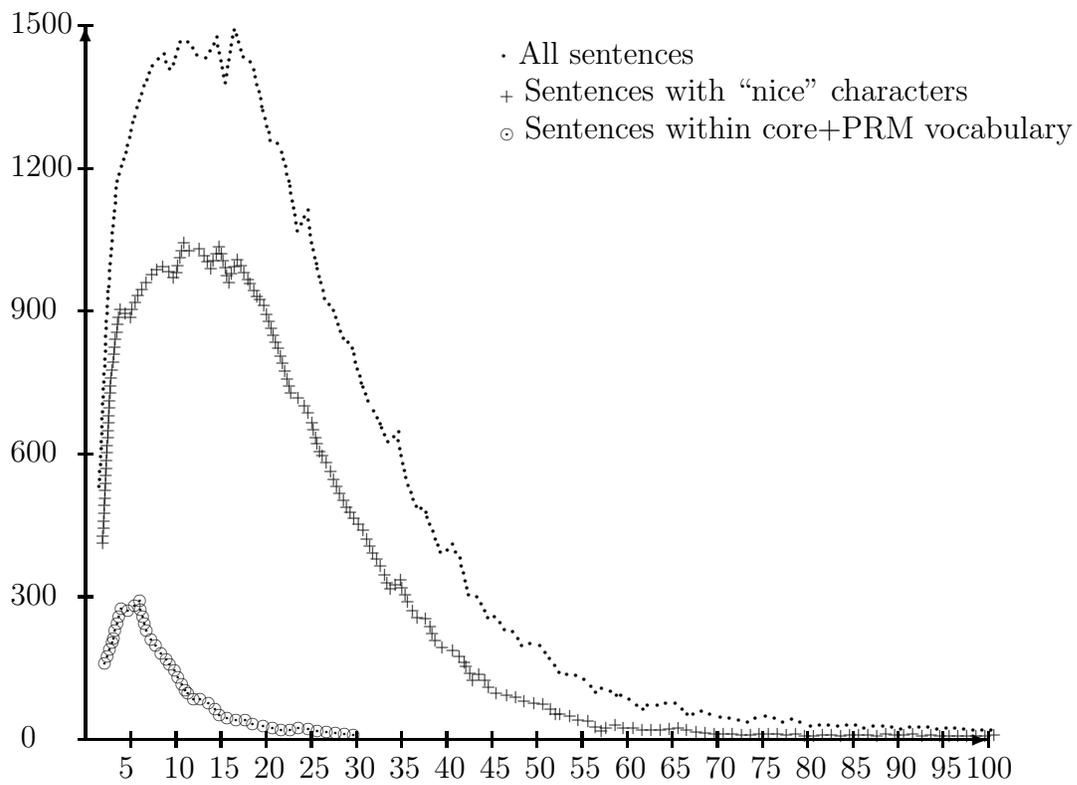

When vocabulary-unlimited text sets were processed, an interface to
the MRC Psycholinguistic Database (Coltheart, 1981; Alshawi, 1992,
p125) was used for words not in the core or PRM lexicons.  It was also
used to provide additional definitions for major categories (parts of
speech) not appearing in the main lexicons. For example, ``burn'' is
defined in the core lexicon as a verb, but in the MRC database as both
a verb and a noun. When this word was used, a noun definition was
created on the basis of the MRC data. Proper name inference (Alshawi
et al, p118) was also used to define capitalized words as potentially
either noun phrases or count nouns.

CLARE-2 and earlier versions interface to the earlier form of
Coltheart's database.  The only useful information in this database is
for major categories of words. Thus the fact that a word can be a verb
is recorded, but distinctions between different kinds of verb are
absent.  For this reason, the derived lexical entries are fairly
crude, representing only the most common behaviours, and many errors
and omissions occur as a result; most other machine-readable lexicons
provide significantly richer information. CLARE-3 uses the version of
the MRC database released by Oxford University Press under the title
of the OUP Psycholinguistic Database (Quinlan, 1992). In this version,
intransitive and transitive verbs are distinguished, as are mass and
count nouns; this led to a noticeable improvement in the accuracy of
QLFs derived using the external lexicon.\footnote{At the time of
writing, the OUP database is available with a Macintosh interface at
\pounds 175 from Oxford University Press. The CLARE-3 release is also
accompanied by a less developed interface to the rather richer
WordNet database (Miller, 1990), which is available by anonymous ftp
from {\tt clarity.princeton.edu}.}

\begin{figure}
\setlength{\unitlength}{0.9mm}
\begin{picture}(170,80){\small
\put(20,46){\vector(1,0){5}}
\put(0,54){\makebox(20,0)[c]{\it LOB}}
\put(0,49){\makebox(20,0)[c]{\it ATIS}}
\put(0,44){\makebox(20,0)[c]{\it Collins}}
\put(0,39){\makebox(20,0)[c]{\it Sun-spots}}
\put(25,40){\framebox(20,12)[c]{}}
\put(25,46){\makebox(20,6)[c]{Character}}
\put(25,40){\makebox(20,6)[c]{filter}}
\put(45,46){\vector(1,2){10}}
\put(47,61){\makebox(0,0)[c]{\it LOB}}
\put(55,58){\framebox(20,18)[c]{}}
\put(55,70){\makebox(20,6)[c]{Normalising}}
\put(55,64){\makebox(20,6)[c]{(400 per}}
\put(55,58){\makebox(20,6)[c]{length)}}
\put(75,67){\vector(1,0){10}}
\put(85,61){\framebox(20,12)[c]{}}
\put(85,67){\makebox(20,6)[c]{Length}}
\put(85,61){\makebox(20,6)[c]{up to 20}}
\put(105,67){\vector(1,0){37}}
\put(145,67){\makebox(20,0)[c]{\it LOB (8000)}}
\put(45,46){\vector(1,0){10}}
\put(55,40){\framebox(20,12)[c]{}}
\put(55,46){\makebox(20,6)[c]{Length}}
\put(55,40){\makebox(20,6)[c]{up to 10}}
\put(45,46){\vector(1,-3){10}}
\put(55,4){\framebox(20,12)[c]{}}
\put(55,10){\makebox(20,6)[c]{Length}}
\put(55,4){\makebox(20,6)[c]{up to 20}}
\put(65,16){\vector(0,1){6}}
\put(55,19){\makebox(20,0)[c]{\it Sun-~spots}}
\put(55,22){\framebox(20,12)[c]{}}
\put(55,28){\makebox(20,6)[c]{Selected}}
\put(55,22){\makebox(20,6)[c]{questions}}
\put(75,30){\vector(3,1){40}}
\put(95,33){\makebox(0,0)[c]{\it SSQ}}
\put(75,26){\vector(1,-1){10}}
\put(84,22){\makebox(0,0)[c]{\it SSQ}}
\put(75,46){\vector(1,0){40}}
\put(115,40){\framebox(20,12)[c]{}}
\put(115,46){\makebox(20,6)[c]{Select}}
\put(115,40){\makebox(20,6)[c]{1000/2000}}
\put(135,46){\vector(1,0){7}}
\put(145,56){\makebox(20,0)[c]{\it LOB (2000)}}
\put(145,51){\makebox(20,0)[c]{\it ATIS (593)}}
\put(145,46){\makebox(20,0)[c]{\it Collins (1000)}}
\put(145,41){\makebox(20,0)[c]{\it Sun-spots (1000)}}
\put(145,36){\makebox(20,0)[c]{\it SSQ (1000)}}
\put(75,10){\vector(1,0){10}}
\put(85,4){\framebox(20,12)[c]{}}
\put(85,10){\makebox(20,6)[c]{Core/PRM}}
\put(85,4){\makebox(20,6)[c]{vocab only}}
\put(105,10){\vector(1,0){10}}
\put(115,4){\framebox(20,12)[c]{}}
\put(115,10){\makebox(20,6)[c]{Select}}
\put(115,4){\makebox(20,6)[c]{1000}}
\put(135,10){\vector(1,0){7}}
\put(145,20){\makebox(20,0)[c]{\it LOB (2423)}}
\put(145,15){\makebox(20,0)[c]{\it ATIS (59)}}
\put(145,10){\makebox(20,0)[c]{\it Collins (1000)}}
\put(145,5){\makebox(20,0)[c]{\it Sun-spots (1000)}}
\put(145,0){\makebox(20,0)[c]{\it SSQ (525)}}
}\end{picture}
\caption{Creation of text sets}
\label{textfiltfig}
\end{figure}
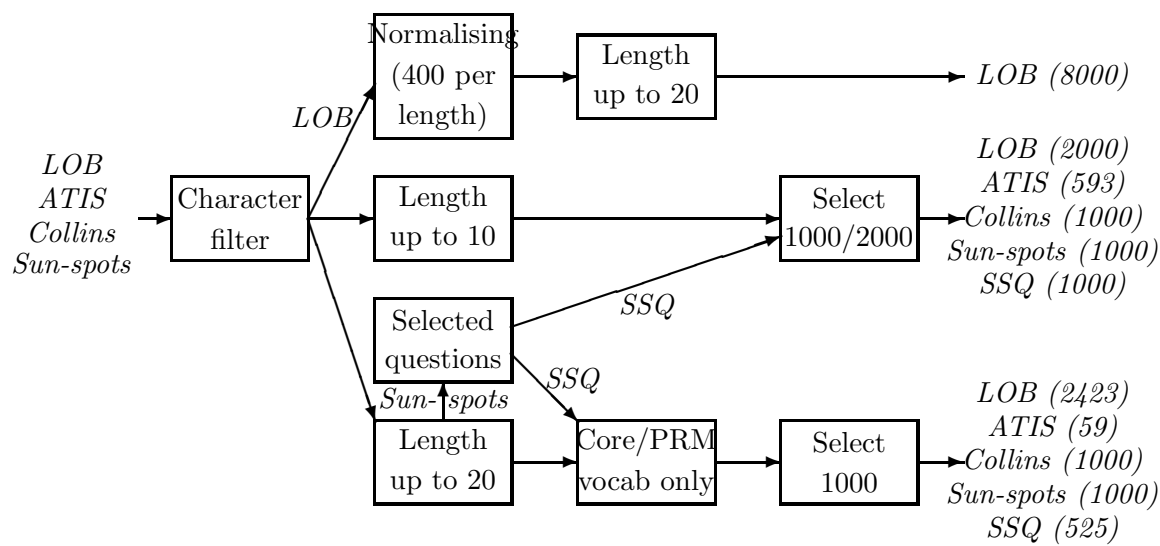

The filtering applied to the LOB corpus, and to the other text sets
which will be described in more detail later, is summarized in figure
\ref{textfiltfig}. Each filter is applied to each possible text set
except where indicated otherwise. In some cases, a random subset of
the data (1000 or 2000 sentences) was selected if the input was larger
than this, to reduce the amount of processing needed to manageable
dimensions.

\section{Development over Time}

In the first set of experiments, versions 0, 1, 2 and 3 of CLARE were
run on the vocabulary-limited subset of the LOB corpus and on the
vocabulary-unlimited (but still legible) subset. In the latter case,
the interface to the MRC Psycholinguistic Database was used to provide
word definitions when necessary.

As expected, later versions showed considerable improvements over
earlier ones.

It was not realistic to process the sentences in context-dependent
ways. This was because
\begin{itemize}
\item modelling context for general language input, unconstrained by
task and domain, is well beyond the state of the art;
\item some of the sentences in the original discourses were ignored
because they were not ``legible'';
\item some of the remaining sentences failed to parse, so that their
contributions to a context model could not in any case have been
evaluated.
\end{itemize}
Thus only stages up to and including the construction of QLFs by
semantic analysis were evaluated.  The criteria used were the
proportion of sentences getting a QLF, and the proportion of sentences
getting a ``good'' first-choice QLF.  The first of these values can be
calculated automatically, but the second involves human inspection,
and hence was only carried out for a small proportion of sentences.

By a ``good'' QLF, we mean a QLF that would be the correct one in some
context judged to be reasonably plausible. Obviously there is an
element of subjectivity in such a judgment. Furthermore, such figures
are only a rough guide to the performance of CLARE in a discourse
application. The production of a good first-choice QLF by the system
in the absence of contextual processing is neither necessary nor
sufficient for the successful processing of that sentence in a full
discourse-based application.  It is not necessary, because contextual
processing may reject the first QLF and accept a later one, or cause
the preference ordering of QLFs to differ so that another is produced
first. And it is not sufficient, because even if the QLF is good in
some context, it may not be correct in the context in which the
application operates; and even if it is, later processing may
introduce errors (in reasoning or reference resolution, for example).
Nevertheless, it is probably safe to assume that the greater the
proportion of good first-choice QLFs produced by a version of CLARE,
the more capable it will be of supporting a full application.

In CLARE-2 and CLARE-3, but not earlier versions, preposition meanings
are represented in QLF by \verb!form!s that do not attempt to
distinguish their senses. To allow comparison between versions,
implausible preposition senses were ignored in assessing the goodness
of a QLF derived from CLARE-0 or CLARE-1. However, all other sense
distinctions were treated as significant.

\subsection{Proportions of QLFs produced}

\input{lobhistqlf}

Figures \ref{lobqlfinvfig} to \ref{lobqlfmrccumfig} show the
percentages of sentences of given lengths (in words) for which QLFs
were produced. Sentence length is shown along the horizontal axis in
each figure. Figure \ref{lobqlfinvfig} is for the limited (core plus
PRM) vocabulary sentences, with no external lexicon access. It shows
that, for CLARE-3, the proportion falls from around 65\% for very
short sentences to around 40\% for sentences of ten words.  For longer
sentences, success rates were under 40\%; it is not possible to be
more exact because the numbers of sentences of these lengths were
fairly small (see figure \ref{loblenfig}).

The performance of CLARE-3 is consistently better than that of CLARE-2
for sentences of up to ten words, just as CLARE-2 improves on
CLARE-1. This reflects the way that coverage extensions were made
during the last two years of the project in response to shortcomings
identified in processing various (non-LOB) texts. CLARE-1 is in turn
superior to CLARE-0 for very short sentences, but only a little better
thereafter. This is because the main coverage extensions in the first
year of the project were for elliptical phenomena, and very
short sentences are often elliptical in nature.

Figure \ref{lobqlfinvcumfig} presents the same data in a cumulative way:
that is, for each length $l$, the percentage of sentences of length
$l$ or less that received a QLF is shown. This again makes the
improvement in the second and third years of the project apparent.

The QLF production rates for this restricted vocabulary set should be
viewed very much as a pessimistic lower bound on the performance CLARE
is likely to exhibit in a domain to which it has been tailored. This
is because in any natural language, core vocabulary items tend to be
those with the most complex behaviour. Evidence for this is the fact
that one third of the Longman Dictionary of Contemporary English
consists of definitions for the two thousand words in the ``basic
vocabulary'' identified by the dictionary writers, although those
words represent only a small minority of the tens of thousands of
words that the dictionary defines. Thus the restricted vocabulary
sentences contain many words with a large range of syntactic and
semantic behaviours, only the most frequently occurring of which are
represented in the core lexicon.

In contrast, sentences pertaining to a specific domain will typically
contain many words with comparatively simple behaviours, and many core
lexicon words will only occur in a subset of their possible uses. If
the domain lexicon is properly set up, it is to be expected that a
much lower proportion of sentences will contain word uses not foreseen
in the lexicon.

Unfortunately, no sufficiently large corpus of sentences is currently
available to assess this likely improvement in the PRM domain.
However, the performance of the various CLARE systems on the unlimited
vocabulary set, with MRC external lexicon access switched on, is
illuminating; see figures \ref{lobqlfmrcfig} and
\ref{lobqlfmrccumfig}.  Under these conditions, the proportion of
words with complex behaviours more closely reflects the language as a
whole and therefore presumably, likely behaviour in applications. The
advantage of domain lexicon tailoring is clearly not available,
however.  Nevertheless, significant improvements are apparent: for
sentences of up to ten words, success rates range from 80\% down to
60\% for CLARE-3, compared to the range of 65\% down to 40\% for the
restricted vocabulary case. Similar improvements are apparent for
earlier versions. The relatively large increase between CLARE-2 and
CLARE-3 is due in part to the introduction of the OUP Database.
Thorough exploitation of a more sophisticated external lexicon would
improve the figures still further.

Performance figures for an early version of an application of CLARE
to an air travel inquiry domain, detailed in Section \ref{slt}, lend
weight to the claim that much better performance can be expected when
the system is properly tailored. In that application, 92\% of
sentences of up to 10 words produced a QLF, as did 91\% of sentences
up to 15 words. These figures are much superior to any derived from
running CLARE on the general LOB corpus.

The LOB figures turn out to be quite sensitive to the range of
characters that are treated as being legible. For example, commas
currently have only a partial treatment in the grammar, and hyphens
often occur (in LOB) in pairs signifying dashes, which are not treated
at all.  If commas and hyphens are excluded, leaving only
alphanumerics, white space and final punctuation characters as
legible, then for the unrestricted vocabulary case the proportion of
sentences of up to ten words that get a QLF from CLARE-2 rises by
5.9\%, from 53.4\% to 59.3\%. For restricted vocabulary, the rise is
more modest, by 2.9\% from 44.5\% to 47.6\%.

Part of these increases are because decreasing the set of legible
characters tends to decrease the average sentence length; however,
this effect on its own should cause no more than a 1\% rise in QLF
rates. In fact, the effect of eliminating commas and hyphens from the
sentences in the ``normalised'' LOB sentence set discussed in section
\ref{covsentlen} below is to increase the percentage of QLFs produced
by an average of 4\% {\it at each sentence length} between 1 and 10,
i.e.\ after removing the bias caused by the length change.

\subsection{Proportions of good QLFs}

The preceding discussion says nothing about the appropriateness of the
QLFs produced. The goodness or otherwise of a QLF has to be assessed
by human inspection, a time-consuming process which meant that only a
small proportion could be inspected.

One hundred QLFs from each of the eight conditions (CLARE-0, CLARE-1,
CLARE-2 and CLARE-3, each for the limited and unlimited vocabulary
cases) were inspected. The results are given in table
\ref{goodqlftab}, along with those from an analogous experiment on
SRI's air travel enquiry application of CLARE (see Section \ref{slt}).
The two ``percent of good QLFs'' columns show the numbers of good QLFs
found in the inspections; the other two columns show these figures
multiplied by the proportions of sentences of ten words or less for
which QLFs were produced at all under the conditions in question.

\begin{table}
\begin{center}
\begin{tabular}{|l|c|c|c|c|} \hline
System & \multicolumn{2}{c|}{\% of good QLFs} &
         \multicolumn{2}{c|}{\% of sentences good} \\  \cline{2-5}
       & Limited & Unlimited  & Limited & Unlimited \\ \hline
CLARE-0 & 89 & 79 & 15 & 12 \\ \hline
CLARE-1 & 94 & 57 & 26 & 22 \\ \hline
CLARE-2 & 85 & 61 & 38 & 33 \\ \hline
CLARE-3 & 78 & 83 & 44 & 58 \\ \hline
Air Travel &\multicolumn{2}{c|}{84} & \multicolumn{2}{c|}{77} \\ \hline
\end{tabular}
\end{center}
\caption{Good QLF percentages for CLARE versions}
\label{goodqlftab}
\end{table}

Because of the small sizes of the samples, these figures should be
interpreted with caution. The margin of error in a count of, say, 90
is around 6 (at the 5\% confidence level) and for a count of 60, it
is around 10.

Nevertheless, it is possible to conclude that a large majority of
first-choice QLFs are good in the limited-vocabulary cases. The
decrease (from 94 to 85 and then to 78) from CLARE-1 to CLARE-3 is due
to the coverage of later versions being so much greater, which
introduces scope for more ``false positive'' interpretations. The
preference mechanisms have an important role to play here in
depreferring readings that are linguistically unlikely (and hence not
really plausible in any context). That they are fulfilling this role
is shown by the increase in the estimate of the proportion of {\it
sentences} producing good QLFs, which rises from 26\% in CLARE-1 to
38\% in CLARE-2 and 44\% in CLARE-3, a marked improvement.

The figures in the vocabulary-unlimited case also show an improvement
for later versions of the system in the proportion of sentences
yielding a good QLF. The proportion of good to bad QLFs fell up to
CLARE-2 but then rose to their highest level in CLARE-3. The fall is
largely because MRC-derived entries tend to be quite general and so,
as the syntactic coverage of the system increases, the risk of a false
parse resulting from overgeneral entries increases. The rise in
CLARE-3 is due mainly to the more discriminating external lexicon and
the improved preference metrics.

The intermediate results for the air travel application, given for
comparison, suggest that the fairly high proportions of good QLFs
(84\%) can be maintained even when domain-specific linguistic coverage
is increased to over 90\% of the inputs, giving an overall 77\% of
sentences with a good first choice QLF. Although it is difficult to
generalize from one domain to another, this figure shows that, for
some domains at least, domain-specific additions to the grammar and
lexicon can greatly improve performance.

\subsection{Development over Time: Conclusions}

It is interesting that in CLARE-3, the proportion of sentences
yielding good QLFs is higher for the vocabulary-unlimited case in
spite of the relative paucity of external lexicon information. This
may partly be because whereas the core and PRM lexicons distinguish
between senses for content words, the MRC-derived entries only do so
where there is a syntactic difference. Thus the MRC-derived QLFs tend
to be vaguer, and hence have a better chance of being judged good.
However, working against this is the fact that the unlimited sentences
were on average longer and therefore should have been harder to
analyse. 

More important, however, is the fact that, as already
discussed, the vocabulary-limited sentences are relatively hard ones
because they contain a preponderance of frequent, and therefore often
idiosyncratic, words. Thus the level of performance in a domain to
which CLARE has been tailored should be significantly better than
either the vocabulary-limited or the vocabulary-unlimited figures
given here, as it has the advantages of both and the disadvantages of
neither.  Domain constraints will often make it possible to do a
thorough job of representing a much greater proportion of the word
behaviours that do occur in the domain in question. The importance of
tailoring CLARE's vocabulary to the domain is therefore underlined.

\section{Dependence on text type}
\label{covtexttype}

In the second experiment, the CLARE-2\footnote{This experiment was not
repeated for the CLARE-3 system. However, informal observations of
performance on some of the test sets involved suggest the CLARE-3
percentages will typically be around 10\% higher. The conclusions are
likely to be unaffected.} system was run on five different text sets,
in order to test the dependence of the system's performance on text
type. The text sets were as follows. The names used to refer to them
elsewhere in the chapter are given in bold text.
\begin{itemize}
\item The {\bf LOB} corpus, described earlier. Because of the nature 
of its original medium, printed text, sentences tend to be fairly
long, and ungrammaticality is rare.
\item A set of spoken enquiries about air travel information (the {\bf ATIS}
task), taken from version 1 of the CD-ROM recently release under the
Association for Computational Linguistics' Text Encoding Initiative.
This material was selected because it represents interactive dialogue,
because the domain is quite constrained, and because, unlike the other
text sets used, it originated in speech.  Sentences are fairly short,
and, as with most spoken material, there are many syntactic and
semantic errors.
\item The example phrases and sentences from the (parsed) version of
the {\bf Collins} English dictionary, provided on the ACL CD. This is
an artificial set, but covers the lexicon of English quite thoroughly.
It contains few or no errors, but fragmentary sentences are common,
and the average sentence length is low.
\item Sentences from the {\bf Sun-spots} electronic bulletin board, whose
purpose is to allow users of Sun Microsystems products to share
problems and solutions. Although this medium for this sentence set is
typed, the material is informal and ``semi-interactive'' in flavour;
there are, for example, many instances of participants quoting
someone's question and then answering it or commenting on it as if the
dialogue were taking place in real time. Sun-spots material contains
many errors and sentence fragments and much specialized vocabulary.
\item A set of Sun-spots sentences identified as questions (henceforth
{\bf SSQ}) by virtue of their terminating with a question mark.
Questions are, perhaps, of particular interest because query answering
is an important possible use for CLARE. In addition, the performance
of the CLARE system on this sentence set has been monitored throughout
the project, and some coverage extensions were made as a result.
\end{itemize}

Table \ref{corpuslentab} shows the number of sentences in each text set,
the percentage of them that were ``legible'', and the average sentence
length of all the sentences and of just the legible ones. Some figures
for the SSQ set are omitted because this set was derived from the
legible part of the Sun-spots set.

\begin{table}
\begin{center}
\begin{tabular}{|l|c|c|c|c|} \hline
Text set  &  Sentences &  \% legible & Avge len all & Avge len legible \\ \hline \hline
LOB       &   45382 &  40.1   &   21.0 &  10.9 \\ \hline
ATIS      &    1015 &  80.3   &    9.9 &   9.0 \\ \hline
Collins   &   18813 &  92.4   &    4.5 &   4.4 \\ \hline
Sun-spots &   24356 &  39.0   &   15.4 &   9.8  \\ \hline
SSQ       &    1800 &         &        &   9.0  \\ \hline
\end{tabular}
\end{center}
\caption{Sentence characteristics of text sets}
\label{corpuslentab}
\end{table}

The results are given in tables
\ref{perflimtab} and \ref{perfunlimtab}, for sentences of
up to ten words. The mean and median lengths of sentences are given
because performance tends, other things being equal, to be better for
shorter sentences. The other three columns give the percentage of
sentences that produced QLFs, the estimated percentage of first-choice
QLFs that were good, and, as the product of these, the estimated
percentage of sentences producing a good first-choice QLF.

\begin{table}
\begin{center}
\begin{tabular}{|l|c|c|c|c|c|} \hline
Text  & Mean & Median & \% QLFs & \% QLFs & \% sents \\
set  & Length & Length &        & good & good \\ \hline
ATIS      & 4.22 & 4 & 27.6 & 100 & 28  \\ \hline
Collins   & 4.59 & 4 & 42.6 & 87  & 37  \\ \hline
LOB       & 5.12 & 5 & 44.5 & 85  & 38  \\ \hline
Sun-spots & 4.54 & 4 & 47.4 & 83  & 40  \\ \hline
SSQ       & 5.54 & 5 & 51.6 & 90  & 46  \\ \hline
\end{tabular}
\end{center}
\caption{CLARE-2 performance for vocabulary-limited case}
\label{perflimtab}
\end{table}

\begin{table}
\begin{center}
\begin{tabular}{|l|c|c|c|c|c|} \hline
Text  & Mean & Median & \% QLFs & \% QLFs & \% sents \\
set  & Length & Length &        & good & good \\ \hline
ATIS      & 6.53 & 7 & 68.5 & 63  & 43   \\ \hline
Collins   & 4.21 & 4 & 74.4 & 53  & 40   \\ \hline
LOB       & 5.95 & 6 & 53.4 & 61  & 33   \\ \hline
Sun-spots & 5.88 & 6 & 58.6 & 57  & 33   \\ \hline
SSQ       & 6.29 & 6 & 59.7 & 77  & 46   \\ \hline
\end{tabular}
\end{center}
\caption{CLARE-2 performance for vocabulary-unlimited case}
\label{perfunlimtab}
\end{table}

Because of the large number of cases, the percentage of good QLFs was
estimated from a sample of only 30 in all but one case. With a sample
of 30, the results can be considered accurate to within about 5\%.
However, only 12 QLFs were available in the ATIS vocabulary-limited
case, so the result of 100\% good is rather unreliable.

With these caveats in mind, the conclusions must be fairly tentative.
However, excluding the ATIS set from consideration, it does not appear
that there is any very strong dependence of performance on the text
types. The SSQ set gives a rather better score than the Sun-spots set;
this is probably due to the addition to the CLARE-2 grammar of
constructions such as ``anyone else'' and ``out there'', which are
very common in SSQ but less so in the main Sun-spots list or in the
other text sets. The difference points to the significant improvements
that can be gained from making grammar development sensitive to the
frequency of occurrence of the phenomena considered.

In the Collins text set, the fact that the sentences tended to be
error-free and fairly short was partly offset for the
vocabulary-limited case by the fact that examples were provided of
word uses irrespective of their frequency.  Thus relatively rare uses,
which were less likely to be defined in the core lexicon, occurred
more often than they would in other kinds of text set.

In the ATIS set, the small percentage of QLFs for the
vocabulary-limited case is due to a high frequency of domain-specific
phrases such as ``class M'' which happen to fall within lexical but
not syntactic coverage. A few extensions to the grammar would improve
matters significantly.

\section{System components and sentence length}
\label{covsentlen}

In the final experiment, CLARE-3 was tested against a subset of the
LOB corpus to determine, for sentences of different lengths, the
success rate for the various system components up to and including
semantic analysis. Because of the numbers of sentences needed to
ensure reliable results, the vocabulary-unlimited condition, with the
MRC external interface (OUP version), was used.

The corpus consisted of 400 sentences, selected at random, of each of
the lengths 1 to 20 inclusive (this is the process referred to as
``normalising'' in figure \ref{textfiltfig}). Because of improvements
in the external lexicon interface and in the TCP-based corpus
processing code, very few sentences timed out or failed at the lexical
level.  Figure \ref{lobratefig} therefore shows the percentages of
sentences of each length for which a full parse and a full semantic
analysis, respectively, were found.
Values are accurate to within about 2.5\% at the usual 5\%
confidence level.

\setlength{\unitlength}{1mm}
\thicklines
\begin{figure}
\begin{picture}(130,100)
\put(10,5){\vector(1,0){120}}
\put(10,5){\vector(0,1){95}}
\put(0,  4.0){  0
}\put(  9.0,  5.0){\line(1,0){2}}
\put(0, 23.0){ 20
}\put(  9.0, 24.0){\line(1,0){2}}
\put(0, 42.0){ 40
}\put(  9.0, 43.0){\line(1,0){2}}
\put(0, 61.0){ 60
}\put(  9.0, 62.0){\line(1,0){2}}
\put(0, 80.0){ 80
}\put(  9.0, 81.0){\line(1,0){2}}
\put(0, 99.0){100
}\put(  9.0,100.0){\line(1,0){2}}
\put( 65.0, 95.0){$\cdot$ Got a parse
}
\put( 65.0, 90.0){{\tiny $+$} Got a QLF
}
\put( 13.0,  0.0){  1}
\put( 16.0,  4.0){\line(0,1){2}}
\put( 16.8, 81.1){$\cdot$}
\put( 17.6, 81.7){$\cdot$}
\put( 18.4, 82.3){$\cdot$}
\put( 19.2, 82.9){$\cdot$}
\put( 20.0, 83.5){$\cdot$}
\put( 20.8, 84.1){$\cdot$}
\put( 21.6, 84.7){$\cdot$}
\put( 16.8, 79.9){{\tiny $+$}}
\put( 17.5, 80.6){{\tiny $+$}}
\put( 18.3, 81.2){{\tiny $+$}}
\put( 19.1, 81.9){{\tiny $+$}}
\put( 19.8, 82.5){{\tiny $+$}}
\put( 20.6, 83.1){{\tiny $+$}}
\put( 21.4, 83.8){{\tiny $+$}}
\put( 22.1, 84.4){{\tiny $+$}}
\put( 19.0,  0.0){  2}
\put( 22.0,  4.0){\line(0,1){2}}
\put( 22.8, 84.4){$\cdot$}
\put( 23.7, 83.9){$\cdot$}
\put( 24.5, 83.3){$\cdot$}
\put( 25.3, 82.8){$\cdot$}
\put( 26.2, 82.2){$\cdot$}
\put( 27.0, 81.7){$\cdot$}
\put( 27.8, 81.1){$\cdot$}
\put( 22.9, 83.8){{\tiny $+$}}
\put( 23.7, 83.3){{\tiny $+$}}
\put( 24.6, 82.8){{\tiny $+$}}
\put( 25.5, 82.3){{\tiny $+$}}
\put( 26.3, 81.8){{\tiny $+$}}
\put( 27.2, 81.3){{\tiny $+$}}
\put( 28.0, 80.8){{\tiny $+$}}
\put( 25.0,  0.0){  3}
\put( 28.0,  4.0){\line(0,1){2}}
\put( 28.7, 80.3){$\cdot$}
\put( 29.4, 79.6){$\cdot$}
\put( 30.1, 78.9){$\cdot$}
\put( 30.9, 78.2){$\cdot$}
\put( 31.6, 77.5){$\cdot$}
\put( 32.3, 76.8){$\cdot$}
\put( 33.0, 76.1){$\cdot$}
\put( 33.7, 75.4){$\cdot$}
\put( 28.6, 80.0){{\tiny $+$}}
\put( 29.2, 79.2){{\tiny $+$}}
\put( 29.9, 78.5){{\tiny $+$}}
\put( 30.5, 77.7){{\tiny $+$}}
\put( 31.1, 76.9){{\tiny $+$}}
\put( 31.7, 76.1){{\tiny $+$}}
\put( 32.3, 75.3){{\tiny $+$}}
\put( 33.0, 74.5){{\tiny $+$}}
\put( 33.6, 73.7){{\tiny $+$}}
\put( 34.2, 73.0){{\tiny $+$}}
\put( 31.0,  0.0){  4}
\put( 34.0,  4.0){\line(0,1){2}}
\put( 34.9, 75.5){$\cdot$}
\put( 35.8, 75.9){$\cdot$}
\put( 36.8, 76.3){$\cdot$}
\put( 37.7, 76.7){$\cdot$}
\put( 38.6, 77.1){$\cdot$}
\put( 39.5, 77.5){$\cdot$}
\put( 40.4, 77.9){$\cdot$}
\put( 34.9, 73.5){{\tiny $+$}}
\put( 35.9, 73.9){{\tiny $+$}}
\put( 36.8, 74.2){{\tiny $+$}}
\put( 37.8, 74.5){{\tiny $+$}}
\put( 38.7, 74.9){{\tiny $+$}}
\put( 39.7, 75.2){{\tiny $+$}}
\put( 37.0,  0.0){  5}
\put( 40.0,  4.0){\line(0,1){2}}
\put( 40.9, 77.3){$\cdot$}
\put( 41.8, 76.9){$\cdot$}
\put( 42.8, 76.5){$\cdot$}
\put( 43.7, 76.1){$\cdot$}
\put( 44.6, 75.7){$\cdot$}
\put( 45.5, 75.3){$\cdot$}
\put( 46.4, 74.9){$\cdot$}
\put( 40.9, 74.8){{\tiny $+$}}
\put( 41.7, 74.3){{\tiny $+$}}
\put( 42.6, 73.8){{\tiny $+$}}
\put( 43.5, 73.4){{\tiny $+$}}
\put( 44.4, 72.9){{\tiny $+$}}
\put( 45.2, 72.4){{\tiny $+$}}
\put( 46.1, 71.9){{\tiny $+$}}
\put( 43.0,  0.0){  6}
\put( 46.0,  4.0){\line(0,1){2}}
\put( 47.0, 75.0){$\cdot$}
\put( 48.0, 74.8){$\cdot$}
\put( 49.0, 74.6){$\cdot$}
\put( 50.0, 74.5){$\cdot$}
\put( 50.9, 74.3){$\cdot$}
\put( 51.9, 74.2){$\cdot$}
\put( 46.9, 71.5){{\tiny $+$}}
\put( 47.8, 71.1){{\tiny $+$}}
\put( 48.7, 70.7){{\tiny $+$}}
\put( 49.6, 70.3){{\tiny $+$}}
\put( 50.5, 69.8){{\tiny $+$}}
\put( 51.4, 69.4){{\tiny $+$}}
\put( 52.3, 69.0){{\tiny $+$}}
\put( 49.0,  0.0){  7}
\put( 52.0,  4.0){\line(0,1){2}}
\put( 52.9, 73.8){$\cdot$}
\put( 53.8, 73.4){$\cdot$}
\put( 54.8, 73.0){$\cdot$}
\put( 55.7, 72.6){$\cdot$}
\put( 56.6, 72.3){$\cdot$}
\put( 57.5, 71.9){$\cdot$}
\put( 53.0, 69.0){{\tiny $+$}}
\put( 54.0, 68.9){{\tiny $+$}}
\put( 55.0, 68.8){{\tiny $+$}}
\put( 56.0, 68.7){{\tiny $+$}}
\put( 57.0, 68.6){{\tiny $+$}}
\put( 58.0, 68.5){{\tiny $+$}}
\put( 55.0,  0.0){  8}
\put( 58.0,  4.0){\line(0,1){2}}
\put( 59.0, 71.5){$\cdot$}
\put( 60.0, 71.3){$\cdot$}
\put( 60.9, 71.0){$\cdot$}
\put( 61.9, 70.8){$\cdot$}
\put( 62.9, 70.6){$\cdot$}
\put( 63.9, 70.4){$\cdot$}
\put( 59.0, 68.3){{\tiny $+$}}
\put( 60.0, 68.1){{\tiny $+$}}
\put( 61.0, 68.0){{\tiny $+$}}
\put( 62.0, 67.8){{\tiny $+$}}
\put( 62.9, 67.7){{\tiny $+$}}
\put( 63.9, 67.5){{\tiny $+$}}
\put( 61.0,  0.0){  9}
\put( 64.0,  4.0){\line(0,1){2}}
\put( 64.7, 69.7){$\cdot$}
\put( 65.5, 69.0){$\cdot$}
\put( 66.2, 68.3){$\cdot$}
\put( 66.9, 67.7){$\cdot$}
\put( 67.7, 67.0){$\cdot$}
\put( 68.4, 66.3){$\cdot$}
\put( 69.2, 65.6){$\cdot$}
\put( 69.9, 64.9){$\cdot$}
\put( 64.6, 66.7){{\tiny $+$}}
\put( 65.3, 66.0){{\tiny $+$}}
\put( 65.9, 65.2){{\tiny $+$}}
\put( 66.6, 64.4){{\tiny $+$}}
\put( 67.2, 63.7){{\tiny $+$}}
\put( 67.8, 62.9){{\tiny $+$}}
\put( 68.5, 62.1){{\tiny $+$}}
\put( 69.1, 61.4){{\tiny $+$}}
\put( 69.8, 60.6){{\tiny $+$}}
\put( 67.0,  0.0){ 10}
\put( 70.0,  4.0){\line(0,1){2}}
\put( 70.8, 64.2){$\cdot$}
\put( 71.6, 63.6){$\cdot$}
\put( 72.4, 63.0){$\cdot$}
\put( 73.2, 62.4){$\cdot$}
\put( 74.0, 61.8){$\cdot$}
\put( 74.8, 61.2){$\cdot$}
\put( 75.6, 60.6){$\cdot$}
\put( 76.4, 60.0){$\cdot$}
\put( 70.7, 59.6){{\tiny $+$}}
\put( 71.4, 58.9){{\tiny $+$}}
\put( 72.2, 58.2){{\tiny $+$}}
\put( 72.9, 57.5){{\tiny $+$}}
\put( 73.6, 56.8){{\tiny $+$}}
\put( 74.3, 56.2){{\tiny $+$}}
\put( 75.1, 55.5){{\tiny $+$}}
\put( 75.8, 54.8){{\tiny $+$}}
\put( 73.0,  0.0){ 11}
\put( 76.0,  4.0){\line(0,1){2}}
\put( 77.0, 60.3){$\cdot$}
\put( 78.0, 60.4){$\cdot$}
\put( 79.0, 60.4){$\cdot$}
\put( 80.0, 60.5){$\cdot$}
\put( 81.0, 60.5){$\cdot$}
\put( 82.0, 60.6){$\cdot$}
\put( 77.0, 54.9){{\tiny $+$}}
\put( 77.9, 55.1){{\tiny $+$}}
\put( 78.9, 55.4){{\tiny $+$}}
\put( 79.8, 55.7){{\tiny $+$}}
\put( 80.8, 56.0){{\tiny $+$}}
\put( 81.8, 56.2){{\tiny $+$}}
\put( 79.0,  0.0){ 12}
\put( 82.0,  4.0){\line(0,1){2}}
\put( 82.7, 59.9){$\cdot$}
\put( 83.4, 59.2){$\cdot$}
\put( 84.1, 58.5){$\cdot$}
\put( 84.8, 57.8){$\cdot$}
\put( 85.5, 57.0){$\cdot$}
\put( 86.2, 56.3){$\cdot$}
\put( 87.0, 55.6){$\cdot$}
\put( 87.7, 54.9){$\cdot$}
\put( 82.7, 55.6){{\tiny $+$}}
\put( 83.3, 54.8){{\tiny $+$}}
\put( 84.0, 54.1){{\tiny $+$}}
\put( 84.7, 53.3){{\tiny $+$}}
\put( 85.3, 52.6){{\tiny $+$}}
\put( 86.0, 51.8){{\tiny $+$}}
\put( 86.7, 51.1){{\tiny $+$}}
\put( 87.4, 50.4){{\tiny $+$}}
\put( 88.0, 49.6){{\tiny $+$}}
\put( 85.0,  0.0){ 13}
\put( 88.0,  4.0){\line(0,1){2}}
\put( 89.0, 54.4){$\cdot$}
\put( 90.0, 54.3){$\cdot$}
\put( 91.0, 54.1){$\cdot$}
\put( 92.0, 54.0){$\cdot$}
\put( 92.9, 53.8){$\cdot$}
\put( 93.9, 53.7){$\cdot$}
\put( 89.0, 49.5){{\tiny $+$}}
\put( 90.0, 49.3){{\tiny $+$}}
\put( 91.0, 49.2){{\tiny $+$}}
\put( 92.0, 49.0){{\tiny $+$}}
\put( 92.9, 48.9){{\tiny $+$}}
\put( 93.9, 48.7){{\tiny $+$}}
\put( 91.0,  0.0){ 14}
\put( 94.0,  4.0){\line(0,1){2}}
\put( 94.9, 53.3){$\cdot$}
\put( 95.9, 52.9){$\cdot$}
\put( 96.8, 52.6){$\cdot$}
\put( 97.7, 52.2){$\cdot$}
\put( 98.7, 51.9){$\cdot$}
\put( 99.6, 51.5){$\cdot$}
\put( 94.8, 48.1){{\tiny $+$}}
\put( 95.6, 47.5){{\tiny $+$}}
\put( 96.4, 46.8){{\tiny $+$}}
\put( 97.1, 46.2){{\tiny $+$}}
\put( 97.9, 45.6){{\tiny $+$}}
\put( 98.7, 45.0){{\tiny $+$}}
\put( 99.5, 44.4){{\tiny $+$}}
\put(100.3, 43.7){{\tiny $+$}}
\put( 97.0,  0.0){ 15}
\put(100.0,  4.0){\line(0,1){2}}
\put(100.9, 50.9){$\cdot$}
\put(101.8, 50.4){$\cdot$}
\put(102.6, 49.9){$\cdot$}
\put(103.5, 49.5){$\cdot$}
\put(104.4, 49.0){$\cdot$}
\put(105.3, 48.5){$\cdot$}
\put(106.2, 48.0){$\cdot$}
\put(100.9, 43.6){{\tiny $+$}}
\put(101.8, 43.2){{\tiny $+$}}
\put(102.8, 42.8){{\tiny $+$}}
\put(103.7, 42.4){{\tiny $+$}}
\put(104.6, 42.0){{\tiny $+$}}
\put(105.5, 41.6){{\tiny $+$}}
\put(106.4, 41.2){{\tiny $+$}}
\put(103.0,  0.0){ 16}
\put(106.0,  4.0){\line(0,1){2}}
\put(106.9, 47.8){$\cdot$}
\put(107.9, 47.5){$\cdot$}
\put(108.8, 47.1){$\cdot$}
\put(109.8, 46.8){$\cdot$}
\put(110.7, 46.5){$\cdot$}
\put(111.7, 46.2){$\cdot$}
\put(107.0, 41.2){{\tiny $+$}}
\put(108.0, 41.0){{\tiny $+$}}
\put(108.9, 40.8){{\tiny $+$}}
\put(109.9, 40.6){{\tiny $+$}}
\put(110.9, 40.4){{\tiny $+$}}
\put(111.9, 40.2){{\tiny $+$}}
\put(109.0,  0.0){ 17}
\put(112.0,  4.0){\line(0,1){2}}
\put(113.0, 46.3){$\cdot$}
\put(113.9, 46.6){$\cdot$}
\put(114.9, 46.9){$\cdot$}
\put(115.8, 47.1){$\cdot$}
\put(116.8, 47.4){$\cdot$}
\put(117.8, 47.7){$\cdot$}
\put(113.0, 40.2){{\tiny $+$}}
\put(114.0, 40.2){{\tiny $+$}}
\put(115.0, 40.2){{\tiny $+$}}
\put(116.0, 40.3){{\tiny $+$}}
\put(117.0, 40.3){{\tiny $+$}}
\put(118.0, 40.3){{\tiny $+$}}
\put(115.0,  0.0){ 18}
\put(118.0,  4.0){\line(0,1){2}}
\put(118.9, 47.4){$\cdot$}
\put(119.9, 47.0){$\cdot$}
\put(120.8, 46.6){$\cdot$}
\put(121.7, 46.3){$\cdot$}
\put(122.6, 45.9){$\cdot$}
\put(123.6, 45.5){$\cdot$}
\put(119.0, 40.4){{\tiny $+$}}
\put(120.0, 40.5){{\tiny $+$}}
\put(121.0, 40.6){{\tiny $+$}}
\put(122.0, 40.7){{\tiny $+$}}
\put(123.0, 40.8){{\tiny $+$}}
\put(124.0, 40.9){{\tiny $+$}}
\put(121.0,  0.0){ 19}
\put(124.0,  4.0){\line(0,1){2}}
\put(124.7, 44.7){$\cdot$}
\put(125.5, 44.0){$\cdot$}
\put(126.2, 43.3){$\cdot$}
\put(126.9, 42.7){$\cdot$}
\put(127.7, 42.0){$\cdot$}
\put(128.4, 41.3){$\cdot$}
\put(129.2, 40.6){$\cdot$}
\put(129.9, 40.0){$\cdot$}
\put(124.8, 40.3){{\tiny $+$}}
\put(125.5, 39.6){{\tiny $+$}}
\put(126.3, 39.0){{\tiny $+$}}
\put(127.0, 38.3){{\tiny $+$}}
\put(127.8, 37.7){{\tiny $+$}}
\put(128.6, 37.0){{\tiny $+$}}
\put(129.3, 36.4){{\tiny $+$}}
\put(130.1, 35.7){{\tiny $+$}}
\put(127.0,  0.0){ 20}
\put(130.0,  4.0){\line(0,1){2}}
\end{picture}
\caption{Sentence fates according to length}
\label{lobratefig}
\end{figure}
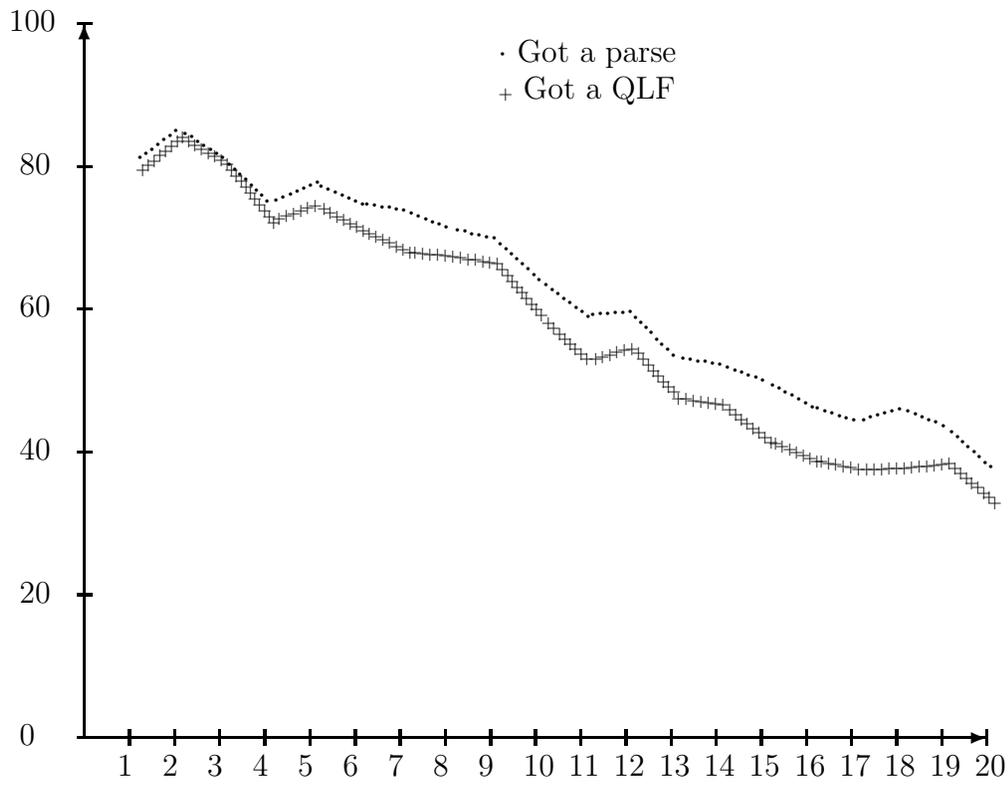

It is clear from the figure that, unsurprisingly, parsing
becomes increasingly likely to fail as sentence length increases.
Semantic analysis only rarely fails when parsing succeeds; most such
failures are on sortal grounds.

Because of the coverage work done during the third year of the
project, the success rate falls off more slowly with sentence length
than it did a year ago. More than half of sentences at all lengths up
to twelve words receive an analysis, and even at twenty words, nearly
a third of sentences do. Again, it should be emphasised that tailoring
to a domain sublanguage would significantly increase these figures.

\section{Tagging for parsing efficiency: a preliminary experiment}

This section is based on a talk given by Stephen Pulman to a
meeting of the DTI Speech and Language Technology Club on
Sublanguages, in Edinburgh, January 1992. We are grateful to the
participants there for their comments, and to David Elworthy for
providing both software and help.

\subsection{Introduction}

One of the crucial factors affecting parsing and analysis speeds for
large rule-based systems is the degree of lexical ambiguity found in
the input sentence. Verbs, in particular, often have many
syntactically distinct entries even for similar senses, because of
their differing subcategorisation requirements. But other ambiguities
like whether something is a verb or a noun can also have a significant
effect on efficiency, because they lead to locally valid parsing
hypotheses that may not be ruled out for some time. The combinatoric
effect of these local ambiguities might be quite large, and in a large
coverage descriptive grammar and lexicon the degree of lexical
ambiguity is usually very large.

Ideally, one would like only to take into account those lexical entries
for words which are capable of  playing a part in the overall parse,
not even looking at those which cannot be integrated into a larger
structure. One might want also want to discriminate against those
lexical entries which {\it can} play a part in an overall parse, 
but which are relatively unlikely to do so. In the first case,
we are interested in  producing all parses, but not wasting time
on local red herrings. In the second case we  might be trying to
tune a general purpose natural language system to some sublanguage,
taking advantage of regularities characteristic of that sublanguage,
so that valid but unlikely analyses will not be produced.

For both these purposes, the most obvious way to achieve the goals in
question is to introduce a statistical element in analysis.  In the
cases we shall be talking about, this will mean statistical information
about the frequency with which a word appears with a particular
syntactic category, and about bi-gram probabilities between pairs of
categories. However, if the necessary data was available, it would be
sensible to do something analogous for semantic information too.
Ignoring or down-grading senses of words that are irrelevant or
unlikely with respect to a given domain, even where there is no
syntactic distinction, is a cheap and cheerful way of accomplishing a
task that could otherwise require sophisticated contextual reasoning.

The best way to introduce statistical information into the analysis
process is to associate statistical information with rules and lexical
entries and make the parsing algorithm sensitive to that information in the
desired way. Ideally, this would be done in a comprehensive and systematic
manner, allowing for information from different sources to be combined.
In the long run,  this will be the best approach. In the short term,
a very simple way to introduce a statistical element into parsing is
to use `tagging' as a way of disambiguating lexical categories.
Tagging is the assignment of lexical categories to a stream of words based
on statistical cooccurrence data of the type that can be modelled using
Hidden Markov Models or the simple subcases of these as described
e.g.\ in DeRose (1988)  or De Marcken (1990).

A tagger of the type described in these references will assign to each
word in a sentence a part of speech in such a way as to maximise the
overall probability of that sequence of assignments in the sentence.
We can then use this information so that the parser can ignore all the
lexical entries for a word that are inconsistent with this assignment
of tags.  (See De Marcken (1990) for a description of this type of
approach in aiding deterministic parsing of a large corpus).  In the
ideal case, we would discover that this information still allowed the
parser to build the most likely overall analysis for the sentence.

This method will not guarantee overall correctness or completeness, of
course. There are several sources of error possible. Firstly, accuracy
rates for taggers are usually in the region of 95\% word accuracy.
This sounds high, but means that every 20 word sentence is likely to
contain at least one error. Secondly, some sentences may be genuinely
intended as ambiguous: many advertising slogans, for example, rely on
this for their effect. A deterministic assignment of tags will mean
that such examples will only get one analysis, if the ambiguity in
question involves different choices of lexical category. (This is
unlikely to happen very frequently in most applications). Thirdly, and
most simply, it may be the case that syntactic probability does not
equate with semantic plausibility.  The most syntactically likely
analysis may not always correspond to the most semantically plausible.

It is clear that any kind of pruning strategy will affect
completeness.  The question is one of whether the trade-off between
completeness and efficiency is an acceptable one. The rest of this
section describes two small experiments in which we attempted to
investigate what this combination of tagging and rule-based analysis
would give rise to: in particular, to measure exactly what the above
trade-off was. This experiment used a tagger and the CLARE system.

\subsection{Tagging}

Since no existing tagger was available to us, we implemented a simple
bigram-based system of the type described in DeRose (1988).
Although his presentation does not make the connection, such a system can be
seen as the bigram case of a Hidden Markov Model, where the bigram
probabilities represent probabilities of transitions between states in
a set of two-state models, his `relative tag probability' measure
corresponds to output probabilities of symbols on transitions, and the
`optimal path' computation described is an application of the Viterbi
algorithm.  

Transition and output probabilities were obtained by processing the
tagged version of the LOB corpus. The resulting tagging system
achieved accuracy rates of about 94\% when tested on closed data, and
about 88\% when tested on open data. This could be improved on by
moving to trigrams, and by doing some preprocessing to identify fixed
phrases, but the necessary resources (of time) were not available to
me.

The LOB tag set is rather large, and does not coincide with the set of
lexical categories used in the CLARE natural language processing
system.  Furthermore, even where the relevant set of tags is
identical, the tagged LOB corpus does not always agree with the CLE
grammar over the allocation of specific words to lexical categories.

In order to use the tagging for the purpose intended, the output of
the tagger was mapped to a set of CLE categories. This process is
relational rather than functional: for example, the CLE grammar does
not distinguish syntactically between quantificational and
non-quantificational determiners, and so `any' and `the' are members
of the same CLE category. LOB distinguishes these two.  Conversely, a
word like `both' in LOB has a special single category, whereas in the
CLE grammar `both' can be either a conjunction or a determiner.

Ideally, the experiments should have used a corpus of sentences
successfully parsed by the CLE to derive the statistical information.
Although we hope soon to be in a position to carry this out, using an
existing tagged corpus was at the time the only way of getting
sufficiently large numbers of examples for the statistics to be
reliable. However, the mismatch between category sets means, among
other things, that the results of any experiments can only be regarded
as suggestive.  On no account should they be taken as conclusive.

\subsection{Experiment One}

The main purpose of this experiment was to use the output of the
tagger to prune the lexical entries seen by the parser. Ideally, one
would hope that processing speeds would be faster, but that accuracy
would not be too badly compromised.

The first issue to be solved is: how do we measure accuracy?  In
CLARE, the output of syntactic and semantic analysis is a set of QLFs
which are ranked according to various measures of plausibility, both
{\it a priori} and empirically derived.  The {\it a priori} estimates
are based on properties of the structure of QLFs, including sortal
restrictions, enabling the system to attach preferences to, for
example, low attachments, properly balanced conjunctions, and so on.
We have found that in practice these rankings are surprisingly
effective in ensuring that the most likely interpretation of a
sentence is the one that is most highly rated.

As a first approximation, therefore, we selected a set of sentences
from the LOB corpus for which the CLE had provided, as its first
choice, a QLF that we regarded as a plausible one in the absence of any
kind of biasing context.  The sentences were up to 12 words long: in
general, the CLE produces QLFs for about 67\% of sentences up to 10
words long in LOB, with coverage gradually declining after that. There
were 124 such sentences (representing about 1 day's manual QLF
inspection effort).

Next 
we arranged for each of these sentences to be tagged, as
described above, the result mapped to a set of CLE tags, and for the
CLE then to be run on the same sentences, using only the non-null
intersection of the lexical entries for each word from these two
sources. In the case where there was no intersection between the two
sets of tags, the CLE tags were left. (The LOB and CLE tag sets are
not subsets of each other). The motivation for this method of pruning
was in order to try to minimise the bad effects of the non-identity of
the tag sets described earlier. Then for each sentence the QLF
produced by the system using the pruned tag set was compared with that
produced earlier. Relative timings and various other statistics were
also collected.

Effects of the sort that we had been hoping to find can be illustrated
by the following sentence, where the lexical entries produced by the
CLE with the 100,000 word lexicon are shown for each 
word in figure~\ref{Tagging-figure-1}.

\begin{figure}
\caption{Pruning by tagging}\label{Tagging-figure-1}
\small
\begin{verbatim}
   these: det(...)
   can: v(...)
   >> can: n(...)
   be: v(...) (3 times)
   read: [read,ed] v_v_ed_Perf v(...) (5 times)
   >> read: [read,ed] v_v_ed_Past v(...) (5 times)
   read: [read,ed] v_v_ed_Passive v(...) (3 times)
   >> read: [read] v_v_Not3ps v(...) (5 times)
   >> read: [read] v_v_Inf v(...) (5 times)
   in: p(...)
   their: possdet(...)
   proper: adj(...)
   place: n(...)
\end{verbatim}
\end{figure}
\normalsize

The verb `read' has 5 distinct subcategorisations in the CLE. Of
these, three are capable of appearing in the passive form. On
morphological grounds, therefore, we have a total of 23 distinct
possibilities for the morpheme `read'. (These entries contain semantic
information: if this were factored out some collapsing of entries
might be possible. But the general point will still remain).

Entries signalled with \verb!>>! are those that were pruned out
because they did not overlap with the tags produced for that sentence.
Singular nouns (`can') do not usually follow plural determiners. It is
statistically very unlikely that a simple past, a
non-third-person-singular present, or an infinitive form of a verb
will be found immediately following `be'. The LOB tag set does not
distinguish passive from perfective forms of verbs (they are all
'VBN') and so both the perfective and the passive CLE forms are
retained.

This illustrates one kind of lack of constraint caused by the mismatch
in tag sets: if the tagger had been trained on CLE categories it would
have eliminated the perfective (`vform=en') alternative as this only
occurs (or should occur) following a form of `have': correspondingly,
following `be' a passive (`vform=passive') is much more likely.

Parsing and semantically analysing the original sentence took
50.4 and 1.1 seconds respectively, on a Sun Sparcstation 1 (a machine
two generations behind current hardware).
Parsing and analysing the pruned version of the sentence took 16.4 and
0.5 secs. In both cases only one QLF was obtained - the correct one.

Unfortunately, this splendid result was not uniformly obtained.  The
averages over all 124 sentences are summarised 
in figure~\ref{Tagging-figure-2}.

\begin{figure}
\caption{Experiment 1}\label{Tagging-figure-2}
\small
\begin{verbatim}
   ACCURACY
   Unaffected  Pruned  No Parse   Bad QLF  %accuracy
      35         89      12          4        87
   
   SPEED    (successful parses only)
   Averages        #Entries  Parsing  #Parses Sem #QLFS
   Without Pruning   17.7     15.3    8.8     5.9  3.6
   With Pruning      13.7     9.9     7.6     4.8  2.8
   % Difference      22       35      13      18   22
\end{verbatim}
\end{figure}
\normalsize
As can be seen, for almost 30\% of the sentences, tagging produces the
same set of entries as the CLE does. In cases where some pruning takes
place, two things can go wrong. Either no parse can be found for the
sentence at all, since an entry crucial to a successful parse has been
pruned out, or a parse and a QLF are found, but the `wrong' one: i.e.
not the one ranked top by the system on the previous run. For the
sentences taken overall this gives an error rate of almost 13\%.

Against this, in those cases where analysis is successful, we get an
improvement in parsing speed of 35\%, and a respectable improvement in
speed of semantic analysis. While useful, my own reaction to this
finding was that the tradeoff between speed and accuracy was that it
was not profitable enough. Losing over 12\% accuracy for only 35\%
speed up does not seem like a very good trade in most applications,
especially when one remembers that we have in any case excluded those
cases where the parser fails even where all the lexical categories are
present.

One optimistic note is that the number of bad QLFs that were obtained
was relatively small. The biggest contribution to overall accuracy are
the large number of cases for which no parse at all can be obtained.
Inspection of a few of these cases revealed that in some of these
cases the tagger had made a decision based on local preferences, that
was not always optimal when the sentence was taken as a whole.  N-gram
tagging techniques are of course always `shortsighted': there will
always be cases where a local cooccurrence possibility is
overwhelmingly more likely than the next best alternative on local
grounds, but actually less plausible when a bigger context is taken
into account.

\subsection{Experiment Two}

One way of overcoming the deficiency noted above would be to move to a
trigram-based system. However, even a 3 word window will not always be
sufficient to make the correct decision.  A simpler alternative would
be to be less ruthless in the elimination of tags, on the assumption
that by relaxing the requirement on local plausibility, enough (more
globally plausible) candidates would be retained so that the correct
parse can still be obtained.

In the first experiment, assignment of tags to a word by the LOB based
tagger was deterministic: one word, one tag. In the mapping to CLE
categories, this determinism was not always preserved, but
nevertheless no `close seconds' were transmitted from the tagger.

De Marcken (1990) describes how to use thresholding to vary the number
of hypotheses delivered by a tagger. The system assigns a tag to a
word in the usual way, and then adds to the chosen tag all those tags
that come within some threshold of the winner's score. The
thresholding is achieved by multiplying the score of potential
alternatives by some factor, and doing a comparison with the score of
the actual winner.

If the factor is 1, the result is as for the deterministic case, since
this will never increase an alternative's score.  By setting the
factor to be very large, no elimination of tags at all will be
achieved, because every alternative tag will then exceed the
threshold.  Between these two extremes it ought to be possible to
retain only those tags for a word that are reasonably likely, while
still eliminating those that have only a very small chance of
contributing to the final parse.

Thus for the second experiment we modified the tagger so as to be able
to vary the factor governing how many tags for each word would be
retained.  Again, the hope was to determine an acceptable tradeoff
between accuracy and efficiency.  Ideally, one would hope to find that
tagging with one particular factor, and then using that for pruning,
would retain all the lexical entries that were necessary for finding
the correct parse, while still filtering out enough to give a useful
gain in parsing speed.

As an illustration of the effects hoped for, consider the example
in figure~\ref{Tagging-figure-3}.
\begin{figure}
\caption{Tagging at different factors}\label{Tagging-figure-3}
\small
\begin{verbatim}
  BEFORE:
   and: conj(...) 
   then: conj(...) 
   then: advp(...)
   all: det(...)
   failed: [fail,ed] v_v_ed_Perf v(...) (3 times)
   failed: [fail,ed] v_v_ed_Past v(...) (3 times)
   failed: [fail,ed] v_v_ed_Passive v(...)
   
  AFTER (Factor= 1):
   and: conj(...) 
   then: advp(...)
   all: det(...)
   failed: [fail,ed] v_v_ed_Perf v(...) (3 times)
   failed: [fail,ed] v_v_ed_Passive v(...)
    
  AFTER (Factor=300):
   and: conj(...) 
   then: advp(...)
   all: det(...)
   failed: [fail,ed] v_v_ed_Perf v(...) (3 times)
   failed: [fail,ed] v_v_ed_Past v(...) (3 times)
   failed: [fail,ed] v_v_ed_Passive v(...)
\end{verbatim}
\end{figure}
\normalsize 
With no pruning we have 2 entries for `then', and 7 entries for
`failed'. Tagging deterministically, with a factor of 1, eliminates
the unlikely entry for `then' (this is not the `then' of `if-then')
but also, for some reason presumably to do with the character of the
LOB corpus, eliminates the simple past entries for `failed',
preferring the entry appropriate in phrases like `all failed
politicians'. (As before, this entry yields two CLE possibilities).
Unfortunately, there is no parse for this sentence with these entries.

Tagging with a factor of 300 retains all those entries which are such
that their score, if multiplied by 300, equals or exceeds that of the
winner. \footnote{This seems like a big number. But probabilities get
very small very quickly. Factors of 50 or more are needed, at least
with the statistics we were using, to make any difference at all.} In
this case, the `conjunction' entry for `then' is correctly eliminated,
while the simple past entry for `failed' is retained.

The experiment was repeated with several different factors. The
results are given here for a factor of 300.  However, while 300 was
the best number we could find, it did not prove to be a trade-off that
was spectacularly better than the results of the previous experiment.
The results are summarised in figure 4, with those of the previous
experiment repeated for easy comparison.

\begin{figure}
\caption{Experiment 2}
\small
\begin{verbatim}
ACCURACY
Factor  Unaffected  Pruned   No Parse  Bad QLF  %accuracy
  1        35         89       12         4        87
 300       41         83       9          3        90

SPEED
Factor=1  (successful parses only)
Averages        #Entries  Parsing  #Parses Sem #QLFS
Without Pruning   17.7     15.3    8.8     5.9  3.6
With Pruning      13.7     9.9     7.6     4.8  2.8
 Difference      22       35      13      18   22

Factor=300 (successful parses only)
Averages        #Entries  Parsing #Parses Sem #QLFS
Without Pruning  17.5      14.5    8.3    5.7   3.5
With Pruning     13.9      10.2    7.2    4.7   2.8
 Difference     20        29      13     17    20
\end{verbatim}
\end{figure}
\normalsize
In general these results are in the right direction, though they do
not go far enough, unfortunately.  Roughly speaking, we get accuracy
of a little over 90\%, with efficiency gains of around 30\%. It is
still the case that the major source of inaccuracy is failure to find
a parse.

These figures conceal one small fact which is worth remarking on. It
was not the case that the parse failures on the second run were a
subset of those on the first run. This somewhat curious state of
affairs is explained by the pruning regime used.  Recall that for each
word, the set of categories derived from the tagger were intersected
with those assigned by the CLE. Where the intersection was not null,
only the intersection was retained.  Where it was null, the CLE
categories were retained.  On several occasions it was the case that a
previously null intersection became, on the second run, a non-null
intersection, and so the set of categories for the word seen by the
CLE was actually smaller on the second run than on the first. It was
thus possible for a sentence which had previously been parsed
correctly to now fail, and this indeed did happen on several
occasions.

Inspection of the remaining parse failures revealed that again the
major cause of error was the shortsightedness of the tagger, even when
relaxed as described. A typical example is shown in figure 5.

\begin{figure}
\caption{Over eager pruning}
\small
\begin{verbatim}  
  BEFORE:
   what: det(...)
   had: v(...) (4 times)
   he: np(...)
   told: [tell,ed] v_v_ed_Perf v(...) (6 times)
   told: [tell,ed] v_v_ed_Past v(...) (6 times)
   told: [tell,ed] v_v_ed_Passive v(...) (6 times)
   them: np(...)
   
  AFTER (Factor=300):
   what: det(...)
   had: v(...) (4 times)
   he: np(...)
   told: [tell,ed] v_v_ed_Past v(...) (6 times)
   them: np(...)
\end{verbatim}
\end{figure}
\normalsize
There are six different subcategorisations for `tell' in the CLE
lexicon. This means that there are 18 semantically and syntactically
distinguishable possibilities for the morpheme `told'. Here, pruning
removes all but the simple past entries for `told', even at the
relatively generous factor of 300. This is incorrect, for the `told'
is linked with `had' and thus is a perfective form.  Unfortunately,
`had' and `told' do not fall within the bigram `window' and so the
local preference of a finite form following a proper name overwhelms
the correct possibility.

Of course, the correct form will be retained if the factor is
increased.  Unfortunately, we discovered that if the factor is
increased to 400 or above, then while this achieves the desired result
of retaining the correct tag, it also achieves the undesired result
that {\it all} the other tags are also retained. Thus (on short
sentences like these, at least) the only way to preserve accuracy is
to abandon any increase in efficiency.

Notice that in the example above (and most of the others that failed
for this reason) the relevant dependency would fall within a trigram
window `... had he told ....'.  In the general case, of course, there
will always be some dependencies that will not do so: a similar
sentence could be constructed with arbitrarily long NPs intervening
between `had' and `told'. Thus although moving to a trigram based
tagger would improve performance on examples like these, there would
always be a residue of cases where analogous myopia would set in. In
these cases the tagger is likely to go wrong systematically and we
will be back in the same position.

\subsection{Conclusions}

This was in every way a preliminary study, and its findings should be
treated with caution.  It should be replicated using a tag set the
same as that used by the parser, and using statistics derived from a
corpus that has been analysed in a way consistent with the way the
parser would analyse it.  This would avoid the problems caused by the
mismatch of tag sets.  A larger number of examples should also be used
so that the results would be more reliable statistically. Sentence
length should also be controlled to see what interactions emerge
by comparing performance of short and long sentences. (We tried to do
this here but the numbers of sentences involved were so small that
nothing reliable emerged.)

We would be very surprised if the results of a more accurate study were
not somewhat more encouraging than those found here.  However, we would
also expect that there would still be a sizeable residue of cases
involving non-local dependencies which would still not be covered
properly. For some corpora or applications this residue might be small
enough that the efficiency/accuracy tradeoff would be favourable.  For
others, this would not be the case.

There are several obvious places where statistical information could
be used to good effect which the experiments described above did not
tackle. The most obvious fact about the few example sentences
discussed is the large number of subcategorisations given for each
verb. Tag sets like that used by LOB and others do not distinguish
between different subcategorisations for verbs, for the good reason
that this is not usually determinable on a local bi- or tri-gram
basis. However, if it were possible to eliminate unlikely
subcategorisations this would have a large impact on efficiency. (Of
course, it might also have a large impact on accuracy).

\chapter{Applications Development Reports}

\section{DRA Malvern}

{\it Sue Browning}

\noindent
The Speech Research Unit at DRA Malvern is primarily concerned with fundamental
research into speech recognition techniques and algorithms. In 1989 we produced
a demonstrator of speaker-dependent, task specific continuous speech
recognition with a vocabulary of around 500 words, and we have just completed a
more advanced, speaker independent version of that system. We are currently
working towards task-independent large vocabulary recognition, among other
things.  As vocabulary size and task complexity grow we are becoming
increasingly concerned with how speech can be integrated into a complex
man-machine interface, and we see NLP as of vital importance to the future
effectiveness of such interfaces.

Our interest in NLP therefore stems from the need to model (spoken) language in
order to both constrain the recognition and to interpret the recognised speech
in terms of the task domain.  Currently most of our direct involvement in NLP
is through projects contracted out to universities and others (including SRI).
These are mainly investigations into the suitability of various techniques for
language analysis and generation, with the emphasis on techniques that deal
with ``real'' language, rather than artificial examples, as well as on those
which are extendable to deal with the spoken, as well as the written, word.  We
are also extending our interest to look at speech dialogue management
techniques, in which it is assumed that ``natural'' language can be both
interpreted and generated.

As an application in which to develop and evaluate future speech and language
technology, we have chosen Autoroute, a commercially available software package
for route planning.  The user supplies the start and destination points of a
car journey in the UK, along with any places to be visited or avoided along the
way, and Autoroute calculates the best route according to other user-specified
variables, such as preferred road types and driving speeds. It was thought that
this application would provide a relatively constrained environment, but one
which would be rich enough to be interesting and challenging. Therefore, it is
around the Autoroute application that most of our involvement and experience
with CLARE revolves.

The first project which involved the use of CLARE at DRA involved the building
of an interface between CLARE and Autoroute.  The project was completed in May
1991 by Gordon Brown, a student working in the SRU.  Gordon spent some time at
SRI learning about the CLARE software and then was able to build a prototype of
the interface fairly easily, although he had had no previous experience of
either Prolog or NLP.

Version 1 of CLARE was used on a DEC 3100 workstation running Quintus Prolog
v2.5, which was connected, by means of a commercially available program called
pcANYWHERE, to Autoroute running on a Dell PC.  It was necessary to extend the
lexicon to cover the many new words in the route planning task, and this was
done quickly and easily using VEX.

The interface took the TRL output from CLARE and searched it for pre-defined
patterns of words to fill in templates containing the information required by
Autoroute to process a query.  These were then passed to the PC as control
codes for Autoroute.  The reply from Autoroute was easily returned because
pcANYWHERE transmits the PC screen back to the Decstation. It was also possible
to change the display type from map to table, but not all of the functionality
of Autoroute was built into the prototype interface.

A small number of genuine queries elicited by questionnaire were collected  as
a sample on which to test the combined system.  Detailed results are available
in Gordon's report, which is part of the CLARE project documentation, but
briefly, only around 9\% of the 146 unique queries submitted got an appropriate
answer from Autoroute, 74\% failed to get through CLARE, and 17\% failed at the
interface stage.  The majority of  CLARE failures were due to there being no
parse for the input,and this was usually caused by the use of non-standard
(telegraphese) and ungrammatical constructions in the queries, and showed up
important areas where the CLE could be extended to cover this type of language.
Interface failures were almost always due to the pattern matching not being
sophisticated enough, and this would clearly need extending in any future
application.

It was clear from this experiment that a much larger body of data about the
kinds of queries that would have to be handled would be crucial to further
development of the Autoroute application.  Therefore, we are currently
producing a corpus of genuine spoken queries to the Autoroute system.
Using ``Wizard of Oz'' techniques we simulate powerful speech recognition,
natural language understanding and speech output in a genuine
telephone-based route planning enquiry system, and have already recorded a
quite sizeable corpus of spoken interactions.  This corpus not only allows
the techniques used in systems like CLARE to be tested on and adapted to
``real'' spoken language (in a restricted domain), but also allows the SRU
to study the speech produced by callers to such systems, both in terms of
what kinds of things people say, and in terms of how they react to such a
system, and therefore how the dialogue can be designed to accommodate them.

A second aspect of our involvement with CLARE stems from our concern with how
speech and language technology can best be integrated. There are two basic ways
in which NLP can aid speech recognition; by working in parallel with the
recognition algorithms (or in some sense integrated into them) to make
predictions that might constrain the recognition task, or by taking the output
of the recogniser and analysing it to find the most likely interpretation or
reject unlikely interpretations.  The ASLU (Architectures for Spoken Language
Understanding) project which the DRA supports at SRI is investigating how CLARE
might be integrated with the SRU's speech recognition techniques and is using
the route planning corpus.

More recently, we have had a very short exploratory project to investigate
techniques for spoken dialogue management.  The exchange between the user and
Autoroute is viewed as a conversation in which the dialogue manager
directs the conversation so as to extract the information  necessary to
construct a valid Autoroute query from the user in as efficient a manner as
possible.  The strategy employed to do this will depend on the user's degree of
familiarity with the task and the system, and the quality of the communications
link etc. The dialogue manager would also control the error handling and
correction facilities which will be needed if the input is naturally spoken
queries, due both to errors in the recognition process, and to the potential
ambiguity of natural language and the inevitable lack of coverage in the NLP
component. The project assumed the availability of a fairly sophisticated NL
analysis and generation system (to generate appropriate queries and
confirmation messages). Ultimately the intention would be to use something like
the CLE, but due to lack of time in this prototype the NL component was
simulated by a very simple pattern matching parser.  Although it was very
simple,  some interesting ideas about generic dialogue management techniques
were developed, which we intend to explore more fully in the future.

As well as further exploiting the route planning corpus to develop and evaluate
generic and task specific speech and language technology, we also foresee that
our interest in CLARE and NLP in general will have a slightly different focus.
Many of our possible speech recognition applications could involve
interrogation, manipulation, and maintenance of large complex databases, such
as are becoming increasingly  available in (military) command and control
situations. Many of these involve quite restricted task domains with
medium-sized vocabularies which have a rather rich structure.  As we extend the
power and flexibility of our speech recognition technology to these new domains
there will be a need to automatically produce linguistic models targeted at
these specific applications (something like sublanguage modelling). Although
each model may be rather simple, the ability to customise the linguistic
analyser automatically to new applications could be crucial to the take up of
our speech technology.

\newpage
\section{BP Research}

{\it Dave Wolstenholme}

\subsection{The context for natural language processing}
\subsubsection{Perceived Business need for natural language processing}
Information is the life-blood of a large organization such as BP.  Without
an adequate supply of good-quality information, individuals, teams and
other organizational units are unable to function to their best ability;
this clearly has a detrimental effect on the functioning of the overall
organization.  In our society, the volume and complexity of information and
data is continually increasing, and it is necessary to make use of
computers for storing it---frequently in databases.  The problem with
storing information in computer databases is that the people who might need
the information frequently have great difficulty in obtaining it.

This problem is, in part, related to a more general problem associated with
user-interfaces: that users of computer systems are often expected to learn
unnatural, formal command or query languages.  This need to communicate in
machine-oriented terms is widely perceived as adversely affecting the
acceptability of computer systems and as reducing the effective use made of
them by users; the means of communication should be more user-oriented,
e.g. through natural spoken or written English.

The problem of interacting with databases is, however, deeper than that.
When using a formal database query language, such as SQL, a user is
expected to have a fairly detailed knowledge of the way in which the data
are stored, such as the names of tables and fields.  This puts an unfair
load on database users, particularly on those who need to access several
databases, possibly using different query languages.  The more casual user
is, of course, at a serious disadvantage, and may, indeed, give up trying
to obtain the data.  There can be very few users, even so-called `computer
literate' ones, who have not been either unable to obtain the information
they require or frustrated and delayed in the attempt.  The information
stored is thus being under-utilized, and is less valuable than it might
otherwise be.

The problems identified above point to the need for a different way of
interacting with databases.  A good model might be the way we would obtain
information from a human intermediary who has that information stored in
some way, whether in paper form or on computer.  We would ask for the
information in English; we would not need to know where, or how, the
information was stored (it may be in more than one physical place)---this
would normally not concern us.  Something the human intermediary also does
is use common sense or specialized knowledge to bring together the various
individual items of data needed to answer a particular question.  So, for
example, if the human intermediary were asked how long Fred's boss had
worked for BP, he might look at a paper-based organigram to see who Fred's
boss was, then query a computer database to see when that person joined BP,
and, finally, use general knowledge of the current date to determine how
long this corresponded to.

If we look at the model of interaction with a human intermediary, we see
that the interaction is seldom a simple ask-question-get-answer type.
Apart from the dialogue required by social conventions---saying hello,
asking after each other's health, discussing the weather, etc.---there will
normally be dialogue preliminary to the main question, concerned with
establishing the background and reason for the question.  This provides the
context in which the question is set, and may be useful for helping the
intermediary to answer the question.  There may also be subsidiary dialogue
associated with the question; this may be instigated by the intermediary in
order to clarify any imprecisions or to sort out any ambiguities in the
question.  The answer too may not be straightforward.  It may be that the
question is outside the scope of the intermediary's knowledge, or that
there is some other reason, e.g. security, why the answer cannot be
given.  This must be conveyed to the questioner.  Even if a real answer can
be given, certain qualifications and provisos may be given with it.  For
example, the intermediary may make explicit certain assumptions upon which
the answer is based.

The CLARE Project, with its emphasis on interactive natural language
processing and on contextual reasoning and cooperative response, was seen
as a good vehicle for implementing NL database interfaces with interaction
modelled on that of a human intermediary.  These would address the two main
problems identified above: first, providing interaction that is clearly
human- rather than machine-oriented, and, second, ensuring that the user
need not be aware of the way in which data are stored.  They would also
enable a human-computer dialogue to exhibit many of the features found in a
complex human-human dialogue, as discussed above, such as using knowledge
to bring together raw data to answer a question, handling context, and
interacting to resolve ambiguities and explain in-built assumptions.

The main perceived benefits that would accrue to BP Businesses from the use
of natural language database interfaces were clear and simple: a)
information stored on computer would be more accessible and, hence, more
useful and valuable; and b) individuals needing to access data would feel
less frustrated, more in control, and less alienated from the
ever-spreading computers.  There would also be direct IT benefits, in that
the separation of the user from the direct form of data representation
would enable modifications to the representation to be made transparently.

\subsubsection{BP involvement in CLARE and NLP}
Those directly involved in the CLARE Project in BP have come from the
Information Management (IM) team, one of several teams in the Information
Science and Engineering (ISE) branch at BP's Research Centre,
Sunbury-on-Thames.  The teams in this branch are involved with advanced IT
research and development.  The IM team is concerned with many areas of
information and knowledge management, including knowledge-based systems,
database interfaces, computer-supported cooperative working, distributed
systems, multimedia and process modelling.  In addition, it employs
psychologists for applied human factors work.

Apart from participation in an earlier Esprit project, LOKI, which had some
NL content, the CLARE Project is BP's only involvement with natural
language processing.

\subsection{BP's experience with the CLARE Project}
\subsubsection{Development of a NL interface to a mock HR database}
BP has three core businesses: exploration, oil and chemicals.  One thing
that all three have in common is human resources.  For this reason, and
because the domain was felt to be more complex than a technical one, and
hence more challenging and more likely to raise interesting issues, it was
decided that a natural language interface to a mock HR database should be
built.

The database is modelled on BP's Sunbury research centre, and contains data
about various object types including divisions, branches, teams, buildings,
rooms, projects and staff.  Details of an individual member of staff
include date of birth, sex, whether married, number of children, address,
skills, and telephone numbers.  In addition, relationships between the
various objects are stored, e.g. organization structure, locations of
individuals and rooms, managerial posts and line management relationships.

The interface was firstly built on top of the result of the earlier Nattie
Project, CLE.  CLE was supplied with some demonstration interfaces,
including one to a Prolog database about Cambridge colleges.  Whilst this
proved a useful basis for developing the interface to our HR database,
which was also to be built first in Prolog, the general approach taken in
it was seen to be inadequate for our intended purpose---to build an
interface to a database held in a commercial DB system, accessed via SQL.
The main reasons for this inadequacy were:

\begin{itemize}
\item identical representation of objects and relations in CLARE and the
Prolog database program.  Whilst useful for building a quick demonstration,
this would be quite unrealistic for a real system, where the database
representation might be quite different from the internal CLARE
representation.

\item the concept of events is important within the CLARE system, where
the first argument of a verb represents an event.  This representation was
carried over directly into the colleges database, where each event was
explicitly named, and the time of each event was recorded separately in the
database.  Again, this would be unrealistic in a real database.

\item clauses for a particular relationship in the Prolog database were a
mixture of general rules and database facts.  In a real system, it is
likely that only raw facts would be stored in the database, while general
rules about them would be stored elsewhere, e.g. in a knowledge base.
\end{itemize}

\noindent
The following clauses from the colleges database demonstrate these three
points.
\begin{quote}
\begin{verbatim}
build_make(Event, Person, Building) :-
           found1(Event, Person, Building).
build_make(event40, wren1, trinity1).

found1(event1, beaufort1, christs1).

in_time(event1, year(1148)).
\end{verbatim}
\end{quote}

\noindent
Therefore, the interface was designed with the following characteristics:
\begin{itemize}
\item the database containing the basic data (e.g. who is in which room,
which room is in which building), was conceptually quite separate from the
knowledge base.  The knowledge base included general rules for inferring
information from the raw data, e.g. which person is in which
building.

\item clean interfaces between CLARE and the knowledge base, and between
the knowledge base and the database.  All objects and relationships were
mapped between different representations at these interfaces.  Although
both the knowledge base and database, as well as CLARE, were initially
implemented in Prolog, this design allowed an easy porting to an Oracle
database later.

\item a function-like representation of events in
arguments of relationships in the knowledge base.  The following clauses
demonstrate these.

\end{itemize}

\begin{quote}
\begin{verbatim}

    clare_KB_pred(work_on_IsEmployedOn, work_on).


    work_on([work_on, Person, Objective], Person, Objective) :-
        isa(Objective, objective),
        'works on'(Person, Objective).
    work_on([work_on, Person, Objective], Person, Project) :-
        isa(Project, project),
        'objective of project'(Objective, Project),
        'works on'(Person, Objective).


    'works on'(Person, B) :-
        db_query('works on_db',
                 [[person, Person], ['work object', B]]).


    db_query(Pred, [[T1, Var1], [T2, Var2]]) :-
        'KB_DB_object'(T1, Var1, V1),
        'KB_DB_object'(T2, Var2, V2),
        Term =.. [Pred, V1, V2], call(Term),
        'DB_KB_object'(T1, V1, Var1),
        'DB_KB_object'(T2, V2, Var2).
\end{verbatim}
\end{quote}

\begin{figure}
\begin{center}
\setlength{\unitlength}{0.0125in}%
\begin{picture}(175,300)
\thicklines
\put(120,296){\oval(92,40)}
\put(120,41){\oval(112,40)}
\put(145,86){\vector(0,1){20}}
\put(145,61){\vector(0,1){20}}
\put(135,86){\line(1,0){20}}
\put(135,81){\line(1,0){20}}
\put(85,81){\line(1,0){20}}
\put(85,86){\line(1,0){20}}
\put(95,81){\vector(0,-1){20}}
\put(95,106){\vector(0,-1){20}}
\put(85,181){\line(1,0){20}}
\put(85,186){\line(1,0){20}}
\put(95,181){\vector(0,-1){20}}
\put(95,206){\vector(0,-1){20}}
\put(145,186){\vector(0,1){20}}
\put(145,161){\vector(0,1){20}}
\put(135,186){\line(1,0){20}}
\put(135,181){\line(1,0){20}}
\put(65,206){\framebox(110,50){}}
\put(65,106){\framebox(110,55){}}
\put(95,276){\vector(0,-1){20}}
\put(145,256){\vector(0,1){20}}
\put(0,70){\makebox(0,0)[lb]{\em SQL Queries}}
\put(0,91){\makebox(0,0)[lb]{\em Prolog Queries}}
\put(0,171){\makebox(0,0)[lb]{\em Prolog Queries}}
\put(0,193){\makebox(0,0)[lb]{\em Logical Forms}}
\put(0,263){\makebox(0,0)[lb]{\em Language}}
\put(0,276){\makebox(0,0)[lb]{\em Natural}}
\put(120,36){\makebox(0,0)[b]{\rm Oracle DB}}
\put(120,123){\makebox(0,0)[b]{\rm Base}}
\put(120,141){\makebox(0,0)[b]{\rm Knowledge}}
\put(120,231){\makebox(0,0)[b]{\rm CLARE}}
\put(120,293){\makebox(0,0)[b]{\rm User}}
\end{picture}

\end{center}
\caption{Architecture of CLARE-Oracle prototype}
\label{BP-figure}
\end{figure}
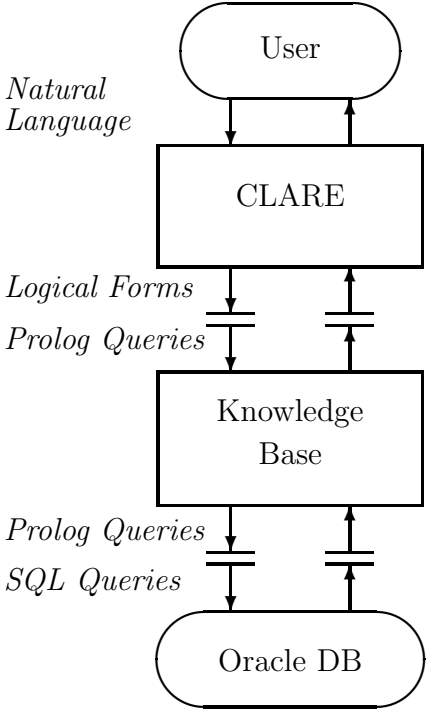

The interface was later modified for a setup where the basic data were held
in an Oracle database (see Figure~\ref{BP-figure}).  Apart from creating
and linking to the database itself, this involved making changes only to
the database query procedure and to the relevant object conversion
procedures.  The Oracle database was resident on an HP machine, while CLARE
and the knowledge base ran under Quintus Prolog on a Sun.  The database
query procedure made use of a Quintus Prolog-Oracle SQL interface
package, ProDBI (now from Keylink Computers Ltd), for generating SQL
queries to Oracle for solving the lowest-level relationships.  The link
across the network between the Sun and the HP was made via SQLNet.  No
optimization of SQL queries was carried out; whenever a knowledge-base
relationship needed information from the database to be solved, an SQL
query was made, in the same way that a Prolog query would normally have
been made.  This approach would not be suitable for large databases.  This
approach contrasts with that taken later in the CLARE Project where several
queries are effectively bundled into one more complex query.

The system was later ported onto CLARE, version 1.  This conversion was not
trivial, as there were many internal changes between CLE and CLARE 1. CLARE
1 allowed handling of ellipsis and paraphrase generation.  A typical
sequence of questions and answers is shown next:

\begin{quote}
{\em Where does Dave Smith work?}\\\nopagebreak
Information Management, ISE, Room 25, Building 200,  Sunbury, etc

{\em Which room does he work in?}\\\nopagebreak Room 25

{\em What year was he born in?}\\\nopagebreak 1953

{\em What is his boss's phone number?}\\\nopagebreak 3256

{\em Who is Humphrey's boss's line manager?}\\\nopagebreak Jim Boon

{\em Who works in Information Management and 
knows about fuel technology?}\\\nopagebreak John Corner

{\em Is he married?}\\\nopagebreak No	

{\em Who is the oldest person in the ISE?}\\\nopagebreak Eric

{\em Who is older than his boss?}\\\nopagebreak Eric

{\em How many women are there in Jim's branch?}\\\nopagebreak 5

{\em In which team are more than two women working?}\\\nopagebreak Information Management

{\em Does Mark know about HCI?}\\\nopagebreak Yes

{\em Logic programming?}\\\nopagebreak No

{\em Does Dave?}\\\nopagebreak Yes

\end{quote}

\noindent
The system has been demonstrated to many parts of BP, and to external
companies, and has been favourably received.  The ability to ask complex
queries with a certain amount of indirection in them, e.g. {\em What
is his boss's phone number?}, and {\it ad hoc} queries for which it would
be unlikely that pre-defined templates/screens would be provided, {\it
e.g.} {\em In which team are more than two women that know about HCI
working?} was particularly liked.  The fact that the architecture does not
restrict the data source to be just one database, so that the knowledge
base could be given access to several databases in order to answer a query,
was also appreciated.

The architecture chosen, where objects and relationships needed to be
represented in several different ways (in English, in CLARE, in the
knowledge base and in the database), led us to develop tools to aid system
construction.  These tools were written in MacProlog, where advantage could
be taken of its multi-windowing environment.  The first tool aided the
development of a `raw database', i.e. definitions of the class hierarchy,
objects, atrributes and basic binary relationships between them.  The
second provided for taking this raw database and producing all the Quintus
Prolog code required for the system, i.e. the CLARE definitions, the
knowledge base, the database and the interface/conversion procedures.  The
code resulting from this second tool was a combination of Prolog code and
CLARE definitions generated from the raw database and directly written
code.  The system required both that a domain model be built and that many
extensions to the lexicon be provided, and the tool greatly aided their
provision.  For example, it provided easy ways of defining synonyms for
nouns, adjectives and verbs, and of defining type declarations and
relational forms of nouns.  The following are example declarations given
within the tool, from which CLARE declarations for the given nouns and
verbs were directly generated.

\begin{verbatim}
type_sort(section_leader, count, section_leader_LeaderOfSection, person).
rel_pred(section_leader, section_leader_SectionLeaderOfPerson, person).
rel_pred(section_leader, section_leader_SectionLeaderOfSection, branch).
synonyms(section_leader, [[section, leader],[leader], [project, leader],
         [group, leader], [team, leader]]).

type_sort(computing, mass, computing_ComputingSubject, subject).
synonyms(computing, [[computing], [computing, science],
                                  [computer, science]]).

clare_verb(know_about, know_about_IsFamiliarWith, [person, subject]).
verb_synonyms(know_about, [v_subj_specpp([know, about]),
                          v_subj_obj([understand])]).
\end{verbatim}

\noindent
Because of the perceived need to generate new code automatically when
changes were made, the VEX tool was not used in developing the system.  The
system did not require changes to the built-in sortal hierarchy, and
changes to the built-in grammar were not made directly.  However,
development of the prototype did highlight several shortcomings of the
grammar, which was modified accordingly by SRI.  For example, the
construction where an object is named by its type followed by a
subname/number, e.g. `building 200', `room 12A', or `project
lionheart', was found to be a very common requirement which could not
originally be handled.

No direct use has been made of versions 2.0 or 2.5 of CLARE, except for
demonstrations, which have, again, been well-received.

\subsubsection{The future of CLARE and NLP in BP}

Although CLARE and the prototypes built
using it have been well-received, none of the BP Businesses have allocated
direct funding for any further development work, or for continued research
into NLP.  The general perception is that

\begin{itemize}
\item the costs of developing NL database interfaces is high, whilst the benefits obtained are difficult to quantify;

\item much work is still required to provide an acceptable HCI;

\item NL database interfaces face competition from other means of access,
e.g. graphical interfaces.  Whilst these may not provide the
naturalness of NL interfaces, and generally do not hide the users from the
database structure, they are found to be acceptable and flexible enough for
the domains of interest to BP.

\end{itemize}
\noindent
Related to this last point, BP do see benefit in NLP related to text files,
where there is less competition from other means.  Potential applications
include mail filtering and summarization, first step translation of foreign
language documents, and checking specifications and operator manuals for
ambiguity.

Future use of CLARE, and of NLP work generally, within BP has been put
further into doubt by recent events.  BP has taken a strategic decision to
concentrate on its core Businesses and to halt all internal IT research and
outsource IT development. This has led to the disbanding of ISE from the
end of 1992.

\newpage
\section{BT Labs}

{\it Gavin Meggs and Sandra Williams}

\noindent
BT Labs. has maintained active support and interest in SRI Cambridge's 
CLARE project throughout its lifetime.  Lack of resources and manpower 
have restricted the amount of work carried out at BT Labs using CLARE, 
except for the three short-term projects described below:

\subsection{Sept - Nov 1990 Geoffrey Sampson's evaluation of the July 1989 
version of the CLE}
Geoffrey Sampson spent 12 weeks working at BT Labs on a short-term 
fellowship.  He evaluated the CLE using a small corpus of naturally 
occurring queries elicited from employees of BT Labs.  Although his 
corpus was too small to obtain statistically significant results, he 
found that the performance of the CLE compared favourably with that of 
other major NLP parsers and semantic analysers.  
He performed an extensive investigation of the coverage of the 
lexical, grammatical and semantic components of the CLE.  He 
criticised the many gaps in coverage, the idiosyncrasies of the 
system, and the lack of robustness, mentioned above.  As quantified
in the coverage evaluation chapter, linguistic coverage has been
considerably extended under CLARE and we hope the ARLA project will bridge 
other gaps.

\subsection{Summer 1990 - April 1991 Interfacing CLARE to BT's SPUD database}

This work was carried out in an attempt to direct some of  the work of 
BT Labs. Natural Language Processing Group towards a practical 
application.  It was a small-scale project carried out by one person 
over 9-10 months and called BARNACLE (Business Application of a 
Relational Network database with the Aid of the Core Language Engine). 
 Since the work was to use a new database which was being developed in 
another part of BT, a great deal of time was spent liaising with the 
database developers and learning about its proposed structure.
BT's SPUD (Service Protection Utilities Database) is a relational 
database implemented in ORACLE.  For the CLARE application, a 
``snapshot'' of the database under development was used which contained 
37 database relations.  The SPUD developers provided a corpus of 40 
database queries in English and the equivalents of the first five in 
SQL
The database ``snapshot'' was simulated using the Quintus Prolog 
database interface to  represent the database relations as Prolog 
clauses.  The BARNACLE system was designed with three major parts: a 
Prolog translator to translate CLARE's LFs (Logical Forms) to RXs 
(relational query expressions);  application-dependent extensions to 
CLARE; and  software to control the system.  CLARE's TRL was not used 
because at the time, summer 1990, it was too ill-defined.
The LF to RX translator was implemented successfully with the LFs 
simplified into as near flat lists as possible prior to translation.  
The control software was never implemented in the short project life.  
 The emphasis of this project was more linguistic than that of the 
project described above, with more work on application-dependent 
extensions to CLARE.  A domain-dependent lexicon was constructed and 
an interface to the Oxford Advanced Learner's Dictionary was 
investigated.  Effort was directed towards trying to resolve gaps in 
CLARE's coverage within the restricted domain. 
The conclusions as a result of this work were that CLARE ought to be 
made more robust.  There are large gaps in the coverage and the system 
(at the time) fell over if it came across anything it did not 
recognise.  Analysis of most sentences in the sample corpus failed.  
The robustness problem will hopefully be addressed in the next phase, 
ARLA.

\subsection{April-May 1992: The Use of CLARE in Intelligent Query}

Research into the field of Intelligent Query necessitates the study of systems
which provide the user with an interface which is easily understood, yet
has the complexity to reason with the query so that relevant information
may be accessed. The most natural choice for a language which the user will
understand is natural language, but until recently natural language systems
have been incapable of providing adequate reasoning capabilities. CLARE,
at first glance, appears to overcome this restriction.

CLARE-2's PRM domain model, and associated literature, shows how a natural
language processing system may be interfaced with a fictional database of
Prolog facts.  Preliminary evaluations of CLARE-2 have shown that it is
also possible to interface CLARE-2 with a relational database and to build
a simple domain model based around the structure of the PRM model.

The PRM data was translated into relations which were stored in a relational
database, using the ORACLE database management system. An interface was
then designed which took the low-level database queries made by CLARE-2
and transformed them into SQL queries. These queries were made via ORACLE
to the (now relational) PRM database. Performance was seen to be almost
identical to that of the original system, with the exception of recursive
table queries which gave error messages. It should be noted that these errors
were not a fault of CLARE-2, but of the method of translation from Prolog
queries to SQL queries and should be overcome by the forthcoming SQL/CLARE-2
interface being developed by SRI.

The idea of Project Resource Management (PRM) was taken up as a suitable
test-bed for building a domain model. Data on local projects and individuals
was collected and stored both as Prolog facts and relations within a relational
database. The sequence of steps outlined in the CLARE-2 software manual
(Oct. 1991) and end of year report (Nov. 1991) were then followed in order
to build a new domain model.  The resultant complete domain model can be
used to extract information about real projects and real individuals in
a real domain using queries typed in English.

The steps for building a domain model explain how to build a declarative
representation of the domain that can be used for reasoning by the CLARE-2
system. The majority of the steps were found to be quite straightforward
and their implementation was simple. Some difficulty was encountered in
defining assumptions for the domain, but by far the most difficult step
was the definition of 'equivalence rules'. For the implementation of this
step it was necessary to make heavy use of the PRM model as a template from
which to build new equivalence rules. It should also be noted that in building
the new domain use was made of the PRM lexicon files and so no measure of
the difficulty of building a domain specific lexicon was made.

Since the development of the above system additional literature has been
provided by SRI which describes in more detail the process for building
a domain model. This information both clarifies the use of assumptions in
query processing and goes some way to simplifying the process of building
equivalence rules. Using the updated 'guide to building a domain model'
we hope to use CLARE to interface with a more complex domain database, which
has yet to be identified.

\newpage
\section{British Aerospace Involvement with the CLARE Project}

{\it Phil Sims}

\label{bae}

This section briefly describes the background to the Sowerby Research
Centre's 
(SRC) involvement with the CLARE project. It also describes the SRC's 
experience to date with the software and the 
future plans to continue work in the area of Natural Language Processing
and  related applications.

British Aerospace joined the CLARE consortium on the basis of its
experience as a member of the previous NATTIE  project. The results gained in
the development of the Core Language Engine (CLE) had been extremely promising,
and the goal of the CLARE project to show the CLE working in an embedded
application, and incorporating further contextual reasoning was in line
with the company's own view of where the next progress should be made. 

Each version of the CLARE software has been installed and runs on the Sun
Network at the Sowerby Research Centre. It has been demonstrated to a
large number of senior management in the course of visits made to the
Centre. An early demonstration of queries to an possible in-house project
management 
system was used to demonstrate the basic principles behind the system, and
time was spent with SRI working on this application.
Further plans have been drawn up to extend this work to a full
demonstration of the 
software's capability by using it to access the Project Management software
which has recently been installed and which is now used throughout the
Research Centre to plan and monitor all of its projects. 

Two other pieces of work currently under discussion with the British
Aerospace  may also lead to a future use of the software:
\begin{description}
\item[Automatic Document Generation] British Aerospace produces a number
of extremely complex systems. The documentation of such systems often runs
into many volumes. Whilst much of this documentation often goes unused it
is a contractual requirement, and represents a significant amount in
project's overall cost. This is particularly true when modifications are
made to the system, or slight variants are made to an existing product
for a new customer. A new set of complete documentation is required and
current methods of production are being investigated to improve cost
efficiency. 
As a result, three lines of research have been proposed in this area:
\begin{itemize}
\item The automatic production of documentation from appropriate models of
the relevant system - Standard Logical Form, and the generation software
in CLARE will be relevant. 
\item Producing documentation in an electronic as well as paper form.
\item Building interfaces to any resulting electronic version of the
documentation - Natural Language will be considered in conjunction with
HyperText interfaces.
\end{itemize} 
\item[Message Interpretation] Another area of British Aerospace business is
in providing communications systems. Recent developments in the area mean
that in order to stay competitive, systems must provide some degree of
extra processing. In the first instance this will be parsing and error
correction in the the structured headers used to introduce and close
messages, but the next step will be to extend this approach to an analysis
of the free text included within the message, for error correction,
paraphrasing and recording in a historical database. Again, we hope to
deploy some of the CLARE based software and technology in this area.
\end{description} 

The Sowerby Research Centre feels that it has gained appreciably through its
involvement with the CLARE consortium, both in terms of the software it has
received, and from the more general transfer of natural language processing
technology from SRI, Cambridge University and the other partners. Although
the SRC has been resource limited throughout the project, thus restricting
its own ability to fully exploit the work, it is hoped that new areas of
work which are taking place in conjunction with the operating companies
will lead to an increased level of commitment to exploiting the obvious
technological lead which CLARE could give the company.

\newpage
\section{SRI and Swedish partners: A speech-to-speech translation system}
\label{slt}

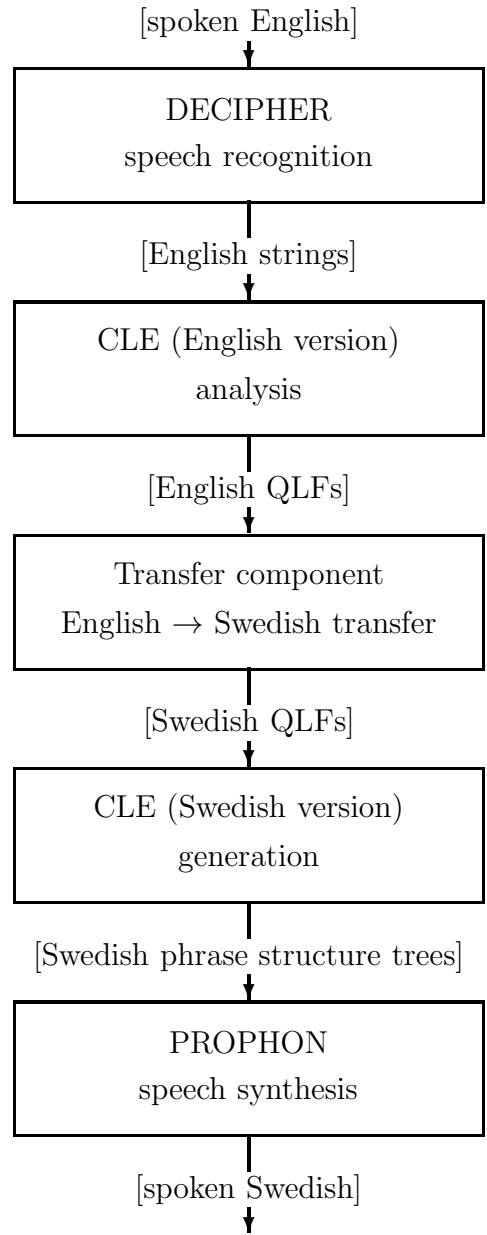
\begin{figure}[h]\centering
\setlength{\unitlength}{.9 ex}
\begin{picture}(40,100)
\thicklines 
\put(1,98){\makebox(38,3)[c]{[spoken English]}}
\put(20,98){\vector(0,-1){2}}
\put(1,85){\framebox(38,10.7)[c]{}}
\put(1,91){\makebox(38,3)[c]{DECIPHER}}
\put(1,87){\makebox(38,3)[c]{speech recognition}}
\put(20,85){\line(0,-1){3}}
\put(1,79){\makebox(38,3)[c]{[English strings]}}
\put(20,79){\vector(0,-1){2}}
\put(1,66){\framebox(38,10.7)[c]{}}
\put(1,72){\makebox(38,3)[c]{CLE (English version)}}
\put(1,68){\makebox(38,3)[c]{analysis}}
\put(20,66){\line(0,-1){3}}
\put(1,60){\makebox(38,3)[c]{[English QLFs]}}
\put(20,60){\vector(0,-1){2}}
\put(1,47){\framebox(38,10.7)[c]{}}
\put(1,53){\makebox(38,3)[c]{Transfer component}}
\put(1,49){\makebox(38,3)[c]{English~$\rightarrow$~Swedish transfer}}
\put(20,47){\line(0,-1){3}}
\put(1,41){\makebox(38,3)[c]{[Swedish QLFs]}}
\put(20,41){\vector(0,-1){2}}
\put(1,28){\framebox(38,10.7)[c]{}}
\put(1,34){\makebox(38,3)[c]{CLE (Swedish version)}}
\put(1,30){\makebox(38,3)[c]{generation}}
\put(20,28){\line(0,-1){3}}
\put(1,22){\makebox(38,3)[c]{[Swedish phrase structure trees]}}
\put(20,22){\vector(0,-1){2}}
\put(1,9){\framebox(38,10.7)[c]{}}
\put(1,15){\makebox(38,3)[c]{PROPHON}}
\put(1,11){\makebox(38,3)[c]{speech synthesis}}
\put(20,9){\line(0,-1){3}}
\put(1,3){\makebox(38,3)[c]{[spoken Swedish]}}
\put(20,3){\vector(0,-1){2}}
\end{picture}
\caption{Architecture of the Spoken Language Translation System}
\label{sltfig}
\end{figure}

In this final section, we briefly report on a new project, aimed at
construction of a prototype speech-to-speech translation system, in which
the Bilingual Conversation Interpreter (BCI) appears as a subsystem. The
BCI itself consists of two copies of the CLE, for the source and target 
natural languages, together with a transfer component for converting between 
QLFs representations for these languages.
The other subsystems of the speech translation
system are the SRI DECIPHER speech recognizer (Murveit, Butzberger and 
Weintraub 1991), and the
Swedish Telecom Prophon speech synthesizer (B\"ackstr\"om, Ceder and Lyberg
1989, Ceder and Lyberg 1990). The project is being sponsored by Swedish
Telecom; it started in August 1992, and is initially planned to run for one
year. At the end of this period, the intention is to be able to demonstrate
a prototype system which will translate spoken English into spoken Swedish,
using a vocabulary of between 700 and 1000 words.  The proposed
architecture of the system is sketched out in Figure~\ref{sltfig}.

The system's domain will be defined by the well-known
ATIS corpus, a collection of about 10000 sentences relating to air
travel information. Typical examples of sentences from the corpus are
displayed below.
\begin{verbatim}
LIST FLIGHTS FROM DENVER TO BALTIMORE.
I'D LIKE TO GO AT FIVE FORTY.
WHAT MEALS ARE SERVED ON EASTERN FLIGHT SEVENTY.
SHOW ME THE MORNING FLIGHTS FROM BOSTON TO PHILADELPHIA.
I WOULD LIKE TO PLAN A FLIGHT ON AMERICAN AIRLINES.
HOW MUCH DOES IT COST TO USE THE AIR TAXI.
WHAT IS THE LEAST EXPENSIVE FLIGHT BETWEEN BOSTON AND SAN FRANCISCO.
SHOW ME THE ROUND TRIP FARE FOR THE U S AIR FLIGHT.
WHAT FLIGHTS GO FROM PHILADELPHIA TO SAN FRANCISCO WITH A 
                                     STOPOVER IN DALLAS.
\end{verbatim}
It is still somewhat early to make any definite promises about the
outcome of the project, but it should be noted that the BCI has several
concrete strengths in this type of application when compared to other
MT architectures. Firstly,
the high quality of the translation output becomes doubly important when
dealing with spoken language, since post-editing is in the nature of
things impractical. Secondly, the fact that the Core Language Engine is
basically a general natural-language processing device greatly simplifies
the task of switching domains (in this case, from car hire to airline
reservations); in fact, the preliminary adaptation of the English CLE to
the ATIS domain already produces plausible analyses for over 80\%
of the corpus sentences under 15 words in length.

The performance of a preliminary version of the system for tests on unseen
material carried out on the 25th of November are shown in the tables in
Figures \ref{ATIS-raw} and \ref{ATIS-cum}.
(These are scores for English analysis only, NOT translation).

\begin{figure}
\begin{tabular}{||c|c|c|c|c|c||} 
\hline
Length & Number of   &  Highest ranked   &  Some other QLF &  All QLFs   & No QLFs  \\
       & sentences   &  QLF correct      &    correct      &  incorrect  &  produced    \\ 
\hline
1-10   &    114      &       88          &       12        &        5    &   9 \\
\hline
11-15  &    62       &       36          &       10        &        10   &   6 \\
\hline
15+    &    24       &        2          &        4        &        5    &  13  \\
\hline
\end{tabular}
\caption{Test of English CLE grammar and preference heuristics on 200
         unseen ATIS sentences, 25/10/92.}
\label{ATIS-raw}
\end{figure}

\begin{figure}
\begin{tabular}{||c|c|c|c|c|c||} 
\hline
Length    &  Some QLFs        &  Some QLF       &  Highest ranked   \\
          &  produced         &  correct        &  QLF correct      \\ 
\hline
1--10     &     92\%          &     88\%        &     77\%        \\
\hline
1--15     &     91\%          &     83\%        &     70\%         \\
\hline
all       &     86\%          &     76\%        &     63\%         \\
\hline
\end{tabular}
\caption{Cumulative scores}
\label{ATIS-cum}
\end{figure}
\clearpage

\chapter{NLP Evaluation Study -- Report Digest}

\vspace*{5mm}

\begin{center}
{\Large\bf Evaluating Natural Language Processing Systems}\\[3mm]
{\large J R Galliers and K Sparck Jones}\\[2mm]
Computer Laboratory, University of Cambridge\\
New Museums Site, Pembroke Street, Cambridge CB2 3QG, UK\\[3mm]
December 1992\\[5mm]
\end{center}

This report is a comprehensive study of NLP system evaluation. It
provides a detailed analysis of the concepts and issues involved,
an extensive review of the state of the art as represented both
by actual evaluations to date and by published discussions of
evaluation methodology in, for example, workshop proceedings, and
suggestions for sensible approaches to evaluation.

Part 1 establishes a descriptive framework for addressing
evaluation. This refers on the one hand to systems and their
subsystems and components, and on the other to systems within
their settings, as together forming 'setups'. These are then
examined in detail for their elements and aspects relevant to
evaluation, via carefully developed examples. This emphasises,
for instance, the importance of system objectives and setup
purposes in relation to evaluation. The report then dissects
evaluation to identify and characterise its many constituent
notions drawing distinctions, for example, between evaluation
criteria, measures and methods, and again illustrating their
application via a set of related examples. The discussion
emphasises the need to distinguish a system and its parameters
from a system's environment and its variables. Part 1 also
includes discussions, for comparative purposes, of evaluation in
information retrieval, in speech processing, and in the social
sciences in general; and it considers the particular problems
presented by 'generic' system evaluation, i.e. how to evaluate
language processors independent of task applications, and by the
provision of general-purpose test data.

Part 2 is devoted to a detailed account of the present situation
in NLP evaluation. Interest and activity in evaluation has grown
rapidly in the last few years, as evinced by the DARPA
Conferences and by a number of workshops specifically addressing
evaluation. The description of actual evaluations covers major or
representative cases and is grouped by four main areas: machine
translation, message understanding, database query, and speech
understanding. One of the most important issues involved is the
extent to which NLP evaluation has to be task specific. The
description of explicit discussions of evaluation methodology (as
opposed to its exhibition in use) is devoted in particular to
accounts of and comments on the 1991 Berkeley and 1992 Edinburgh
Workshops; and the DARPA evaluations, which are having a
noticeable impact on evaluation, are collectively considered for
their overall contribution to evaluation methodology. This part
also briefly reviews the current state of speech processing
evaluation and of the design and provision of test resources like
corpora and toolkits. The report lists the main points emerging
from this review, for instance that evaluation has so far
focussed on systems without sufficient regard for their
environments, and finds that NLP evaluation methodology is still
relatively underdeveloped, for instance in frequently failing to
be clear about the precise remit and bounds of evaluation, and in
not addressing factor decomposition with enough rigour.

Part 3 addresses the future. The suggested approach to evaluation
is based on two major grounds: the first is that NLP evaluation
narrowly conceived is, if not a chimaera, of very limited value;
the second is that evaluation techniques cannot be at once
specific and generally applicable. The only approach to NLP
evaluation is through a proper evaluation {\it strategy}, which has two
main components: a complete statement of the evaluation {\it remit} and
a thorough decomposition of the whole system/setup into all its
{\it factors}. These together imply a systematic {\it grid} design for
evaluation and thus in general not a single test but a test
{\it programme}. Particular test and evaluation {\it techniques} can only be
chosen and applied after this explicit and detailed
characterisation of what is given and what is required has been
provided. These points are again illustrated by worked examples.
Thus the report conclusion is that while there are no magic
bullets for NLP evaluation, and the present standard must be
raised, this can be done by wholehearted pursuit of the strategy
described. In the process this will provide the more specific,
applicable knowhow needed to carry out actual evaluations that
are well-conducted and thus informative ones advancing NLP
itself.

\vspace*{5mm}
\noindent
(Approx, 175 pages, with Bibliography and Figures; available from
the Bookshop, Computer Laboratory, University of Cambridge, early
1993)

\newpage
\chapter*{References}

\newenvironment{reverseindent}%
{\begin{list}{}{\setlength{\labelsep}{0in} 
	        \setlength{\labelwidth}{0in}
	        \setlength{\itemindent}{-\leftmargin}}}%
{\end{list}}

\begin{reverseindent}

\item\pagebreak[3]
Abe, M., Y.~\={O}shima, K.~Yuura and N.~Takeichi. 1986. ``A
Kana-Kanji Translation System for Non-Segmented Input Sentences Based
on Syntactic and Semantic Analysis'', {\it Proceedings of the Eleventh
International Conference on Computational Linguistics}, 280--285.

\item\pagebreak[3]
Alshawi, H. and D.M. Carter. 1992. ``Optimal Scaling of Preference
Metrics''. Paper submitted for presentation at the 31st Annual Meeting of
the Association for Computational Linguistics.

\item\pagebreak[3]
Alshawi, H. 1990. ``Resolving Quasi Logical Forms''. 
{\it Computational Linguistics} 16:133--144.

\item\pagebreak[3]
Alshawi, H., ed. 1992. {\it The Core Language
Engine}. Cambridge, Massachusetts: The MIT Press.

\item\pagebreak[3]
Alshawi,~H., D.M.~Carter, M.~Rayner and B.~Gamb\"ack. 1991.
``Translation by Quasi Logical Form Transfer'', in {\it Proc.
29th ACL}, Berkeley, California.

\item\pagebreak[3]
Alshawi, H., Rayner, M., and Smith, A.G. 1991. ``Declarative Derivation of
Database Queries from Meaning Representations'', in Society for Worldwide
Interbank Financial Telecommunications S.C. (ed.)  {\it Proceedings of the
1991 BANKAI Workshop on Intelligent Information Access}, Elsevier,
Amsterdam.

\item\pagebreak[3]
Appelt,~D. and J.R.~Hobbs. 1990.
``Making Abduction more Efficient'', in {\it Proc. DARPA
Workshop on Speech and Natural Language}, Morgan Kaufmann,
San Mateo.

\item\pagebreak[3]
B\"ackstr\"om,\ M., K.\ Ceder and B.\ Lyberg 1989. 
``Prophon -- an Interactive Environment for Text-to-Speech
Conversion'', {\it Proceedings of the
European Conference on Speech, Communication and
Technology}, Paris, CPC Consultants Ltd, Edinburgh, UK.

\item\pagebreak[3]
Barwise, J., and R.~Cooper. 1981. ``Generalized Quantifiers and Natural 
Language''. {\it Linguistics and Philosophy} 4:159--219.

\item\pagebreak[3]
van Berkel, B., and K.~De Smedt. 1988. ``Triphone Analysis: A Combined
Method for the Correction of Orthographical and Typographical
Errors'', {\it Proceedings of the Second Conference on Applied
Natural Language Processing}, pp.~77--83.

\item\pagebreak[3]
Bronneberg, W.J.H.J., H.C. Bunt, S.P.J. Landsbergen, R.J.H. Scha,
W.J. Schoenmakers and E.P.C. van Utteren. 1980. ``The Question Answering
System PHLIQA1''. In L. Bolc (ed.), {\it Natural Language Question
Answering Systems}. Macmillan.

\item Calzolari, N. and R. Bindi. 1990. ``Acquisition of Lexical Information
from a Large Textual Italian Corpus''. {\it Proceedings of the 13th 
International Conference on Computational Linguistics}, 3:54--59.

\item\pagebreak[3]
Church, K.W. and P. Hanks. 1990. ``Word Association Norms, Mutual Information,
and Lexicography''. {\it Computational Linguistics} 16:22--30.

\item\pagebreak[3]
Carter, D.M. 1987. ``An Information-theoretic Analysis of Phonetic
Dictionary Access'', {\it Computer Speech and Language}, 2:1--11.

\item\pagebreak[3]
Carter, D.M. 1989. ``Lexical Acquisition in the Core Language
Engine'', {\it Proceedings of the Fourth Conference of the European
Chapter of the Association for Computational Linguistics},
137--144.

\item\pagebreak[3]
Ceder,\ K. and B.\ Lyberg 1990. ``The Integration of 
Linguistic Levels in a Text-to-Speech Conversion System'',
{\it Proceedings of the International Conference on Spoken Language Systems},
Kobe, Japan.

\item\pagebreak[3]
Clark, K.L.  1978.  ``Negation as Failure''.  In H.  Gallaire and J.
Minker (editors), {\it Logic and Data Bases}, Plenum, New York.

\item\pagebreak[3]
Coltheart, M. 1981. ``The MRC Psycholinguistic Database''. {\it
Quarterly Journal of Experimental Psychology}, 33A:497--505.

\item\pagebreak[3]
Dalrymple, M., S.~M.~Shieber, and F.~C.~N.~Pereira. 1991.
``Ellipsis and Higher-Order Unification''. {\it Linguistics and Philosophy},
14:399--452.

\item\pagebreak[3]
Davidson, D. 1967.
``The Logical Form of Action Sentences''. 
Reprinted in Davidson, 1980, {\it Essays on Actions and Events}, 
Oxford University Press.

\item\pagebreak[3]
De Marcken, C.~G.. 1990. ``Parsing the LOB corpus'', {\it Proceedings
of the 28th ACL},
Pittsburgh, 243--251.

\item\pagebreak[3]
DeRose, S. 1988. ``Grammatical Category Disambiguation by Statistical
Optimisation'', {\it Computational Linguistics}, 14:31--39.

\item\pagebreak[3]
Emirkanian, L., and L.H.~Bouchard. 1988. ``Knowledge Integration in a
Robust and Efficient Morpho-syntactic Analyser for French'', {\it
Proceedings of the Twelfth International Conference on Computational
Linguistics}, pp~166--171.
 
\item\pagebreak[3]
Futrelle, R.P., C.E.~Dunn, D.S.~Ellis and M.J.~Pescitelli, Jr. 1991.
``Preprocessing and Lexicon Design for Parsing Technical Text'', {\it
Proceedings of the Second International Workshop on Parsing
Technologies}, pp.~31--40.
 
\item\pagebreak[3]
Grosz, B.~J., D.~E.~Appelt, P.~Martin, and F.~Pereira. 1987. ``TEAM:  An
Experiment in the Design of Transportable Natural-Language Interfaces''. {\it
Artificial Intelligence} 32: 173--243.

\item\pagebreak[3]
Hirschb\"uhler, P. 1982. ``VP Deletion and Across the Board Quantifier Scope''.
{\it Proceedings of NELS 12}, GLSA, University of Massachusetts.

\item Hindle, D. and M. Rooth. 1990. ``Structural Ambiguity and Lexical
Relations''. {\it Proceedings of the DARPA Speech and Natural Language 
Workshop, Morgan Kaufman}, 257--262.

\item\pagebreak[3]
Hobbs, J.~R. and J.~Bear. 1990. ``Two Principles of Parse Preference''. Vol. 3,
Proceedings of the 13th International Conference on Computational Linguistics,
Helsinki, 162--167.

\item\pagebreak[3]
Hobbs,~J.R., M.~Stickel, P.~Martin and D.~Edwards. 1988.
``Interpretation as Abduction'', in {\it Proceedings of the 26th ACL},
Buffalo, New York.

\item\pagebreak[3]
Hobbs, J.~R., and S.~M.~Shieber. 1987. ``An Algorithm for Generating
Quantifier Scopings''. {\it Computational Linguistics} 13:47--63.

\item\pagebreak[3]
Huet, G. 1975. ``A Unification Algorithm for Typed Lambda Calculus''.
{\it Theoretical Computer Science} 1:27--57.

\item\pagebreak[3]
Kamp, H. 1981. ``A Theory of Truth and Semantic Representation''. 
In {\it Formal Methods in the Study of Language}, edited by 
J.~A.~G. Groenendijk, T.~M.~V. Janssen, and M.~B.~J. Stokhof. 
Mathematisch Centrum, Amsterdam.

\item\pagebreak[3]
Kernighan, M.D.,~K.W.~Church,~and~W.A. Gale. 1990. ``A Spelling
Correction Program Based on a Noisy Channel Model'',  {\it
Proceedings of the Thirteenth International Conference on Computational
Linguistics}, pp~205--210.

\item\pagebreak[3]
Konolige, K. 1981. {\it The Database as Model: A Metatheoretic
Approach}, SRI technical note 255.  

\item\pagebreak[3]
Korf, R. 1986. ``Search: A Survey of Recent Results'' in 
{\it Exploring Artificial Intelligence: Survey Talks
from the National Conferences on Artificial Intelligence}, Morgan
Kaufmann, San Mateo.

\item\pagebreak[3]
Koskenniemi, K. 1983. {\it Two-level morphology: a general
computational model for word-form recognition and production}.
University of Helsinki, Department of General Linguistics,
Publications, No.~11.

\item\pagebreak[3]
Lewin, I. 1990.
``A Quantifier Scoping Algorithm without a Free Variable Constraint'',
Proceedings of COLING 1990.

\item\pagebreak[3]
McCord, M.C. 1989. ``A New Version of Slot Grammar''. IBM Research 
Report RC 14506, IBM Thomas J. Watson Research Center, Yorktown Heights,
New York.

\item\pagebreak[3]
McCord,~M.C. 1987. ``Natural Language Processing in Prolog'', in
A. Walker (ed.) {\it Knowledge Systems and Prolog}. Addison-Wesley,
Reading, MA.

\item\pagebreak[3]
Miller, G.A., ed. (1990). ``WordNet: An On\-Line Lexical Database'',
{\it International Journal of Lexicography}, 3:4.

\item\pagebreak[3]
Moran, D.~B. 1988. ``Quantifier Scoping in the SRI Core Language Engine''.
Proceedings of the 26th Annual Meeting of the Association for Computational 
Linguistics, State University of New York at Buffalo, Buffalo, 
New York, 33--40.

\item\pagebreak[3]
Murveit, H., J. Butzberger and M. Weintraub (1991). 
``Speech Recognition in SRI's Resource Management and ATIS Systems''.
{\it Proceedings of the Speech and Natural Language Workshop}, Morgan
Kaufmann.

\item\pagebreak[3]
Pereira, F.~C.~N. 1990.
``Categorial Semantics and Scoping'', {\it Computational Linguistics}
16:1 1--10.

\item\pagebreak[3]
Pereira,~F.C.N. and M.E.~Pollack. 1991. ``Incremental Interpretation'',
{\it Artificial Intelligence} {\bf 50} 37-82.

\item\pagebreak[3]
Pereira,~F.C.N. and S.M.~Shieber. 1985. {\it Prolog and Natural-Language
Understanding}, CLSI Lecture Notes.

\item\pagebreak[3]
Perrault, C. Raymond and Barbara J. Grosz. 1988. ``Natural Language
Interfaces'' in {\it Exploring Artificial Intelligence: Survey Talks
from the National Conferences on Artificial Intelligence}, Morgan
Kaufmann, San Mateo.

\item\pagebreak[3]
Pollock, J.J., and A.~Zamora. 1984. ``Automatic Spelling Correction in
Scientific and Scholarly Text'', {\it Communications of the ACM},
27:4, pp~358--368.

\item\pagebreak[3]
Quinlan,~P. 1992. {\it The Oxford Psycholinguistic Database}. 
Oxford University Press.

\item\pagebreak[3]
Quirk,~R., S.~Greenbaum, G.~Leech and J.~Svartvik. 1985. {\it A Comprehensive
Grammar of the English Language}. Longman.

\item\pagebreak[3]
Rayner,~M. and S.~Janson. 1987. ``Epistemic Reasoning, Logic
Programming, and the Interpretation of Questions''. In {\it Proc. 
2nd International Workshop on Natural Language Understanding 
and Logic Programming}, 
North-Holland.

\item\pagebreak[3]
Rayner,~M. and S.~Janson. 1989. ``Finding Out = Achieving
Decidability'', in {\it Working notes: IJCAI workshop on 
Knowledge, Planning and Perception}, Detroit, Michigan. 
Also available as Research
Report R89015, Swedish Institute of Computer Science, Kista, Sweden.

\item\pagebreak[3]
Scha,~R.J.H. 1983. {\it Logical Foundations for Question Answering},
Ph.D. Thesis, University of Groningen, the Netherlands.

\item\pagebreak[3]
Sekine, S., J.J. Carroll, S. Ananiadou, and J. Tsujii. 1992. ``Automatic
Learning for Semantic Collocation''. {\it Proceedings of the Third Conference
on Applied Natural Language Processing}, ACL, 104--110.

\item\pagebreak[3]
Shieber, S.~M., G.~van~Noord, F.~C.~N.~Pereira, and R.~C.~Moore. 1990.
``Semantic-Head-Driven Generation''. 
{\it Computational Linguistics} 16:30--43.

\item\pagebreak[3]
Stallard,~D.G. 1986. ``A Terminological Simplification Transformation
for Natural Language Question-Answering Systems''. {\it Proceedings of the 
24th Annual Meeting of the Association for Computational Linguistics}, ACL,
241--246.

\item\pagebreak[3]
Stickel, M. E. 1986. ``A Prolog technology theorem prover: implementation by an
extended Prolog compiler.'', {\it Journal of Automated Reasoning}, 4,
353-380.

\item\pagebreak[3]
Sterling,~L. and S.~Shapiro. 1985. {\it The Art of Prolog}, Addison-Wesley,
Reading, MA.

\item\pagebreak[3]
Veltman, F. 1990.
``Defaults in Update Semantics'', in H. Kamp (ed), {\it Conditionals, Defaults
and Belief Revision}, DYANA deliverable R2.5.A.

\item\pagebreak[3]
Veronis, J. 1988. ``Morphosyntactic Correction in Natural Language
Interfaces'', {\it Proceedings of the Twelfth International Conference
on Computational Linguistics}, pp~708--713.

\item\pagebreak[3]
Webber,~B.L. 1983. ``So What Shall we Talk About Now''. Reprinted in
Grosz, Sparck-Jones and Webber (eds), 1986, {\it Readings in Natural Language
Processing}. Morgan Kaufmann.
\end{reverseindent}

\end{document}